\documentclass[journal]{IEEEtran}
\usepackage{graphicx}
\usepackage{booktabs}
\usepackage{subfig}
\usepackage{enumitem}
\usepackage{algorithmic}
\usepackage{algorithm}
\usepackage{multirow}
\usepackage{tikz}
\usetikzlibrary{cd, decorations.pathmorphing}
\usepackage[accsupp]{axessibility}  % Improves PDF readability for those with disabilities.
\usepackage{amsmath,amssymb} % define this before the line numbering.

\newcommand{\longlongleftrightarrow}{\begin{tikzcd}[cramped,sep=10em,ampersand replacement=\&]{}\arrow[r, leftrightarrow]\&{}\end{tikzcd}} % 長い単射

\usepackage{orcidlink}

\usepackage{cite}
\ifCLASSINFOpdf
\else
\fi
\usepackage{amsmath}
\usepackage{algorithmic}
\usepackage{array}
\usepackage{url}
\begin{document}
\title{PHISWID: Physics-Inspired Underwater Image Dataset Synthesized from RGB-D Images}

\author{Reina Kaneko,~\IEEEmembership{Student Member,~IEEE,}
        Takumi Ueda,
        Hiroshi Higashi,~\IEEEmembership{Member,~IEEE,}
        and~Yuichi Tanaka,~\IEEEmembership{Senior Member, IEEE} %<-this % stops a space
\thanks{Reina Kaneko, Hiroshi Higashi, and Yuichi Tanaka are with the Graduate School of Engineering, The University of Osaka, Suita, Osaka 565--0861, Japan (e-mail: r.kaneko@sip.comm.eng.osaka-u.ac.jp; higashi@comm.eng.osaka-u.ac.jp; ytanaka@comm.eng.osaka-u.ac.jp).}
\thanks{Takumi Ueda was with the Graduate School of BASE, Tokyo University of Agriculture and Technology, Koganei 184--8588, Japan.}}
% \thanks{Hiroshi Higashi and Yuichi Tanaka are with the Graduate School of Engineering,
% The University of Osaka, Suita, Osaka 565-0861, Japan (e-mail: higashi@comm.eng.osaka-u.ac.jp; ytanaka@comm.eng.osaka-u.ac.jp).}% <-this % stops a space
% \thanks{The preliminary version of this paper was presented in \cite{ueda2019underwater}.}}% <-this % stops a space
% \thanks{Manuscript received April 19, 2005; revised August 26, 2015.}}


\markboth{Journal of \LaTeX\ Class Files,~Vol.~14, No.~8, August~2015}%
{Shell \MakeLowercase{\textit{et al.}}: Bare Demo of IEEEtran.cls for IEEE Journals}



\maketitle

\begin{abstract}
This paper introduces the physics-inspired synthesized underwater image dataset (PHISWID), a dataset tailored for enhancing underwater image processing through physics-inspired image synthesis.
For underwater image enhancement, data-driven approaches (e.g., deep neural networks) typically demand extensive datasets, yet acquiring paired clean atmospheric images and degraded underwater images poses significant challenges.
Existing datasets have limited contributions to image enhancement due to lack of physics models, publicity, and ground-truth atmospheric images.
% While several underwater image datasets have been proposed using physics-based synthesis, a publicly accessible collection has been lacking.
% Additionally, most underwater image synthesis approaches do not intend to reproduce atmospheric scenes, resulting in limited enhancement.
PHISWID addresses these issues by offering a set of paired atmospheric and underwater images.
Specifically, underwater images are synthetically degraded by color degradation and marine snow artifacts
%, a composite of organic matter and sand particles that considerably impairs underwater image clarity.
% Generating underwater images 
from atmospheric RGB-D images.
It is enabled based on a physics-based underwater image observation model.
% al models provides pairs of real-world ground-truth and degraded images.
% The dataset applies these degradations to atmospheric RGB-D images, enhancing the dataset’s realism and applicability.
Our synthetic approach generates a large quantity of the pairs, enabling effective training of deep neural networks and objective image quality assessment.
%PHISWID is particularly valuable for training deep neural networks in a supervised learning setting and for objectively assessing image quality in benchmark analyses.
Through benchmark experiments with some datasets and image enhancement methods, we validate that our dataset can improve the image enhancement performance.
%Our results reveal that even a basic deep neural network, when trained with PHISWID, substantially outperforms existing methods specifically designed for underwater image enhancement.
Our dataset, which is publicly available, contributes to the development in underwater image processing.
% PHISWID is publicly available.
% , contributing a significant resource to the advancement of underwater imaging technology.

\end{abstract}


% \fi
\IEEEpeerreviewmaketitle

\section{Introduction}
% % The very first letter is a 2 line initial drop letter followed
% % by the rest of the first word in caps.
% % 
% % form to use if the first word consists of a single letter:
% % \IEEEPARstart{A}{demo} file is ....
% % 
% % form to use if you need the single drop letter followed by
% % normal text (unknown if ever used by the IEEE):
% % \IEEEPARstart{A}{}demo file is ....
% % 
% % Some journals put the first two words in caps:
% % \IEEEPARstart{T}{his demo} file is ....
% % 
% % Here we have the typical use of a "T" for an initial drop letter
% % and "HIS" in caps to complete the first word.
% for IEEE journal papers produced under \LaTeX\ using
% IEEEtran.cls version 1.8b and later.
% % You must have at least 2 lines in the paragraph with the drop letter
% % (should never be an issue)
Image restoration and enhancement \cite{zhangGaussianDenoiserResidual2017a,chan2016plug,romano2017little,xie2012image,chen2018deep,xu2014deep,yeh2017semantic} has been a long-studied topic for decades, with substantial advancements in the deep learning era. This progress is largely due to the availability of extensive datasets comprising pairs of ground-truth and degraded images. However, the situation changes dramatically when dealing with extreme image restoration scenarios, including underwater, satellite, and medical imaging \cite{akkaynak2018revised,berman2017diving,li2019underwater,ueda2019underwater,wangDeepCNNMethod2017,li2020underwater,jiang2020novel,chen2014vehicle,zhang2016deep,zhu2017deep,maggiori2016convolutional,shen2017deep,litjens2017survey,weigert2018content,anwarDivingDeeperUnderwater2020}. These contexts present atypical degradation patterns that diverge from conventional Gaussian models, and they suffer from a scarcity of training images for algorithm refinement, particularly for deep neural networks.

\begin{figure}[t]
\begin{center}
   \subfloat[]{\includegraphics[width =0.4\linewidth]{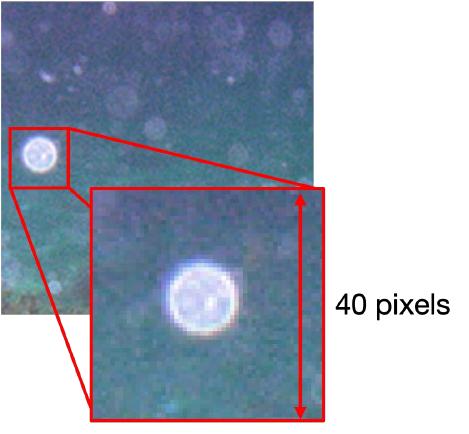}}
   \subfloat[]{\includegraphics[width =0.43 \linewidth]{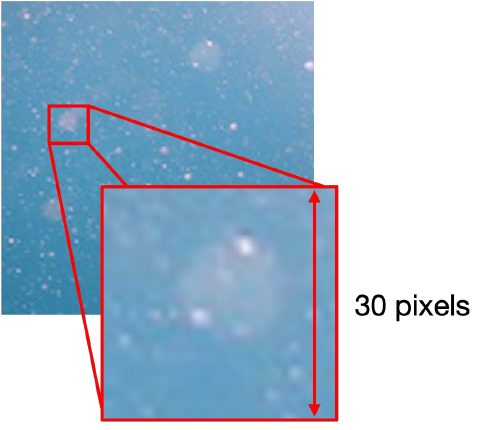}}
\end{center}
   \caption{Marine snow artifacts in real underwater images.}
\label{realMarineSnow}
\end{figure}

Underwater image enhancement is one of the extreme image restoration challenges.
In underwater, light (especially their red components) is strongly absorbed and scattered, rather than that in atmospheric scenes.
This is due to two ``depths'' in underwater scenes: 1) horizontal distances between the camera and scene (this is the typical ``depth'' in regular image processing) and 2) water depth.
Both of them affect light absorption and scattering properties, which result in shifting the color of images in underwater \cite{jerlov}.
Images are also degraded due to \textit{marine snow} \cite{MSRBDataset}, whose artifacts are visible as white dots due to light reflecting off particles like organic matter and sand.
Its example is shown in Fig. \ref{realMarineSnow}.
As a result, underwater image degradation is significantly different from atmospheric images, which results in requirements for image enhancement methods designed for underwater scene.

\begin{figure}[t]
    \centering
    % \subfloat{\includegraphics[width = .2 \linewidth]{image/ori.pdf}}
    % \subfloat{\includegraphics[width = .2 \linewidth]{image/deg.pdf}}
    % \subfloat{\includegraphics[width = .2\linewidth]{image/PHISWID.pdf}}
    % \subfloat{\includegraphics[width = .2\linewidth]{image/LSUI.pdf}}
    % \subfloat{\includegraphics[width = .2 \linewidth]{image/UIEB.pdf}}\\\vspace{-0.15in}
    \setcounter{subfigure}{0}
    \subfloat[Original image]{\includegraphics[width = .4 \linewidth]{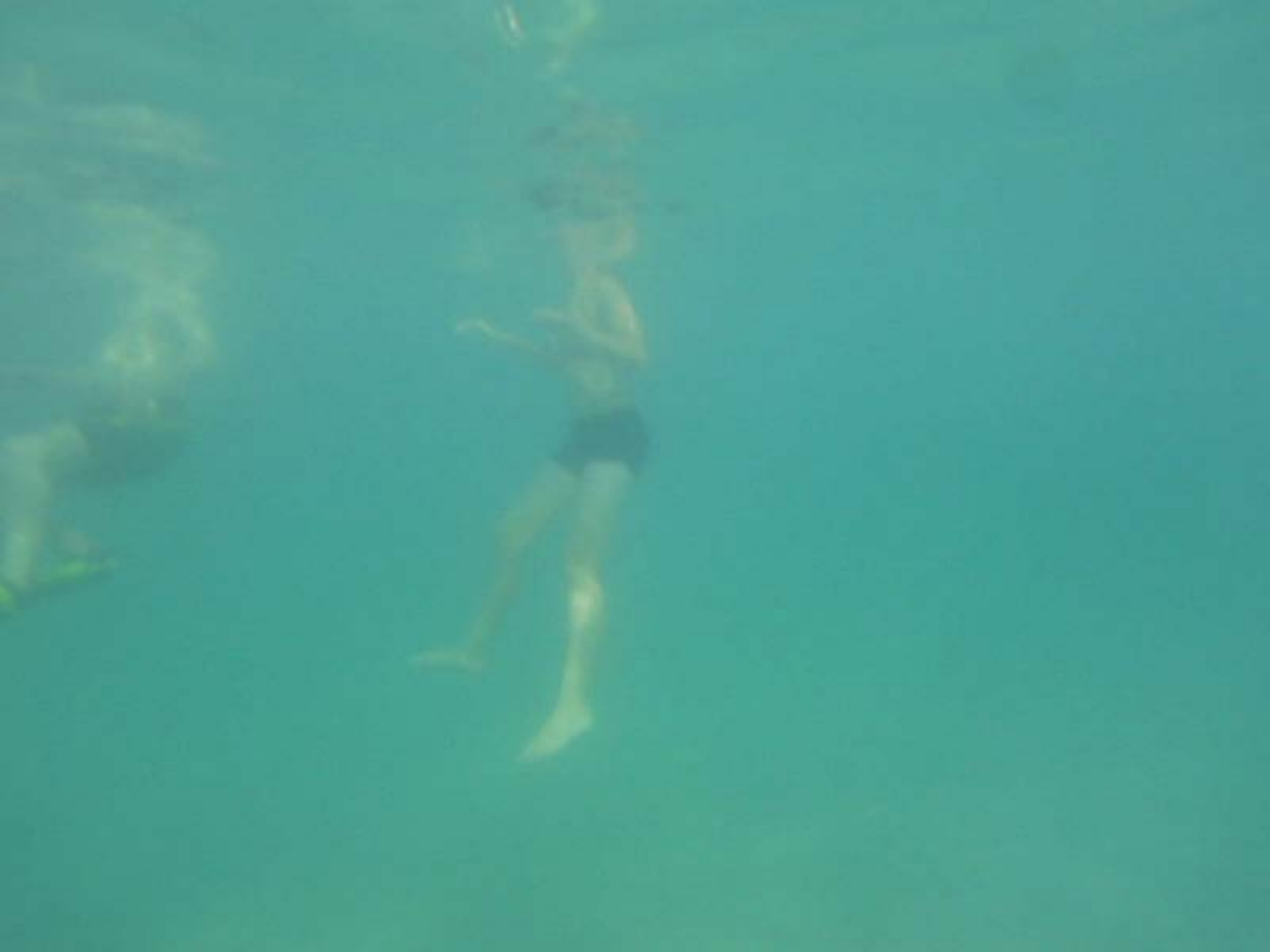}}
    % \subfloat[Test image in PHISWID]{\includegraphics[width = .2 \linewidth]{image/deg.pdf}}
    \subfloat[Enhanced result of PHISWID.]{\includegraphics[width = .4\linewidth]{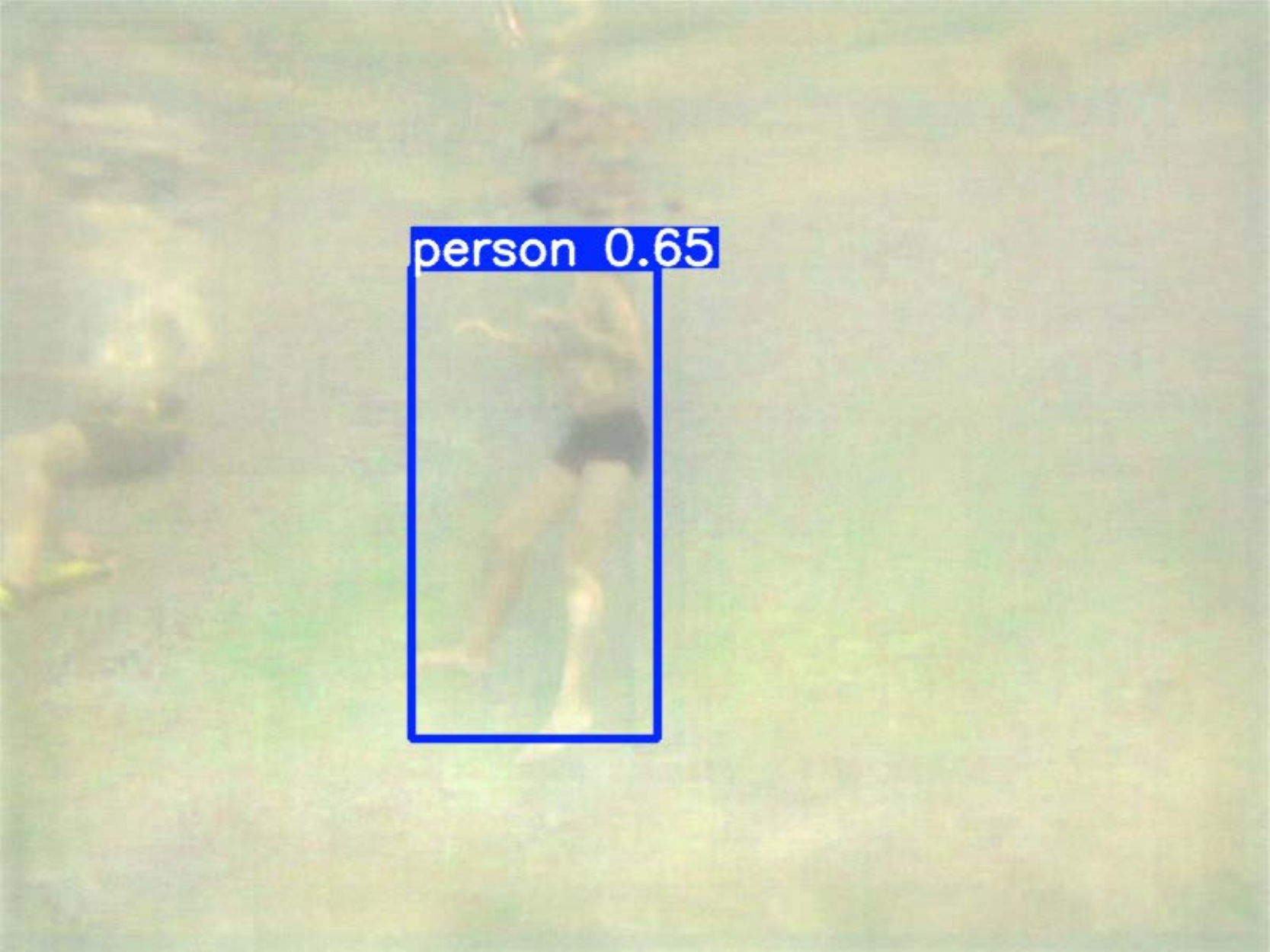}}\\
    \subfloat[Enhanced result of LSUI.]{\includegraphics[width = .4\linewidth]{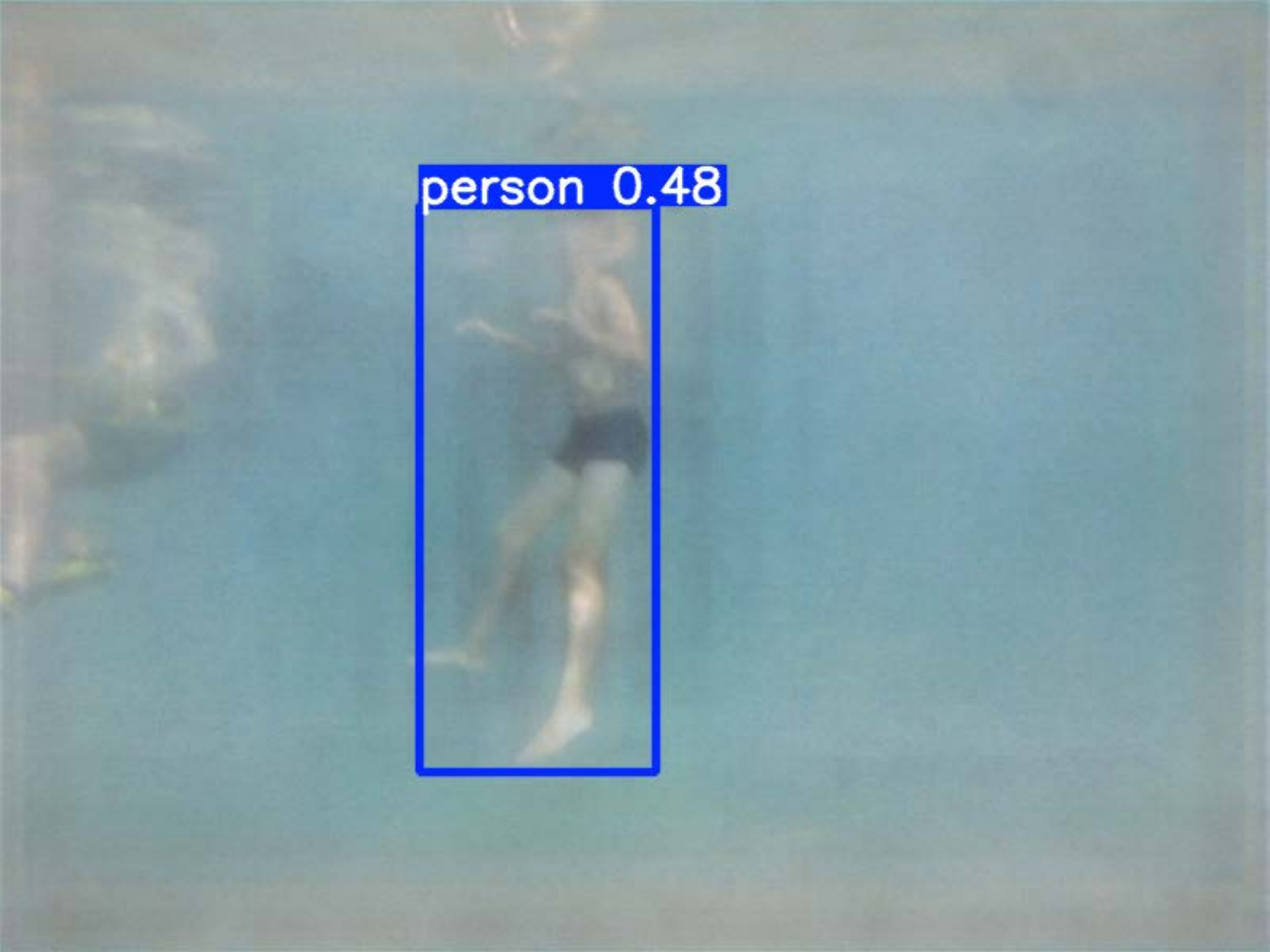}}
    \subfloat[Enhanced result of UIEB.]{\includegraphics[width = .4 \linewidth]{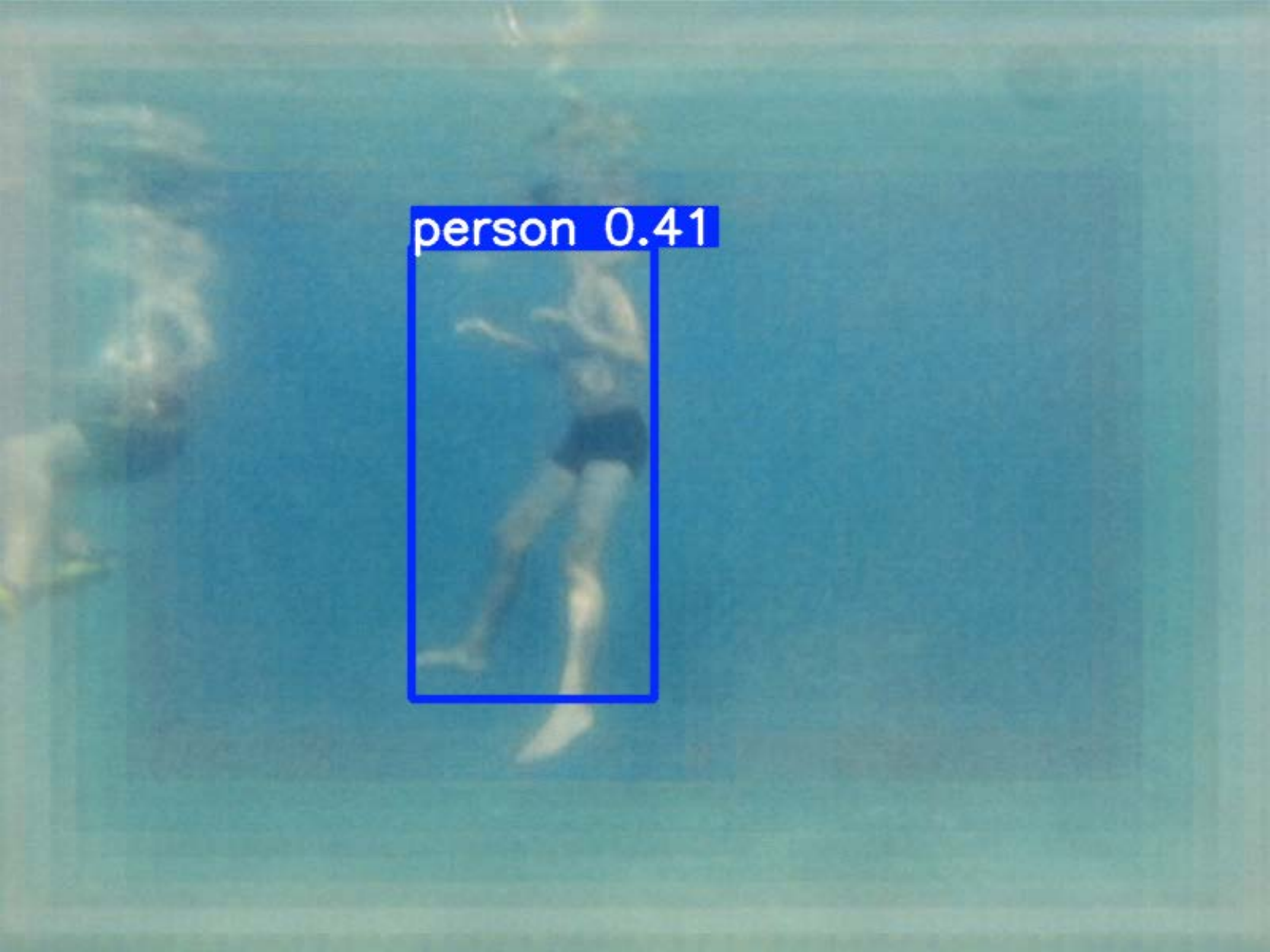}}
    \caption{Object detection results of a test image in the EUVP dataset \cite{islamFastUnderwaterImage2020}. We use a YOLOv5 pretrained model \cite{hussainYOLOv5YOLOv8YOLOv102024} for the object detection experiment. The blue squares are estimated bounding boxes and the numbers are confidence score ($[0, 1]$).  (a) original image (an object is not detected). (b) enhanced image based on PHISWID. (c) enhanced image based on LSUI. (d) enhanced image based on UIEB. The enhancement method is the same as that in the experiment in Section \ref{sec:benchmark}.}
    \label{objectdetection}
\end{figure}

Many underwater image enhancement methods, both model- and deep learning-based ones, have been proposed \cite{sharma2023wavelength,li2019underwater,pengUShapeTransformerUnderwater2023e,wangMetalantisComprehensiveUnderwater2024,tolieDICAMDeepInception2024,Zhao_2024_CVPR,fabbriEnhancingUnderwaterImagery2018,islamFastUnderwaterImage2020}.
While their building blocks vary, most works aim to the same goal---producing \textit{clean underwater images}.
They may be enough for entertainment purposes, however, in many situations like rescue, defense, bioprospecting/resource exploration, we often need \textit{enhanced underwater images looks like atmospheric images}.
For example, we can use an object detection system designed for regular atmospheric images as-is when underwater images are restored like atmospheric ones.
% ...
Fig. \ref{objectdetection} shows an example of object detection for a real underwater image and enhanced ones.
% It is taken from a test image of EUVP dataset \cite{islamFastUnderwaterImage2020}. 
We simply use a YOLOv5 pretrained model \cite{hussainYOLOv5YOLOv8YOLOv102024} for the object detection experiment. 
As can be seen, a person cannot be detected with the original underwater image.
All of the enhanced images help to detect the person, but the confidence scores are different: 
%The image without the blueish region works the best.
Reducing blueish color shift will contribute the accurate detection.
% The method trained with PHISWID restores an image like their atmospheric counterparts, and therefore it shows a good score in object detection compared with other datasets.

The primary hurdle for underwater image enhancement, especially for yielding atmospheric-like images, lies in the absence of ground-truth images.
Unlike regular image processing, obtaining clean atmospheric counterparts for underwater images is not feasible.
% Namely, only degraded images compounded by varying conditions such as sea areas, lighting, depths, and particulate matter are available. 
% This absence of ground truth makes it difficult to objectively assess restoration quality.
%obstructs an objective assessment of image quality and restoration effectiveness. 
The lack of pairs of ground-truth and degraded images hampers neural network training, as traditional loss functions (e.g., peak signal to noise ratio (PSNR) and structual similarity (SSIM)) evaluate the error between these pairs \cite{zhouwangImageQualityAssessment2004}.

%To tackle the lack of ground-truth images, 
Various underwater image datasets and enhancement methods have tackled to address this lack of ground-truth images \cite{Ancuti2018ColorBA,ueda2019underwater,li2020underwater,dudhane2020deep,jiang2020novel,9775132,9426457,zhuangUnderwaterImageEnhancement2022a}.
However, we face the following three problems:
\begin{enumerate}
    \item Many datasets are not publicly available \cite{dudhane2020deep,wangDeepCNNMethod2017,anwarDeepUnderwaterImage2018}.
    \item They often produce clean images as \textit{reverse engineering}: Ground-truth images are obtained by (manual or automatic) enhancement of underwater images. This process does not take into account the above-mentioned two ``depths'' information in underwater since it is not accompanied with the images.
    Moreover, enhanced underwater images, i.e., cleaned underwater images, are used as ground-truth \cite{li2019underwater,pengUShapeTransformerUnderwater2023e,jiangUnderwaterImageEnhancement2022a}.
    \item Many methods overlook a critical aspect of marine snow. This phenomenon degrades overall image quality, yet its removal techniques are often limited to classical median-filtering approaches \cite{banerjeeEliminationMarineSnow2014,cyganekRealtimeMarineSnow2018,koziarskiMarineSnowRemoval2019}.
\end{enumerate}

To tackle the above-mentioned challenges, this paper propose a new dataset, the physics-inspired synthesized underwater image dataset (PHISWID), providing pairs of \textit{real atmospheric ground-truth images} and synthesized underwater images with color degradation and marine snow artifacts taking into account the two-depths information \cite{roomdataset,tartanair2020iros}.
%containing both color degradation and marine snow artifacts \cite{roomdataset,tartanair2020iros}.
%First, we mathematically model color shift based on 
Our dataset is based on physical observation models which enable to synthesis underwater image from real-world RGB-D images. For the physical observation models, the light absorption and scattering model in underwater is used to model color shift \cite{jerlov,akkaynakwhatis2017} 
% because existing datasets often oversimplify color degradation processes.
%Second, we model the 
and the light scattering model for underwater particle to marine snow \cite{mcglamery}.
%Finally, we synthesize color shift and marine snow artifacts to clean atmospheric RGB-D images.
We apply these models to an atmospheric RGB-D image to generate its synthetic underwater image, obtaining a pair of the ground-truth and underwater images.
% With PHISWID, we aim to advance underwater image enhancement to a level where images appear as if captured in pristine atmospheric conditions.

%Our benchmarking with PHISWID reveals intriguing results. 
Our dataset is validated with various image enhancement methods and some existing datasets. We show that the learning-based methods trained with our dataset outperforms the ones trained with the other datasets \cite{li2019underwater,pengUShapeTransformerUnderwater2023e}, even with basic Transformer \cite{wang_2022_CVPR}.
%trained with our PHISWID  outperforms specialized neural networks designed for underwater image enhancement \cite{wang_2022_CVPR,sharma2023wavelength,pengUShapeTransformerUnderwater2023e,zhuangUnderwaterImageEnhancement2022a,li2019underwater}.
% , applicable to both synthesized and real images. 
Our benchmark results suggest a potential contribution of our dataset, which is made publicly available, to development of the underwater image processing\footnote{https://github.com/reina0112/PHISMID-PHISWID.}.

\begin{table*}[t]
\caption{List of underwater image datasets.}
\label{tab:dataset}
\centering
\begin{tabular}{c|c|c|c|c}
\hline
Dataset                                                              & Target                                                               & Number of Images & Method                                                                         & Availability        \\ \hline\hline
PHISWID (proposed)                                                   & Color shift \& Marine snow & 4195                                                       & Physics-based synthesis              & \checkmark     \\ \hline
Dataset used in \cite{dudhane2020deep}              & Color shift \& Haze    & 10143                                                      & Physics-based synthesis \& Filtering &  \\ \hline
UIEB \cite{li2019underwater}                                                                 & Color shift                                                          & 860                                                        & Voting                                                                         & \checkmark    \\ \hline
LSUI \cite{pengUShapeTransformerUnderwater2023e}                                                                & Color shift                                                          & 4279                                                       & Voting                                                                         & \checkmark     \\ \hline
SAUD \cite{jiangUnderwaterImageEnhancement2022a}                                                                & Color shift                                                          & 100                                                        & Voting                                                                         & \checkmark     \\ \hline
EUVP \cite{islamFastUnderwaterImage2020}                                                                           & Color shift                                                          & 12000                                                      & Cycle GAN                                                                      & \checkmark     \\ \hline
Dataset used in \cite{wangDeepCNNMethod2017}        & Color shift                                                          & 200                                                        & Physics-based synthesis           &  \\ \hline
Dataset used in \cite{anwarDeepUnderwaterImage2018} & Color shift                                                          & 1449                                                       & Physics-based synthesis              &  \\ \hline
% PHISMID (proposed)                                                   & Marine snow                                                          & 400                                                        & \begin{tabular}[c]{@{}c@{}}Physics-based\\ synthesis\end{tabular}              & \checkmark     \\ \hline
MSRB \cite{MSRBDataset}                                                                & Marine snow                                                          & 2700                                                       & Observation-based synthesis    & \checkmark     \\ \hline
\end{tabular}
\end{table*}

The preliminary version of this paper was presented in \cite{ueda2019underwater}.
This paper extends it by 1) incorporating marine snow artifacts with physics-based observation models in the dataset, 2) improving the image synthesis process and enhancing resolutions, and 3) providing comprehensive comparisons with alternative datasets and restoration methods.

The remainder of this paper is organized as follows:
Section \ref{sec:relatedworks} reviews related studies for underwater image enhancement.
We propose underwater image synthesis methods along with the marine snow observation model and the color shift model in Section \ref{sec:marine_snow_models}.
The PHISWID specifications are introduced in Section \ref{sec:dataset}.
The benchmarking results with PHISWID are shown in Section \ref{sec:benchmark}.
Real-world examples and limitations are described in Section~\ref{sec:real_msr}.
Finally, we provide some concluding remarks in Section \ref{sec:conclusion}.

\section{Related Work}
\label{sec:relatedworks}
% \subsection{Marine Snow Model}

In this section, we introduce previous works on underwater image enhancement.
We first introduce underwater image enhancement and available datasets, then focus on marine snow removal.
Currently available datasets are listed in Table~\ref{tab:dataset}.

\subsection{Underwater Image Enhancement and Available Datasets}
\label{subsec:underwater}
The current trend of underwater image enhancement is based on deep learning as with the other image enhancement/restoration tasks\cite{sharma2023wavelength,li2019underwater,pengUShapeTransformerUnderwater2023e,wangMetalantisComprehensiveUnderwater2024,tolieDICAMDeepInception2024}.
In \cite{sharma2023wavelength}, a wavelength-attributed deep neural network is proposed.
% It first extracts red, green, and blue color channels from the input image.
% Then, three convolutional neural networks individually perform a wavelength-specific enhancement on each channel.
% The outputs are fused to reconstruct the final color image.
% By leveraging the wavelength information in this way, the model's generalization ability is improved. 
In \cite{pengUShapeTransformerUnderwater2023e}, a transformer designed for underwater image enhancement is presented.
%A color correction module followed by an optical model-based dehazing module is also proposed in \cite{li2019underwater}. 
A color correction module followed by an optical model is used in \cite{li2019underwater,zhuangUnderwaterImageEnhancement2022a}. 
% The color correction module uses a retinal mechanism inspired color correction model to adjust brightness and chromaticity.
% The dehazing module derives an optical model of light attenuation and scattering to estimate scene depth and recovers the latent clean image. 
%These methods were trained with a well-aligned dataset for underwater image enhancement.
These methods were trained with datasets including pairs of ground-truth (clean) and underwater (degraded) images for underwater image enhancement.
However, collecting extensive pairs in underwater situations is still a challenge.
This limits the performance of deep learning-based techniques.

An approach for creating the image pairs is that
a clean (ground-truth-like) image is selected among images processed by various enhancement methods by human voting.
We call it \textit{voting-based approach}.
%It enhances the real-world underwater images with several model-based methods and one of the enhanced images is manually chosen as a ground truth.
% Many methods including \cite{sharma2023wavelength, li2019underwater} are trained with
% To train the network, the 
UIEB dataset \cite{li2019underwater}, which is widely utilized for developing many enhancement methods \cite{sharma2023wavelength, li2019underwater}, provides 860 real-world underwater images and corresponding ground-truth-like images across various water conditions.
LSUI dataset \cite{pengUShapeTransformerUnderwater2023e} is also widely used  \cite{pengUShapeTransformerUnderwater2023e,Zhao_2024_CVPR}. It contains 4279 real-world underwater images and the corresponding reference images.
SAUD dataset \cite{jiangUnderwaterImageEnhancement2022a} contains 100 raw underwater images and enhanced images used some existing methods.
% These datasets, however, make corresponding clean images by enhancing the real-world underwater images with several \textit{model-based methods}.
% One of the enhanced images is manually chosen as a ground-truth.
In the voting-based datasets, there is a possibility that no enhancement method works well on a certain underwater image. %and the selected ground-truth-like image would fairly different to the true ground-truth image. This issue could ruin the reliability of the datasets.
%Unfortunately, if none of the model-based methods works well for the enhancement, the resulting image pair may be inappropriate.
%Furthermore, 
%neural networks trained with such hand-crafted voting-based datasets 
%since the selected ground-truth-like image is cleaned by an enhancement method,
Therefore, they do not guarantee that the clean image is close to the \textit{true} ground-truth image which is taken under the atmospheric environment.
This gap in the ground-truth images could ruin the reliability of the datasets and lead to limited performance.
%since ``clean'' images in the dataset are still processed underwater images, not atmospheric images taken on the ground.
% Several methods including \cite{pengUShapeTransformerUnderwater2023e,Zhao_2024_CVPR} use LSUI dataset \cite{pengUShapeTransformerUnderwater2023e}. LSUI dataset was created in a similar way to UIEB dataset.

\textit{Generative adversarial network} (GAN)-based approach \cite{jiang2020novel,liWaterGANUnsupervisedGenerative2018,fabbriEnhancingUnderwaterImagery2018,islamFastUnderwaterImage2020} is also used to generate clean, ground-truth-like images.
The GANs are trained with sets of clean and degraded underwater images.
%It uses GANs to make clean underwater images from degraded underwater images.
A critical limitation of this approach, which does not provide pairs of ground-truth and degraded images, is that well-known full-reference image quality metrics like PSNR and SSIM cannot be computed to evaluate the enhancement/restoration accuracy.
% Furthermore, some methods use clean underwater images and degraded underwater images to pair, therefore these methods may not remove color shift completely.
% And a number of challenges to improving the restoration performance still exist, some promising results have been presented.
Moreover, many datasets are not publicly available and their contribution to image restoration development is limited.
Furthermore, there is no dataset including both marine snow artifacts and color shift degradation.

Unlike the voting-based and GAN-based approaches, which generate the ground-truth-like images from underwater images, there is an approach to generate underwater images from ground-truth images which are taken under the atmospheric environment.
We call it \textit{physics-based synthesis approach}.
% It is the opposite way to the voting-based approach: They synthetically generate
In this approach, degraded underwater images are generated from atmospheric images by simulating the degradation processes of absorption and scattering \cite{wangDeepCNNMethod2017,anwarDeepUnderwaterImage2018,ueda2019underwater,dudhane2020deep}.
This approach requires precise models of degradation.
However, some datasets \cite{wangDeepCNNMethod2017,anwarDeepUnderwaterImage2018} are based on oversimplified color degradation processes.
They ignore the light scattering and the radiance ratio of the scene reaching the camera.
This synthesis may be incomplete for underwater image enhancement purposes. Furthermore, they are not available online (see Table \ref{tab:dataset}). It leads to a limited contribution. 

\subsection{Marine Snow Removal}
Marine snow is one of the main sources of underwater image degradation but has been underaddressed so far. Here, we briefly review some existing marine snow removal approaches.

A widely used method for marine snow removal is a median filter (MF) \cite{brownrigg1984weighted,huang1979fast,banerjeeEliminationMarineSnow2014,farhadifardSingleImageMarine2017}.
If marine snow artifacts are assumed sufficiently small (typically $1$--$3$ pixels in diameter), MF works well since small marine snow artifacts are similar to salt-and-pepper noise.
However, as shown in Fig. \ref{realMarineSnow}, marine snow artifacts often become large.
As we set a large filter size for MF to remove such artifacts, MF significantly blurs the entire image.
% , particularly when using a large filter kernel size.
% Specific to marine snow removal, a modified version of MF is proposed in \cite{banerjeeEliminationMarineSnow2014,farhadifardSingleImageMarine2017}.
% This method applies the MF selectively if the target pixel has a higher intensity than the surrounding pixels. However, it remains difficult to remove large marine snow artifacts.

% A few marine snow removal methods for video sequences have also been proposed.
% They utilize the fact that marine snow artifacts continuously move in consecutive video frames.
% In \cite{farhadifardMarineSnowDetection2017}, background modeling is used to remove marine snow artifacts from a static scene.
% Marine snow artifacts are tracked and a customized MF is applied to the detected artifacts in \cite{cyganekRealtimeMarineSnow2018}.
% However, they are not applicable to a single underwater image or moving backgrounds.

% Importantly, all the above methods use a model-based approach.
% This is mainly due to a lack of high-quality datasets for marine snow removal, as mentioned in Section \ref{sec:intro}.

So far, a few deep learning-based marine snow removal methods have been proposed \cite{9775132,MSRBDataset,WANG2021106182}.
A typical training dataset consists of underwater images and their corresponding version with synthesized marine snow artifacts.
% synthesized marine snow images and corresponding ground truth images, marine snow masks. This dataset also contains real underwater images with marine snow (without ground truth). 
% In \cite{9775132}, marine snow particles are manually made with Adobe Photoshop.
% This results in a lack of randomness and scalability.
% In \cite{MSRBDataset}, marine snow is created by looking at the pixel values of marine snow and adding them to the image. It is not based on a physical model.
% The seminal study of marine snow removal is described in \cite{MSRBDataset,WANG2021106182,9775132}.
% This method assumes the particles behave like white Lambertian scatters for simplicity.
In \cite{9775132}, marine snow particles are manually appended to underwater images with Adobe Photoshop.
This results in a lack of randomness and scalability.
In \cite{MSRBDataset}, shapes of synthesized marine snow artifacts in the dataset are modeled with ellipses.
% randomly rotated around the ellipse and the transparency is also randomly chosen. Based on plots of real marine snow, the marine snow is classified into two types.
% An marine snow artifact is modeled by a Gaussian function.
% This study was the first in this field and good results. 
% However, its pixel values are determined in an ad-hoc manner and is not based on a physical measurement model of a particle in underwater.
% As a result, there were many marine snows that could not be removed.\\
However, 
%the above-mentioned datasets for marine snow removal have the following issues: 1) shapes and pixel values are 
this marine snow synthesis is not based on a physical measurement model of a particle in underwater and the reproduction accuracy of marine snow is limited.
%; and 2) there are no ground truth atmospheric images, which limits applications.
% \begin{enumerate}
%     \item Shapes and pixel values are not based on a physical measurement model of a particle in underwater.
%     \item There are no ground truth atmospheric images, which limits applications.
% \end{enumerate}

% Note that both marine snow datasets have no ground truth atmospheric images as well as existing underwater image datasets.
\hfill

As a result, for underwater image enhancement, we need a dataset including the following images:
\begin{itemize}
    \item Ground-truth clean atmospheric images, i.e., target images.
    \item Underwater images corresponding to the atmospheric images degraded by absorption, scattering, and marine snow.
\end{itemize}

Creating such a dataset by taking many pictures in the real world in various water types is infeasible.
% We utilize a physics-based model in \cite{ueda2019underwater} for color degradation because of its generality but we also need to mathematically model marine snow effects for our PHISWID.
Hence, we address this problem by synthesizing images based on physics-guided models for color shift and marine snow.

\section{Underwater Image Synthesis}
\label{sec:marine_snow_models}
\begin{figure*}[t]
        \centering
        \includegraphics[width=.8\linewidth]{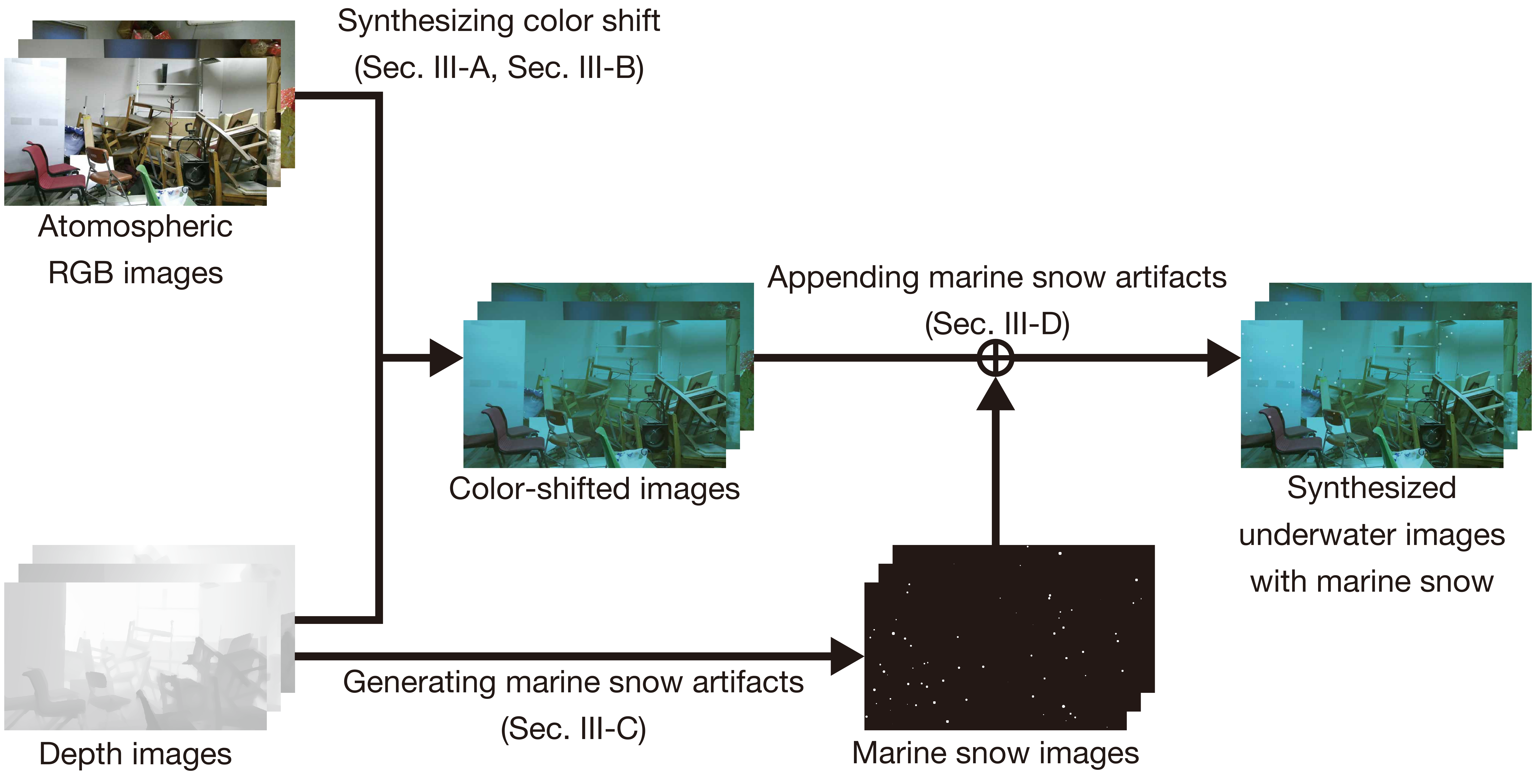}
        \caption{Flow of underwater image synthesis for PHISWID. First, RGB and depth images are used to synthesize underwater images based on \cite{ueda2019underwater}. Then, marine snow artifacts are added with the method introduced in Section \ref{sec:marine_snow_models}.
        % The upper left are the original images and the lower left are the depth images. The middle images are underwater images generated by underwater image degradation model, and the right images are images added marine snow artifacts.
        }
        \label{image_flow}
\end{figure*}
%In this section, we describe our method to synthesize underwater images with marine snow artifacts from clean RGB-D atmospheric images.
In this section, we describe our method for generating underwater images from clean RGB-D atmospheric images by synthesizing degradation caused by absorption, scattering, and marine snow.
 
The flow for image synthesis is illustrated in Fig. \ref{image_flow}. We first synthesize color-shifted images based on a 
%by using RGB-D atmospheric images.
% to generate synthesized underwater images. 
%Then, we add marine snow artifacts to the synthesized underwater images.
% to generate synthesized underwater images with marine snow.
%In the following, first, we introduce the 
physics-inspired color shift model proposed in Section \ref{sec:color} 
%Second, marine snow models are introduced with visual comparison between real and synthesized marine snow artifacts. 
%The process of synthesizing color shift is explained 
through the procedure described in Section \ref{sec:syn_color}. Then, we also synthesize marine snow artifacts based on a model proposed in Section \ref{sec:marine} %The marine snow artifact synthesis is explained
through the procedure described in Section \ref{syn_marine}.
%Finally, we compare synthesized marine snow artifacts to the real ones in Section \ref{sec:visual}.

%Since we utilize the physics-based model in \cite{ueda2019underwater} for the color shift, the main challenge of the synthesis is to model marine snow artifacts with a physics-inspired approach.
%There are various sources of marine snow \cite{trudnowska2021marine} and it is impractical to estimate the sources of all particles from a single underwater image.
% Therefore, marine snow modeling from the sources are impractical.
% Instead of the estimation of the sources, we model the pixel value distributions of marine snow artifacts from observations of underwater images.
%In this paper, we mathematically model the color shift and light scattering of a small particle via a physical measurement process as a marine snow artifact.
% In this paper, we mathematically model color shift and marine snow artifacts via a physical measurement process.

\subsection{Color Shift Model}
\label{sec:color}
Here, we first describe the models simulating color shift of underwater images.

\subsubsection{Algorithm Overview}
\begin{algorithm}[t]
 \caption{Color shift synthesis algorithm}
 \label{al:color}
 \begin{algorithmic}[1]
 \renewcommand{\algorithmicrequire}{\textbf{Input:}}
 \renewcommand{\algorithmicensure}{\textbf{Output:}}
 % \REQUIRE Graph: $G$, clusters in the $l$th scale: $\{S_j^{(l)}\}_{j=1,\ldots,N^{(l)}}$, the numbers of clusters at $l$ and $(l+1)$th scales: $N^{(l)}$ and $N^{(l+1)}$
 \REQUIRE
 Clean RGB-D images : $D_{\text{RGB-D}}$\\
 Attenuation coefficients for each water type $w$ : $D_{w,\beta(\lambda)}$\\
 Camera spectral response : $D_{\mathrm{SR}(\lambda),\text{camera}}$\\
 Illumination irradiance on the water surface : $\Theta_0(\lambda)$\\
 %\# underwater images to be synthesized : $N$\\
 $d_{\text{vert},\text{max}}$, $B_{c,\text{min}}$, $B_{c,\text{max}}$, $\rho(\lambda)=1$
 % \ENSURE  Clusters in the $(l+1)$th scale: $S_1^{(l+1)}, \ldots, S_{N^{(l+1)}}^{(l+1)}$
 \ENSURE
 Synthesized image set corresponding : $D_{\mathcal{W},\text{synthesized}}$
 \\
  \STATE $\mathcal{W}$ $\gets$ \{I, IA, IB, II, III, 1C, 3C\}%, 5C, 7C, 9C\}
  \FOR {$t = 1$ to $7$}
  \STATE $\beta(\lambda)$ $\gets$ $D_{\mathcal{W}_t,\beta(\lambda)}$
  \REPEAT
  \STATE $\Phi$ $\gets$ Clean image in $D_{\text{RGB-D}}$
  \STATE $z$ $\gets$ Depth map in $D_{\text{RGB-D}}$
  \STATE Adjust the dynamic range of $D_h$ into $d_{\text{horiz}} \in [0.5, 14]$
  % \FOR{$k=1$ to $10$}
  \STATE Set $d_{\text{vert}}$ within $0.5 \leq d_{\text{vert}} \leq d_{\text{vert},\text{max}}$ randomly\\
  \STATE Set $B_c$ within $B_{c,\text{min}} \leq B_c \leq B_{c,\text{max}}$ randomly\\
  \STATE Randomly select a camera $r$\\
  \STATE $\mathrm{SR}(\lambda)$ $\gets$ $D_{\mathrm{SR}(\lambda),r}$\\
  \STATE Calculate equation \eqref{degradationmodel1}--\eqref{degradationmodel3} for all pixels and get $U_c$\\
  \STATE Put $U_c$ into $D_{\mathcal{W}_t,\text{synthesized}}$
  % \ENDFOR
  \UNTIL Generating $U_c$ from all images in $D_{\text{RGB-D}}$
  \ENDFOR
 \end{algorithmic} 
\end{algorithm}
% \begin{table}[t]
% \centering
% \caption{Ten Water types by Jerlov \cite{jerlov}.}
% \label{tab:watertype}
% \begin{tabular}{c|c}
% \hline
% Ocean water type   & Coastal water type \\ \hline
% I, IA, IB, II, III & 1C, 3C, 5C, 7C, 9C \\ \hline
% \end{tabular}
% \end{table}
\begin{table}[t]
\centering
\caption{Ten Water types by Jerlov \cite{jerlov}.}
\label{tab:watertype}
\begin{tabular}{c|c}
\hline
Ocean water type   & Coastal water type \\ \hline
I, IA, IB, II, III & 1C, 3C, 5C, 7C, 9C \\ \hline
\end{tabular}\vspace{5pt}  % 表と矢印の間に少し余白を追加
\parbox{0.4\textwidth}{\centering \textbf{Low attenuation} \hfill \longlongleftrightarrow \hfill \textbf{High attenuation}}
\end{table}
% We synthesize color-shifted images from clean RGB-D images based on ten water types proposed by 
Our color shift model considers attenuation of light due to underwater condition and spectral difference of camera property. 
For modeling the attenuation, we use a model proposed by Jerlov \cite{jerlov} which presents attenuation of light at each wavelength (400 to 700 nm).
%with attenuation coefficients of the extent to which attenuation of light at each wavelength (400 to 700 nm). 
The attenuation coefficients are varied for water types, which Jerlov classified into ten as shown in Table \ref{tab:watertype}.
We omit three coastal water types, 5C, 7C, and 9C, from our dataset because objects are almost invisible for these types.
%according to the wavelength from 400 to 700 nm.
%
In addition to the attenuation property, the color also depends on the spectrum response which is an inherent property of a camera and lens.
%For acquiring an image by a digital camera, its spectral responses for red, green, and blue pixels are different according to the camera and lens used as well as the water type. 
Therefore, our color shift synthesis takes into account the attenuation and spectral response to simulate the color shift in underwater environment.
%image synthesis process includes the spectral responses of cameras to yield more accurate synthesized images. 

Algorithm \ref{al:color} describes the overview of our color shift synthesis method. The required inputs are listed as follows:
\begin{itemize}
    \item Clean RGB-D images: $D_\text{RGB-D}$.
    \item Set of attenuation coefficients $\beta(\lambda)$ for each water type $w$, in which $\lambda$ is the wavelength: $D_{w,\beta(\lambda)}$.
    \item Set of RGB spectral responses $\mathrm{SR}(\lambda)$ for several digital cameras: $D_{\mathrm{SR}(\lambda),\text{camera}}$.
    \item Illumination irradiance on the water surface: $\Theta_0(\lambda)$.
    %\item The number of underwater images to be synthesized from a clean RGB-D image for each water type: $N$.
    \item Maximum vertical distance from water surface\footnote{In this paper, we do not use ``depth'' for water depth not to confuse readers.}: $d_{\text{vert},\text{max}}$
    \item Maximum and minimum values of the background light: $B_{c, \text{max}}$ and $B_{c,\text{min}}$.
    % \item The camera coordinate: $(x,y)$.
\end{itemize}
Our algorithm creates seven sets of synthetic images, each corresponds to one water type.

\subsubsection{Underwater Image Degradation Model}
In our color shift model, we consider two main sources of the degradation process: absorption and scattering. The underwater color shift model is represented as follows: 
\begin{equation}
\label{degradationmodel1}
    U_c(x,y) = T_{c,d_{\text{vert}}}T_{c,d_{\text{horiz}}}\Phi(x,y) + B_c T_{c,d_{\text{vert}}}(1 - T_{c,d_{\text{horiz}}}),
\end{equation}
where $c \in \{\text{R}, \text{G}, \text{B}\}$, $U_c(x,y)$ is the pixel values at $(x,y)$ of an underwater image, $\Phi(x,y)$ is the pixel values of the corresponding clean RGB-D image, $T_{c,d_{\text{vert}}}$ is the transmission map of color at the vertical distance $d_{\text{vert}}$, and $T_{c,d_{\text{horiz}}}$ is that in the horizontal direction (i.e., the distance between the camera and scene).

The first term in \eqref{degradationmodel1} represents the process of color absorption both in the vertical and the horizontal directions. The second term corresponds to the background color at the vertical distance $d_{\text{vert}}$ and the horizontal distance $d_{\text{horiz}}$.
Further, $T_{c,d_{\text{vert}}}$ and $T_{c,d_{\text{horiz}}}$ is obtained by
\begin{align}
    T_{c,d_{\text{vert}}} &= \text{exp}(-\beta_{c,d_{\text{vert}}} \cdot d_{\text{vert}})\\
    T_{c,d_{\text{horiz}}} &=  \text{exp}(-\beta_{c,d_{\text{horiz}}}\cdot d_{\text{horiz}}),
\end{align}
where $\beta_{c,d_{\text{vert}}}$ and $\beta_{c,d_{\text{horiz}}}$ are the attenuation coefficients along the vertical and horizontal directions, respectively.
% Clearly, we need to predefine  $\beta_{c,d}$ and $\beta_{c,d_{horiz}}$ for the above equations.

Based on \cite{akkaynakwhatis2017}, $\beta_{c,d}$ and $\beta_{c,d_{\text{horiz}}}$ is given as
\begin{align}
\beta_{c, d_{\text{vert}}}&=d_{\text{vert}}^{-1} \ln \left[\frac{\int \mathrm{SR}(\lambda) \rho(\lambda) \Theta_0(\lambda) d \lambda}{\int \mathrm{SR}(\lambda) \rho(\lambda) \Theta_0(\lambda) \exp [-\beta(\lambda) d_{\text{vert}}] d \lambda}\right]\label{degradationmodel2}\\
\beta_{c,d_{\text{horiz}}}&=\frac{\ln \left[\frac{\int \mathrm{SR}(\lambda) \rho(\lambda) \Theta_0(\lambda) \exp [-\beta(\lambda) d_{\text{vert}}] d \lambda}{\int \mathrm{SR}(\lambda) \rho(\lambda) \Theta_0(\lambda) \exp \left[-\beta(\lambda)\left(d_{\text{vert}}+d_{\text{horiz}}\right)\right] d \lambda}\right]}{d_{\text{horiz}}},\label{degradationmodel3}
\end{align}
where $\rho(\lambda)$ is the reflectance spectrum of the object surface. By using \eqref{degradationmodel2} and \eqref{degradationmodel3}, we obtain attenuation coefficients to calculate $T_{c,d_{\text{vert}}}$ and $T_{c,d_{\text{horiz}}}$. 
% Furthermore, we assume that there are two types of Marin snow from the plot of real Marin snow. 
% Actual underwater images with marine snow are carefully examined to specify three typical mathematical models of marine snow.
% \begin{figure*}[t]
%         \centering
%         \includegraphics[width=.8\linewidth]{image/physwid_flow.pdf}
%         \caption{Flow of underwater image synthesis for PHISWID. First, RGB and depth images are used to synthesize underwater images based on \cite{ueda2019underwater}. Then, marine snow artifacts are added with the method introduced in Section \ref{sec:marine_snow_models}.
%         % The upper left are the original images and the lower left are the depth images. The middle images are underwater images generated by underwater image degradation model, and the right images are images added marine snow artifacts.
%         }
%         \label{image_flow}
% \end{figure*}
\subsection{Synthesizing Color Shift}
\label{sec:syn_color}
By using the models introduced in Section \ref{sec:color}, we synthesize the color shift. We use the attenuation coefficients in all frequency ranges in contrast to the existing work \cite{anwarDeepUnderwaterImage2018} using only three coefficients.

The attenuation coefficients for seven water types $D_{w,\beta(\lambda)}$ (see Algorithm \ref{al:color}), are given by
%that have been obtained from the plot in 
\cite{moserSpectralTransmissionLight1992}. We also use the data of solar irradiance on the water surface \cite{ReferenceAirMass} as $\Theta_0(\lambda)$. The camera spectrum responses are also given from the dataset of 28 digital cameras \cite{jiangWhatSpaceSpectral2013}. The other parameters are set to $d_{\text{vert},\text{max}} = 1$, $\{B_{R,\text{min}}, B_{G,\text{min}}, B_{B,\text{min}}\} = \{0.4, 0.7, 0.7\}$, and $\{B_{R, \text{max}}, B_{G, \text{max}}, B_{B, \text{max}}\} = \{0.5, 0.8, 0.8\}$ \cite{ueda2019underwater}.
We randomly determine one of $B_c$ and $d_{\text{vert}}$ from the uniform distribution in the range of $B_{c, \text{min}} \leq B_c \leq B_{c, \text{max}}$ and $0.5 \leq d_{\text{vert}} \leq d_{\text{vert},\text{max}}$, respectively.
As a result, 7 underwater images are generated from a single atmospheric image with the seven water types and the one set of $B_c$ and $d_{\text{vert}}$.
%In PHISWID, we use NYD-RGB dataset \cite{roomdataset} and an outdoor image dataset \cite{tartanair2020iros} as clean RGB-D images. 
\subsection{Marine Snow Model}
\label{sec:marine}
\subsubsection{Light Scattering of Marine Snow}
\label{light scattering}
%There are various sources of marine snow \cite{trudnowska2021marine} and it 
Although a precise model of marine snow artifacts requires to identify particles, the identification is impractical since any particle drifting in underwater could be the sources.
%of all particles from a single underwater image. 
Therefore, we consider light scattering in water for small particles based on the Jaffe-MacGlamery model \cite{mcglamery,Fahimeh}, which is common for underwater imagery \cite{liWaterGANUnsupervisedGenerative2018,1707999}.

\begin{figure}[t]
 \centering
 \subfloat[Original Model]{\includegraphics[width = .35 \linewidth]{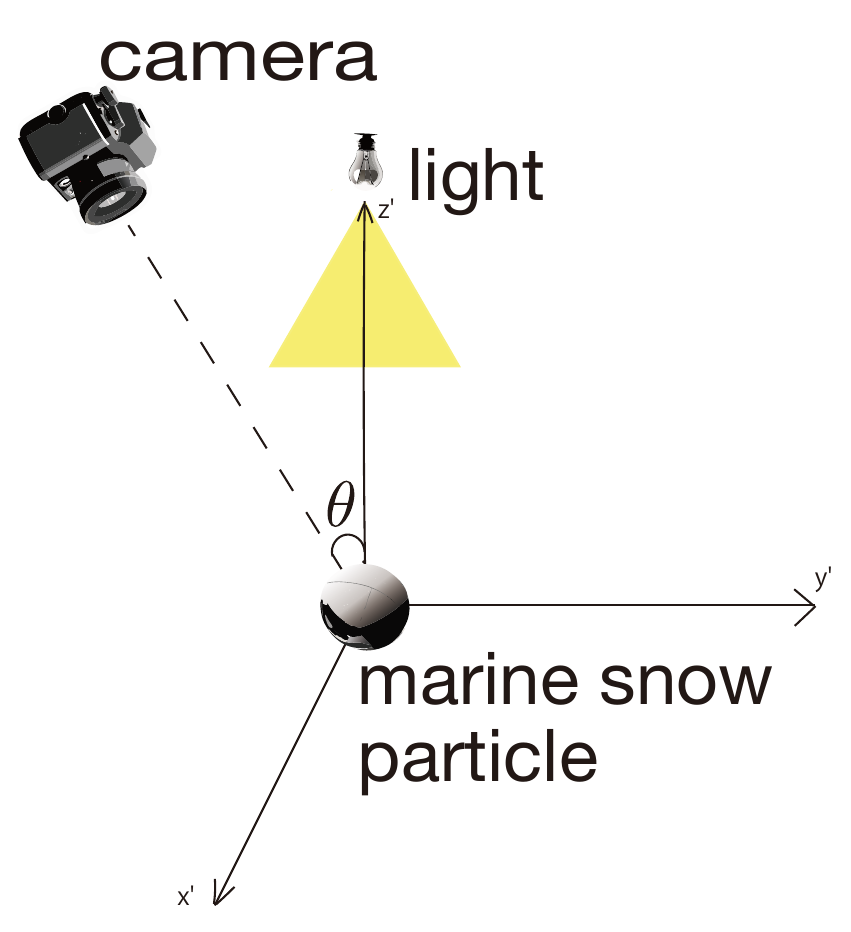}}
 \subfloat[Simplified Model]{\includegraphics[width = .35\linewidth]{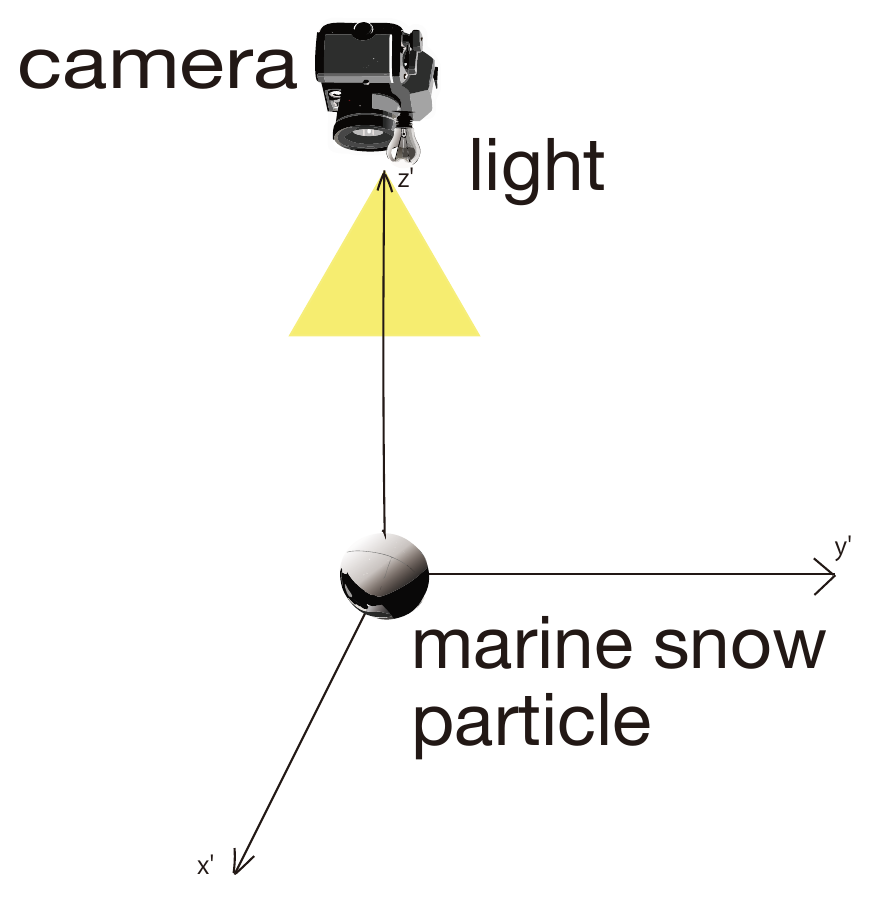}}\\
 \caption{Jaffe-MacGlamery model.}
 \label{sec3:model}
\end{figure}

The Jaffe-MacGlamery model of a small object is illustrated in Fig. \ref{sec3:model}(a).
%First of all, we assume the direct-scattered components only contribute to marine snow artifacts in underwater images.
%In the original physical model 
The model considers direct-scattered and back-scattering components \cite{mcglamery,Fahimeh}.
%direct-scattering and back-scattering are both considered because the target object is relatively larger than surrounding objects.
% was considered from an object and back-scattered was considered from an marine snow. However, this time, 
%However, here, we focus on a marine snow particle in this paper: 
% This can be a small single object and it is enough to only consider the direct-scattering effect to simplify the model.
Although the back-scattering components should be considered if a object is relatively large, we can ignore these components because objects caused marine snow should be small.
Therefore, we simplify the model only with the direct-scattered components.
Another simplification is that there is a single light source in a scene.
We assume that a camera's flash is the single source.
The model with these two simplifications is illustrated in Fig. \ref{sec3:model}(b).
% we think marine snow as an object. Therefore, we focus only on direct-scattered. 
% We also assume that the light source and the camera are in the same three-dimensional position because marine snow is thought to be largely due to the influence of the camera's flash.
% Tha camera flash assumed to have a significant effect on marine snow. Therefore, in the present study, the light source and the camera are assumed to be at the same 3D position.
% The observation model is illustrated in Fig. \ref{model}.
% , the following equation holds in Figs.\ref{model}.

% \begin{equation}
%     \theta = \gamma = 0
% \end{equation}

% The Jaffe-MacGlamery model of a small object is illustrated in Fig. \ref{model}(a).
%We assume the camera flash is the main light source for a marine snow particle for simplicity.
%In the following, we assume that xy transforms to x'y'.
% We also assume the light source and the camera are assumed to be in the same 3D position.
% This simplifies the observation model as in Fig. \ref{model}(b).
\begin{table}[t]
     \caption{Parameter settings for light scattering in \eqref{model_direct1} and \eqref{model_direct2}.}%where $\mathcal{U}(v_{\min}, v_{\max})$ is a continuous uniform distribution between $v_{\min}$ and $v_{\max}$.
    \label{light}
    \centering
    \begin{tabular}{c|c}
    \hline
    Parameter & Setting\\\hline
    $\theta$   &  0 \\
    $E_{\mathrm{d}}(x, y)$ & Direct-scattered component\\
    %$R$  & Distance from the source\\
    $F_l$  & Focal length of camera\\
    $T_l$  & Transmattance of the lens\\
    $R$  & Distance from camera to marine snow\\
    $\beta_c$  & Attenuation along the vertical and horizontal directions\\
    $p(\cdot)$  & Point spread function\\\hline
    \end{tabular}
\end{table}
The direct-scattered component $E_{\mathrm{d}}(x, y)$ at the coordinate $(x,y)$ in the observed image is thus represented as%follows:
\begin{equation}
\label{model_direct1}
E_{\mathrm{d}}(x, y)=E_{\mathrm{I}}\left(x, y, 0\right)e^{-\beta_c R} M\left(x, y\right) \frac{\cos ^4 \theta T_l}{4 f_n}\left[\frac{R-F_l}{R}\right]^2,
\end{equation}
where $E_{\mathrm{I}}\left(x, y, 0\right)$ is defined as
\begin{equation}
    E_{\mathrm{I}}(x,y,0) = E^{\prime}_{\mathrm{I}}(x,y,0)*p(x,y|R,G,\beta_c,B) + E^{\prime}_{\mathrm{I}}(x,y,0),
\end{equation}
and $*$ is convolution, $M(x,y)$ is the reflectance values of the reflectance map for oceanographic objects.
The other symbols are defined in Table \ref{light}.
% $\theta$ is zero is that the camera flash is the cause of the marine snow and we can assume that the light source is in the same position as the camera.
% $F_l$ is the camera of focal length. $T_l$ is the transmattance of the lens. $c$ is the attenuation. $R$ is the distance from camera to marine snow. $p$ is the point spread function. 

Here, we further assume the light source and the camera are located in the same 3D position, which results in $\theta = 0$.
This simplifies the observation model as in Fig. \ref{sec3:model}(b).
By the above assumption, \eqref{model_direct1} can be rewritten as
\begin{equation}
\label{model_direct2}
E_{\mathrm{d}}(x, y) = \left(A\frac{e^{-2\beta_cR}}{R^2}*p + A\frac{e^{-2\beta_cR}}{R^2}\right)\left[\frac{R - F_l}{R}\right]^2,
\end{equation}
where $A$ is a constant coming from those in \eqref{model_direct1}.
The simplified model in \eqref{model_direct2} indicates that the light scattering of a marine snow particle only depends on $R$, i.e., the distance between the camera and the particle. The derivation of the equation is shown in Appendix \ref{app:model}.
% Since the point spread function $p(R)$ also depends on the distance,
% we thought it would be a good idea to change the variance of gaussian filtering depending on the distance from the camera.

% \subsection{Real Marine Snow Examples}
% First, we start by showing marine snow examples in real underwater images.
% Figs. \ref{crater_real_img}(a) and (c) show enlarged portions of representative marine snow artifacts cropped from real underwater images.
% % Clearly, they do not have a bell shape.
% Although they look similar, their pixel value distributions are slightly different.

% Taking a closer look, the 3D plots of Figs. \ref{crater_real_img}(a) and (c) are shown in Figs. \ref{crater_real_img}(b) and (d), respectively.
% As clearly observed, they do \textit{not} have a shape like a Gaussian function in contrast to the conventional assumption in \cite{boffety2012color,boffety2012phenomenological}.
% Rather than a Gaussian function, these 3D plots are similar to \textit{elliptic conical frusta}, i.e., sliced elliptic cones.
% Furthermore, the to surfaces of the frusta have different characteristics between Figs. \ref{crater_real_img}(b) and (d).
% In our preliminary observation, most marine snow artifacts can be classified into these two representative shapes.
% In the following, we present the physical models of marine snow artifacts that reflect the above-mentioned observations.
\subsubsection{Modeling Marine Snow Artifacts}
\label{Modelingmarinesnowart}
Based on the observation model shown in Section \ref{light scattering}, we model the type H and V\footnote{``H'' and ``V'' refer to ``highland'' and ``volcanic crater'', respectively, as in \cite{MSRBDataset}.} marine snow artifacts.
The type H artifacts correspond to generic marine snow particles. The type V artifacts are often observed as an edge-enhanced version of the type H caused by automatic image processing in cameras (like those in smartphones). An example of type H and V marine snow is shown in Fig. \ref{crater_real_img}.
% The type H marine snow corresponds to Fig. \ref{crater_real_img}(a) .
%We think type H is only a gaussian filter version. 

\begin{figure}[t]
    \centering
    \subfloat[Enlarged portion]{\includegraphics[width = .20 \linewidth]{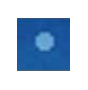}}
    \subfloat[3D plot]{\includegraphics[width = .20 \linewidth]{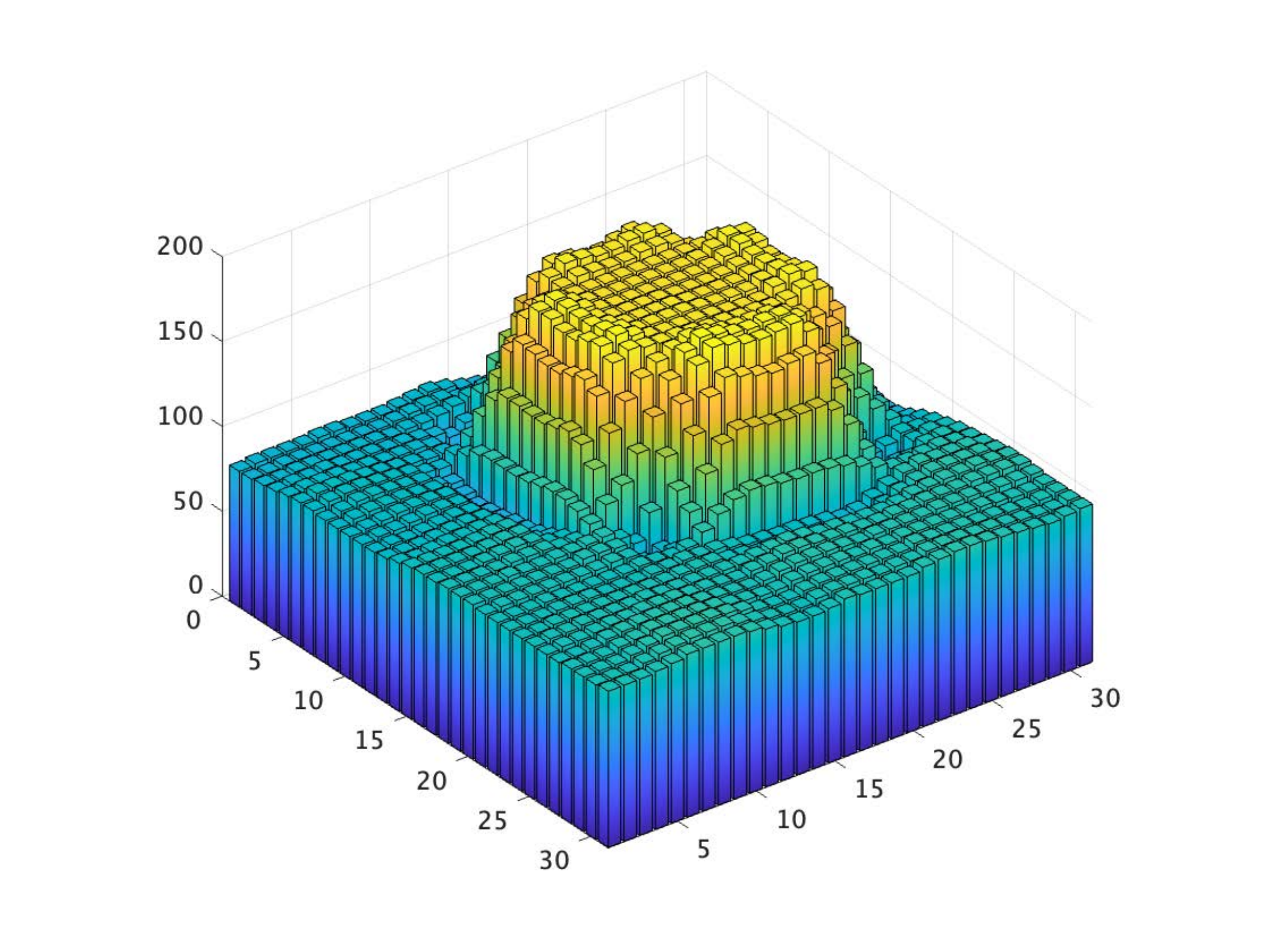}}
%     \caption{Highland type marine snow. a : real image. b : 3D plot.}
%     \label{highland_real_img}
% \end{figure}
% \begin{figure}[t]
%     \centering
    \subfloat[Enlarged portion]{\includegraphics[width = .20\linewidth]{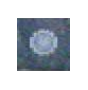}}
    \subfloat[3D plot]{\includegraphics[width = .20 \linewidth]{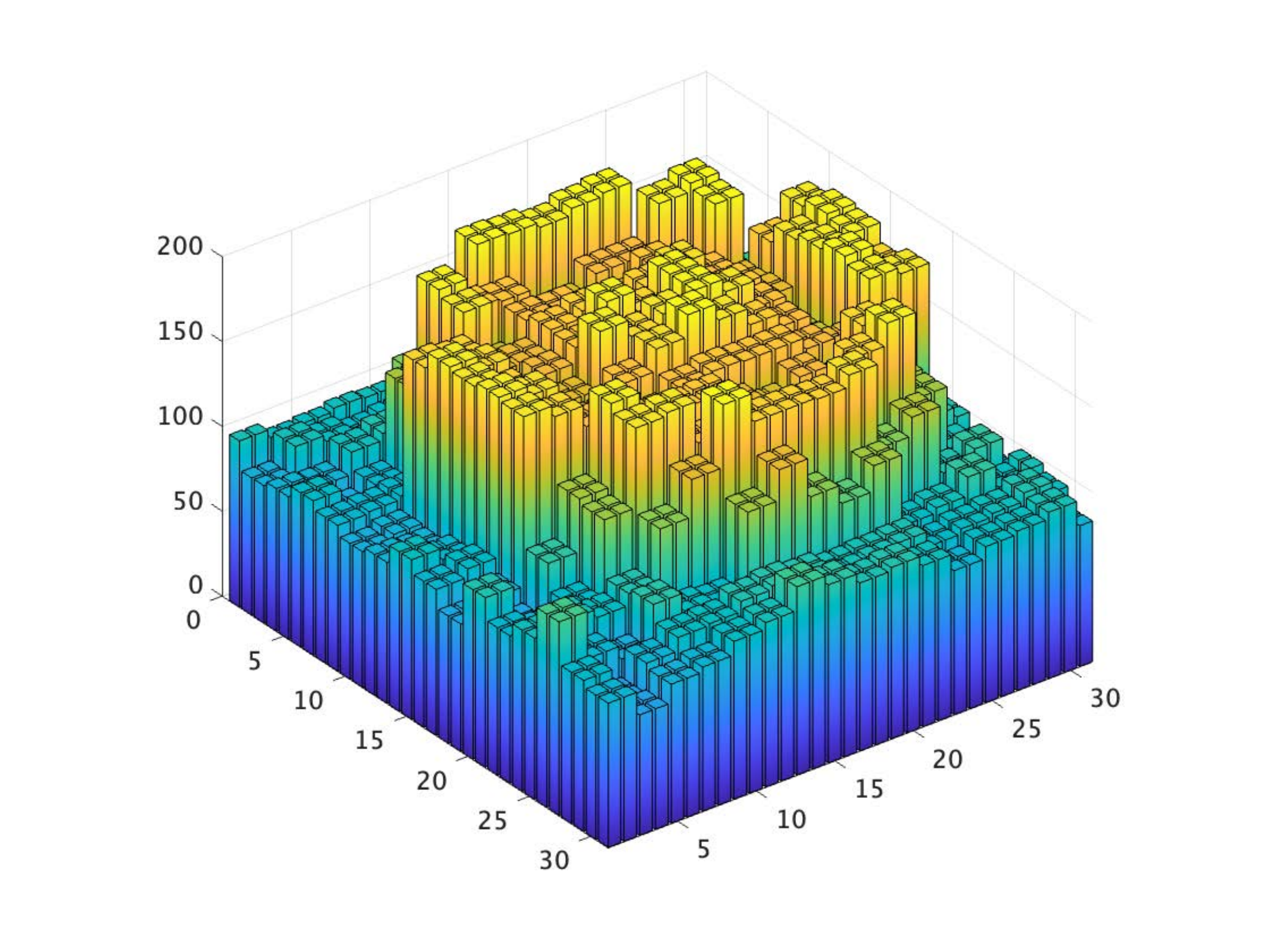}}\\
    \caption{Synthesized marine snow artifact examples. Left: Type H. Right: Type V.}
    \label{crater_synthe_img}
\end{figure}

We model a type H marine snow artifact as a Gaussian filtered version of a point source according to \eqref{model_direct2}. The type V is the slightly modified version of the type H.
% We found the Type H marine snow was a point source with a gaussian filter that changed its variance with distance.
Since the luminance of a particle depends on $R$, i.e., the distance between the camera and the particle, we model marine snow particles with the following steps:
\begin{enumerate}
    \item \textbf{Particle Setting:} As shown in Fig. \ref{model_scatter}, suppose that the camera is positioned parallel to the $(x,y)$-plane and at a sufficiently large distance from the scene located in the three-dimensional Euclidean space.
    The three-dimensional coordinates of the particles are uniform-randomly set.
    The number of marine snow particles is given for each image. 
    
    A sparse matrix $\tilde{S} \in \mathbb{R}^{H \times W}$ for the particle coordinates, where $H$ and $W$ are the height and width of the target image, is created as
    \begin{equation}
        \tilde{S}(x,y) = \begin{cases}
        R & \text{if a particle is located at } (x,y)\\
        0 & \text{otherwise,}
        \end{cases}
    \end{equation}
    where $R$ is randomly set.
    
    We then make a set of matrices $\{E_n\}_n$ such that $\tilde{S} = \sum_{n=0}^{N-1} E_n$ based on $\tilde{S}$ as follows:
    \begin{equation}
        E_n(x,y) = \begin{cases}
        D & \text{if } d_{\text{horiz},n-1} < \tilde{S}(x,y) \le d_{\text{horiz},n}\\
        0 & \text{otherwise,}
        \end{cases}
    \label{5}
    \end{equation}
    where $D$ is a constant corresponding to the brightness of a particle and $d_{\text{horiz},n}$ is the pre-determined horizontal distance from the camera where $d_{\text{horiz},-1} = 0$.
    This helps to reduce the computational cost for the following Gaussian filtering.
    \begin{figure}[t]
        \centering
        \subfloat[Enlarged potion]{\includegraphics[width = .20 \linewidth]{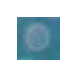}}
        %\hspace{0.1in}
        \subfloat[3D plot]{\includegraphics[width = .23 \linewidth]{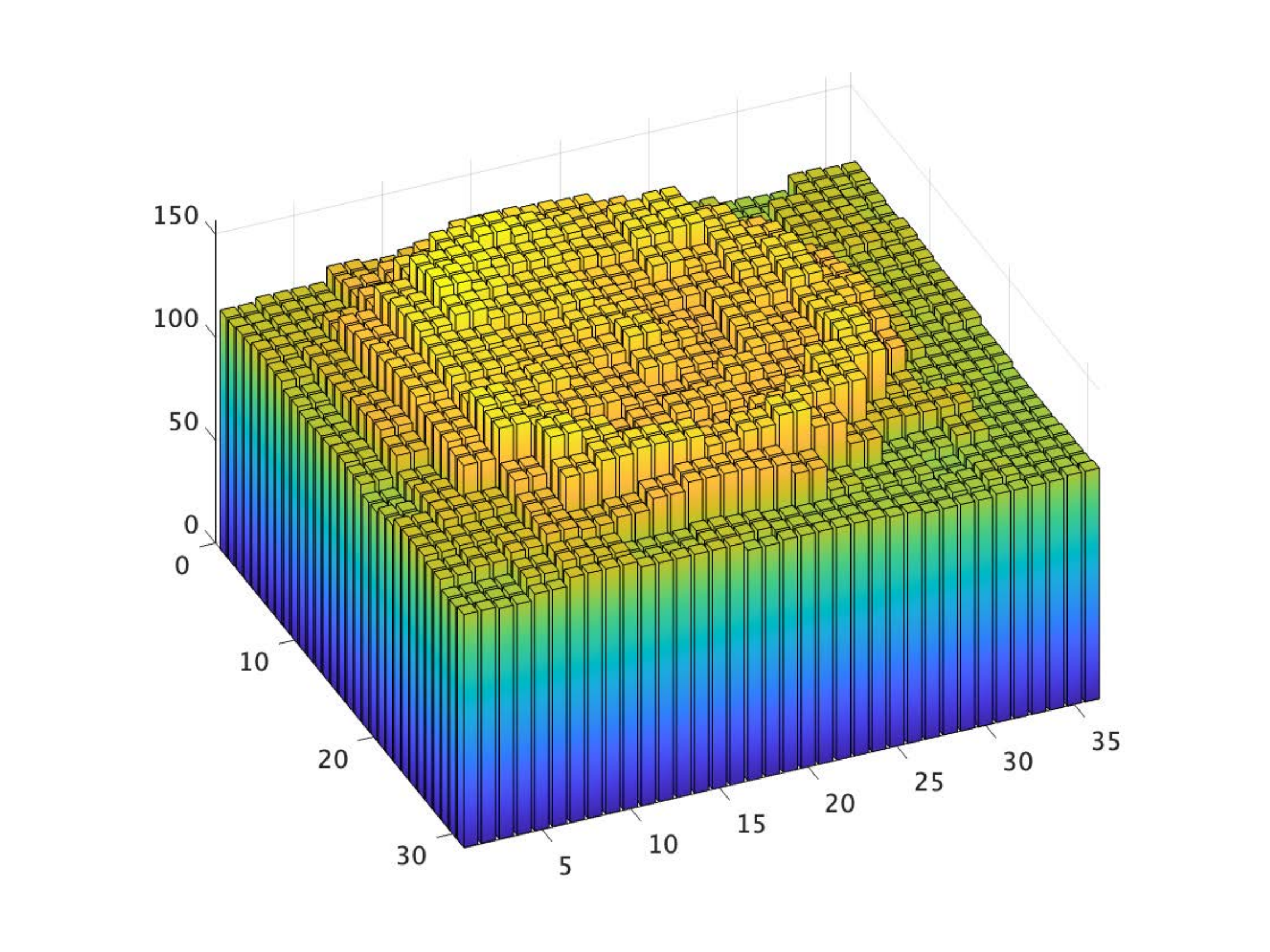}}
        %\hspace{0.1in}
        \subfloat[Enlarged potion]{\includegraphics[width = .20 \linewidth]{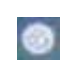}}
        %\hspace{0.1in}
    %     \caption{Highland type marine snow. a : real image. b : 3D plot.}
    %     \label{highland_real_img}
    % \end{figure}
    % \begin{figure}[t]
    %     \centering
        % \subfloat[3D plot]{\includegraphics[width = .16 \linewidth]{image/real_marine_snow/realH.pdf}}
        \subfloat[3D plot]{\includegraphics[width = .25 \linewidth]{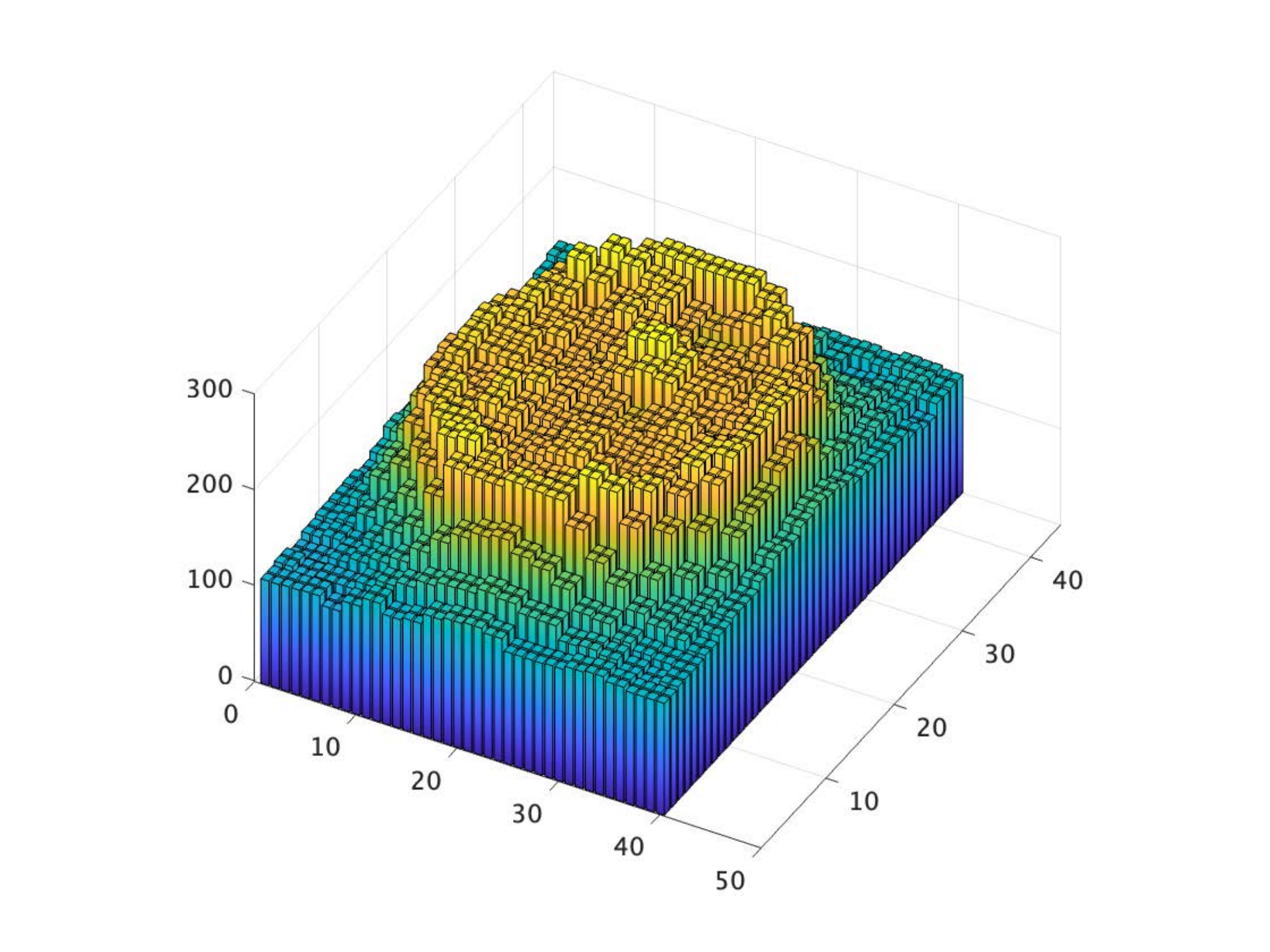}}\\
        \caption{Marine snow examples in real underwater images. Left: Type H. Right: Type V.}
        %Left to right: Highland type (type H). Bottom: Volcanic Crater type (type V). The 3D plots correspond to the bar images.}
        \label{crater_real_img}
    \end{figure}
    \begin{figure}[t]
        \centering
        \includegraphics[width=1.0\linewidth]{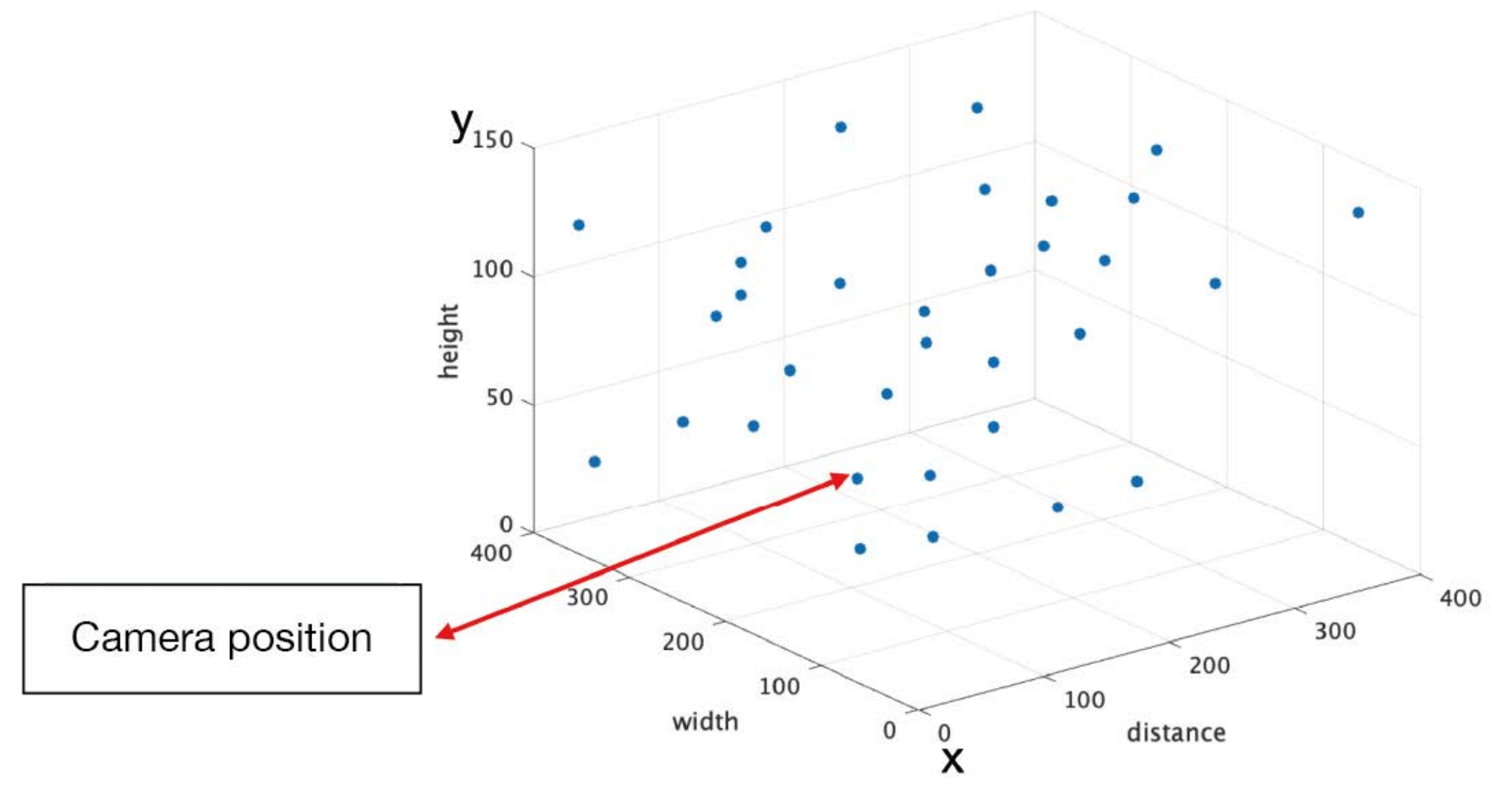}
        \caption{Illustrative example of marine snow particles. Particles are synthesized at the blue dot locations.}
        \label{model_scatter}
    \end{figure}
    % $S_1$ is divided into $k$ parts according to the distance of the camera.
    % \begin{equation}
    %     S_1 = E_n \quad n=1 ,\ldots, 4 \label{1}
    % \end{equation}
    % $S_1$ is a sparse matrix. $E_n$ is the matrix after the partition. 
    \item \textbf{Gaussian filtering to particles:} Let $g(x,y;\sigma_n) \label{gauspara_h}$ be a Gaussian filter centered at $(x,y)$ with the parameter $\sigma_n$. We apply $g(x,y;\sigma_n)$ to $E_n$ for yielding $S_{n}$ for the prototype of the marine snow artifacts.
    \begin{equation}
        S_n(x,y) = g(x,y;\sigma_n) * E_n(x,y).
    \end{equation}
    %Figs. \ref{crater_synthe_img}(a) and (b) are examples of synthesized type H marine snow artifacts.
    % \begin{equation}
    %     S_{2,n}(k,l) = g(x,y;\sigma_n)*E_n(k,l). \label{2}
    % \end{equation}
    % \begin{figure}[h]
    % \begin{center}
    %    \subfloat{\includegraphics[width =0.18 \linewidth]{image/gaussianfilter1.pdf}}
    %    \hspace{0.04\columnwidth}
    %    \subfloat{\includegraphics[width =0.18 \linewidth]{image/gaussianfilter2.pdf}}
    % \end{center}
    %     \vspace{-0.15in}
    %    \caption{Examples of synthesized marine snow artifacts.}
    % \label{gaussian_filtering_img1}
    % \end{figure}\\
    \item \textbf{Thresholding:} To avoid overshooting of the luminance of marine snow artifacts, thresholding is performed as
        % To change the luminance according to the distance from the camera, multiply each by a constant.
    \begin{equation}
        H_n(x,y) = \begin{cases}
    a_n S_{n}(x,y) & S_{n}(x,y) < \text{threshold} \\
    \text{constant} & S_{n}(x,y) \geq \text{threshold},
    \end{cases}
    \label{7}
    \end{equation}
    where $a_n$ is a pre-determined constant to strengthen or weaken the luminance of the artifacts.
    
    Finally, an image only containing marine snow artifacts is obtained by $H = \sum_n H_n$. 
    %Figs. \ref{crater_synthe_img}(a) and (b) are examples of synthesized type H marine snow artifacts.
    The synthesized type H marine snow artifact is shown in Figs. \ref{crater_synthe_img}(a) and (b), which are similar to the real version in Figs. \ref{crater_real_img}(a) and (b).
    % \begin{figure}[h]
    %     \centering
    %     \subfloat[]{\includegraphics[width = .2 \linewidth]{image/gaussianfilter1.pdf}}
    %     \hspace{0.04\columnwidth}
    %     \subfloat[]{\includegraphics[width = .2 \linewidth]{image/gaussianfilter2.pdf}}\\
    % %     \caption{Highland type marine snow. a : real image. b : 3D plot.}
    % %     \label{highland_real_img}
    % % \end{figure}
    % % \begin{figure}[t]
    % %     \centering
    %     \subfloat[]{\includegraphics[width = .2 \linewidth]{image/threshold1.pdf}}
    %     \hspace{0.04\columnwidth}
    %     \subfloat[]{\includegraphics[width = .2 \linewidth]{image/threshold2.pdf}}\\
    %     \caption{Examples of synthesized marine snow artifacts.((a),(b)),Images only containing marine snow artifacts((c),(d))}
    %     \label{scatter_img}
    % \end{figure}
    % \begin{figure}[h]
    % \begin{center}
    %    \subfloat{\includegraphics[width =0.18 \linewidth]{image/threshold1.pdf}}
    %    \hspace{0.04\columnwidth}
    %    \subfloat{\includegraphics[width =0.18 \linewidth]{image/threshold2.pdf}}
    % \end{center}
    %     \vspace{-0.15in}
    %    \caption{Images only containing marine snow artifacts}
    % \label{gaussian_filtering_img}
    % \end{figure}\\
\end{enumerate}
We model the type V artifacts by applying the Laplacian filter to the type H artifacts.
For type V, steps i) and ii) are the same as type H.
Step iii) is different as:
% Type V is modeled in the same way as Type H with an additional step as follows:
% \begin{equation}
%     S' = E'_i\quad i=1,\ldots,4 \label{4}
% \end{equation}
% S' is a sparse matrix.E' is the matrix after the partition. Let $g(x,y;\sigma)$ denote the Gaussian filter.Apply a Gaussian filter with different sigma to each of E'.
% \begin{equation}
%     GS' = g(x,y;\sigma_i)*E'_i \quad i=1,\ldots,4 \label{5}
% \end{equation}
% Third, to change the luminance according to the distance from the camera, multiply each by a constant.
% \begin{equation}
%     V = c_i(g(x,y;\sigma_i)*E'_i) \quad i=1,\ldots,4 \label{6}
% \end{equation}
% $c_i$ is constant. Furthermore, apply a laplasian filter.
\begin{itemize}
    \item[iii')] \textbf{Laplacian filtering and thresholding:}
\begin{equation}
    V_n (x,y) = 
\begin{cases}
%L(\sigma_n)*{a'_nS'_2 (k,l)} \quad n=1,\ldots,m & V < \text{threshold} \\
L(\sigma_n)*{a'_nS'_n (x,y)} & V < \text{threshold} \\
\text{constant} & V \geq \text{threshold},
\end{cases}
\label{8}
\end{equation}
where $L(\sigma_n)$ is a Laplacian filter for edge enhancement with the parameter $\sigma_n$ controlling sharpness and $a'_n$ is similar to $a_n$ in \eqref{7}.

Finally, an image only containing marine snow artifacts is obtained by $V = \sum_n V_n$. The synthesized type V marine snow artifact is shown in Figs. \ref{crater_synthe_img}(c) and (d).
Like the type H, it is also similar to the artifacts in the real-world image in Figs. \ref{crater_real_img}(c) and (d).
% \begin{figure}[h]
% \begin{center}
%    \subfloat{\includegraphics[width =0.18 \linewidth]{image/lap1.pdf}}
%    \hspace{0.04\columnwidth}
%    \subfloat{\includegraphics[width =0.18 \linewidth]{image/lap2.pdf}}
% \end{center}
%     \vspace{-0.15in}
%    \caption{Marine snow with threshold processing inside the camera. The volcanic crater type is generated by the effect of edge enhancement in the camera.}
% \label{realMarineSnow}
% \end{figure}\\
% $S'_2$ and threshold of type V are slightly different from $S_2$.
\end{itemize}

\subsection{Synthesizing Marine Snow Artifacts}
\label{syn_marine}
By using the models introduced in Section \ref{sec:marine}, we synthesize the marine snow artifacts with underwater images or synthesized underwater images.

Let $U_c \in \mathbb{R}^{H\times W \times 3}$ be an underwater image obtained by the model introduced in Section \ref{sec:color}.
To synthesize marine snow artifacts, we simply overlay $H(x, y)$ and $V(x, y)$ on $U_c(x, y)$ as:
\begin{equation}
    I_c(x,y) = U_c(x,y) + H(x,y) + V(x,y),
\end{equation}
where $H(x, y)$ and $V(x, y)$ are images with the type H and V marine snow artifacts shown in \eqref{7} and \eqref{8}, respectively.

%This visually validates the appropriateness of our marine snow observation model.
\section{PHISWID Specifications}
\label{sec:dataset}
% In this section, we present the specifications of PHISWID.
% We also propose two benchmarking tasks using the dataset.
% \begin{description}
%     %\item[PHISMID\footnotemark:] Designed for marine snow removal.
%     \item[PHISWID:] Designed for underwater image enhancement/restoration as well as marine snow removal.
% \end{description}
%\footnotetext{PHysics-Inspired Synthesized Marine snow Image Dataset}
%Clearly, image enhancement/restoration with PHISWID is more difficult than PHISMID, but PHISMID itself is also beneficial for the overlooked marine snow removal task.
%First, setups shared in both tasks are presented.
% Second, we introduce specifications for PHISWID.

\begin{figure}[t]
\centering
    \subfloat{\includegraphics[width = 0.2\linewidth]{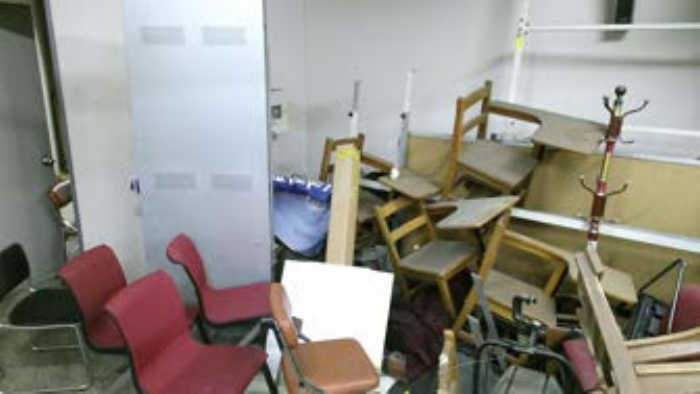}} 
    \subfloat{\includegraphics[width = 0.2\linewidth]{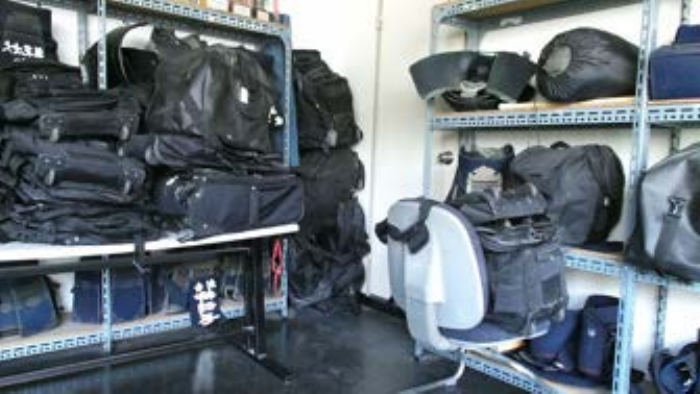}} 
    \subfloat{\includegraphics[width = 0.2\linewidth]{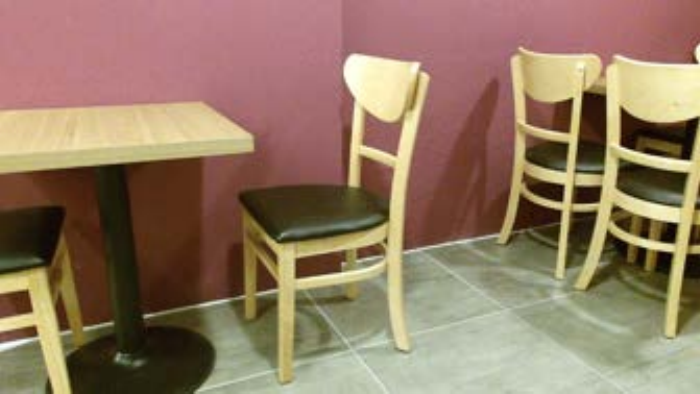}}
    \subfloat{\includegraphics[width = 0.2\linewidth]{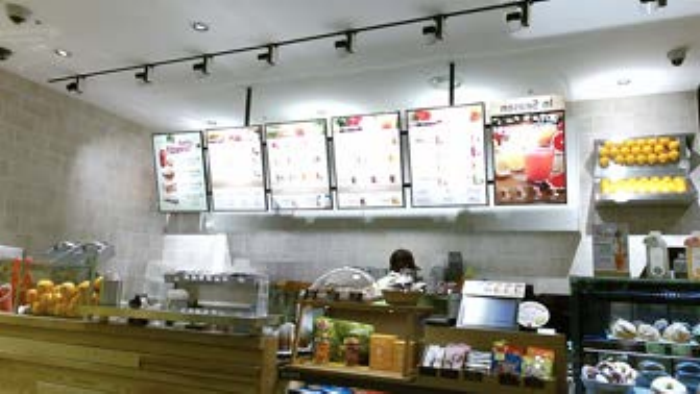}} 
    \subfloat{\includegraphics[width = 0.2\linewidth]{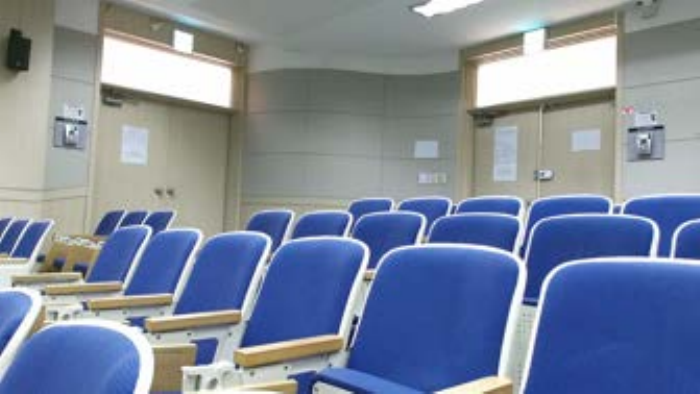}}\\\vspace{-0.15in}
    \subfloat{\includegraphics[width = 0.2\linewidth]{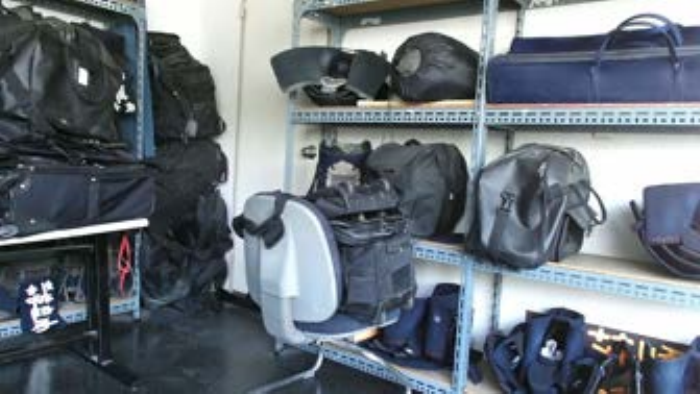}} 
    \subfloat{\includegraphics[width = 0.2\linewidth]{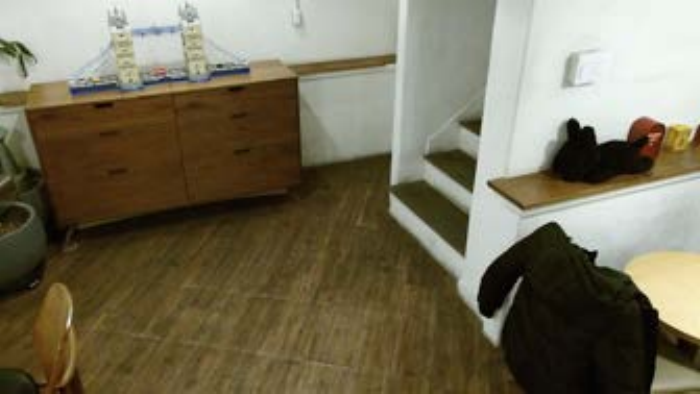}} 
    \subfloat{\includegraphics[width = 0.2\linewidth]{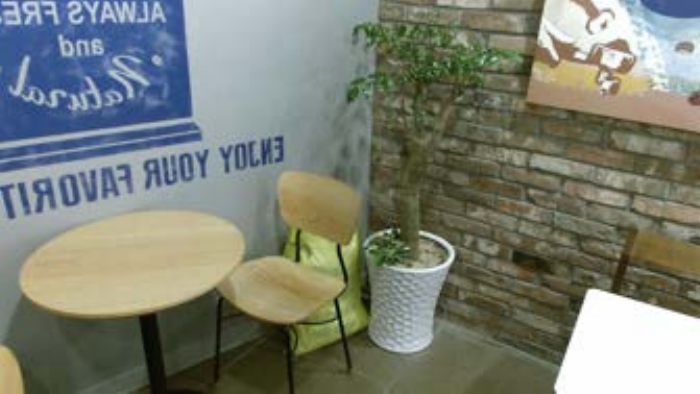}}
    \subfloat{\includegraphics[width = 0.2\linewidth]{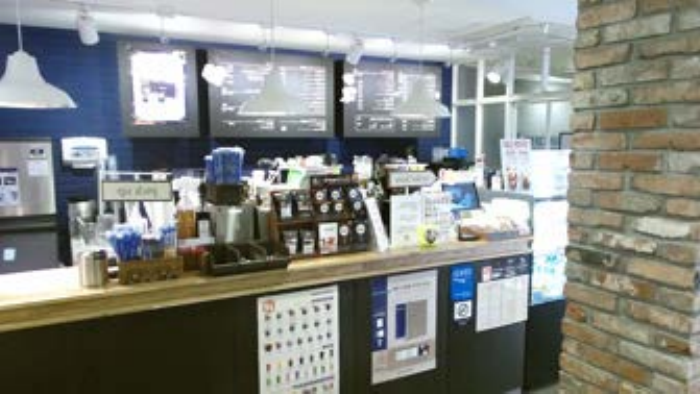}} 
    \subfloat{\includegraphics[width = 0.2\linewidth]{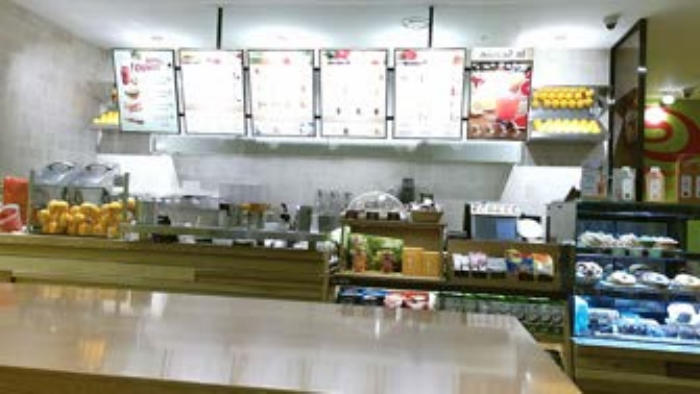}}\\\vspace{-0.15in}
    \subfloat{\includegraphics[width = 0.2\linewidth]{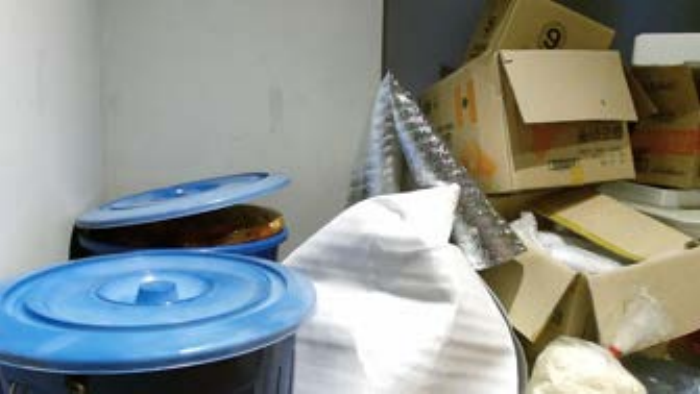}} 
    \subfloat{\includegraphics[width = 0.2\linewidth]{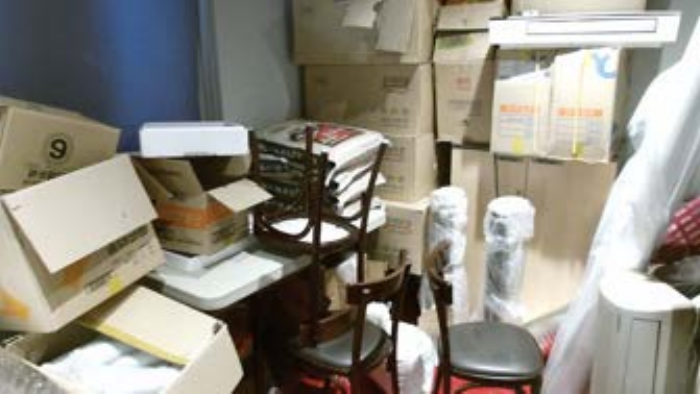}} 
    \subfloat{\includegraphics[width = 0.2\linewidth]{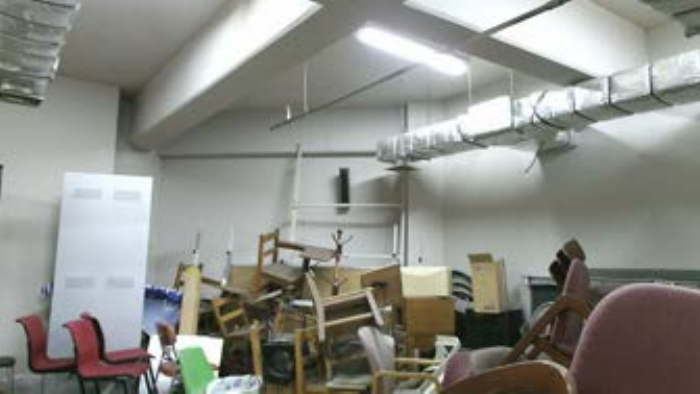}}
    \subfloat{\includegraphics[width = 0.2\linewidth]{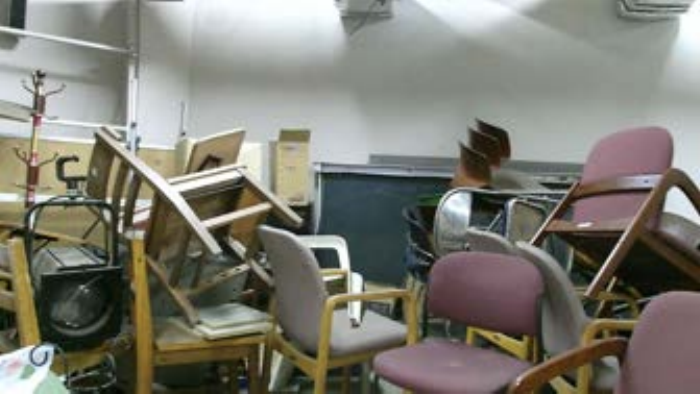}} 
    \subfloat{\includegraphics[width = 0.2\linewidth]{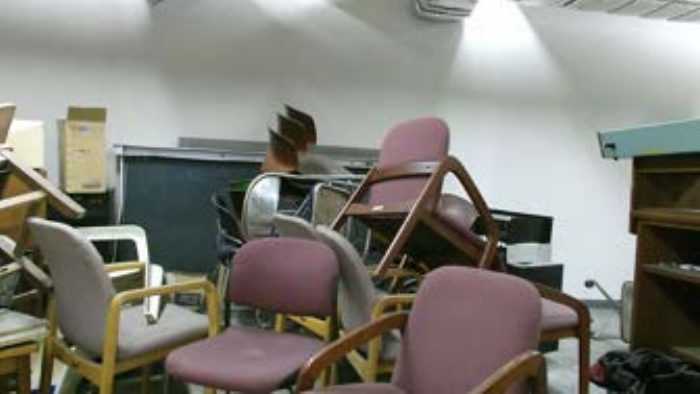}}\\\vspace{-0.15in}
    \subfloat{\includegraphics[width = 0.2\linewidth]{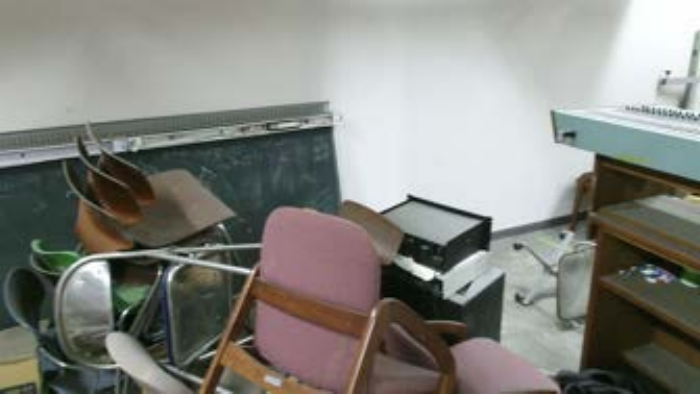}} 
    \subfloat{\includegraphics[width = 0.2\linewidth]{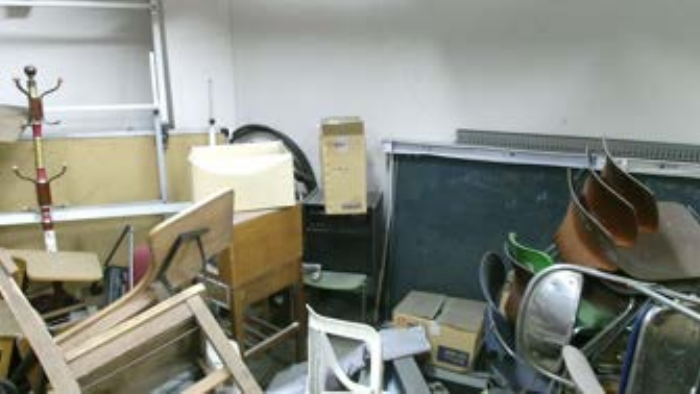}} 
    \subfloat{\includegraphics[width = 0.2\linewidth]{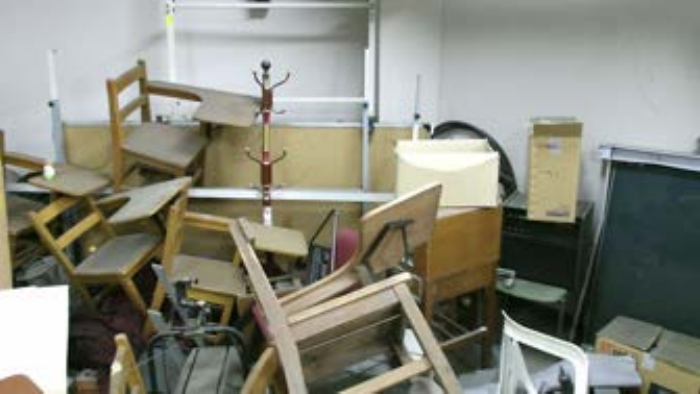}}
    \subfloat{\includegraphics[width = 0.2\linewidth]{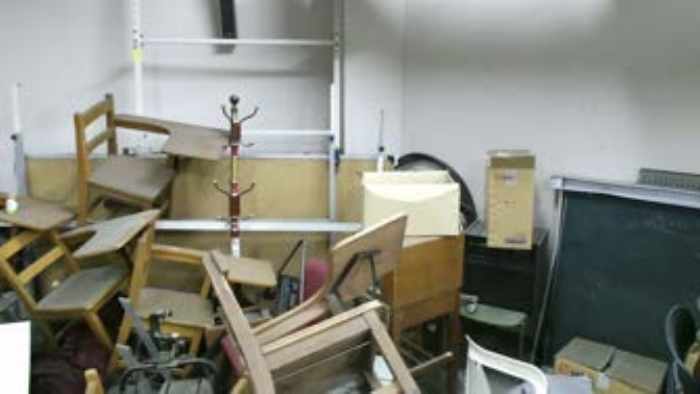}} 
    \subfloat{\includegraphics[width = 0.2\linewidth]{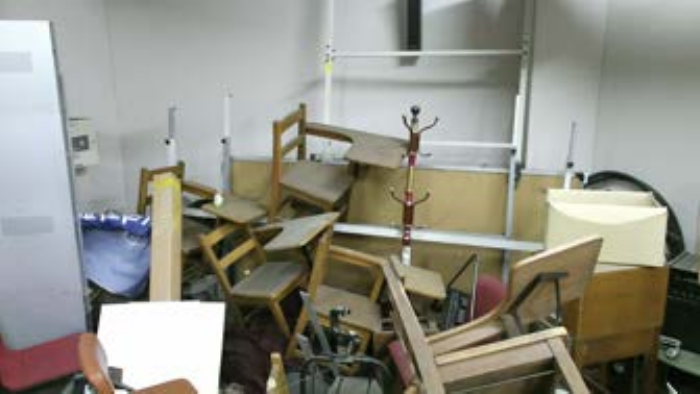}}
    %\vspace{-0.15in}
    \caption*{Original images.}
    \subfloat{\includegraphics[width = 0.2\linewidth]{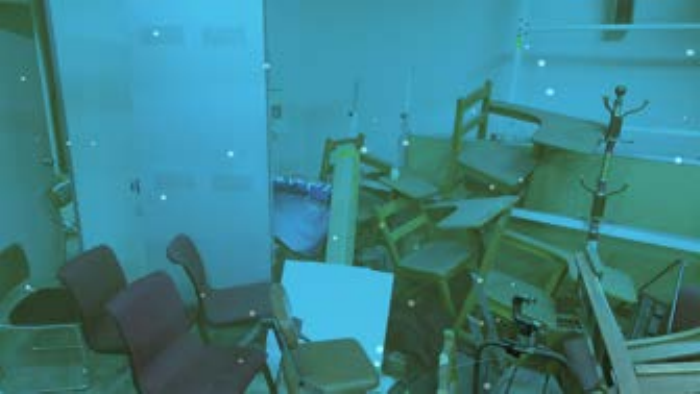}} 
    \subfloat{\includegraphics[width = 0.2\linewidth]{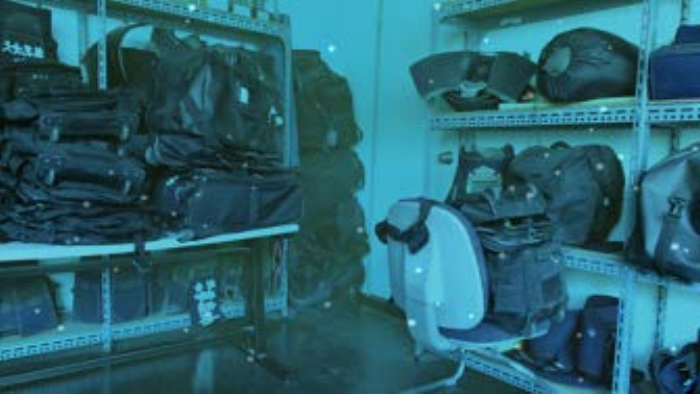}} 
    \subfloat{\includegraphics[width = 0.2\linewidth]{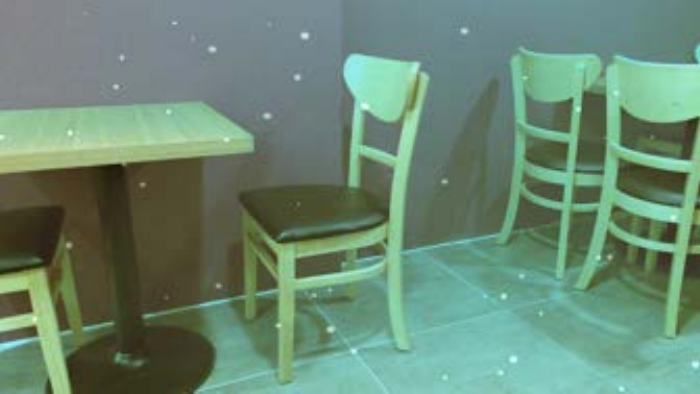}}
    \subfloat{\includegraphics[width = 0.2\linewidth]{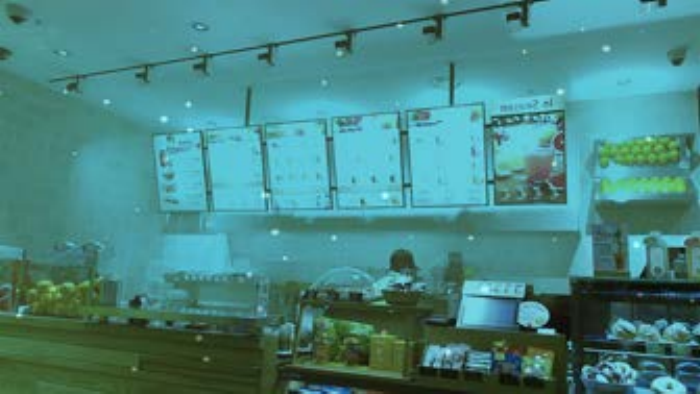}} 
    \subfloat{\includegraphics[width = 0.2\linewidth]{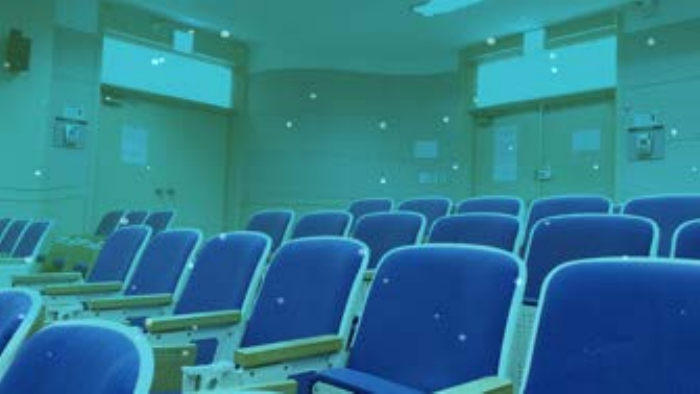}} \\\vspace{-0.15in}
    \subfloat{\includegraphics[width = 0.2\linewidth]{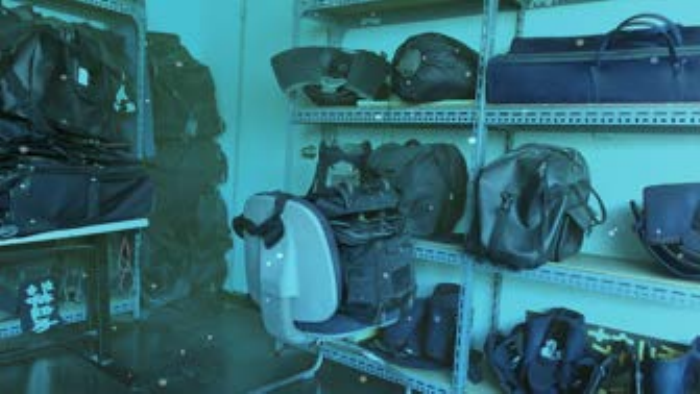}} 
    \subfloat{\includegraphics[width = 0.2\linewidth]{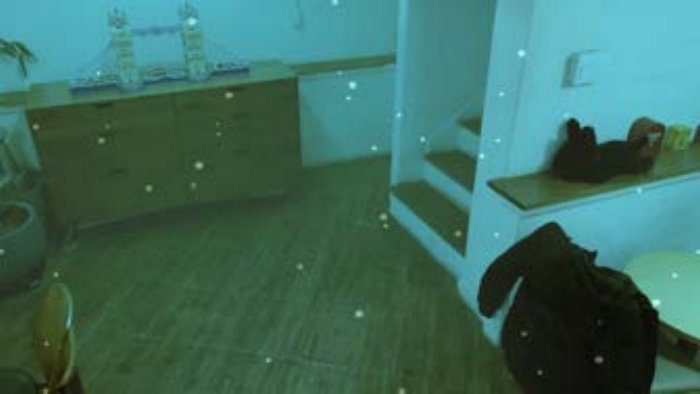}} 
    \subfloat{\includegraphics[width = 0.2\linewidth]{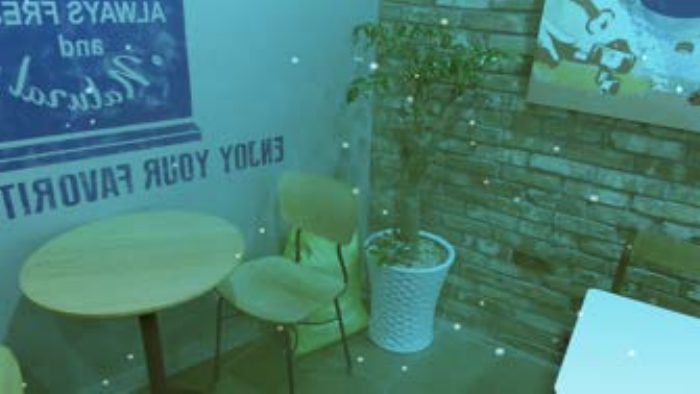}}
    \subfloat{\includegraphics[width = 0.2\linewidth]{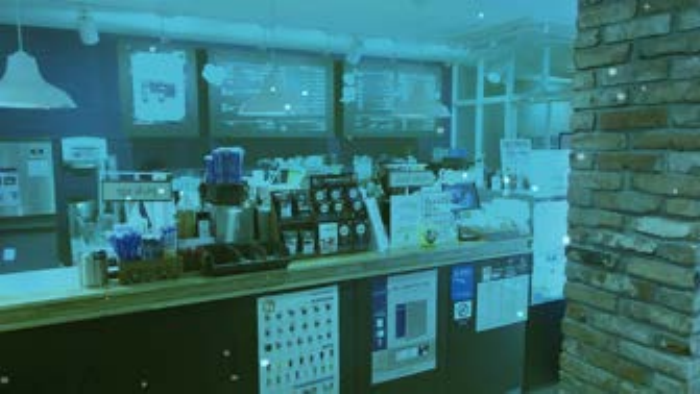}} 
    \subfloat{\includegraphics[width = 0.2\linewidth]{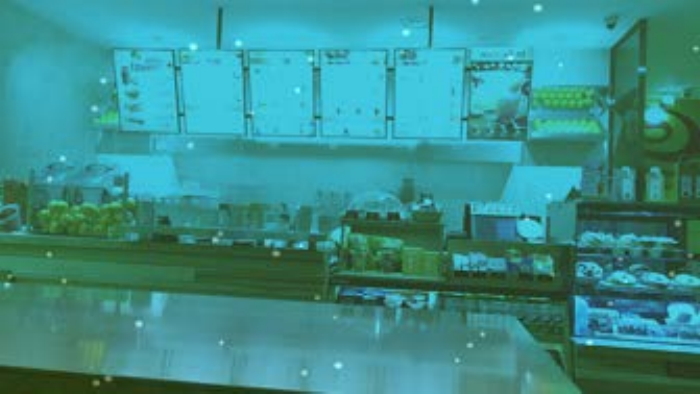}}\\\vspace{-0.15in}
    \subfloat{\includegraphics[width = 0.2\linewidth]{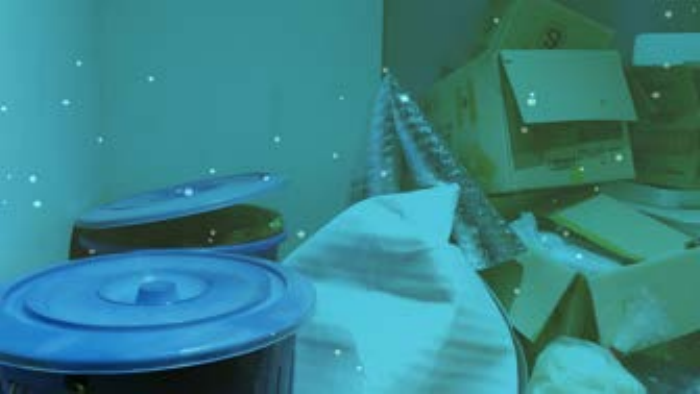}}
    \subfloat{\includegraphics[width = 0.2\linewidth]{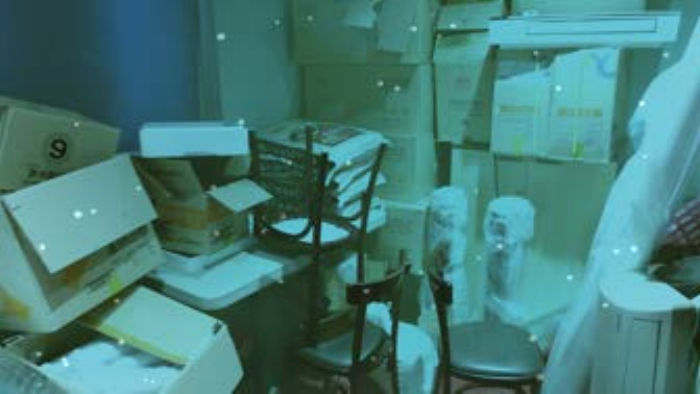}}
    \subfloat{\includegraphics[width = 0.2\linewidth]{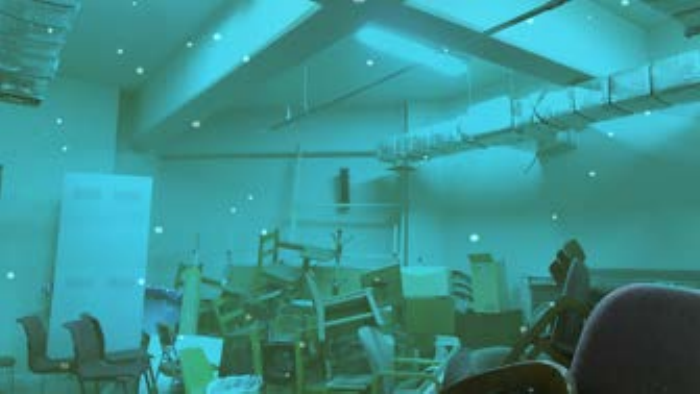}}
    \subfloat{\includegraphics[width = 0.2\linewidth]{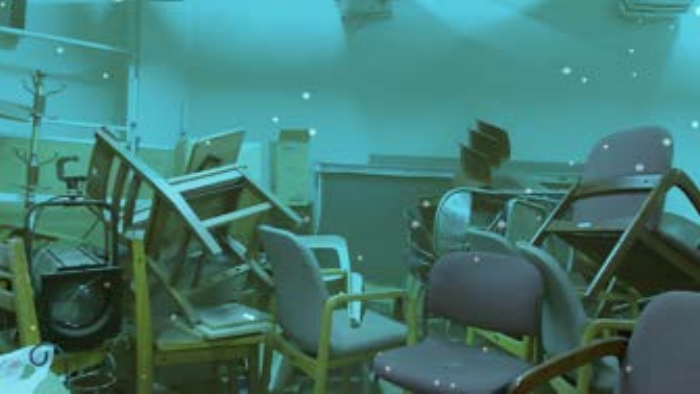}}
    \subfloat{\includegraphics[width = 0.2\linewidth]{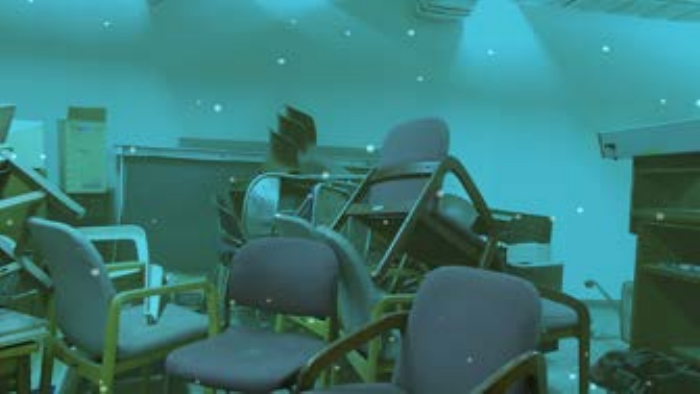}}\\\vspace{-0.15in}
    %\vspace{-0.15in}
    \subfloat{\includegraphics[width = 0.2\linewidth]{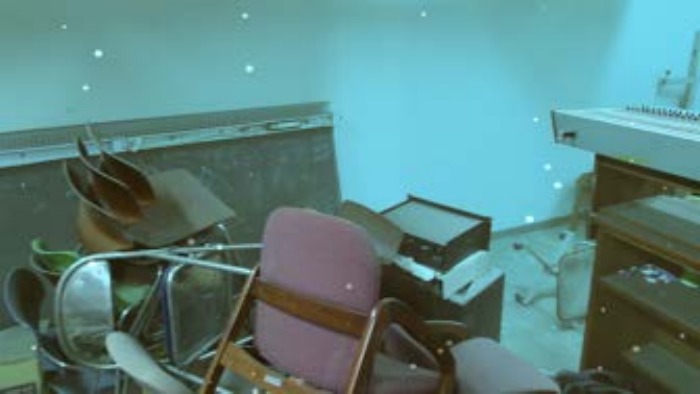}}
    \subfloat{\includegraphics[width = 0.2\linewidth]{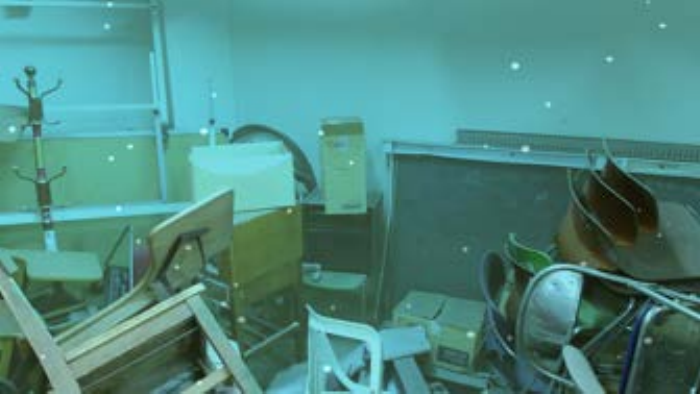}}
    \subfloat{\includegraphics[width = 0.2\linewidth]{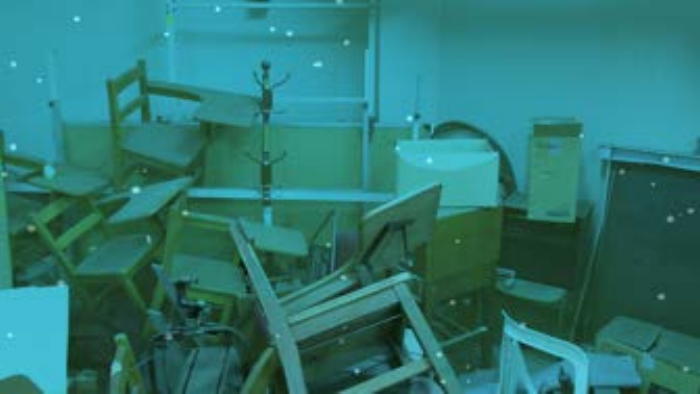}}
    \subfloat{\includegraphics[width = 0.2\linewidth]{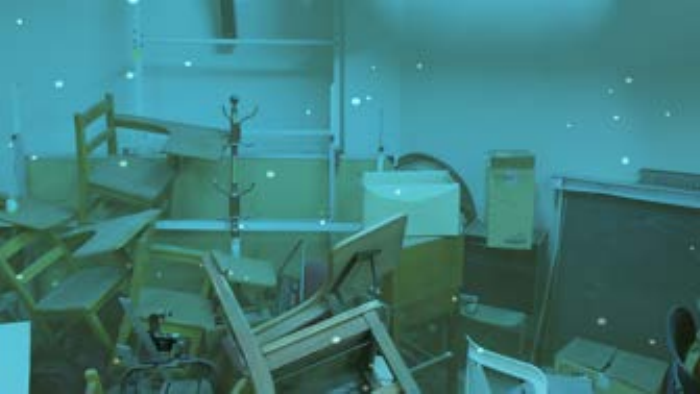}}
    \subfloat{\includegraphics[width = 0.2\linewidth]{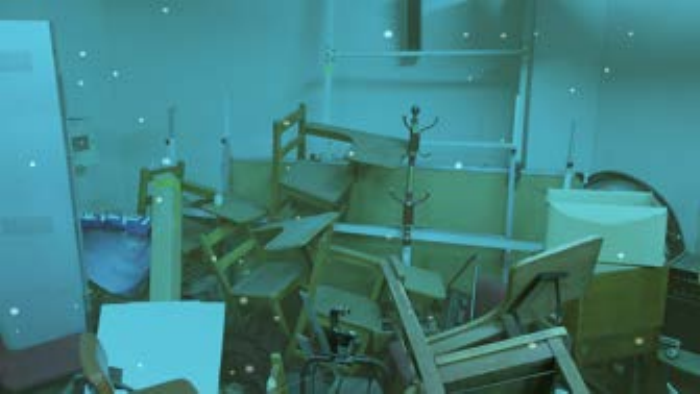}}
    %\vspace{-0.15in}
    \caption*{Synthesized images.}
  \caption{Example images in PHISWID.}
  \label{MSRCCdataset_img}
\end{figure}

%\subsection{PHISWID Specifications}
%\label{PHISWID_spe}
Our developed dataset, PHISWID, contains $4195$ image pairs, all having a pixel resolution of $1344 \times 756$. All original atmospheric RGB-D images used are collected from the large-scale RGB-D database \cite{kimDeepMonocularDepth2018,kimStructureSelectiveDepth2016,choDIMLCVLRGBD2021,choDeepMonocularDepth2021,kimDeepStereoConfidence2017}.
An image pair contains one original atmospheric image and one synthesized underwater image degraded by color shift and marine snow artifacts. We randomly set water type for each image.
Image examples in PHISWID are shown in Fig. \ref{MSRCCdataset_img}.

Each synthesized image contains marine snow artifacts. The number of the type H marine snow particles is randomly generated with a Gaussian distribution of mean 40 and variance 5, and that of the type V marine snow particles conforms to a Gaussian distribution of mean 30 and variance 5.

Most parameters are determined according to our preliminary observations of real underwater images: They are summarized in Appendix \ref{app:param}. 
\begin{figure*}[t]
%\vspace{-0.2in}
\centering
    \subfloat{\includegraphics[width = 0.09\linewidth]{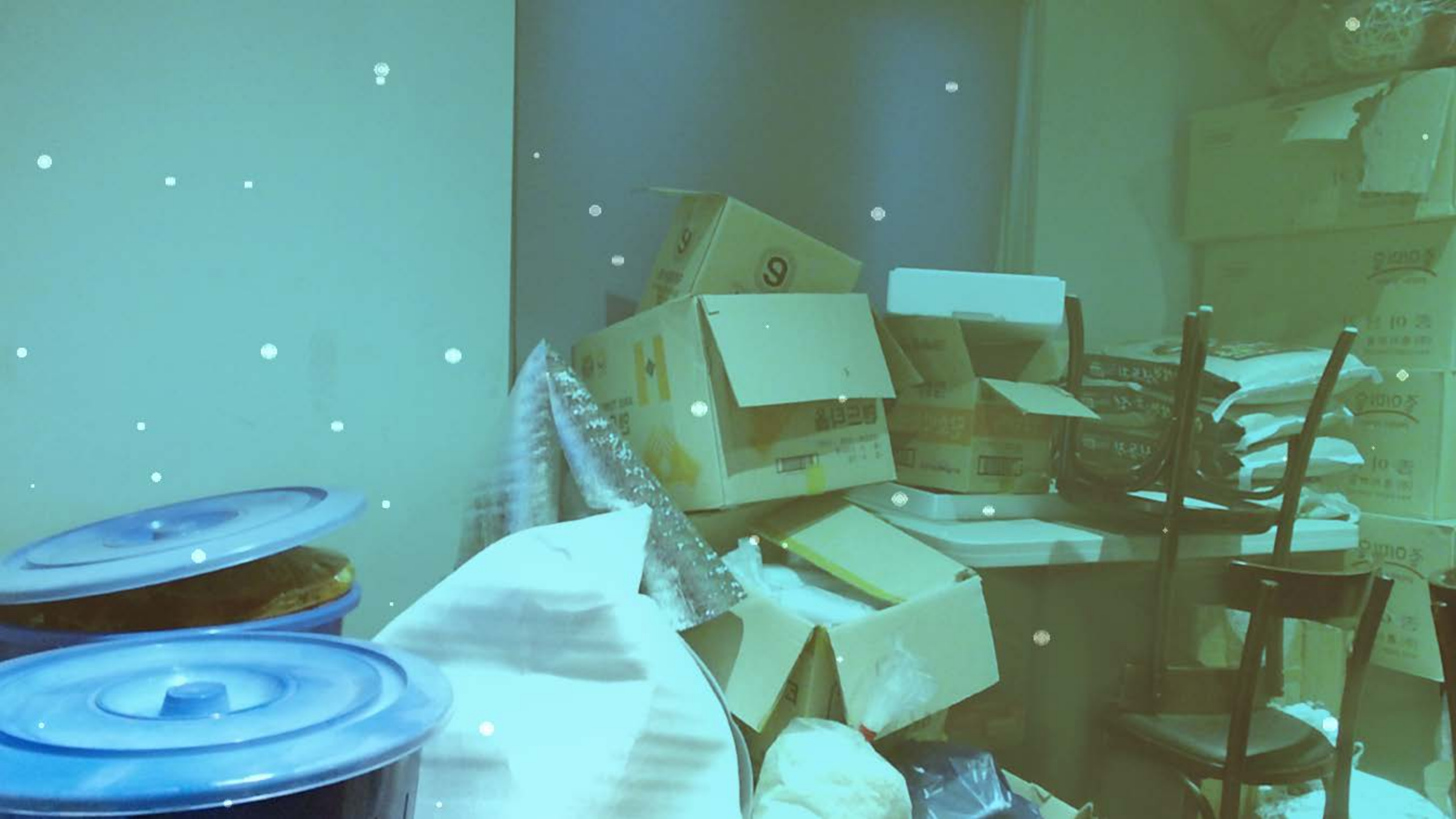}} 
    \subfloat{\includegraphics[width = 0.09\linewidth]{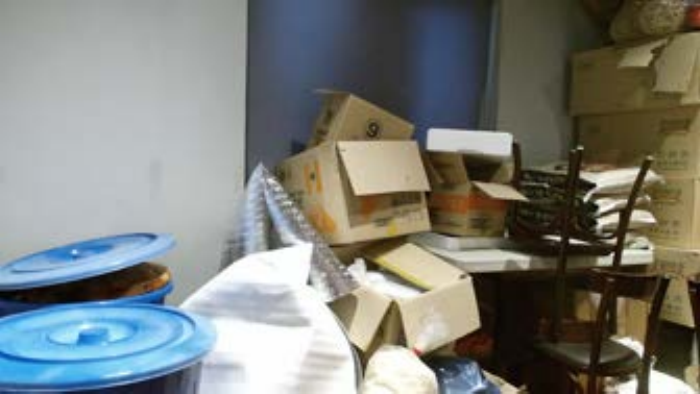}}
    \subfloat{\includegraphics[width = 0.09\linewidth]{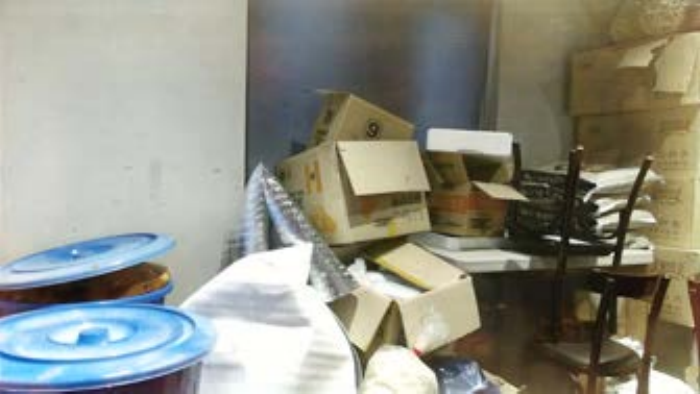}}
    \subfloat{\includegraphics[width = 0.09\linewidth]{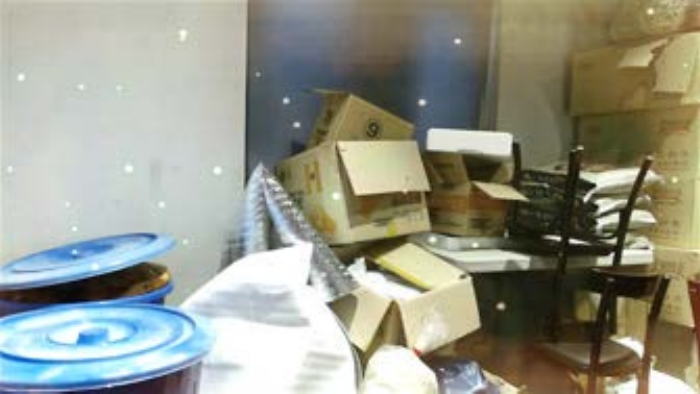}}
    \subfloat{\includegraphics[width = 0.09\linewidth]{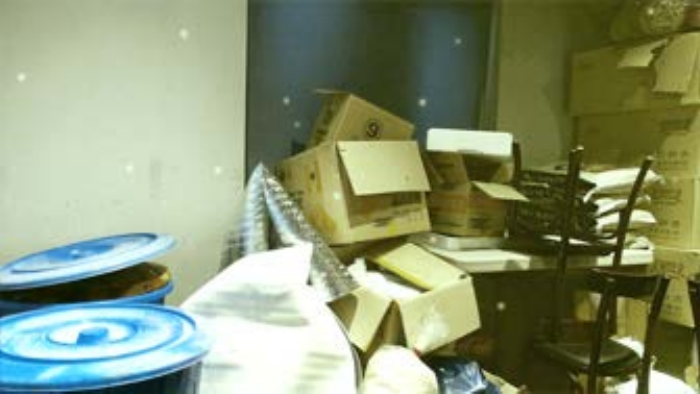}} 
    \subfloat{\includegraphics[width = 0.09\linewidth]{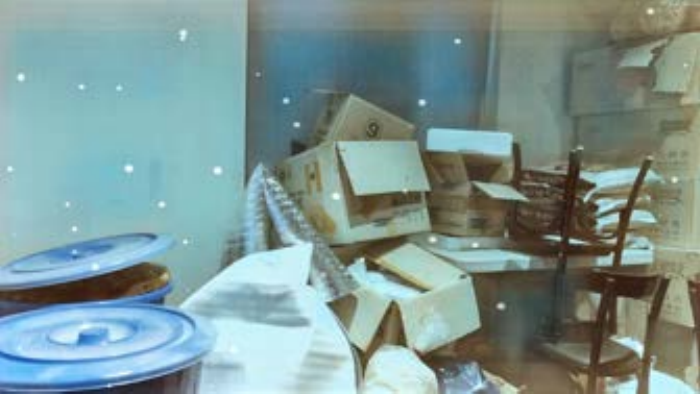}}
    \subfloat{\includegraphics[width = 0.09\linewidth]{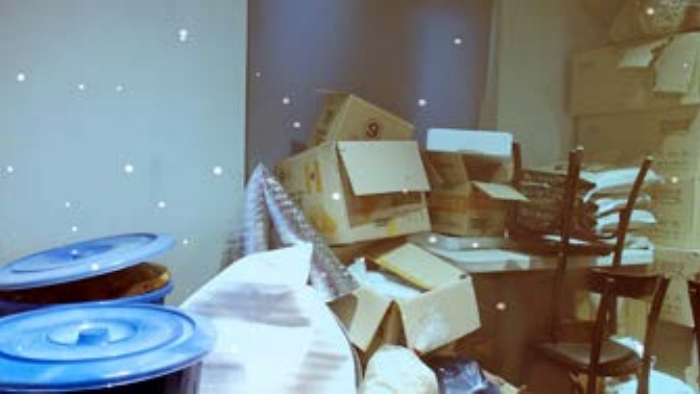}}
    \subfloat{\includegraphics[width = 0.09\linewidth]{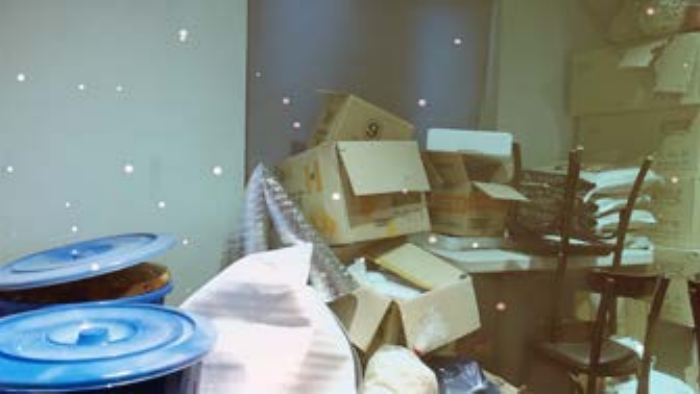}}
    \subfloat{\includegraphics[width = 0.09\linewidth]{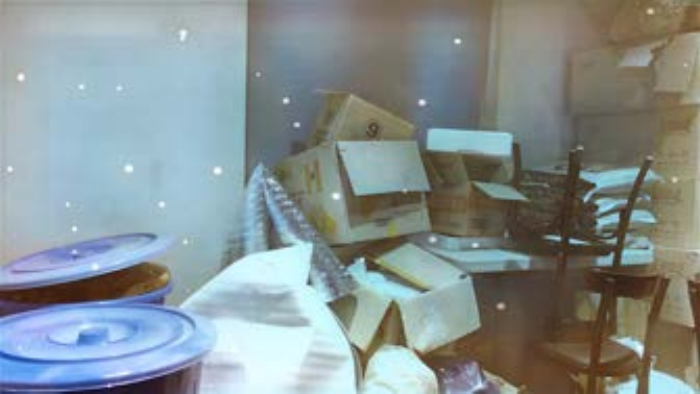}}
    \subfloat{\includegraphics[width = 0.09\linewidth]{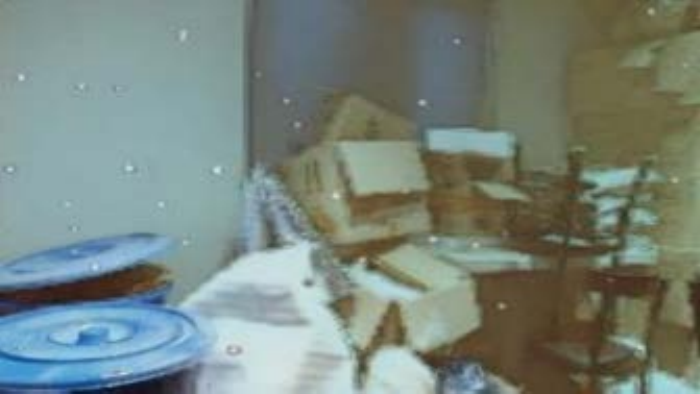}}
    \subfloat{\includegraphics[width = 0.09\linewidth]{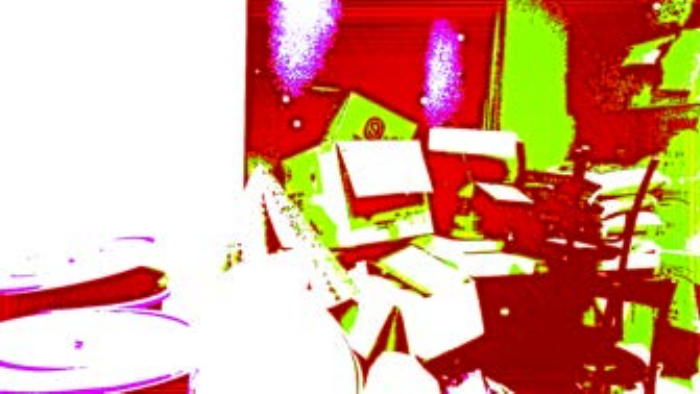}}\\
    \vspace{-0.15in}
    \subfloat{\includegraphics[width = 0.09\linewidth]{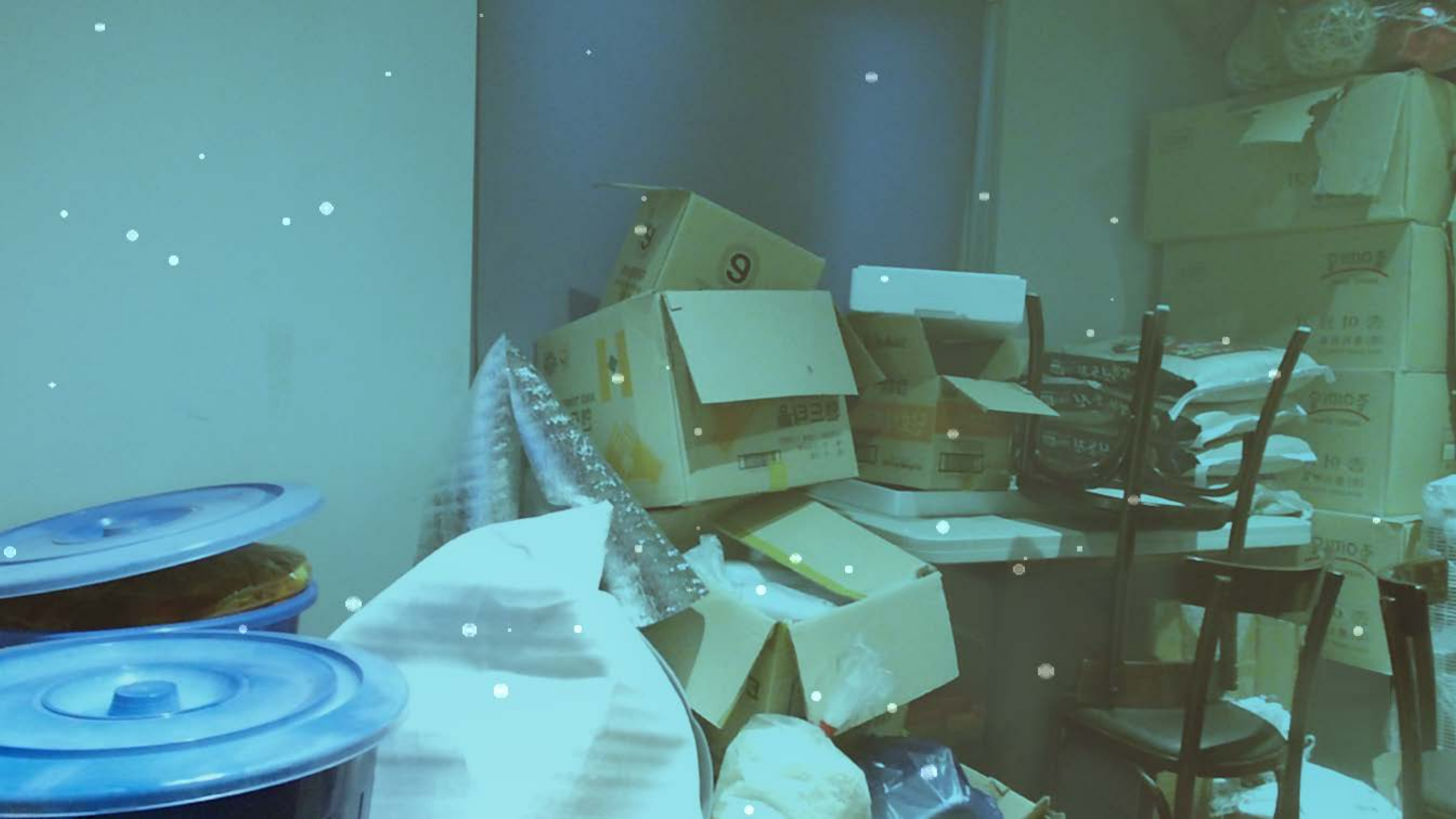}} 
    \subfloat{\includegraphics[width = 0.09\linewidth]{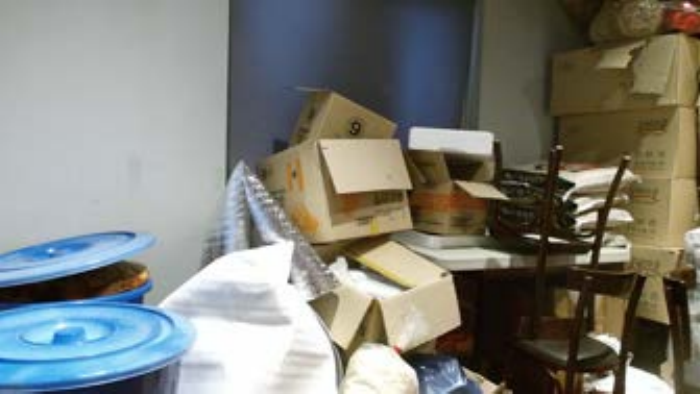}}
    \subfloat{\includegraphics[width = 0.09\linewidth]{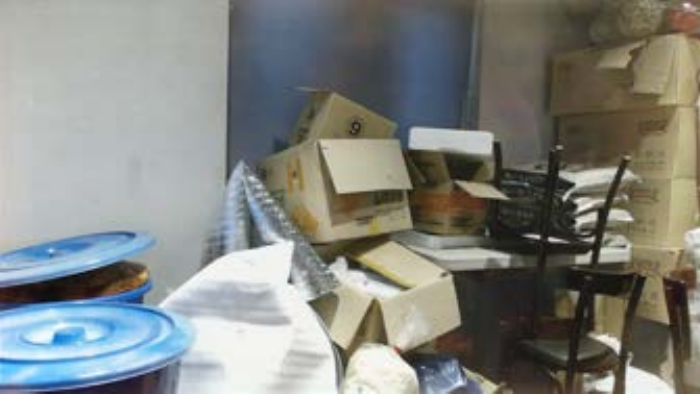}}
    \subfloat{\includegraphics[width = 0.09\linewidth]{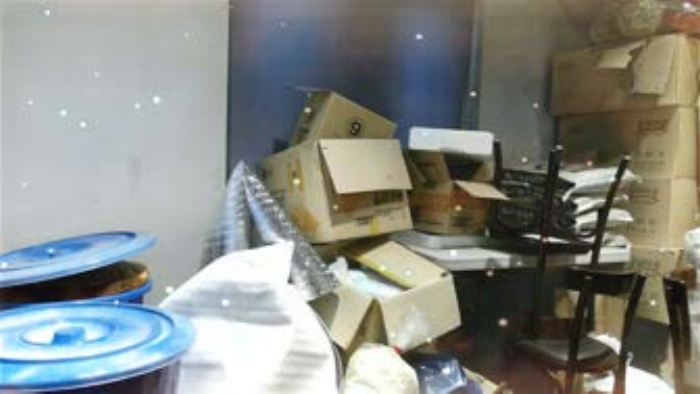}}
    \subfloat{\includegraphics[width = 0.09\linewidth]{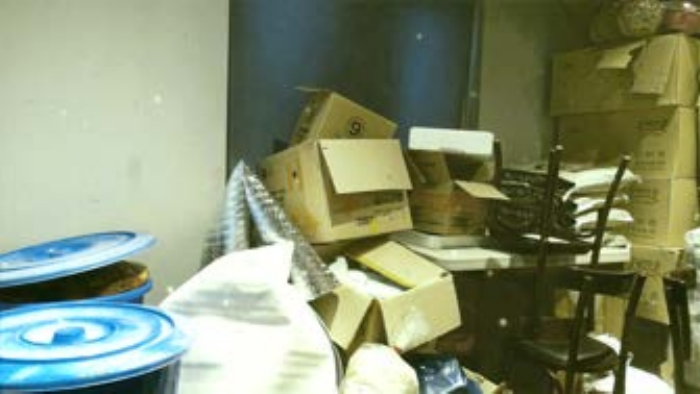}} 
    \subfloat{\includegraphics[width = 0.09\linewidth]{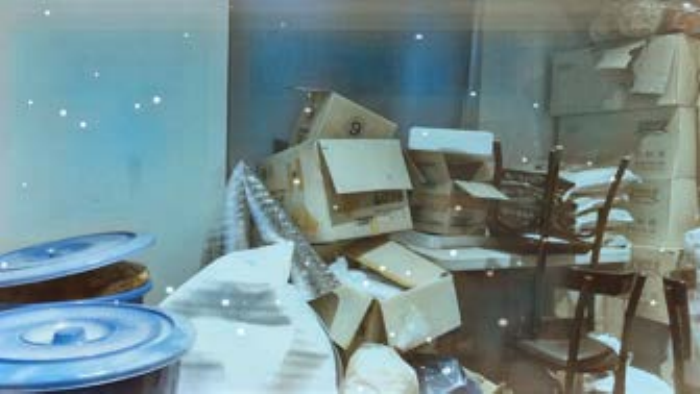}}
    \subfloat{\includegraphics[width = 0.09\linewidth]{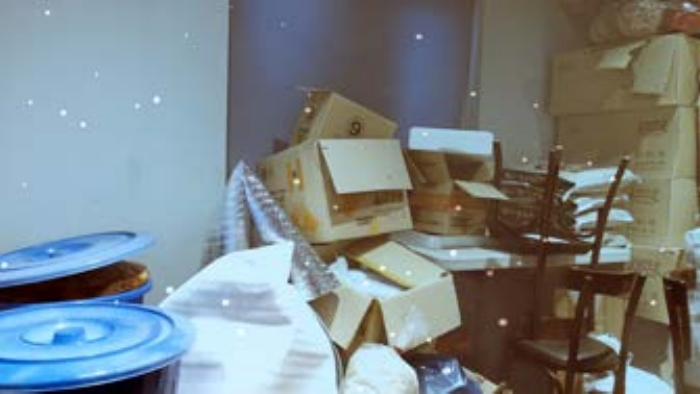}}
    \subfloat{\includegraphics[width = 0.09\linewidth]{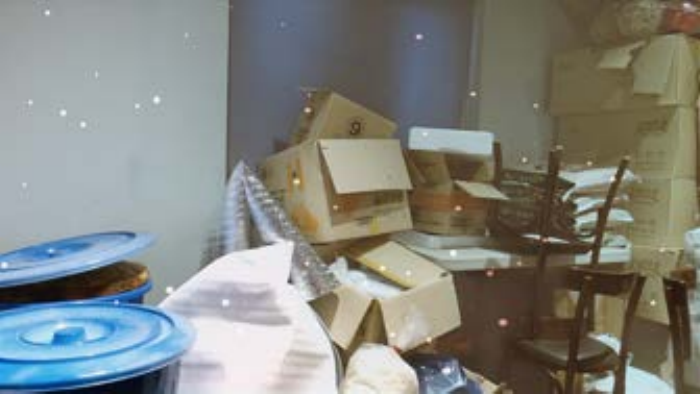}}
    \subfloat{\includegraphics[width = 0.09\linewidth]{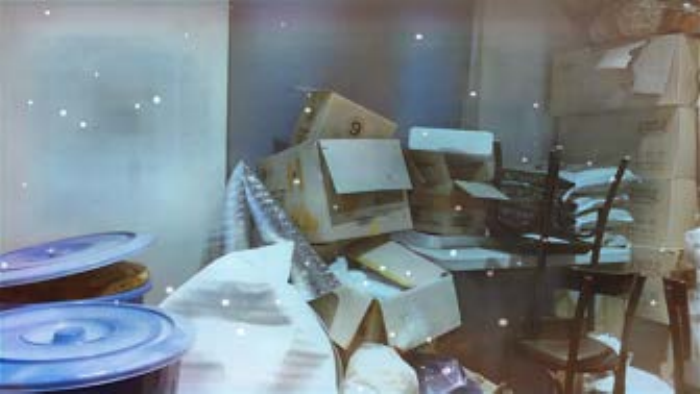}}
    \subfloat{\includegraphics[width = 0.09\linewidth]{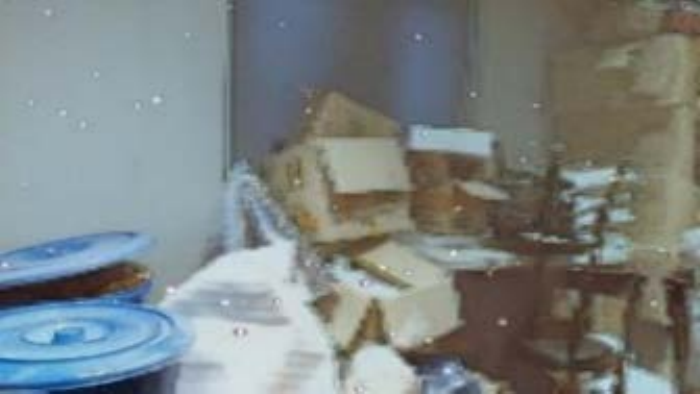}}
    \subfloat{\includegraphics[width = 0.09\linewidth]{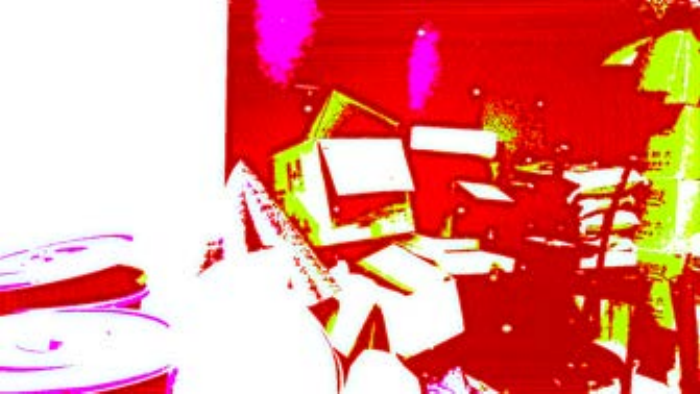}}\\
    % \subfloat{\includegraphics[width = 0.15\linewidth]{image/dataset/MSR_and_CC/206snow.pdf}} 
    % \subfloat{\includegraphics[width = 0.15\linewidth]{image/dataset/MSR_and_CC/207snow.pdf}}
    % \subfloat{\includegraphics[width = 0.15\linewidth]{image/dataset/MSR_and_CC/208snow.pdf}} 
    % \subfloat{\includegraphics[width = 0.15\linewidth]{image/dataset/MSR_and_CC/209snow.pdf}}\\
    \vspace{-0.15in}%\caption*{Synthesized test images.}%\vspace{-0.1in}
    % \subfloat{\includegraphics[width = 0.15\linewidth]{image/dataset/MSR_and_CC/PHISWID_room1.pdf}} 
    \subfloat{\includegraphics[width = 0.09\linewidth]{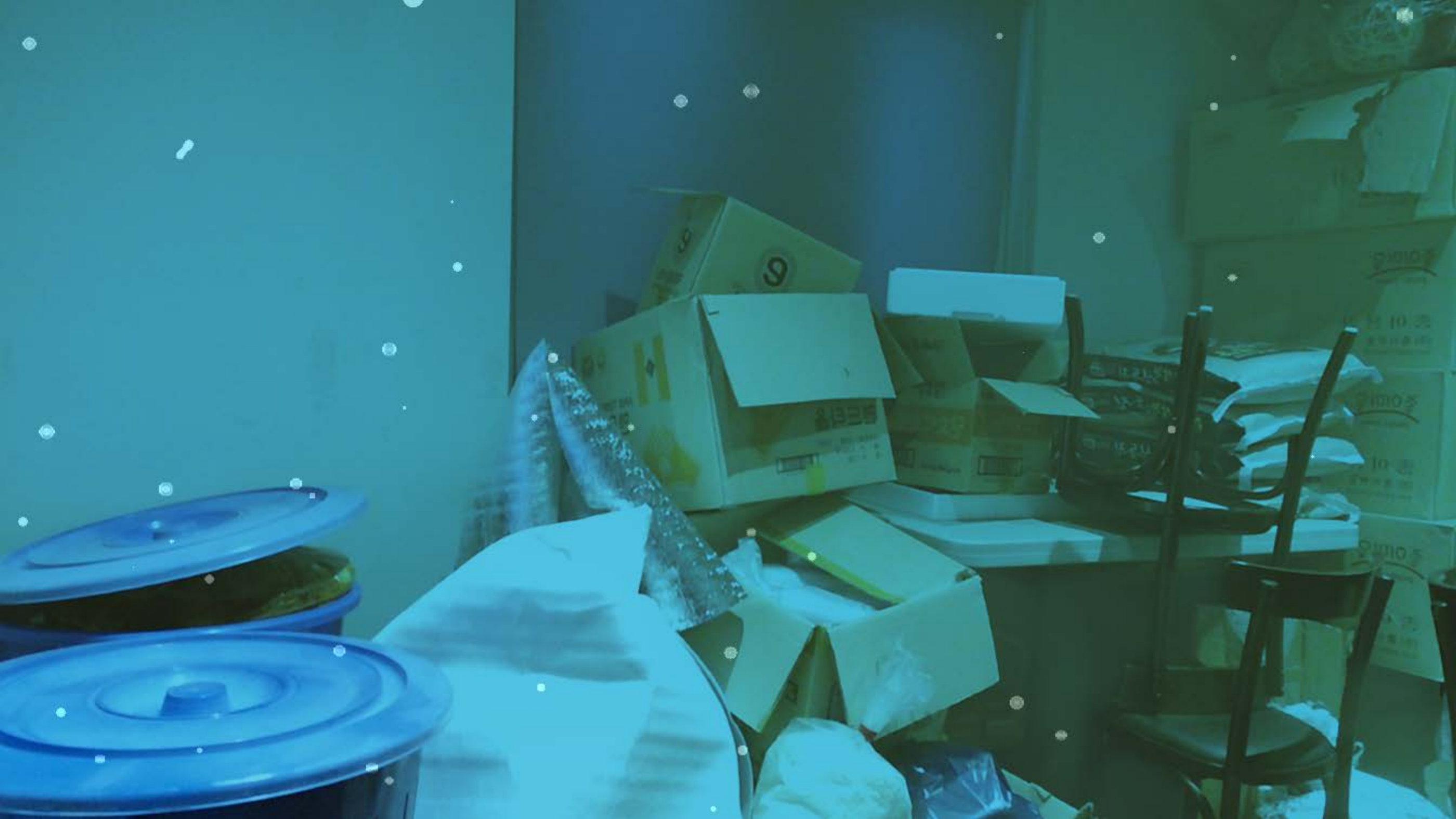}} 
    \subfloat{\includegraphics[width = 0.09\linewidth]{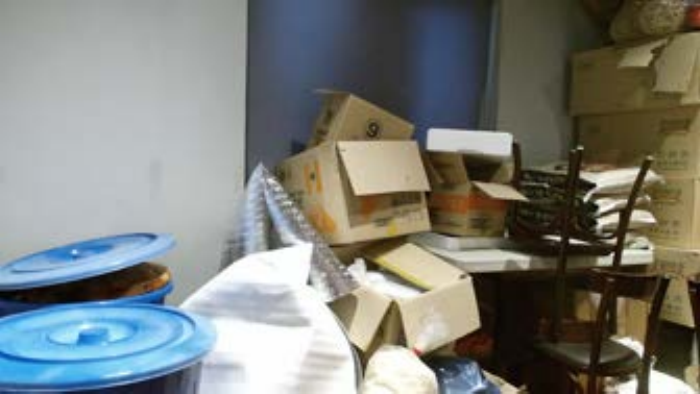}}
    \subfloat{\includegraphics[width = 0.09\linewidth]{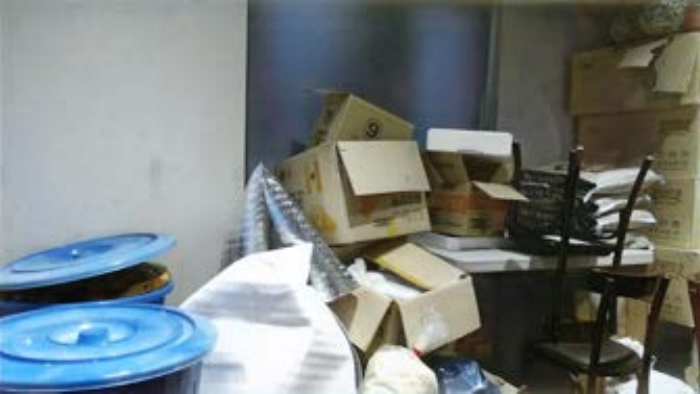}}
    \subfloat{\includegraphics[width = 0.09\linewidth]{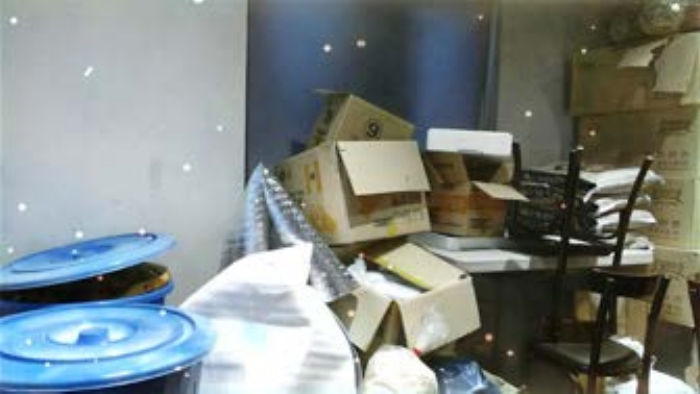}}
    \subfloat{\includegraphics[width = 0.09\linewidth]{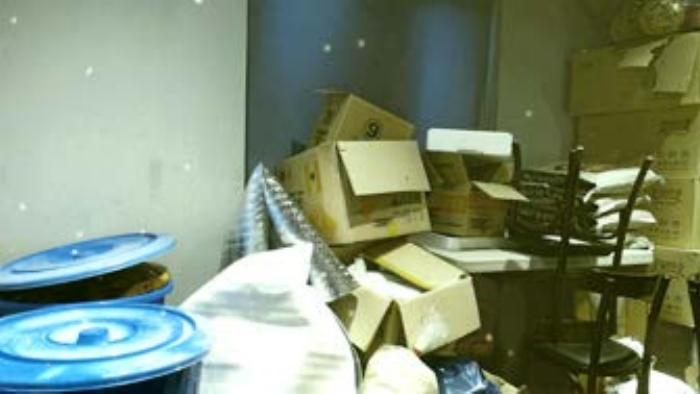}} 
    \subfloat{\includegraphics[width = 0.09\linewidth]{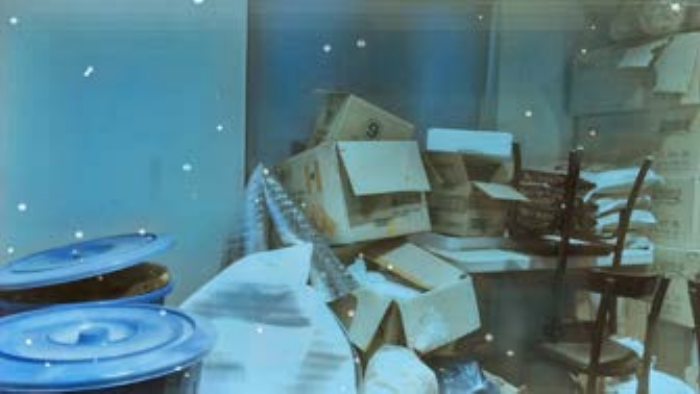}}
    \subfloat{\includegraphics[width = 0.09\linewidth]{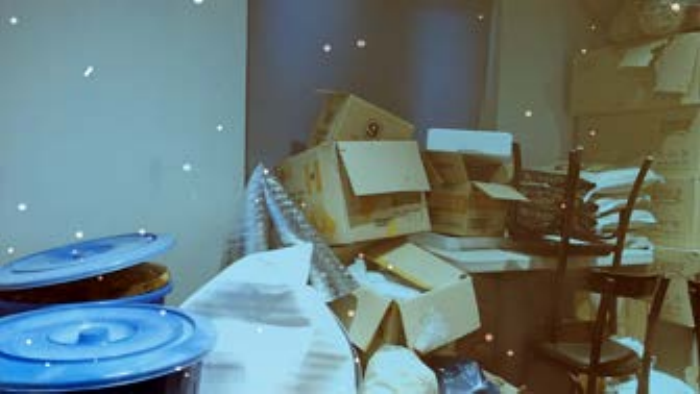}}
    \subfloat{\includegraphics[width = 0.09\linewidth]{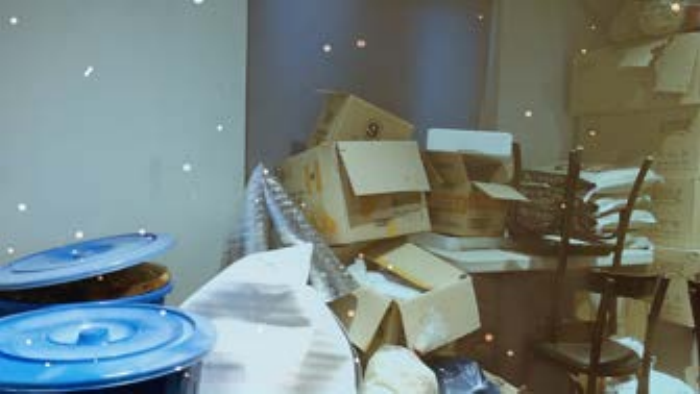}}
    \subfloat{\includegraphics[width = 0.09\linewidth]{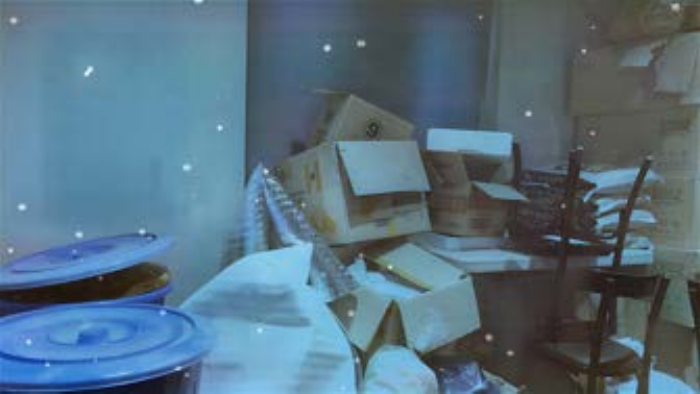}}
    \subfloat{\includegraphics[width = 0.09\linewidth]{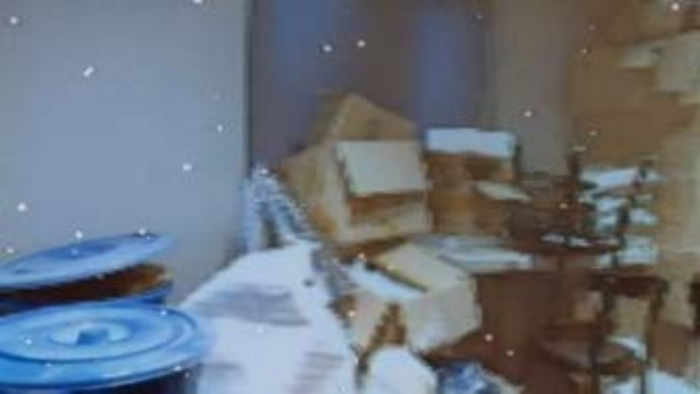}}
    \subfloat{\includegraphics[width = 0.09\linewidth]{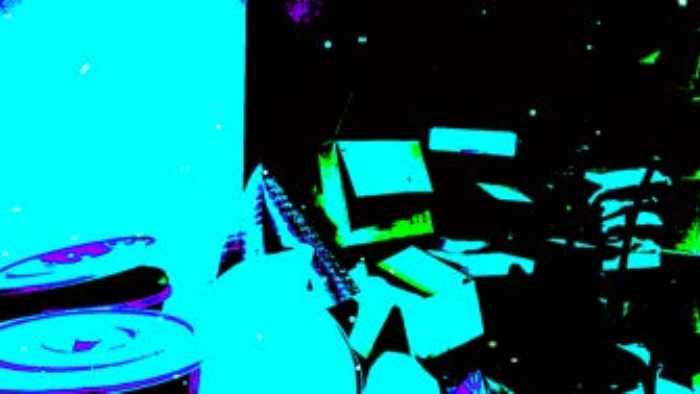}}\\
    \vspace{-0.15in}
    \subfloat{\includegraphics[width = 0.09\linewidth]{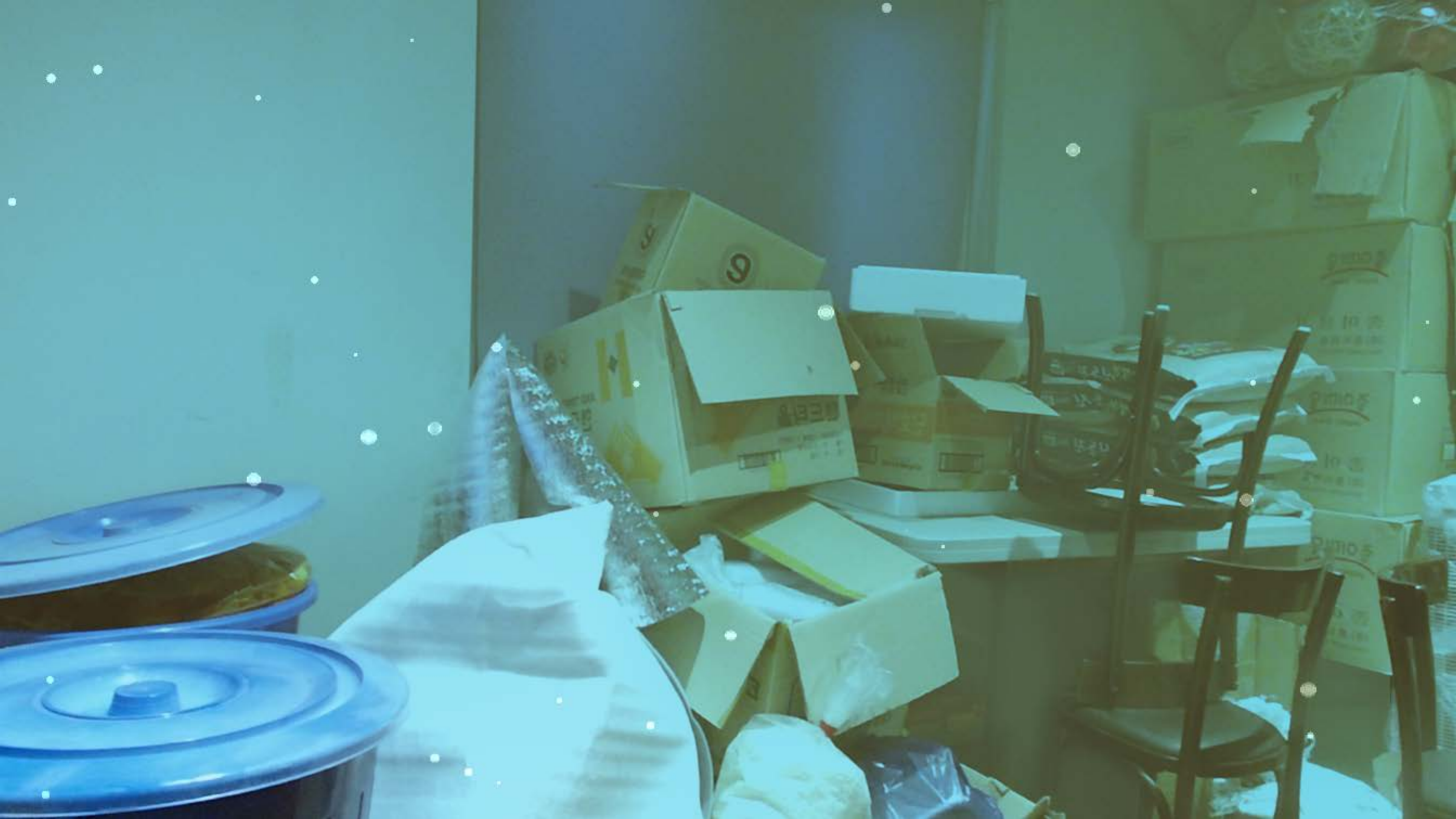}} 
    \subfloat{\includegraphics[width = 0.09\linewidth]{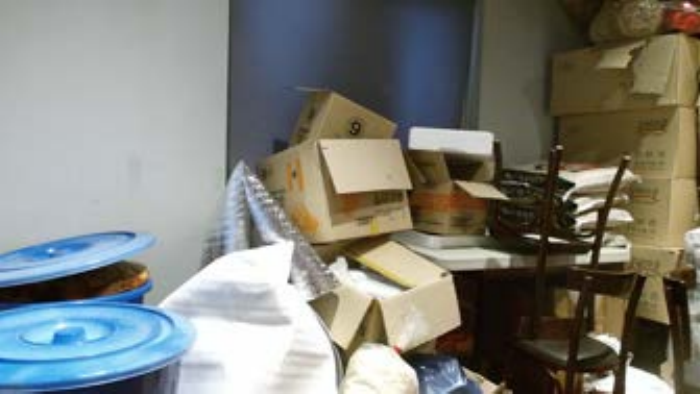}}
    \subfloat{\includegraphics[width = 0.09\linewidth]{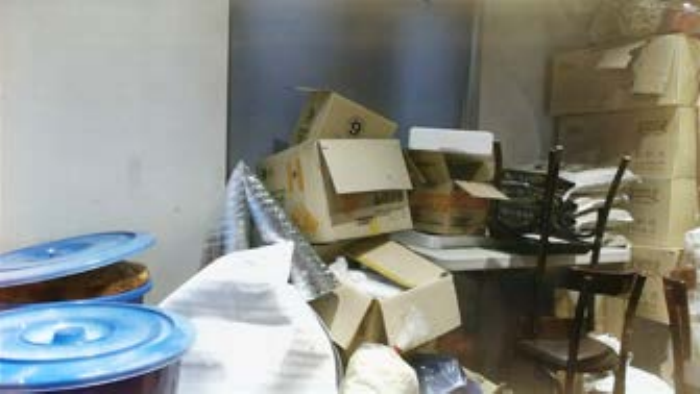}}
    \subfloat{\includegraphics[width = 0.09\linewidth]{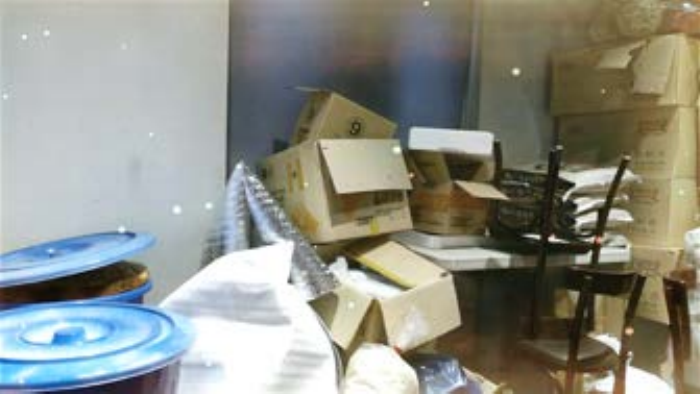}}
    \subfloat{\includegraphics[width = 0.09\linewidth]{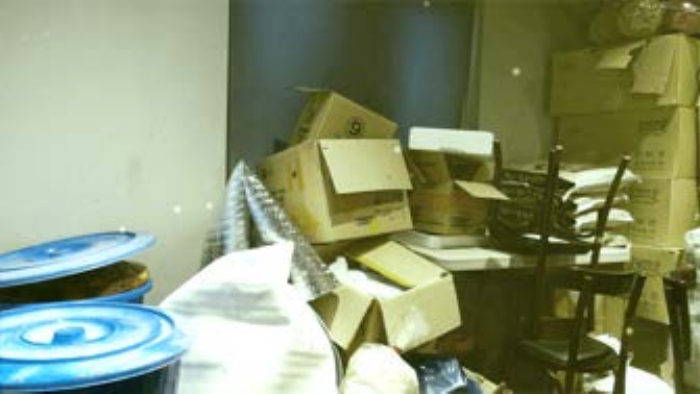}} 
    \subfloat{\includegraphics[width = 0.09\linewidth]{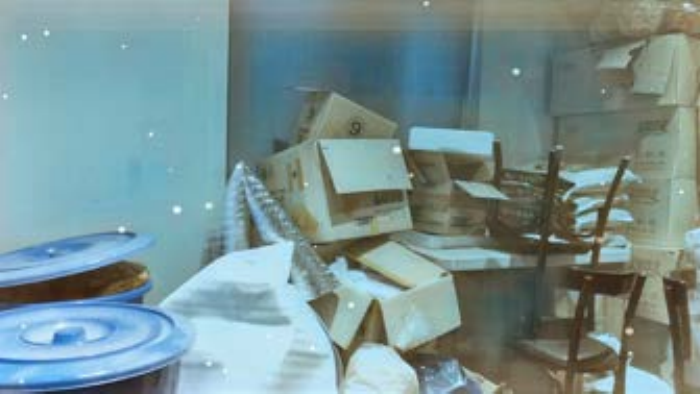}}
    \subfloat{\includegraphics[width = 0.09\linewidth]{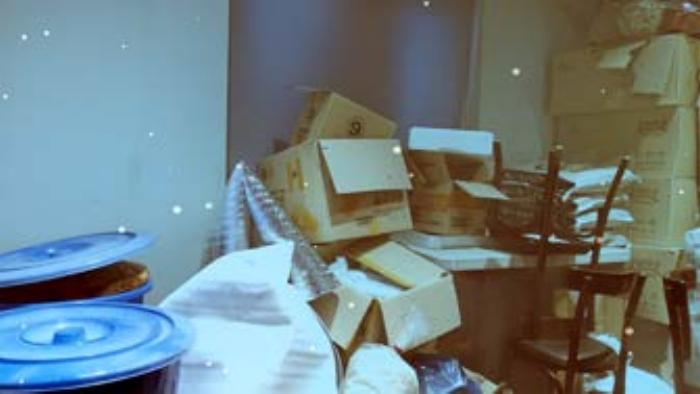}}
    \subfloat{\includegraphics[width = 0.09\linewidth]{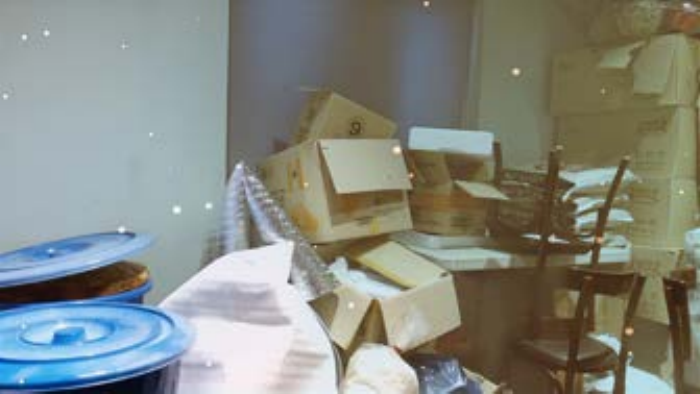}}
    \subfloat{\includegraphics[width = 0.09\linewidth]{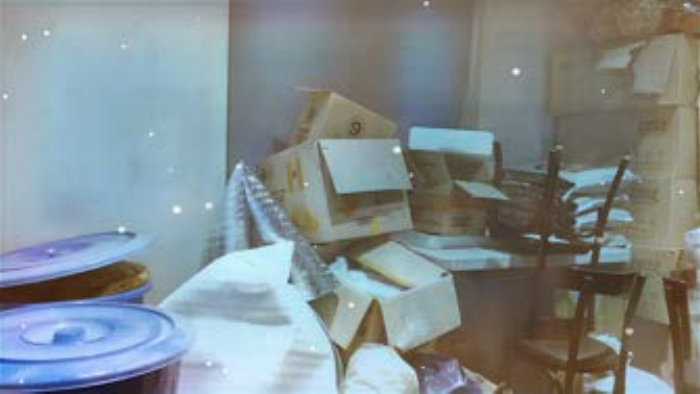}}
    \subfloat{\includegraphics[width = 0.09\linewidth]{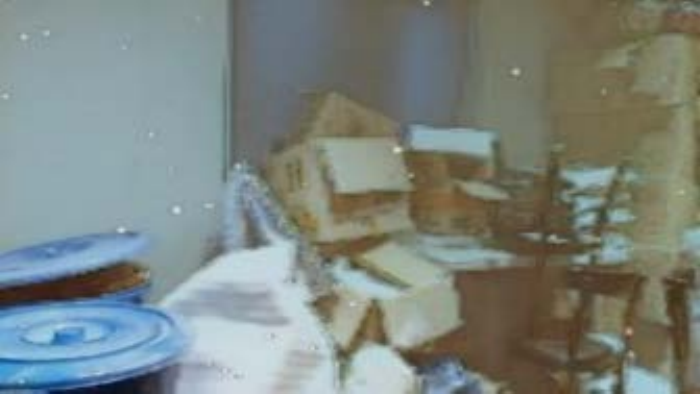}}
    \subfloat{\includegraphics[width = 0.09\linewidth]{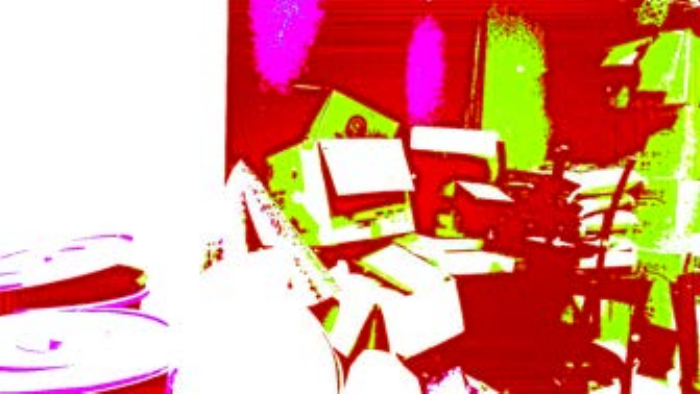}}\\
    % \subfloat{\includegraphics[width = 0.15\linewidth]{image/dataset/MSR_and_CC/PHISWID_room3.pdf}}
    % \subfloat{\includegraphics[width = 0.15\linewidth]{image/dataset/MSR_and_CC/PHISWID_room4.pdf}} 
    % \subfloat{\includegraphics[width = 0.15\linewidth]{image/dataset/MSR_and_CC/PHISWID_room5.pdf}}\\
    \vspace{-0.15in}%\caption*{Restoration results by Transformer.}%\vspace{-0.1in}
    % \subfloat{\includegraphics[width = 0.15\linewidth]{image/dataset/DeepWaveNet/Deepwavenet_205.pdf}} 
    % \subfloat{\includegraphics[width = 0.15\linewidth]{image/dataset/DeepWaveNet/Deepwavenet_206.pdf}} 
    % \subfloat{\includegraphics[width = 0.15\linewidth]{image/dataset/DeepWaveNet/Deepwavenet_207.pdf}}
    \subfloat{\includegraphics[width = 0.09\linewidth]{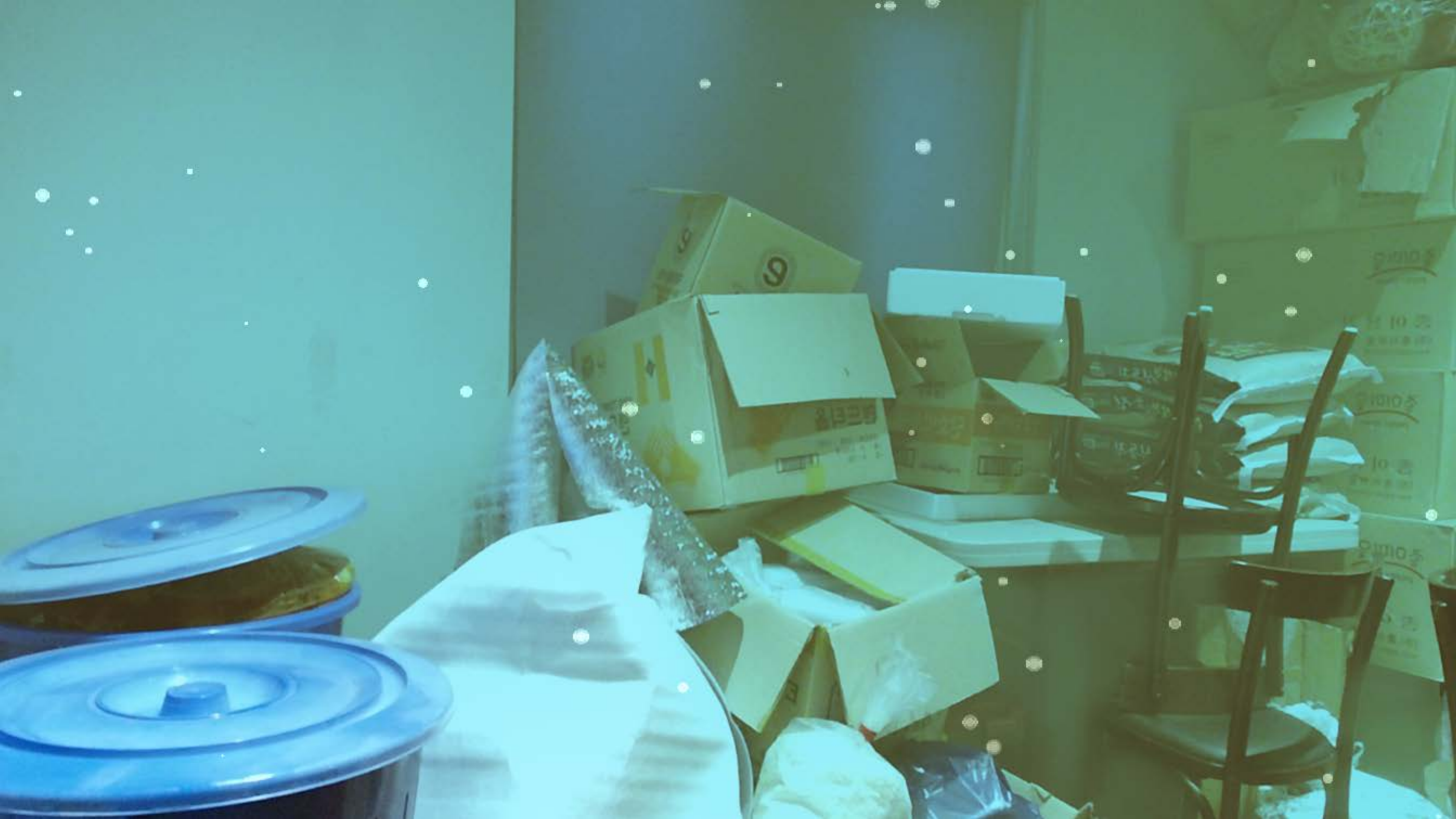}} 
    \subfloat{\includegraphics[width = 0.09\linewidth]{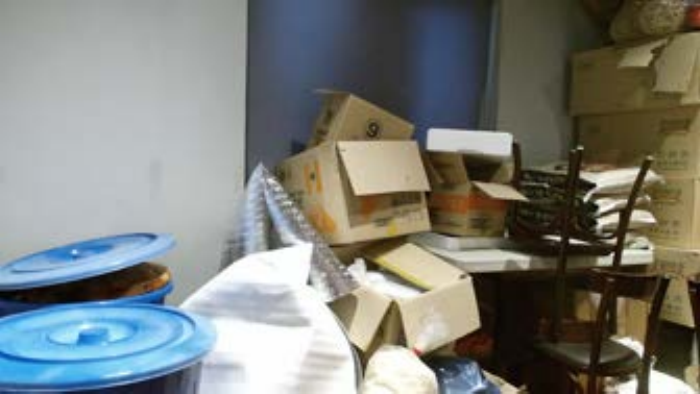}}
    \subfloat{\includegraphics[width = 0.09\linewidth]{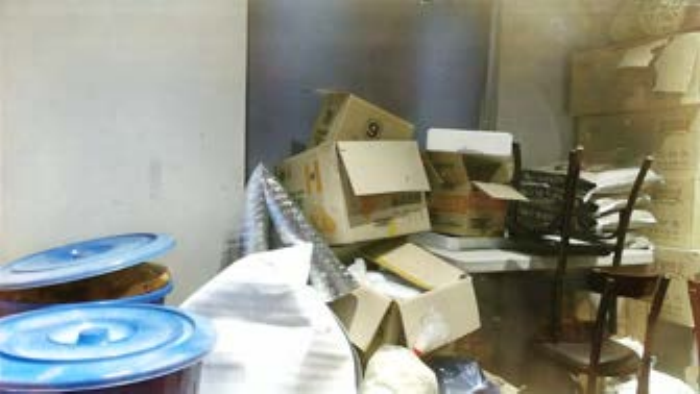}}
    \subfloat{\includegraphics[width = 0.09\linewidth]{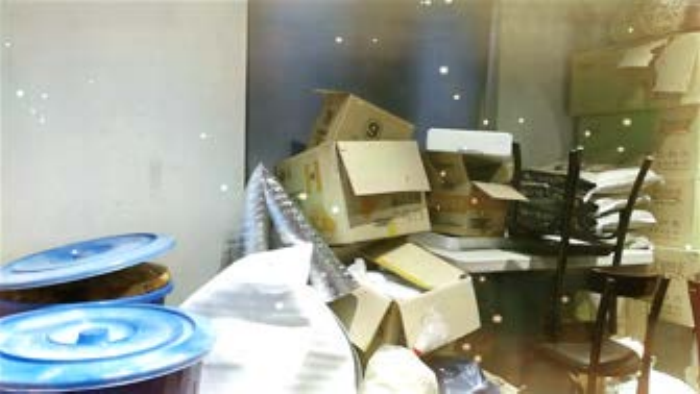}}
    \subfloat{\includegraphics[width = 0.09\linewidth]{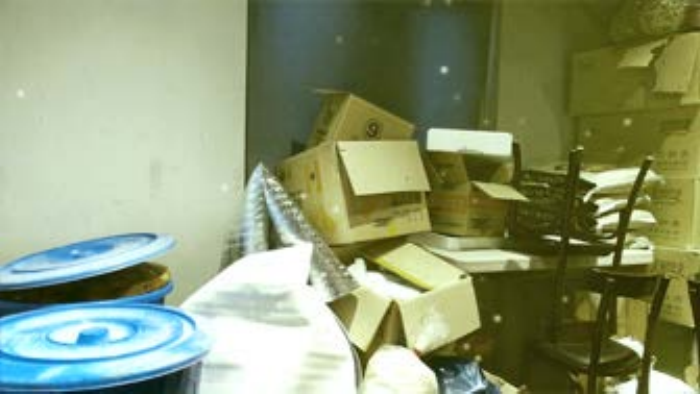}} 
    \subfloat{\includegraphics[width = 0.09\linewidth]{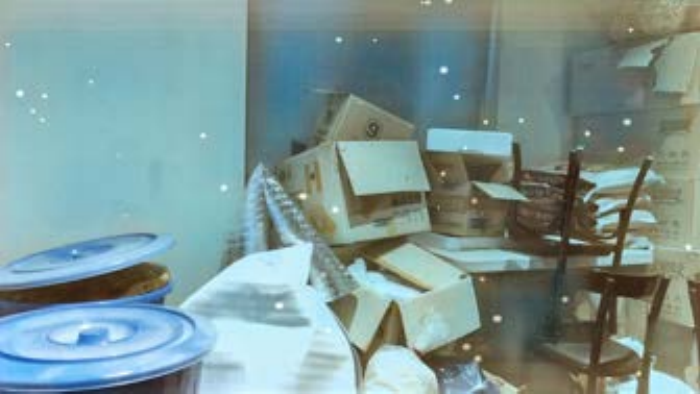}}
    \subfloat{\includegraphics[width = 0.09\linewidth]{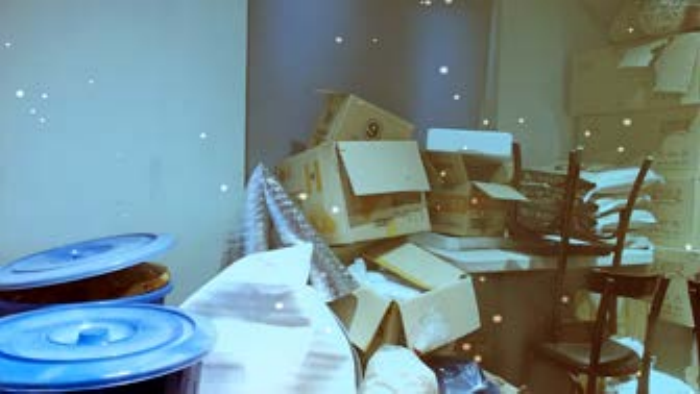}}
    \subfloat{\includegraphics[width = 0.09\linewidth]{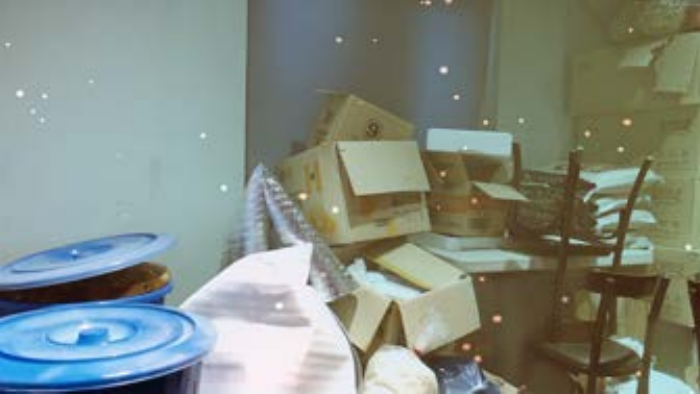}}
    \subfloat{\includegraphics[width = 0.09\linewidth]{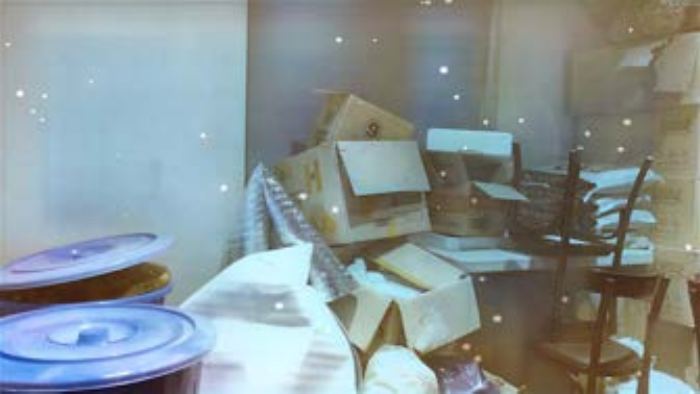}}
    \subfloat{\includegraphics[width = 0.09\linewidth]{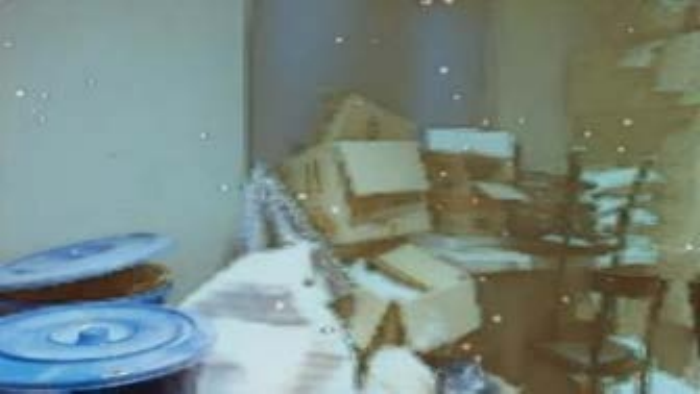}}
    \subfloat{\includegraphics[width = 0.09\linewidth]{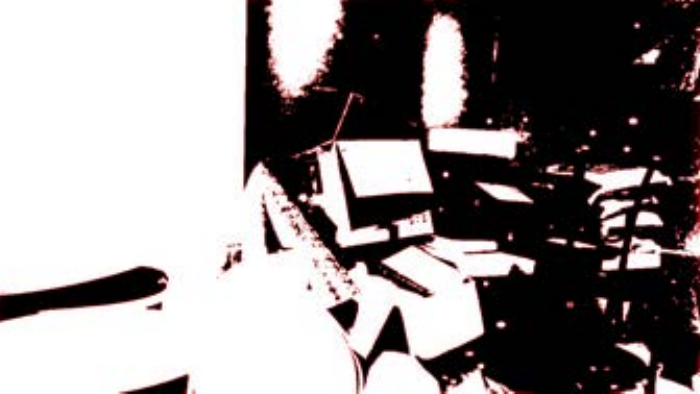}}\\
    \vspace{-0.15in}
    \subfloat{\includegraphics[width = 0.09\linewidth]{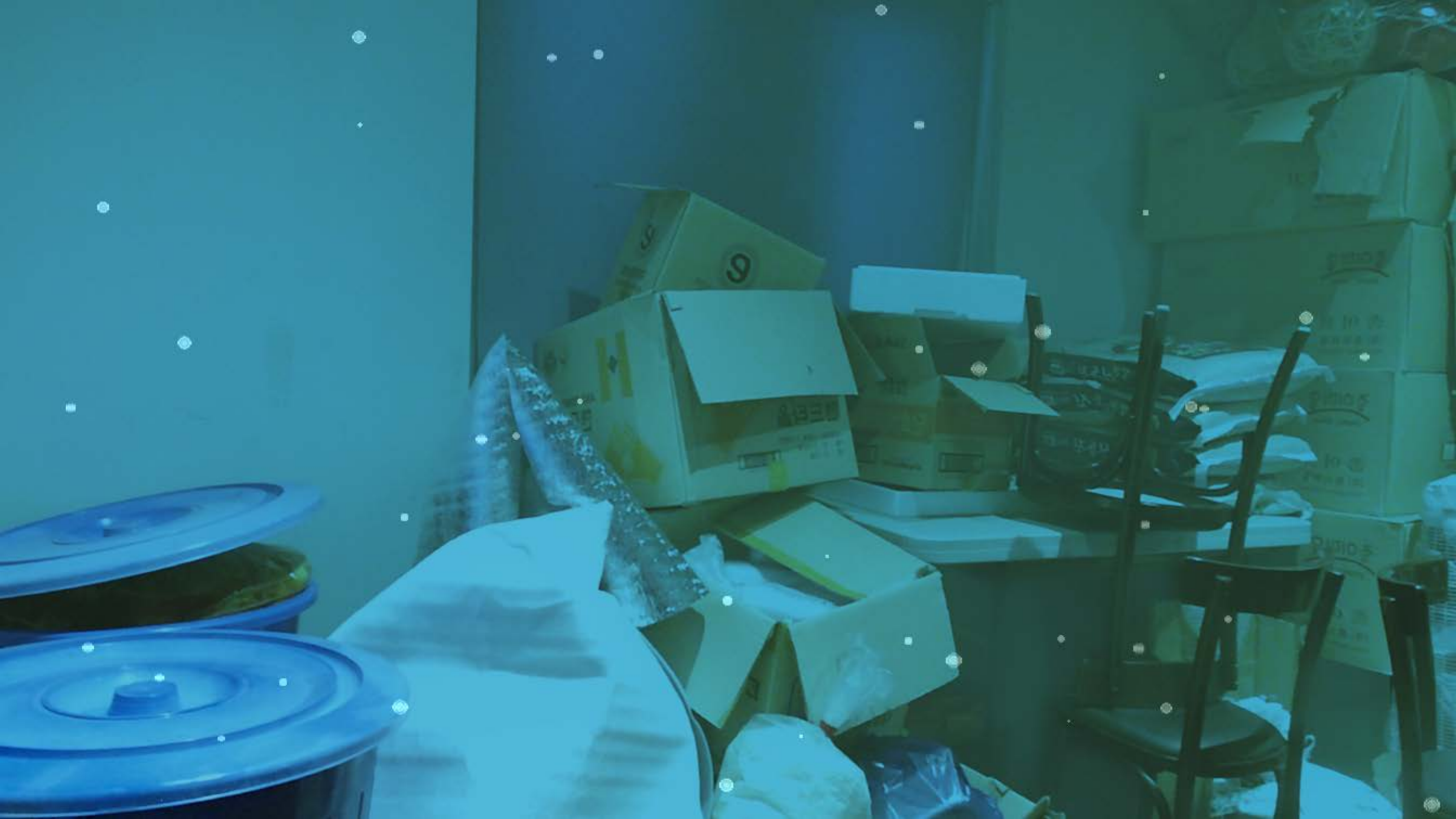}} 
    \subfloat{\includegraphics[width = 0.09\linewidth]{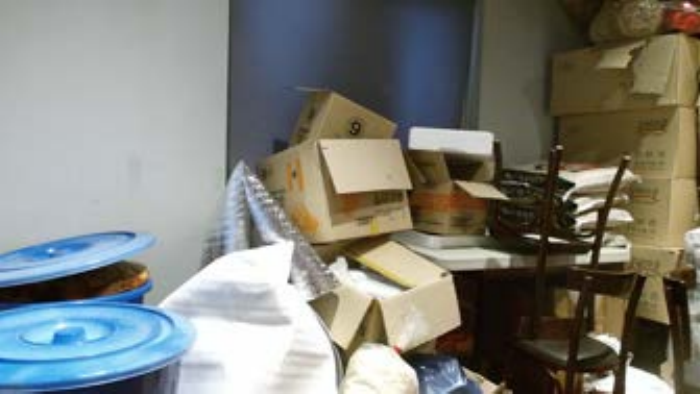}}
    \subfloat{\includegraphics[width = 0.09\linewidth]{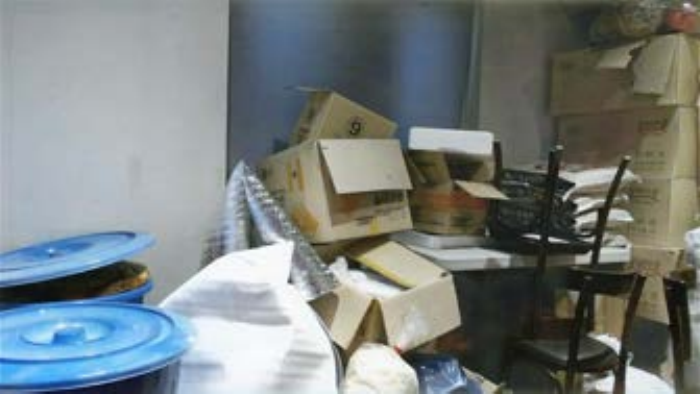}}
    \subfloat{\includegraphics[width = 0.09\linewidth]{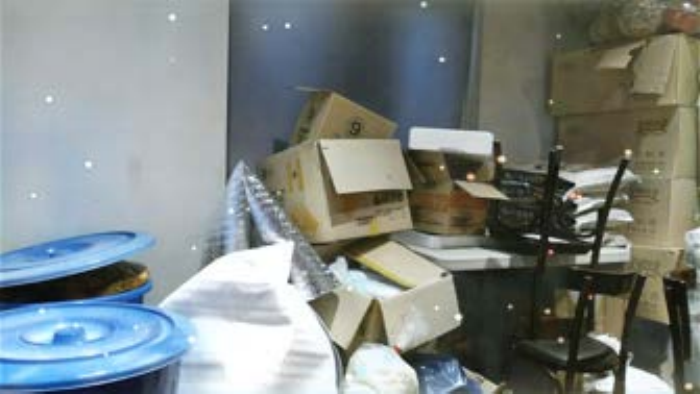}}
    \subfloat{\includegraphics[width = 0.09\linewidth]{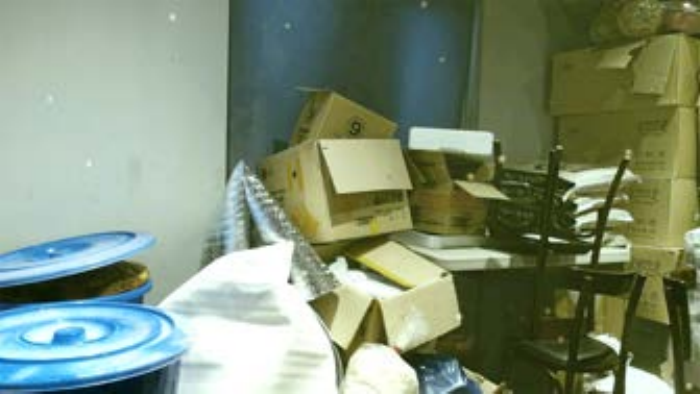}} 
    \subfloat{\includegraphics[width = 0.09\linewidth]{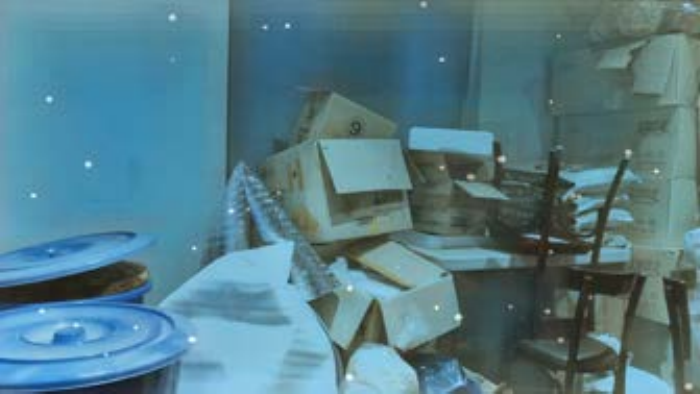}}
    \subfloat{\includegraphics[width = 0.09\linewidth]{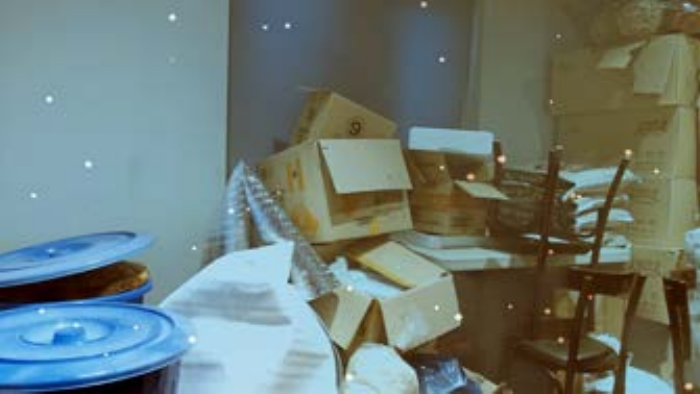}}
    \subfloat{\includegraphics[width = 0.09\linewidth]{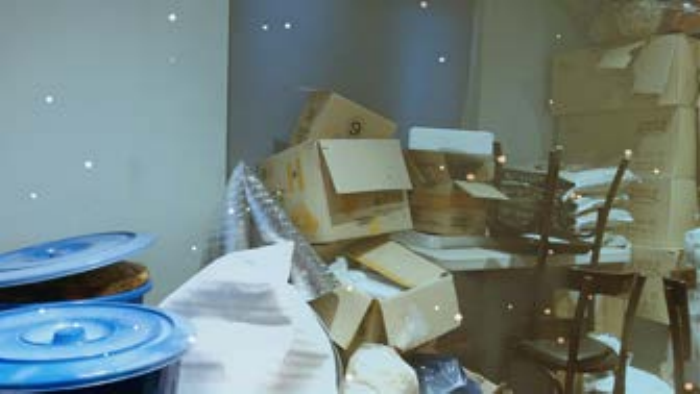}}
    \subfloat{\includegraphics[width = 0.09\linewidth]{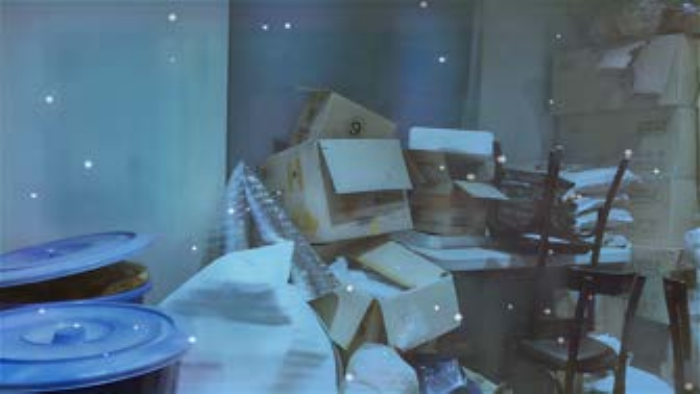}}
    \subfloat{\includegraphics[width = 0.09\linewidth]{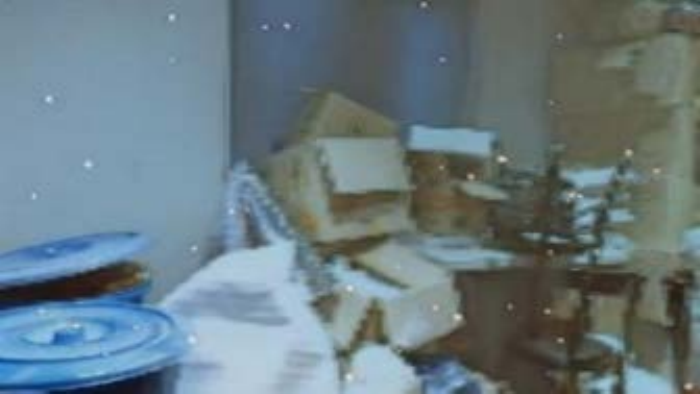}}
    \subfloat{\includegraphics[width = 0.09\linewidth]{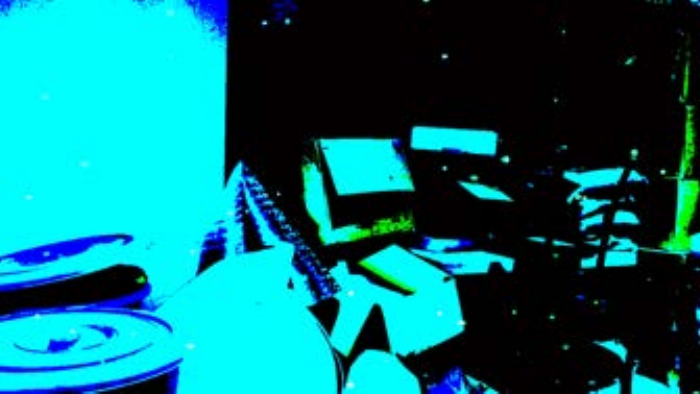}}\\
    % \subfloat{\includegraphics[width = 0.15\linewidth]{image/dataset/DeepWaveNet/Deepwavenet_208.pdf}}
    % \subfloat{\includegraphics[width = 0.15\linewidth]{image/dataset/DeepWaveNet/Deepwavenet_209.pdf}} \\
    \vspace{-0.15in}%\caption*{Restoration results by Deep WN.}%\vspace{-0.1in}
    % \subfloat{\includegraphics[width = 0.15\linewidth]{image/dataset/Waternet/waternet_205.pdf}} 
    % \subfloat{\includegraphics[width = 0.15\linewidth]{image/dataset/Waternet/waternet_206.pdf}} 
    % \subfloat{\includegraphics[width = 0.15\linewidth]{image/dataset/Waternet/waternet_207.pdf}}
    \subfloat{\includegraphics[width = 0.09\linewidth]{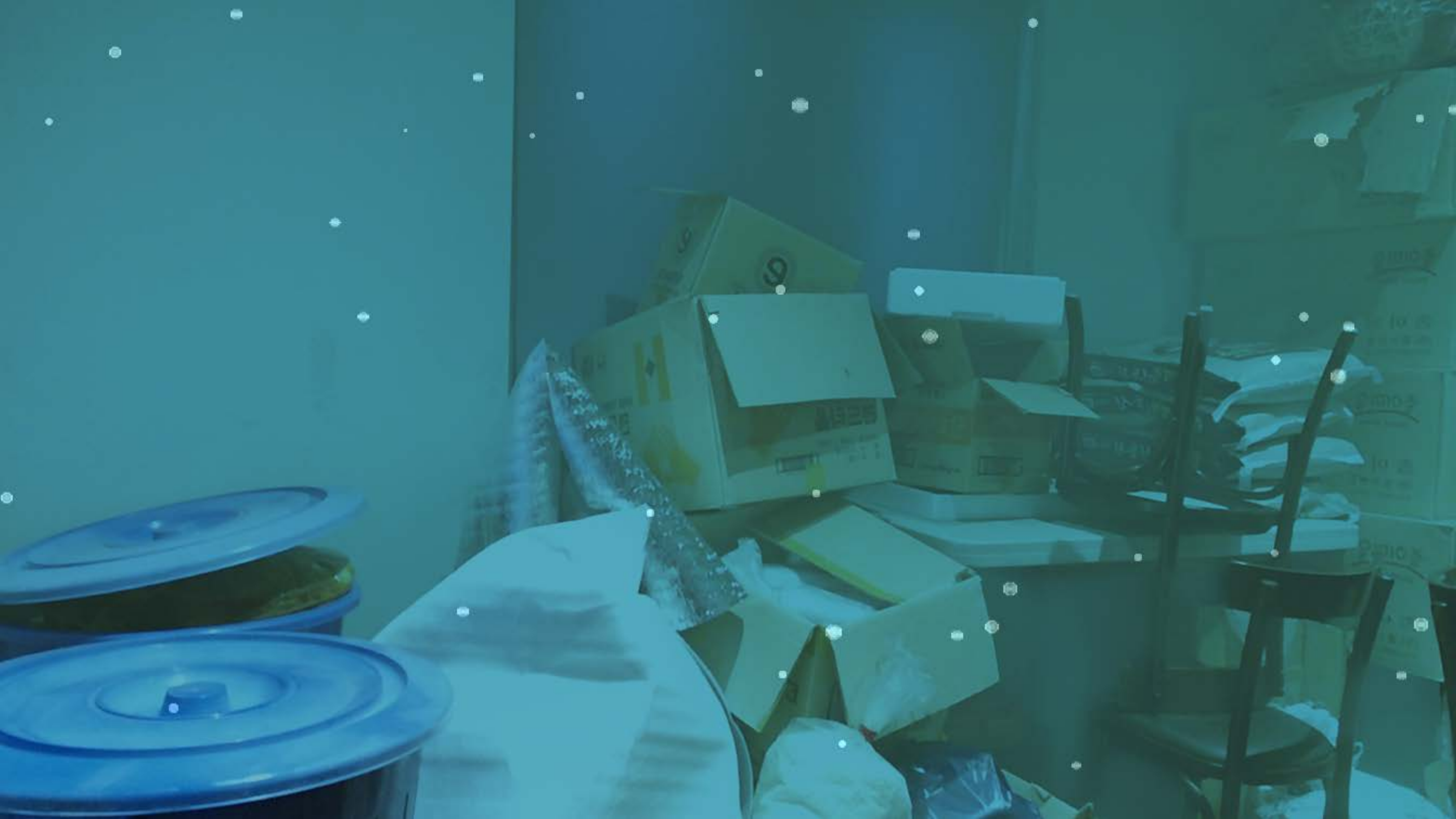}} 
    \subfloat{\includegraphics[width = 0.09\linewidth]{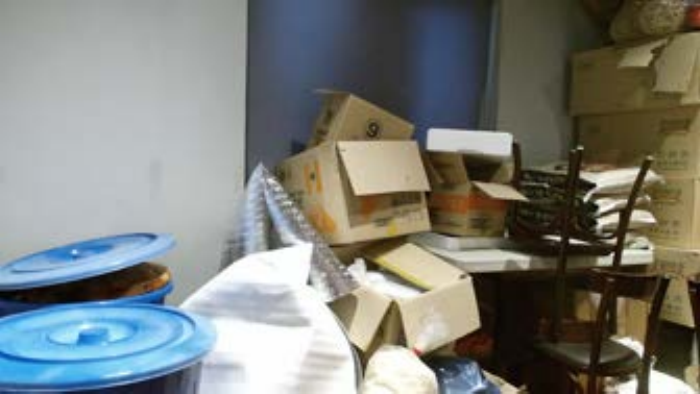}}
    \subfloat{\includegraphics[width = 0.09\linewidth]{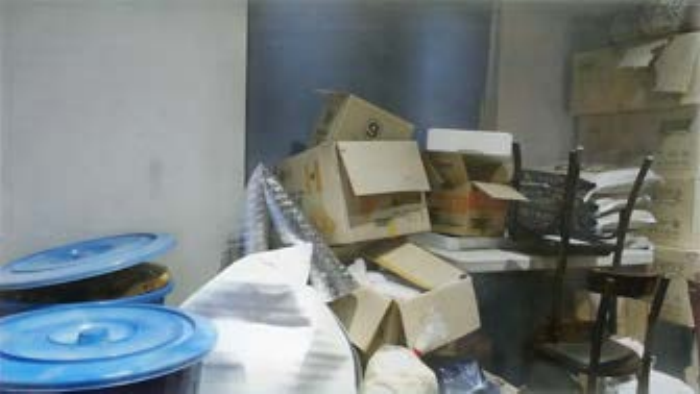}}
    \subfloat{\includegraphics[width = 0.09\linewidth]{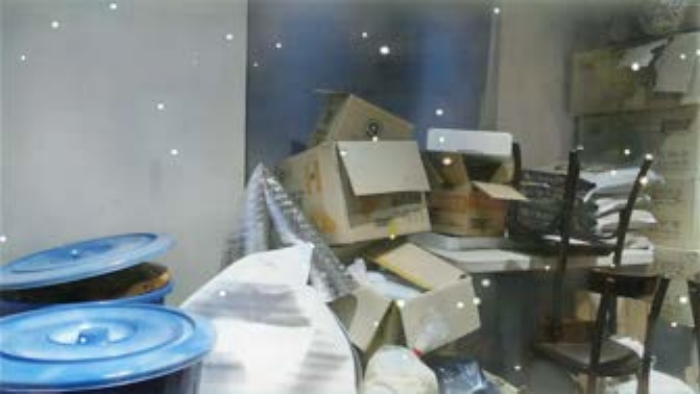}}
    \subfloat{\includegraphics[width = 0.09\linewidth]{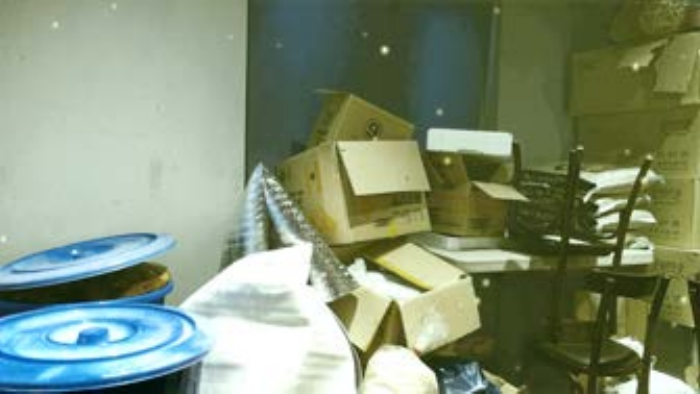}} 
    \subfloat{\includegraphics[width = 0.09\linewidth]{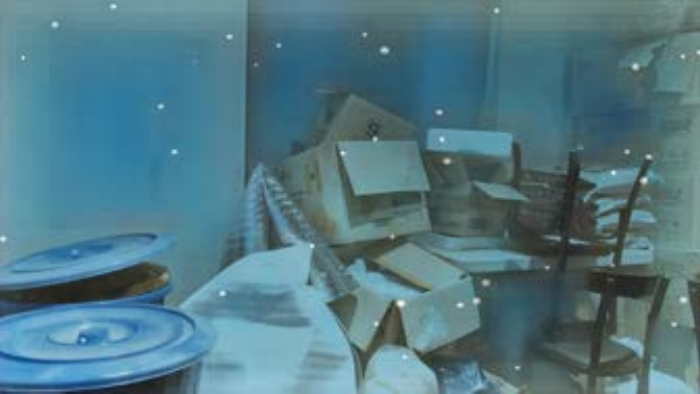}}
    \subfloat{\includegraphics[width = 0.09\linewidth]{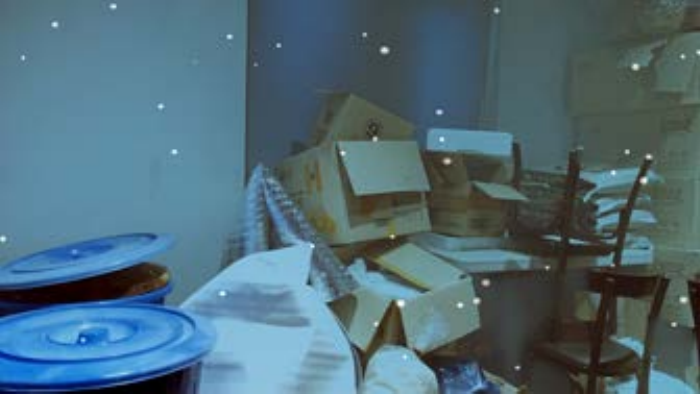}}
    \subfloat{\includegraphics[width = 0.09\linewidth]{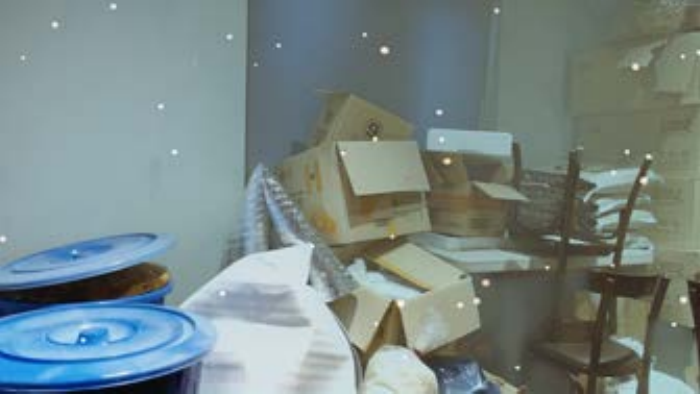}}
    \subfloat{\includegraphics[width = 0.09\linewidth]{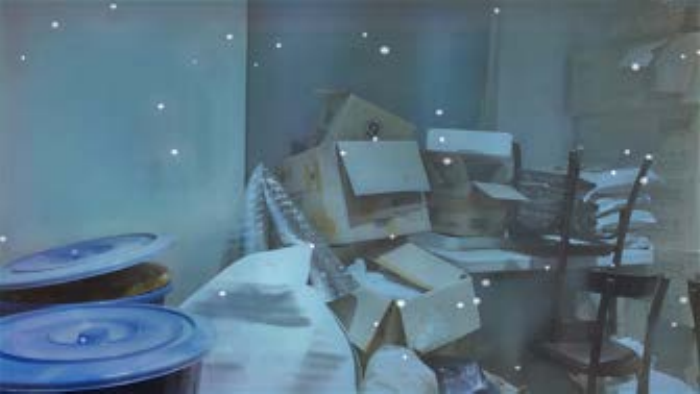}}
    \subfloat{\includegraphics[width = 0.09\linewidth]{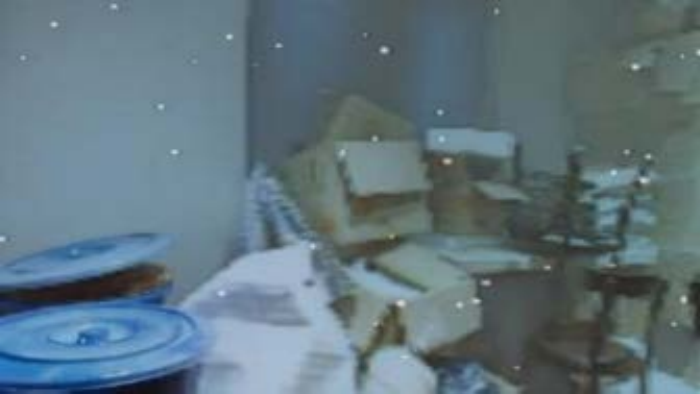}}
    \subfloat{\includegraphics[width = 0.09\linewidth]{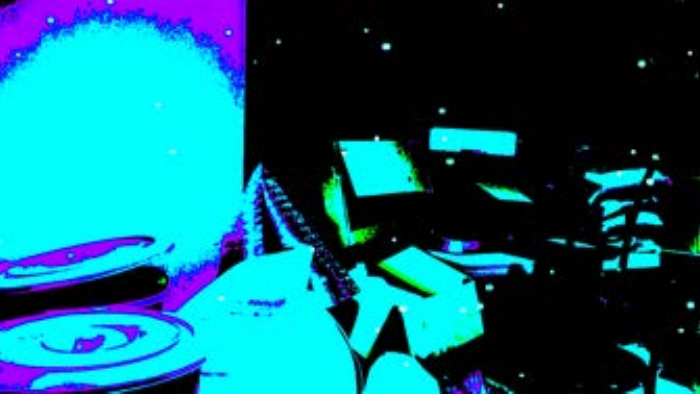}}\\
    \vspace{-0.15in}
    \subfloat{\includegraphics[width = 0.09\linewidth]{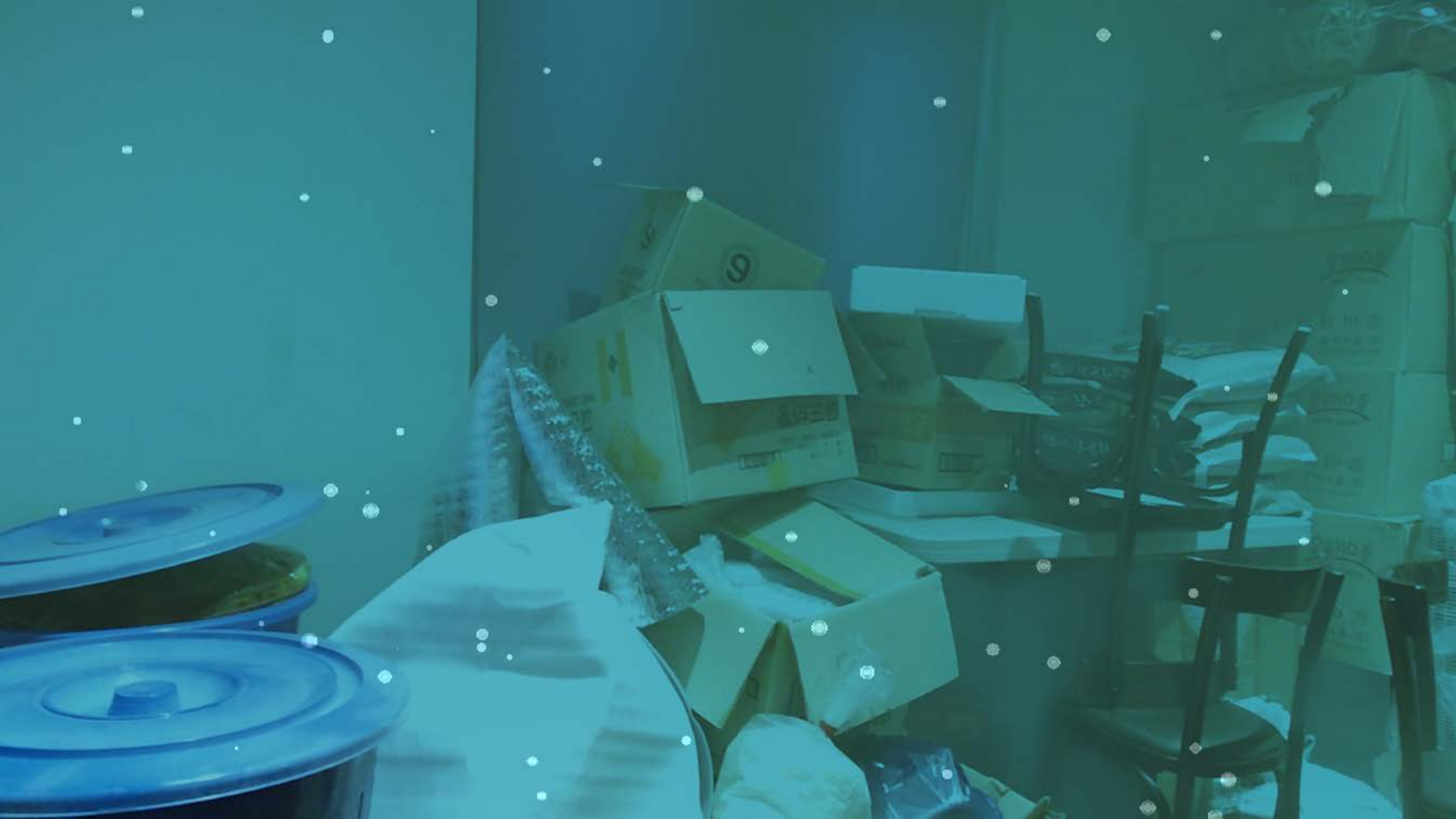}} 
    \subfloat{\includegraphics[width = 0.09\linewidth]{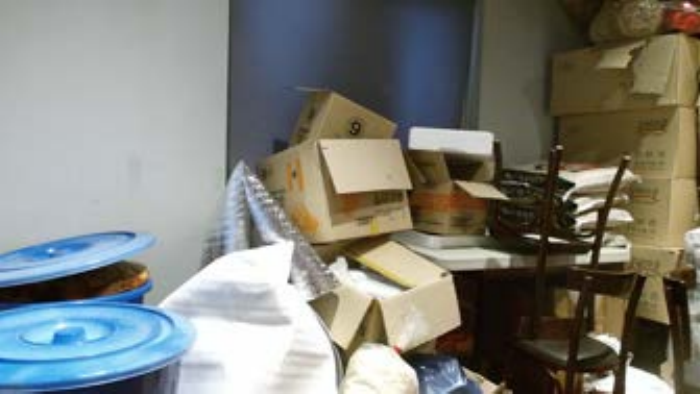}}
    \subfloat{\includegraphics[width = 0.09\linewidth]{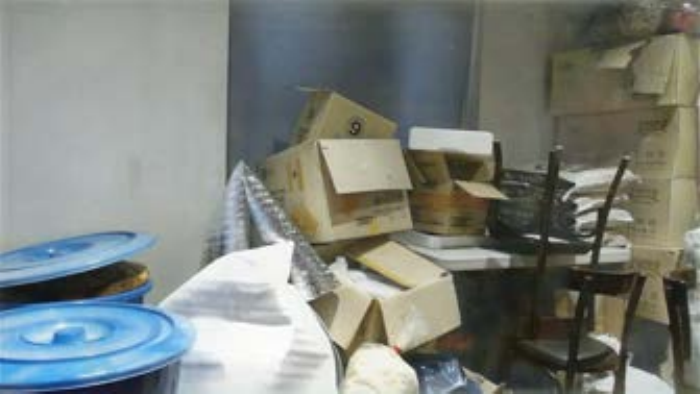}}
    \subfloat{\includegraphics[width = 0.09\linewidth]{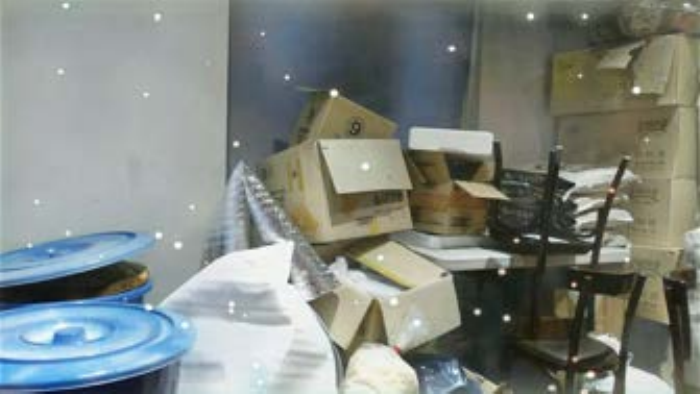}}
    \subfloat{\includegraphics[width = 0.09\linewidth]{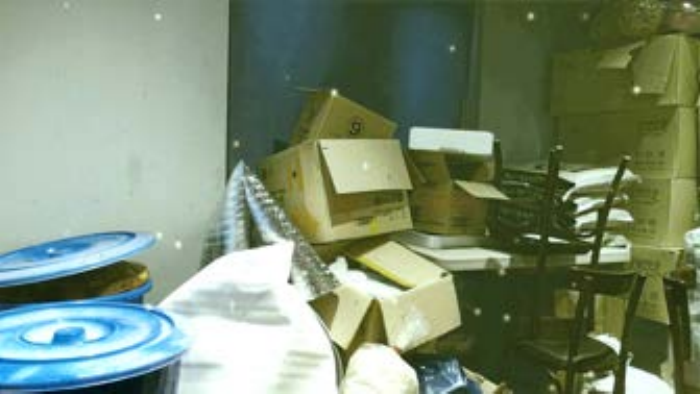}} 
    \subfloat{\includegraphics[width = 0.09\linewidth]{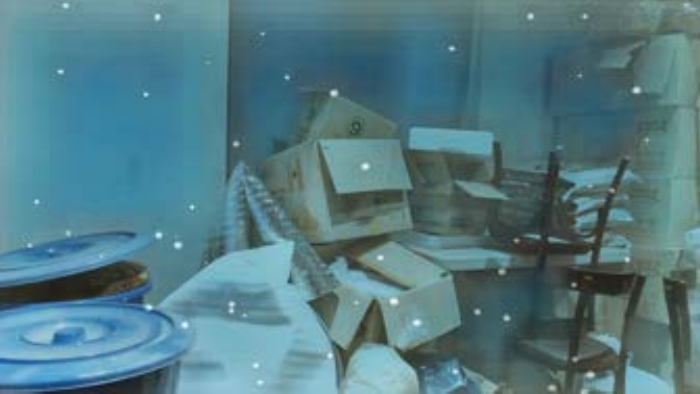}}
    \subfloat{\includegraphics[width = 0.09\linewidth]{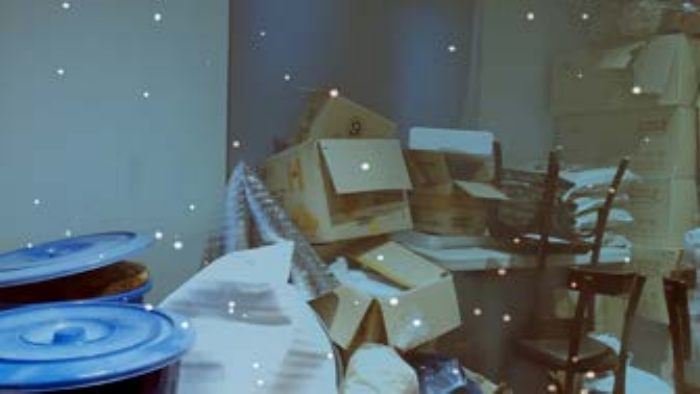}}
    \subfloat{\includegraphics[width = 0.09\linewidth]{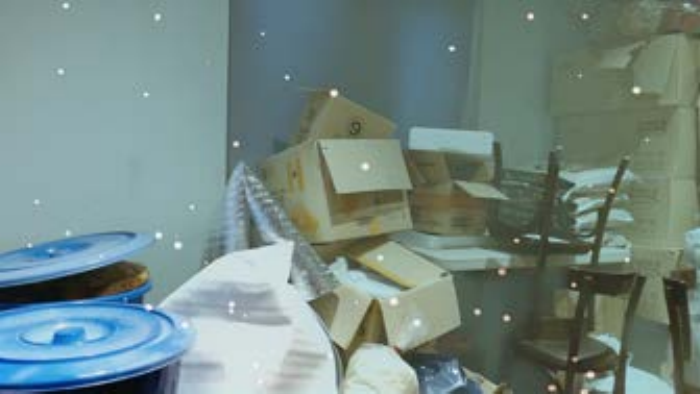}}
    \subfloat{\includegraphics[width = 0.09\linewidth]{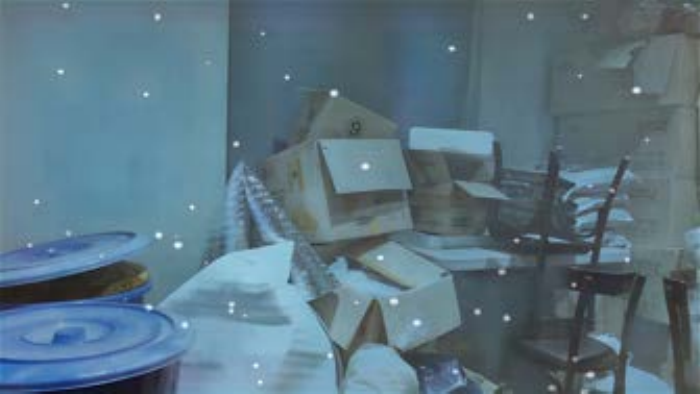}}
    \subfloat{\includegraphics[width = 0.09\linewidth]{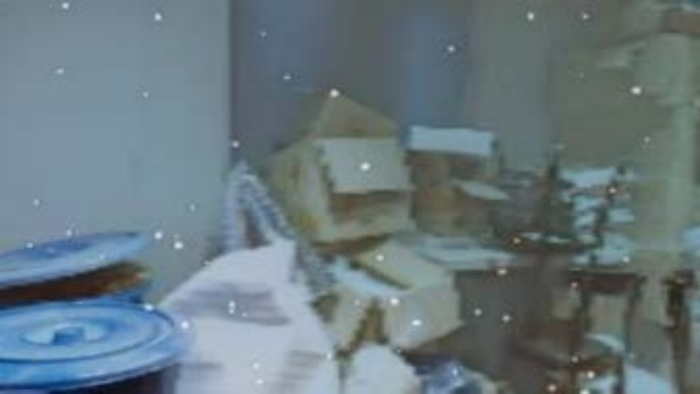}}
    \subfloat{\includegraphics[width = 0.09\linewidth]{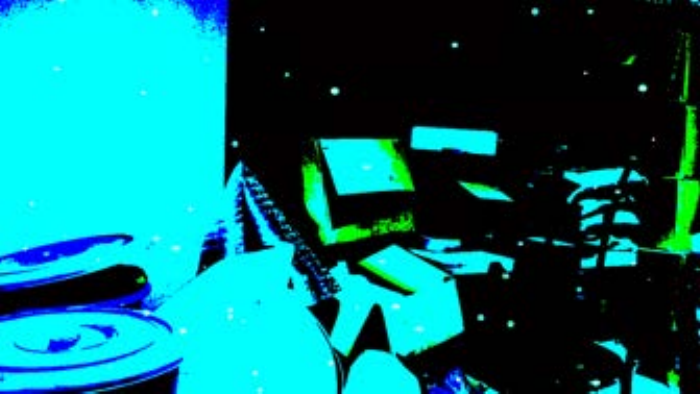}}\\
    % \subfloat{\includegraphics[width = 0.15\linewidth]{image/dataset/Waternet/waternet_209.pdf}} \\
     \vspace{-0.15in}%\caption*{Restoration results by Deep WN.}%\vspace{-0.1in}
    % \subfloat{\includegraphics[width = 0.15\linewidth]{image/dataset/Waternet/waternet_205.pdf}} 
    % \subfloat{\includegraphics[width = 0.15\linewidth]{image/dataset/Waternet/waternet_206.pdf}} 
    % \subfloat{\includegraphics[width = 0.15\linewidth]{image/dataset/Waternet/waternet_207.pdf}}
    \subfloat{\includegraphics[width = 0.09\linewidth]{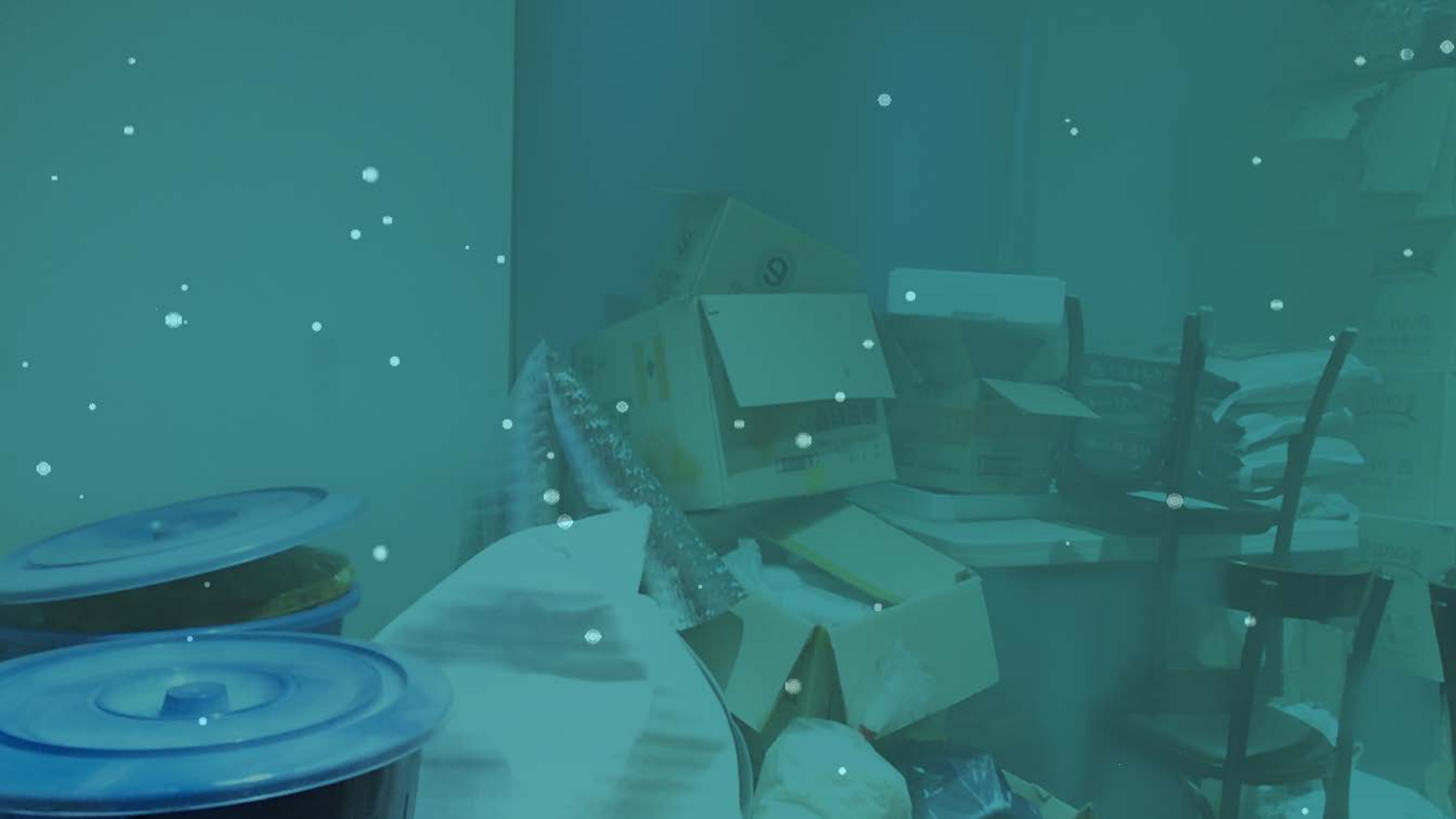}}
    \subfloat{\includegraphics[width = 0.09\linewidth]{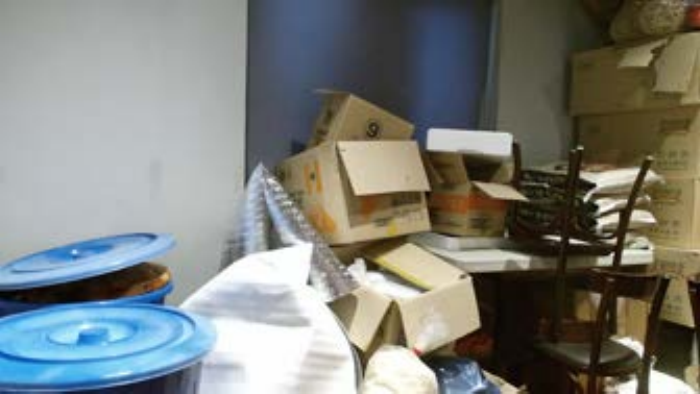}}
    \subfloat{\includegraphics[width = 0.09\linewidth]{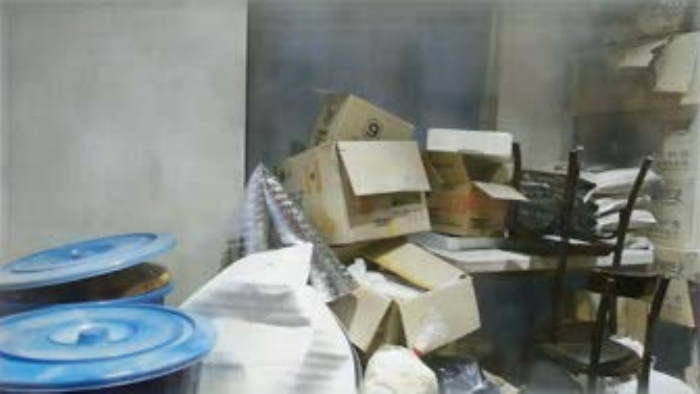}}
    \subfloat{\includegraphics[width = 0.09\linewidth]{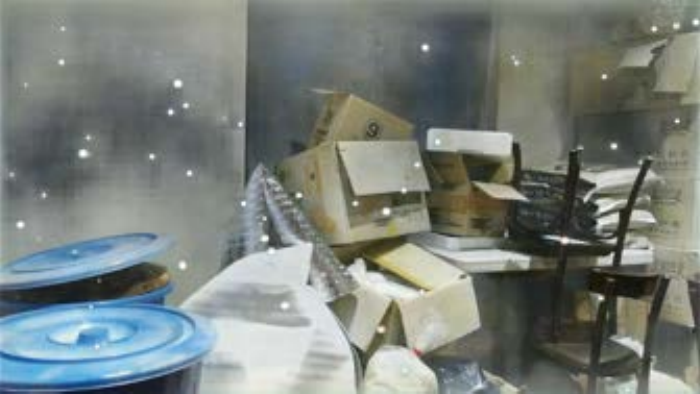}}
    \subfloat{\includegraphics[width = 0.09\linewidth]{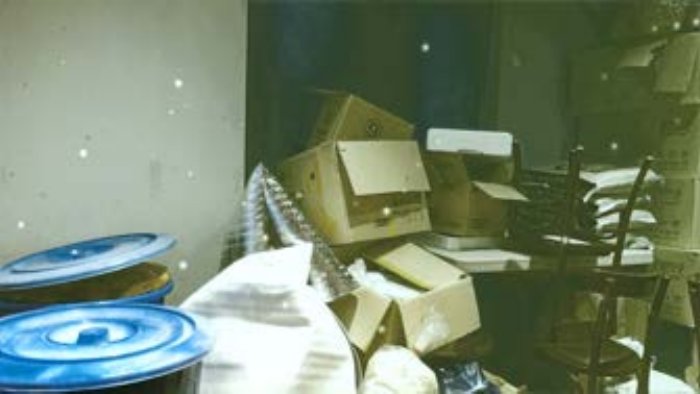}} 
    \subfloat{\includegraphics[width = 0.09\linewidth]{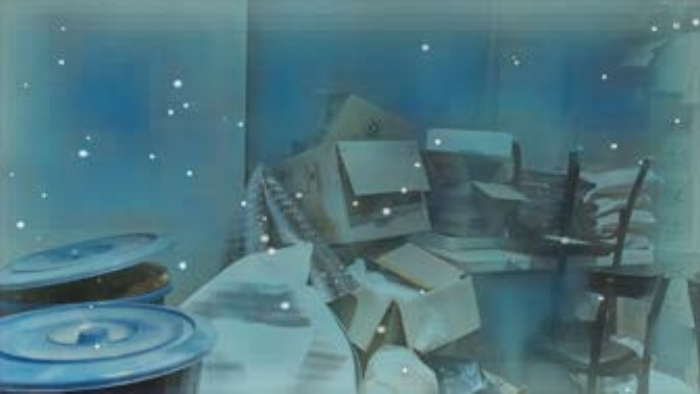}}
    \subfloat{\includegraphics[width = 0.09\linewidth]{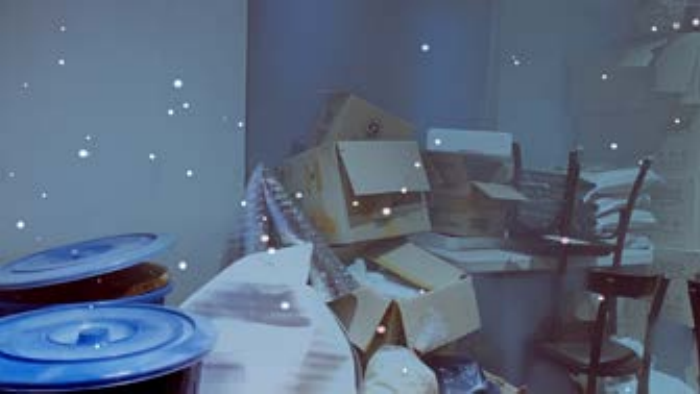}}
    \subfloat{\includegraphics[width = 0.09\linewidth]{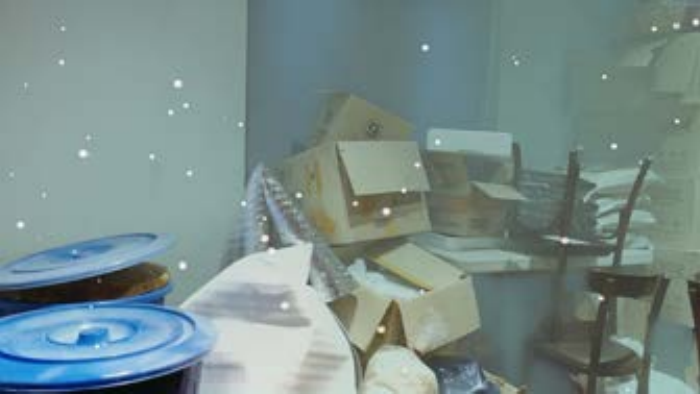}}
    \subfloat{\includegraphics[width = 0.09\linewidth]{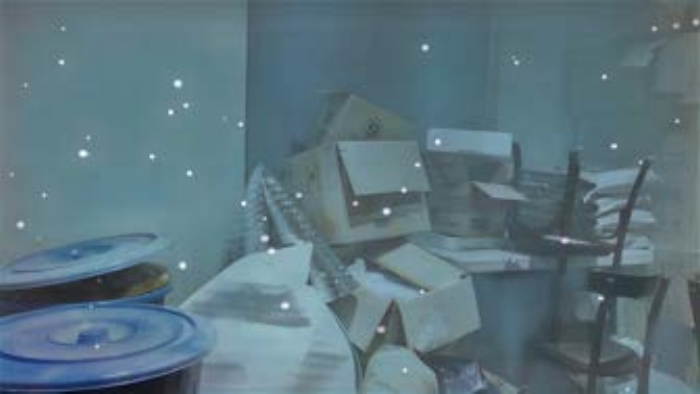}}
    \subfloat{\includegraphics[width = 0.09\linewidth]{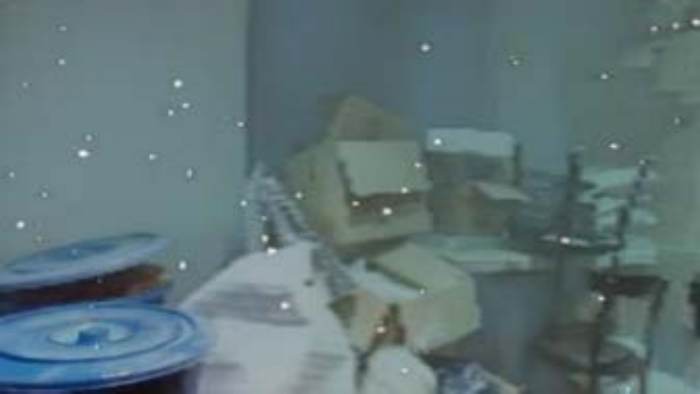}}
    \subfloat{\includegraphics[width = 0.09\linewidth]{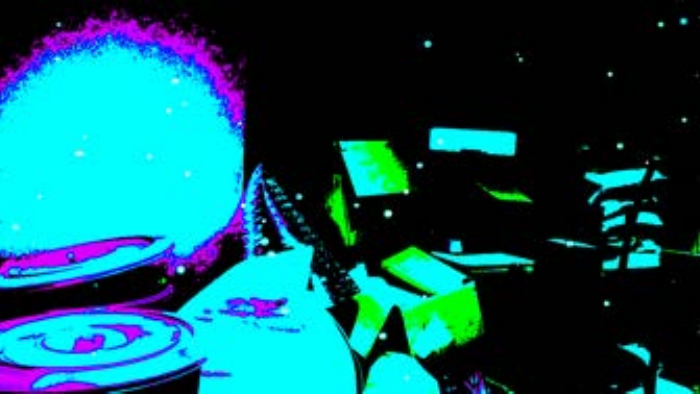}}\\
    \vspace{-0.15in}
    \subfloat{\includegraphics[width = 0.09\linewidth]{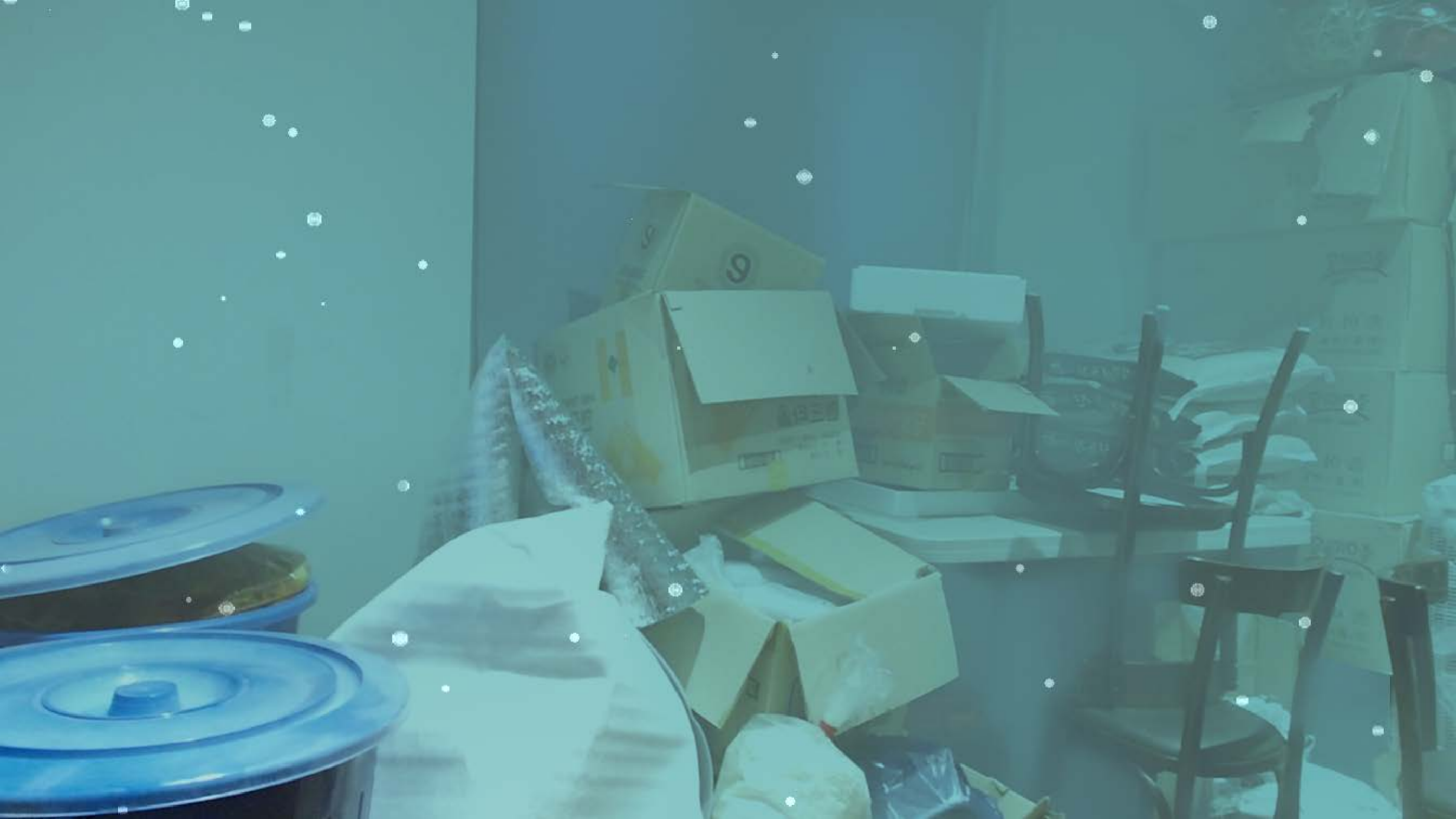}} 
    \subfloat{\includegraphics[width = 0.09\linewidth]{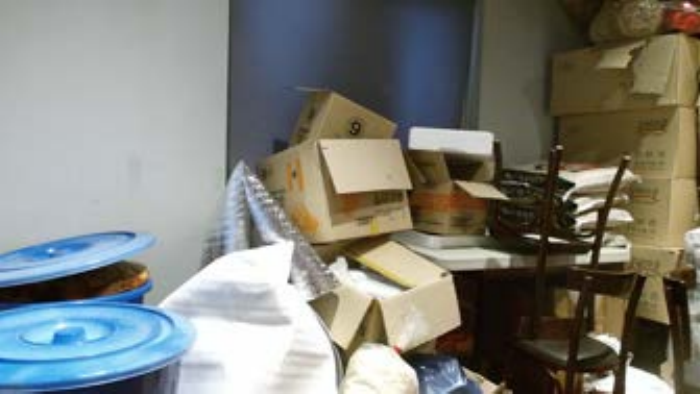}}
    \subfloat{\includegraphics[width = 0.09\linewidth]{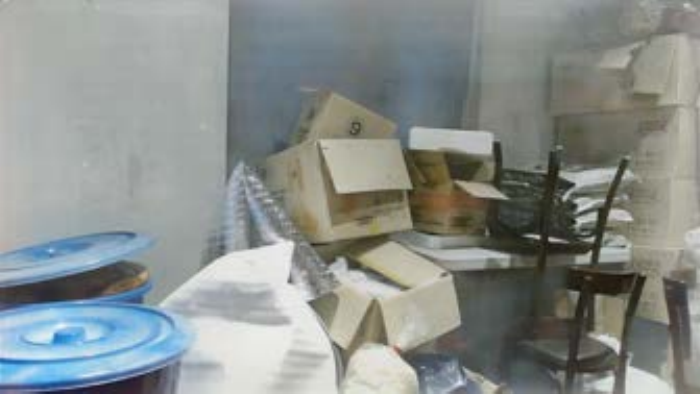}}
    \subfloat{\includegraphics[width = 0.09\linewidth]{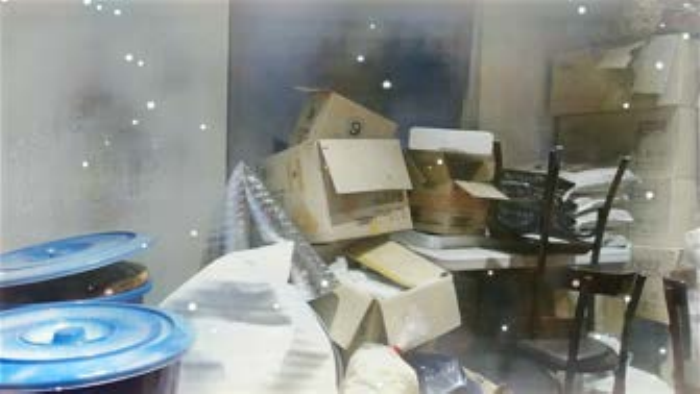}}
    \subfloat{\includegraphics[width = 0.09\linewidth]{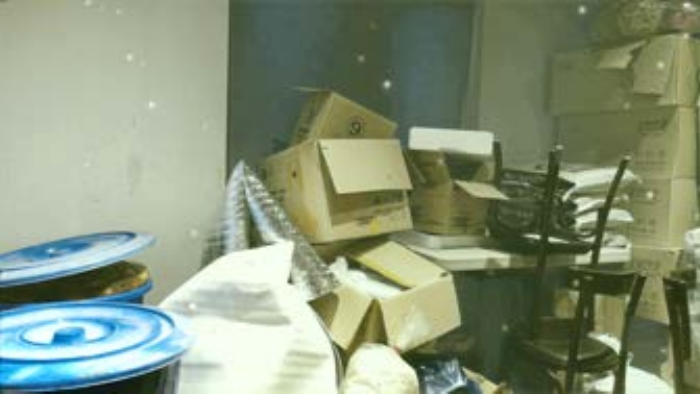}} 
    \subfloat{\includegraphics[width = 0.09\linewidth]{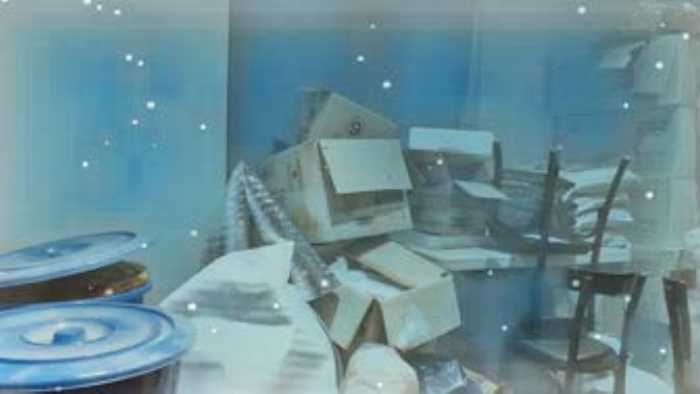}}
    \subfloat{\includegraphics[width = 0.09\linewidth]{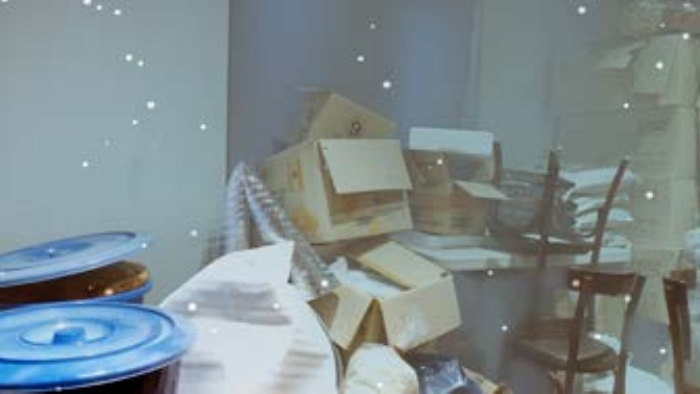}}
    \subfloat{\includegraphics[width = 0.09\linewidth]{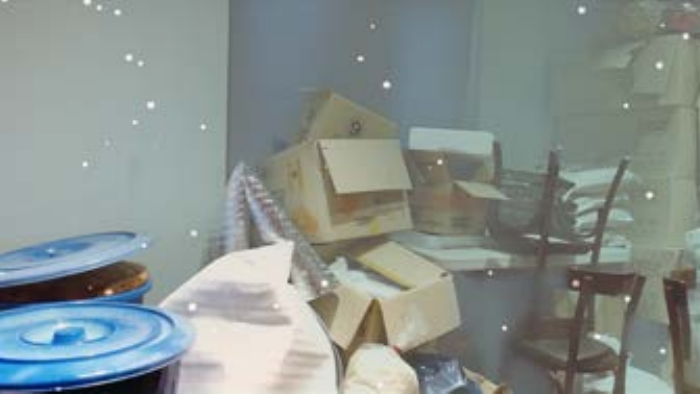}}
    \subfloat{\includegraphics[width = 0.09\linewidth]{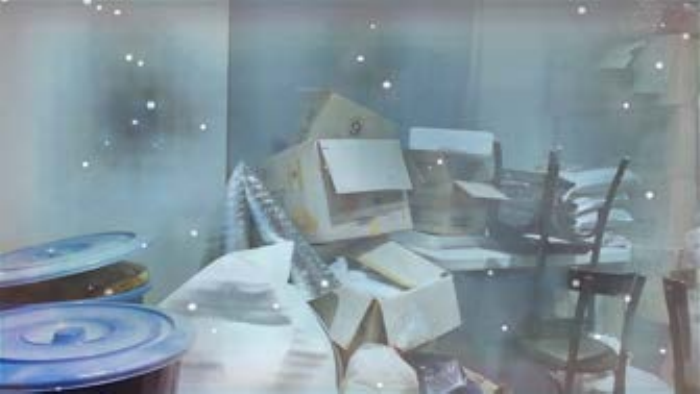}}
    \subfloat{\includegraphics[width = 0.09\linewidth]{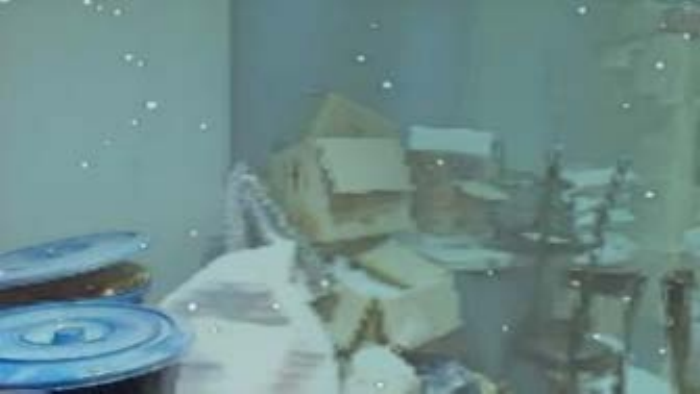}}
    \subfloat{\includegraphics[width = 0.09\linewidth]{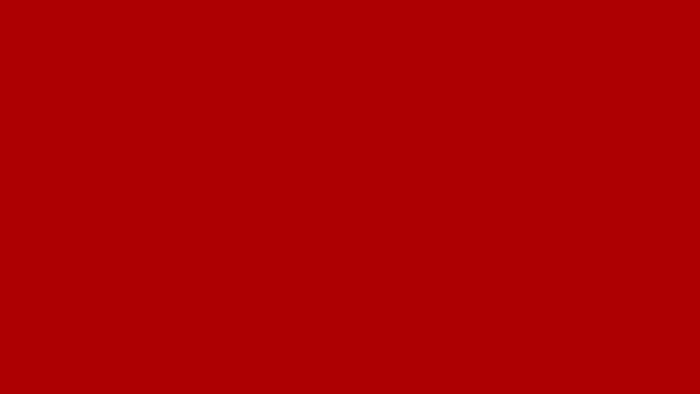}}\\
    \vspace{-0.15in}
    \subfloat{\includegraphics[width = 0.09\linewidth]{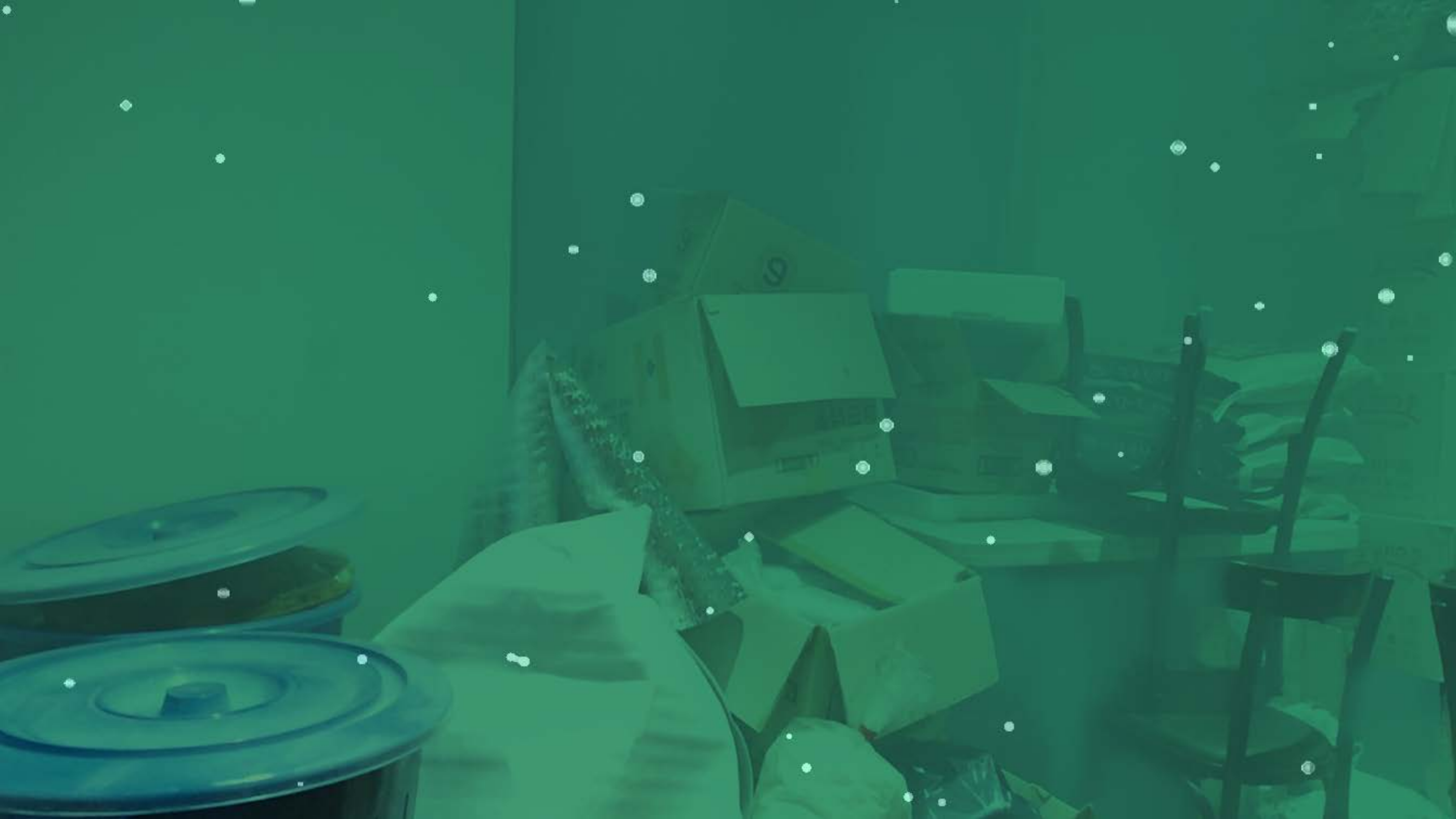}} 
    \subfloat{\includegraphics[width = 0.09\linewidth]{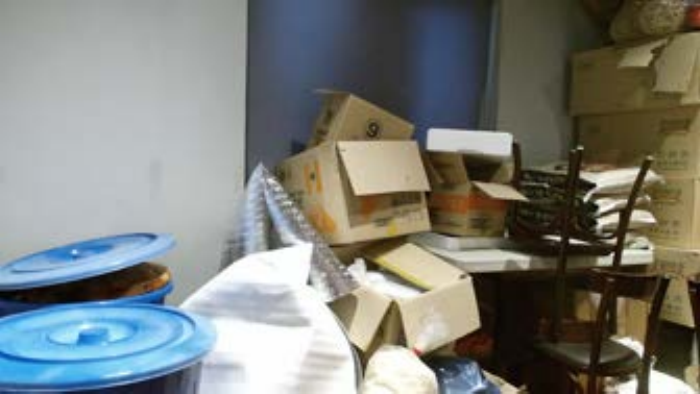}}
    \subfloat{\includegraphics[width = 0.09\linewidth]{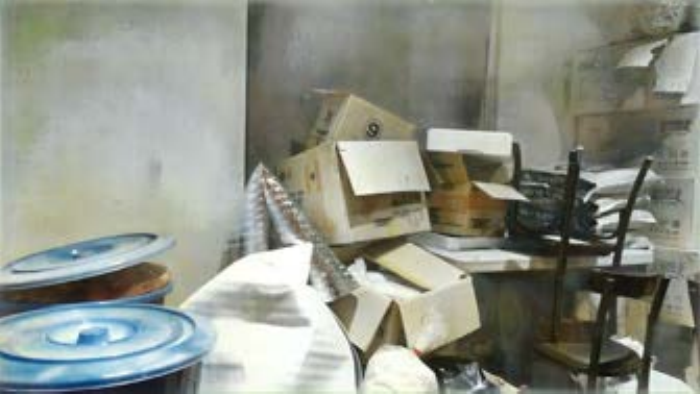}}
    \subfloat{\includegraphics[width = 0.09\linewidth]{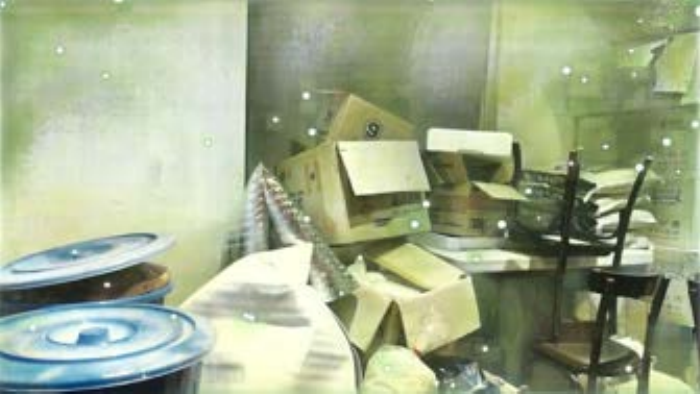}}
    \subfloat{\includegraphics[width = 0.09\linewidth]{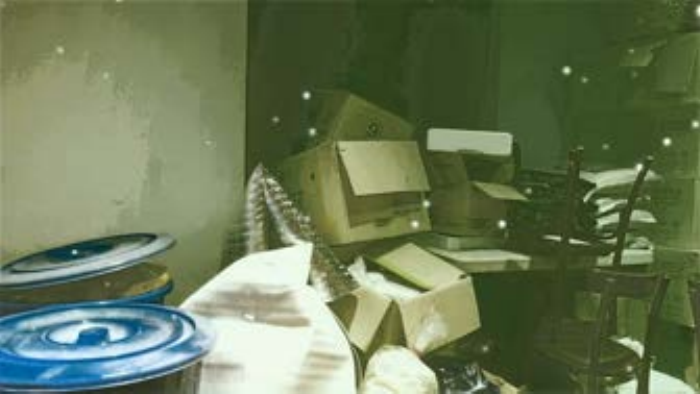}} 
    \subfloat{\includegraphics[width = 0.09\linewidth]{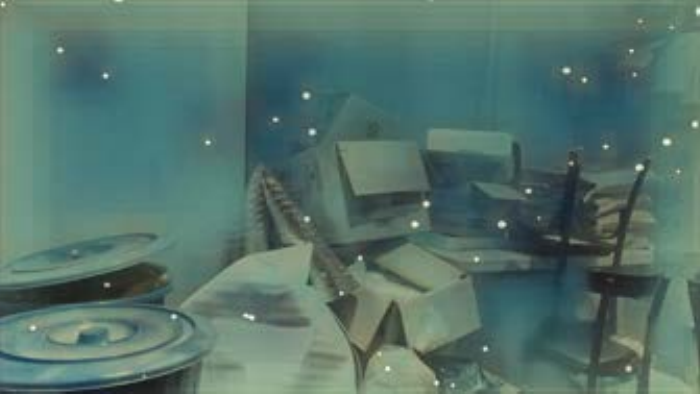}}
    \subfloat{\includegraphics[width = 0.09\linewidth]{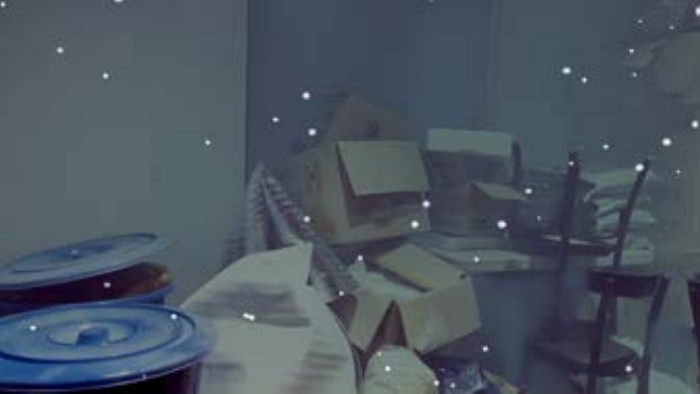}}
    \subfloat{\includegraphics[width = 0.09\linewidth]{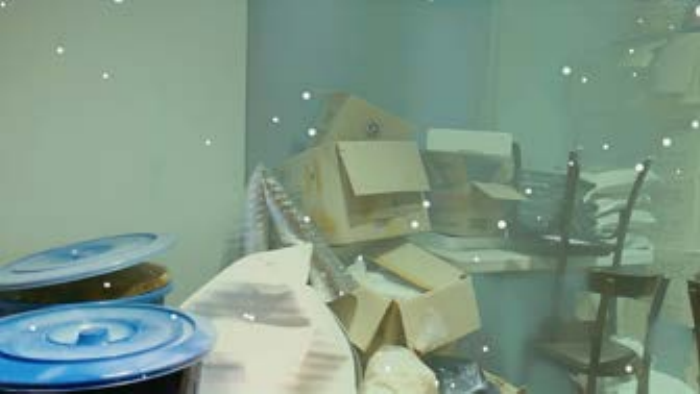}}
    \subfloat{\includegraphics[width = 0.09\linewidth]{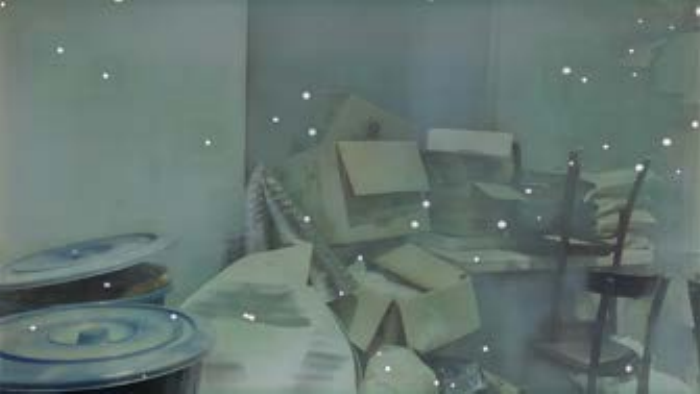}}
    \subfloat{\includegraphics[width = 0.09\linewidth]{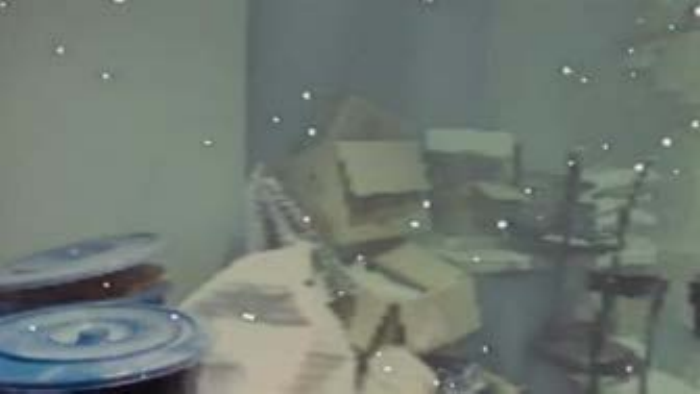}}
    \subfloat{\includegraphics[width = 0.09\linewidth]{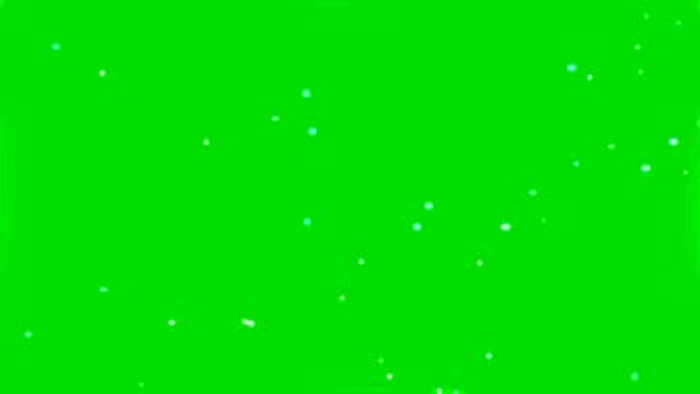}}\\
    \vspace{-0.15in}%\caption*{Restoration results by WaterNet.}
    \setcounter{subfigure}{0}
    \subfloat[Test images]{\includegraphics[width = 0.09\linewidth]{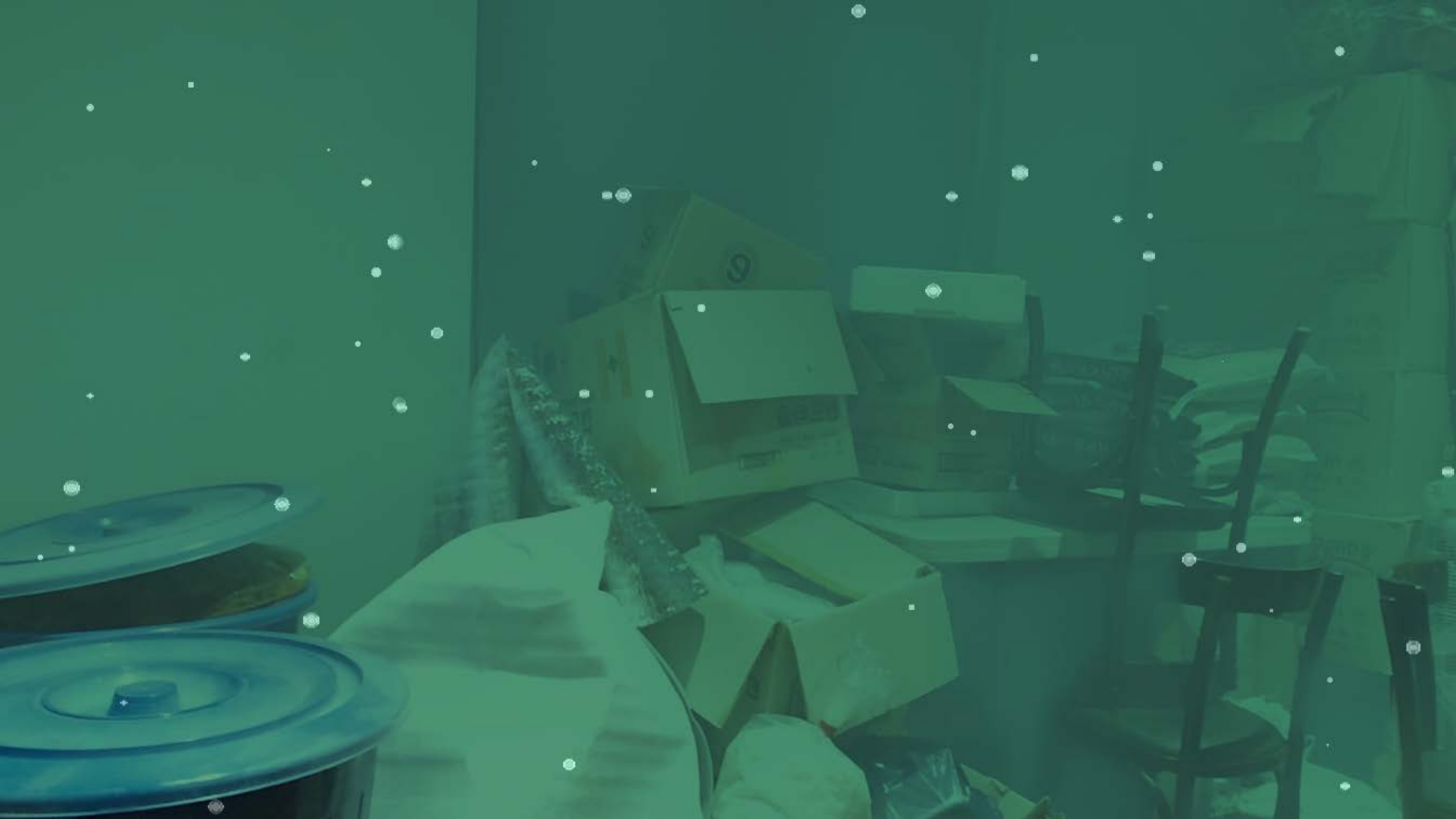}}
    \subfloat[Ground-truth images]{\includegraphics[width = 0.09\linewidth]{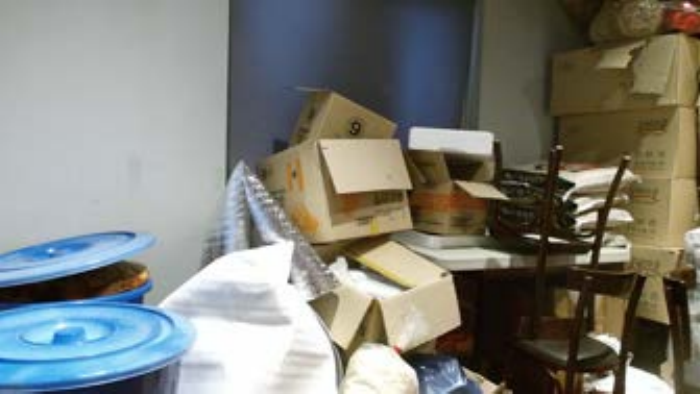}}
    \subfloat[Trans(P)]{\includegraphics[width = 0.09\linewidth]{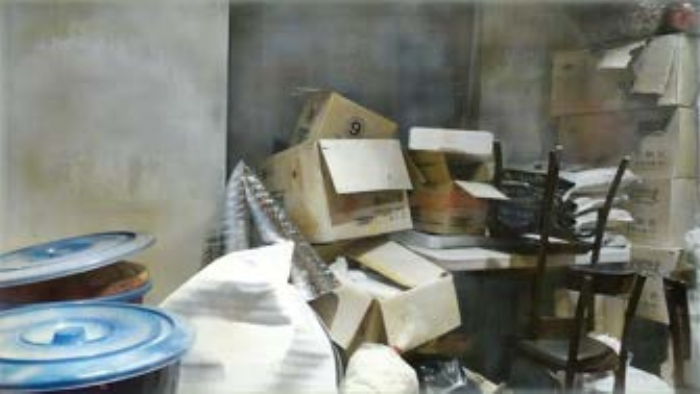}}
    \subfloat[Trans(P')]{\includegraphics[width = 0.09\linewidth]{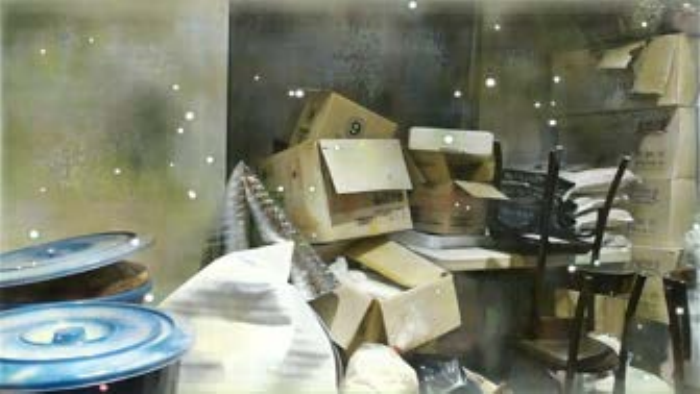}}
    \subfloat[DWN(P)]{\includegraphics[width = 0.09\linewidth]{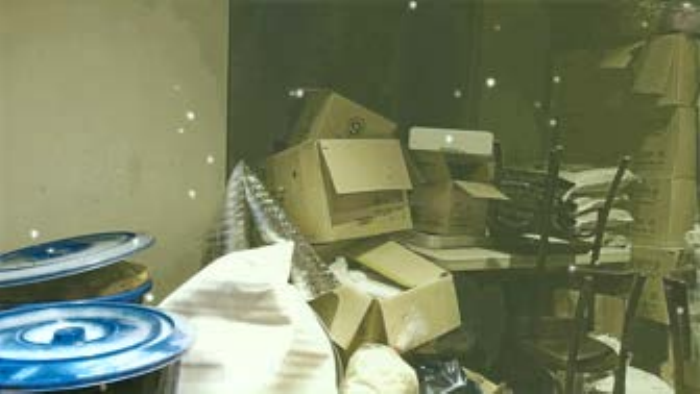}} 
    \subfloat[Trans(U)]{\includegraphics[width = 0.09\linewidth]{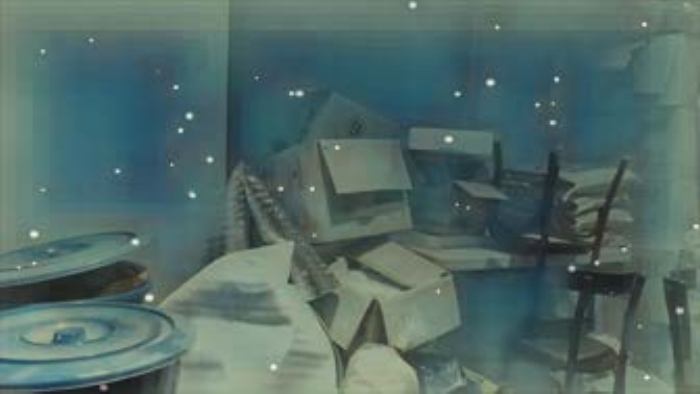}}
     \subfloat[DWN(U)]{\includegraphics[width = 0.09\linewidth]{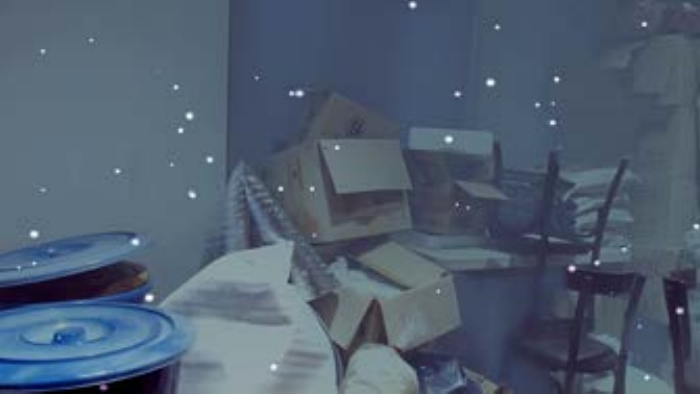}} 
     \subfloat[WN]{\includegraphics[width = 0.09\linewidth]{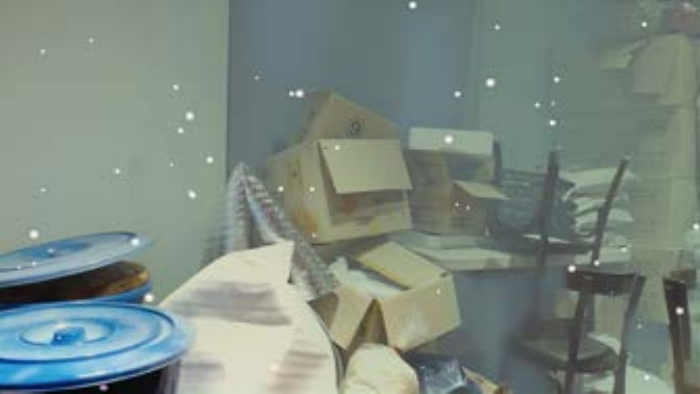}}
    \subfloat[Trans(L)]{\includegraphics[width = 0.09\linewidth]{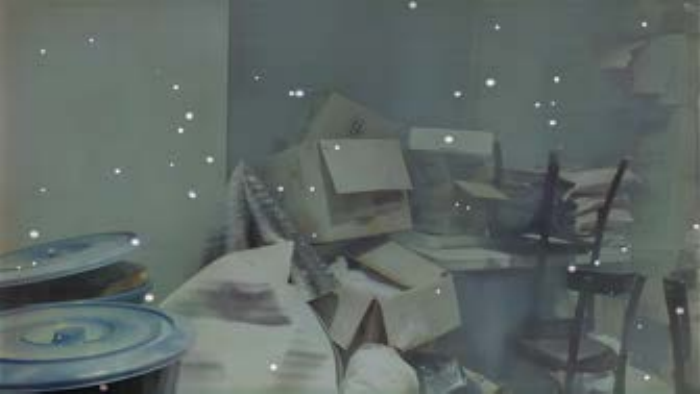}}
     \subfloat[U-shape]{\includegraphics[width = 0.09\linewidth]{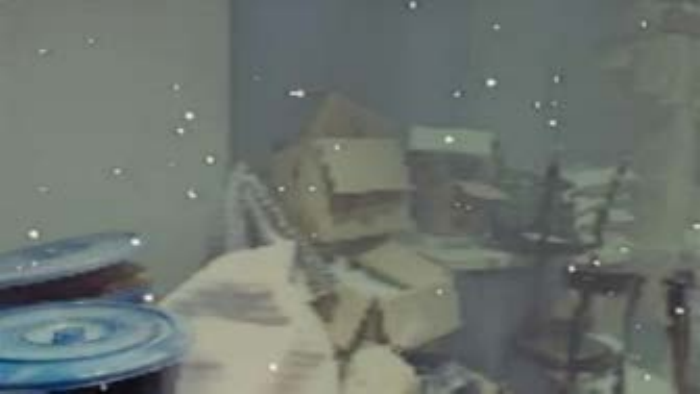}}
     \subfloat[HLRP]{\includegraphics[width = 0.09\linewidth]{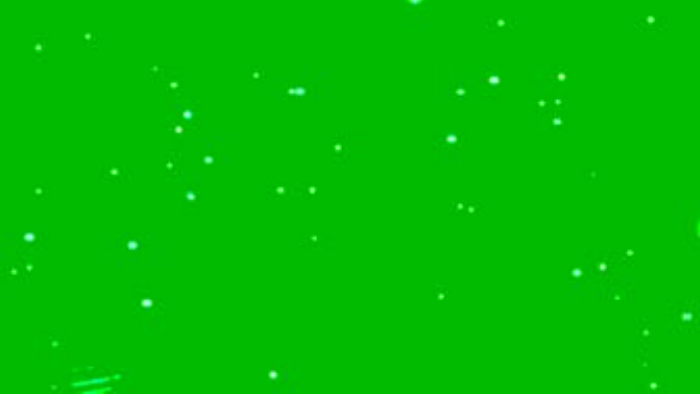}}\\
    \caption{Image enhancement results with PHISWID test images. From up to down, we use different water types}
    % We use super-resolution to visualize clearly \cite{Wang_2021_ICCV}.}
    %From left to right: Synthesized test images, restoration results by Transformer-P, restoration results by Transformer-C, restoration results by Transformer-U, restoration results by Transformer-L, restoration results by Deep WN-U, restoration results by Deep WN-P, restoration results by WaterNet, restoration results by U-shape, and restoration results by HLRP.}
  \label{dataset2result}
\end{figure*}
\section{Benchmarking Results on Synthesized Images}
\label{sec:benchmark}
In this section, 
%we present a benchmarking results with PHISWID.
we compare three datasets (PHISWID, UIEB, and LSUI) using the same module to demonstrate the contribution of our dataset to underwater image enhancement. In addition, we compare other modules using different datasets. %The results indicate that the dataset plays the most critical role in underwater image enhancement.
\subsection{Setup}
\label{subsec:generalUWIE}
\begin{table}[t]
    \centering
    \caption{Comparison methods.}
    \begin{tabular}{c|c|c}\hline
        Abbreviation & Baseline module & Training dataset\\\hline\hline
        Trans(P) & Transformer & PHISWID\\
        Trans(P') & Transformer & PHISWID w/o marine snow\\
        DWN(P) & Deep WaveNet & PHISWID\\\hline
        Trans(U) & Transformer & UIEB\\
        DWN(U) \cite{sharma2023wavelength} & Deep WaveNet & UIEB\\
        WN \cite{li2019underwater} & WaterNet & UIEB\\\hline
        Trans(L) & Transformer & LSUI\\
        U-shape \cite{pengUShapeTransformerUnderwater2023e} & Transformer & LSUI\\\hline
        % Transformer(P') \cite{ueda2019underwater} & Transformer & PHISWID w/o marine snow\\\hline
        % Transformer-c & Transformer & only color\\\hline
        HLRP \cite{zhuangUnderwaterImageEnhancement2022a} & \begin{tabular}[c]{@{}c@{}}Model-based\\enhancement\end{tabular} & --\\\hline
    \end{tabular}
    \label{tab:compare}
\end{table}
We divided PHISWID to 4125 training image pairs and 70 test image pairs.
% , all having a pixel resolution of 384 × 384. An image pair contains one original atmospheric image and one image containing synthesized color shift and marine snow artifacts.
Underwater image enhancement performance is compared with various methods summarized in Table \ref{tab:compare}.
% 1) Deep WN \cite{sharma2023wavelength}, 2) WaterNet \cite{li2019underwater}, 3) U-shape Transformer \cite{pengUShapeTransformerUnderwater2023e}, 4) Transformer \cite{wang_2022_CVPR}, and 5) HLRP \cite{zhuangUnderwaterImageEnhancement2022a}.
% \begin{enumerate}
%     \item Deep WN \cite{sharma2023wavelength}
%     \item WaterNet \cite{li2019underwater}
%     \item Transformer \cite{wang_2022_CVPR}
% \end{enumerate}
% 1) MF \cite{huang1979fast}, 2) adaptive MF \cite{banerjeeEliminationMarineSnow2014}, and 3) Transformer \cite{wang_2022_CVPR}.
% The kernel size of MFs is set to $3\times 3$ or $5 \times 5$ pixels.
% The results of dataset 2 is shown in Figs. \ref{dataset2result}.
% In this paper, we compared DeepWaveNet \cite{sharma2023wavelength}, and waternet \cite{li2019underwater}. Furthermore, we compared Transformer trained only color correction.
The parameters of the alternative methods are basically followed to the original researches.
%taken from the original authors' repository.
We use WaterNet (abbreviated as WN) \cite{li2019underwater}, Deep WaveNet (abbreviated as DWN) \cite{sharma2023wavelength}, U-shepe Transformer (abbreviated as U-shape) \cite{pengUShapeTransformerUnderwater2023e} and HLRP \cite{zhuangUnderwaterImageEnhancement2022a}, which are widely used for underwater image enhancement and restoration.
WN, DWN(U), and U-shape are trained with their own datasets.

To compare dataset qualities, we trained DWN with the proposed PHISWID.
Furthermore, we train Transformers with PHISWID (abbreviated as Trans(P)), UIEB (abbreviated as Trans(U)) \cite{li2019underwater}, and LSUI (abbreviated as Trans(L)) \cite{pengUShapeTransformerUnderwater2023e}. The structure of the Transformer is shown in Appendix \ref{app:unet}. 
% We also train transformer \cite{wang_2022_CVPR} with PHISWID, UIEB \cite{li2019underwater}, and LSUI \cite{pengUShapeTransformerUnderwater2023e}.
%We use the currently available well-known datasets (UIEB \cite{li2019underwater} and LSUI \cite{pengUShapeTransformerUnderwater2023e} as well as PHISWID) with the same setup as Section \ref{sec:dataset}.
%This is to compare the significance between the enhancement models and the datasets.
For a comparison purpose, we also train the same Transformer with the dataset \textit{without} marine snow artifacts, i.e., the dataset only applied our color shift model (abbreviated as Trans(P')).
% called Physics-Inspired Synthesized Color shift Image Dataset (PHISCID).%proposed in \cite{ueda2019underwater}.
% We abbreviated this as 6) Transformer-c.
% We also train a Transformer with UIEB dataset \cite{li2019underwater} or LSUI dataset \cite{pengUShapeTransformerUnderwater2023e} for a comparison purpose. We abbreviated these as 7) Transformer-U and 8) Transformer-L.
% Furthermore, we trained a Deep WN with PHISWID for a comparison purpose too. We abbreviated this as 9) Deep WN-P.
% for removing marine snow artifacts.
%For simplicity, we use the network structure same as Transformer \cite{}.
% In this paper, we compared DeepWaveNet \cite{sharma2023wavelength}, and waternet \cite{li2019underwater} and proposed methods. In addition, we compared Transformer trained only color correction and others.
% As clearly observed, Deep WN and Water Net and Transformer trained only color correction cannot remove marine snow artifact. Only the proposed method removes marine snow artifact. In terms of color correction, the other methods appear to have a bluish tint compared to the proposed method.

For objective quality comparison, we compute the average PSNRs and SSIMs over the test datasets.
We also show naturalness image quality evaluator (NIQE) \cite{mittalMakingCompletelyBlind2013} as a non-reference quality metric.
NIQE fits our purpose of the paper because it returns a low score when images are \textit{natural taken in atmospheric scenes}.

\subsection{Results}
% In this paper, we compared DeepWaveNet \cite{sharma2023wavelength}, and waternet \cite{li2019underwater} and proposed methods. In addition, we compared Transformer trained only color correction and others.
The experimental results are shown in Fig. \ref{dataset2result} and the objective performances are summarized in Table \ref{dataset2_psnr}.
All of the metrics (PSNRs, SSIMs, and NIQEs) of the neural networks trained with PHISWID outperform those with the existing datasets.
Fig. \ref{dataset2result} visualizes such differences: The methods trained with PHISWID restore images like their atmospheric counterparts. On the other hand, marine snow artifacts remain and blueish images are still generated by methods trained with the alternative datasets. Trans(P') can remove bluish color shift, however, it cannot remove marine snow artifacts.
\begin{table}[t]
  \centering
  \caption{Average PSNRs and SSIMs over the test dataset of PHISWID.}
  \label{dataset2_psnr}
  \begin{tabular}{l|ccc} \hline
  %& \multicolumn{2}{c}{Dataset 2}\\\hline
  Method & PSNR$\uparrow$ & SSIM$\uparrow$ & NIQE$\downarrow$ \\ \hline\hline
  Trans(P) & 21.00 & \bf{0.793} & \bf{2.86}\\
  Trans(P') & 19.10 &0.748& 3.69\\
  DWN(P) & \bf{21.10} & 0.671& 3.11\\\hline
  Trans(U) & 14.69 & 0.286& 4.13\\
  DWN(U) & 16.60 & 0.359& 3.67\\
  WN & 17.06&0.359& 3.76\\\hline
  Trans(L) & 15.06 & 0.363& 4.05\\
  U-shape & 14.88 & 0.368& 3.11\\\hline
  HLRP & 7.10 & 0.062& NaN\\\hline
  Synthesized image& 13.67 & 0.120& 4.02\\\hline
  \end{tabular}
\end{table}
\begin{figure*}[t]
%\vspace{-0.2in}
\centering
    \subfloat{\includegraphics[width = 0.1\linewidth]{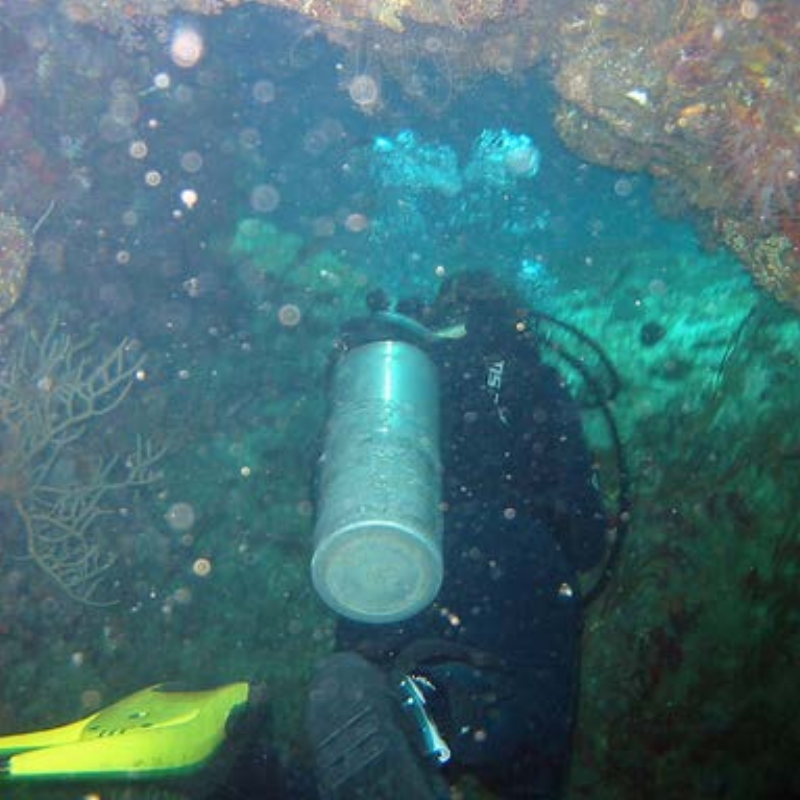}} 
    \subfloat{\includegraphics[width = 0.1\linewidth]{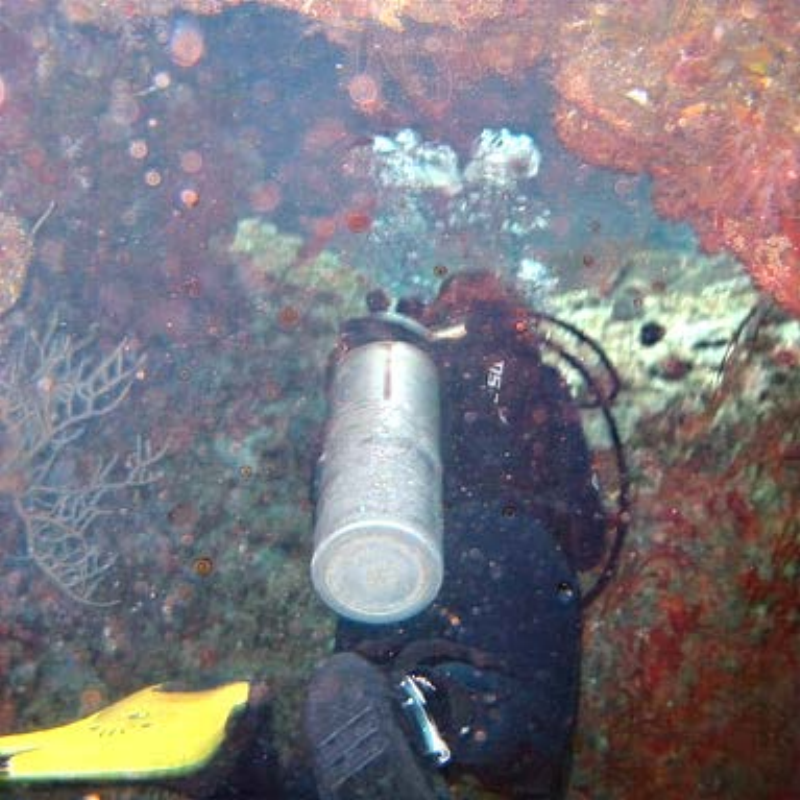}}
    \subfloat{\includegraphics[width = 0.1\linewidth]{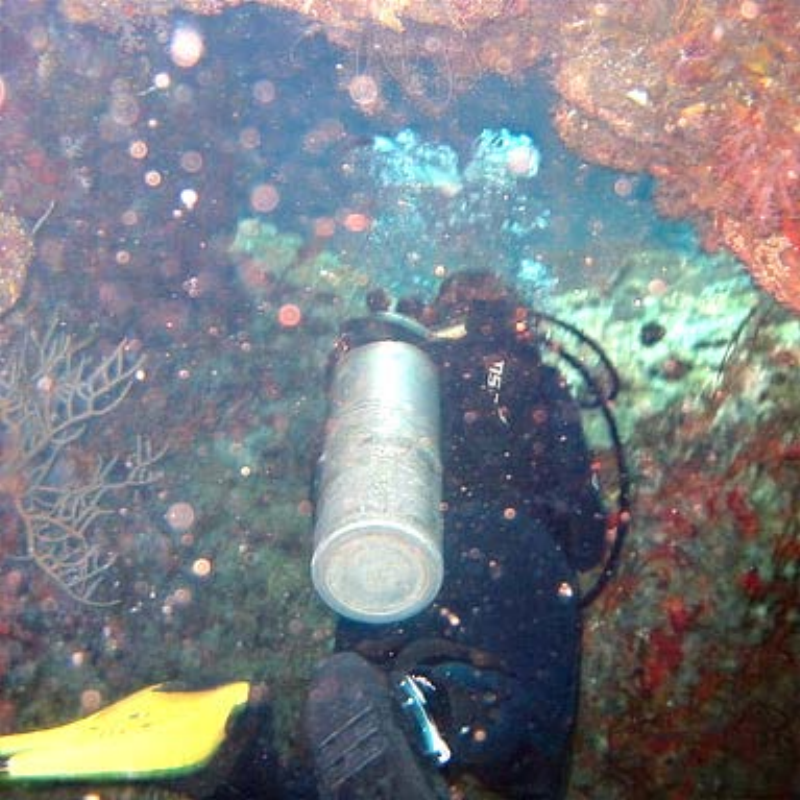}}
    \subfloat{\includegraphics[width = 0.1\linewidth]{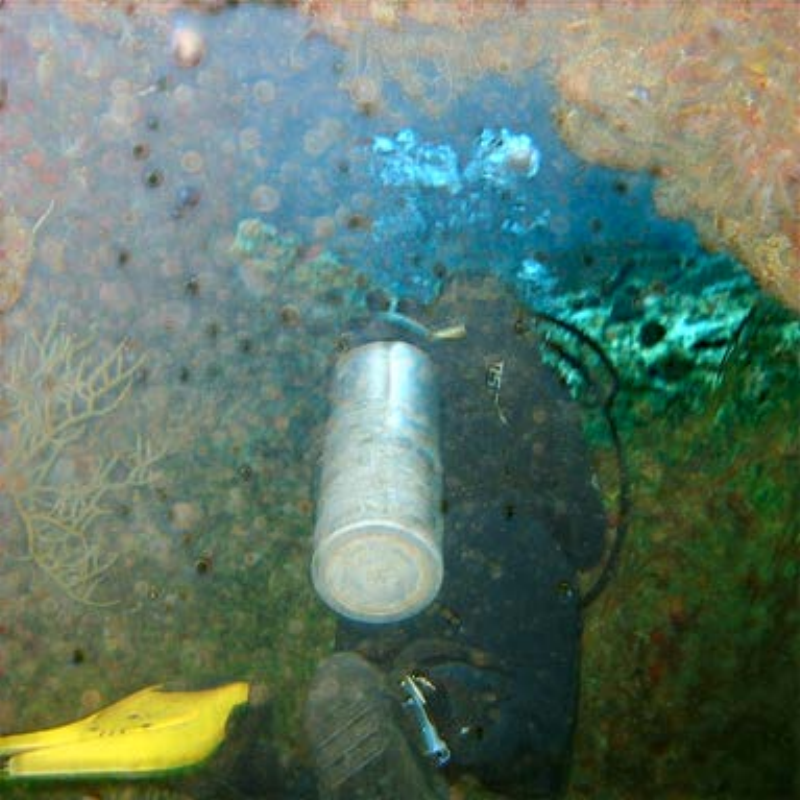}}
    \subfloat{\includegraphics[width = 0.1\linewidth]{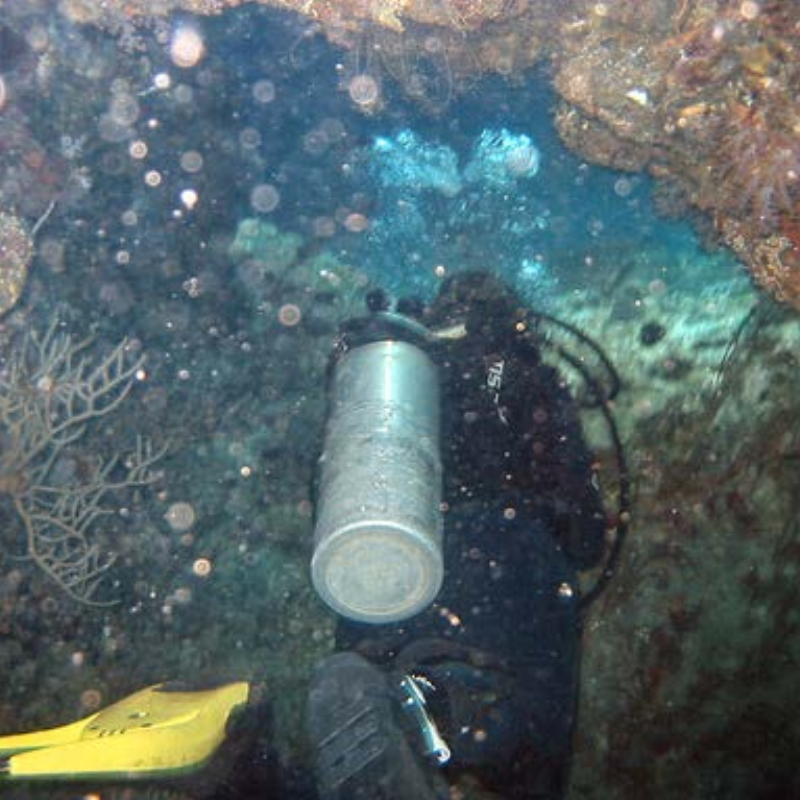}}
    \subfloat{\includegraphics[width = 0.1\linewidth]{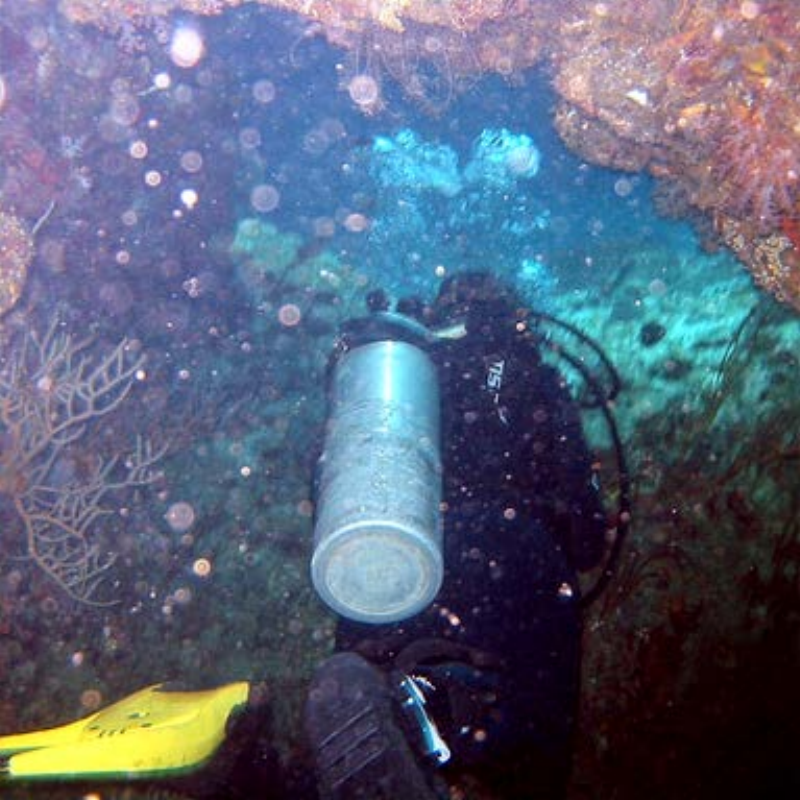}}
    \subfloat{\includegraphics[width = 0.1\linewidth]{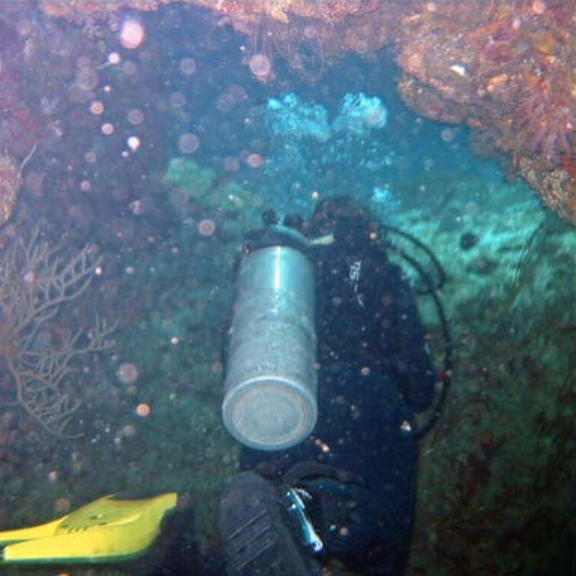}}
    \subfloat{\includegraphics[width = 0.1\linewidth]{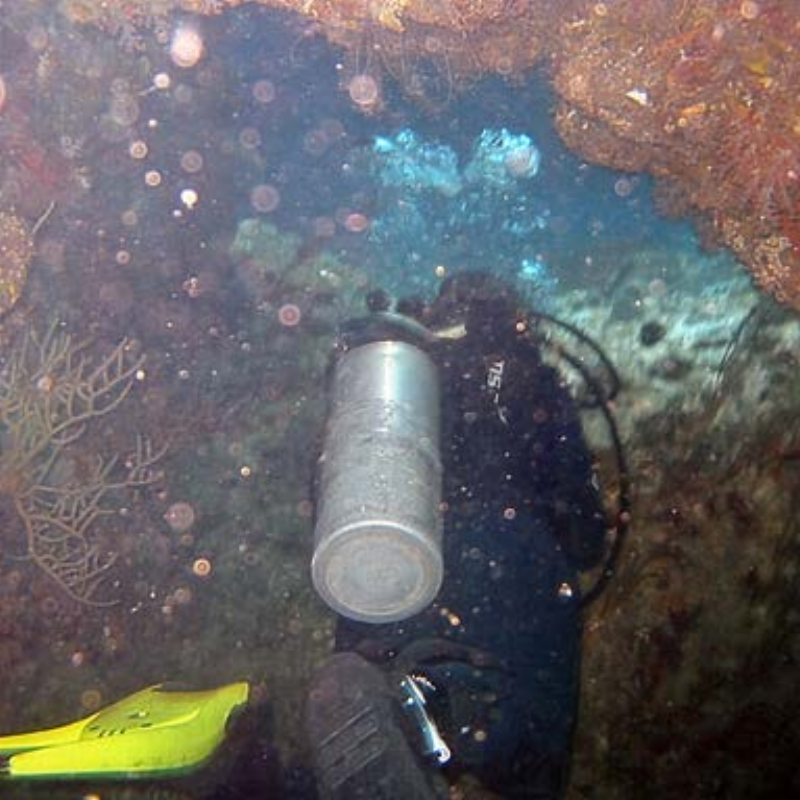}}
    \subfloat{\includegraphics[width = 0.1\linewidth]{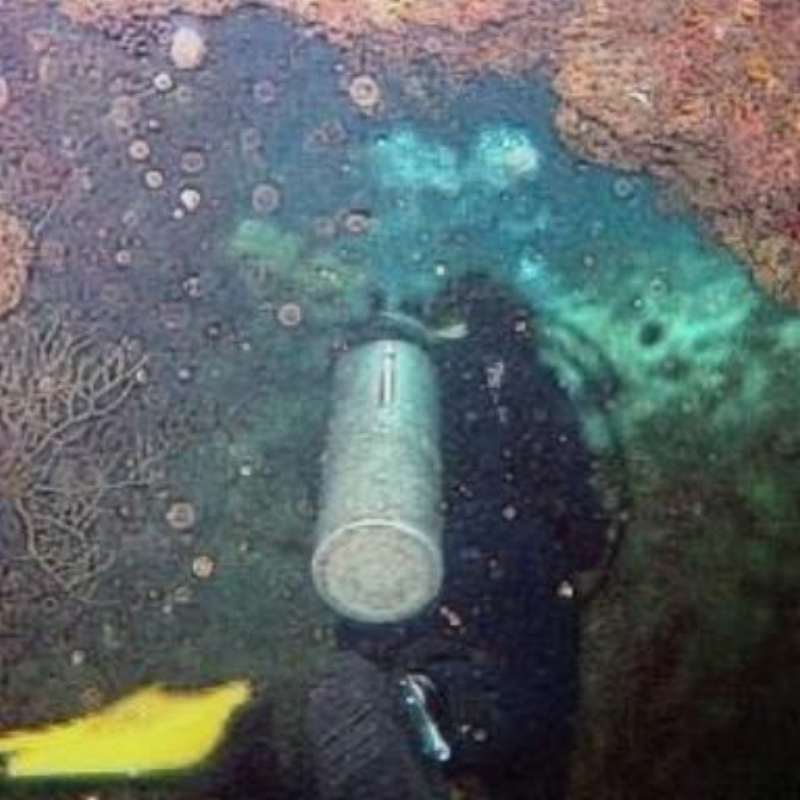}}
    \subfloat{\includegraphics[width = 0.1\linewidth]{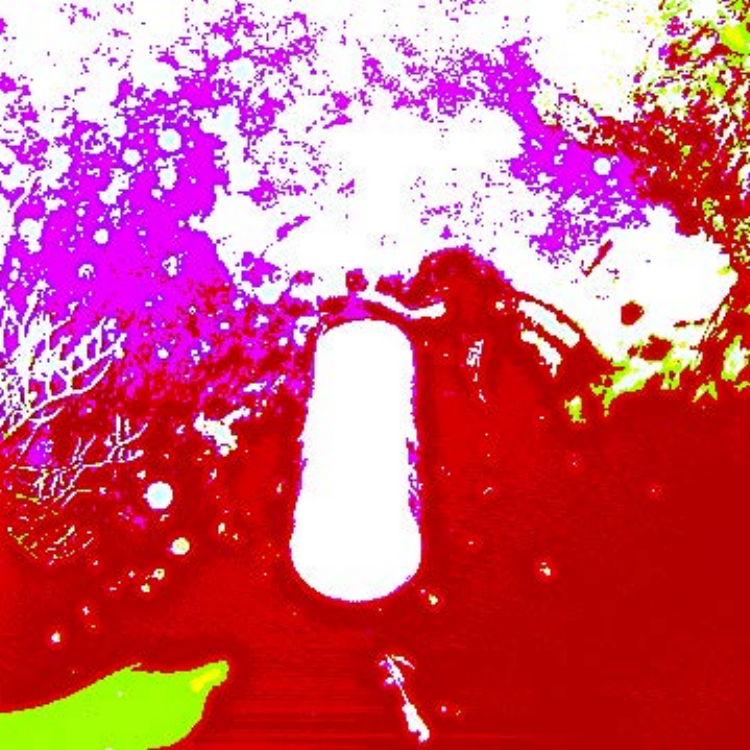}}\\
    % \subfloat{\includegraphics[width = 0.15\linewidth]{image/dataset/MSR_and_CC/206snow.pdf}} 
    % \subfloat{\includegraphics[width = 0.15\linewidth]{image/dataset/MSR_and_CC/207snow.pdf}}
    % \subfloat{\includegraphics[width = 0.15\linewidth]{image/dataset/MSR_and_CC/208snow.pdf}} 
    % \subfloat{\includegraphics[width = 0.15\linewidth]{image/dataset/MSR_and_CC/209snow.pdf}}\\
    \vspace{-0.14in}%\caption*{Synthesized test images.}%\vspace{-0.1in}
    % \subfloat{\includegraphics[width = 0.15\linewidth]{image/dataset/MSR_and_CC/PHISWID_room1.pdf}} 
    \subfloat{\includegraphics[width = 0.1\linewidth]{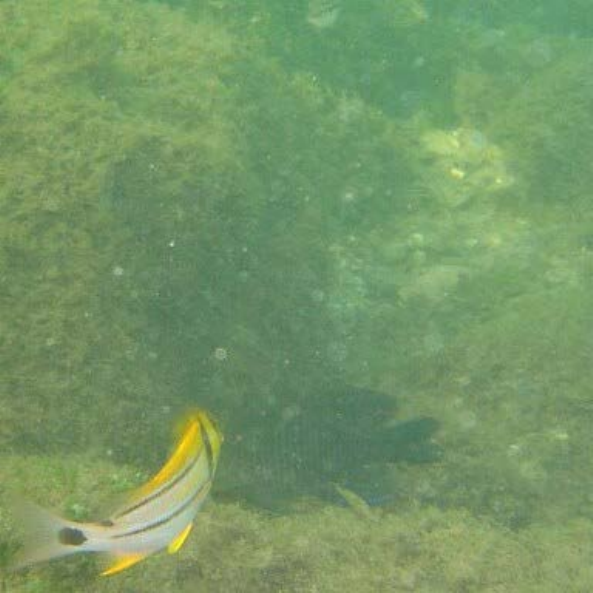}} 
    \subfloat{\includegraphics[width = 0.1\linewidth]{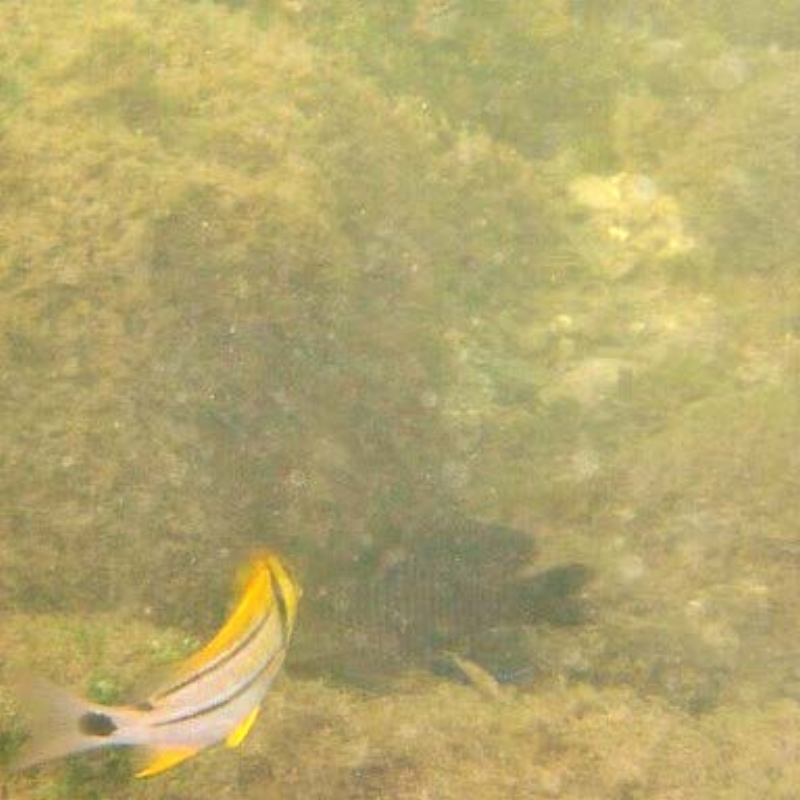}}
    \subfloat{\includegraphics[width = 0.1\linewidth]{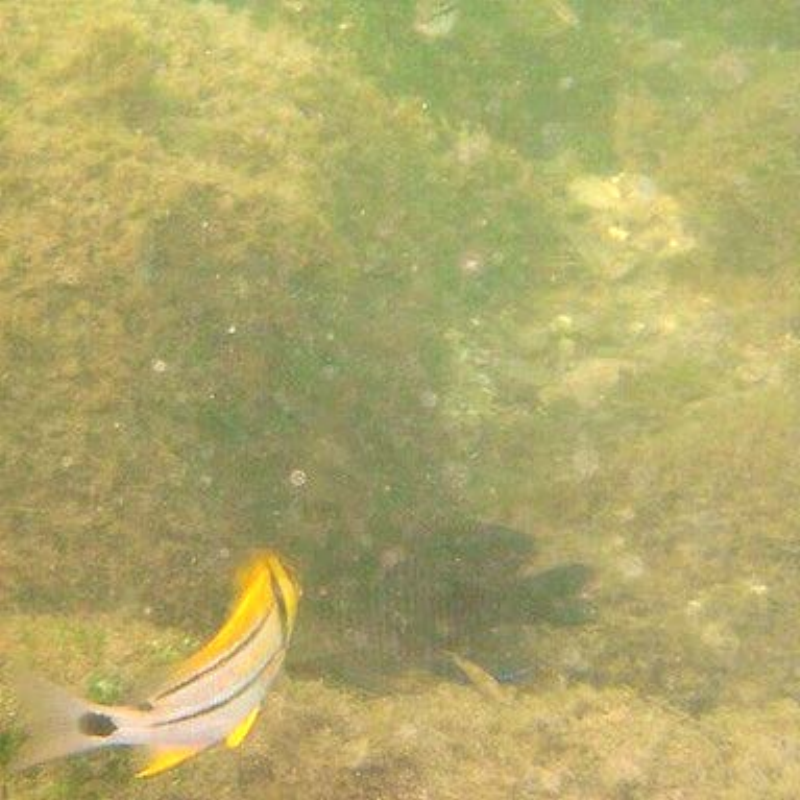}}
    \subfloat{\includegraphics[width = 0.1\linewidth]{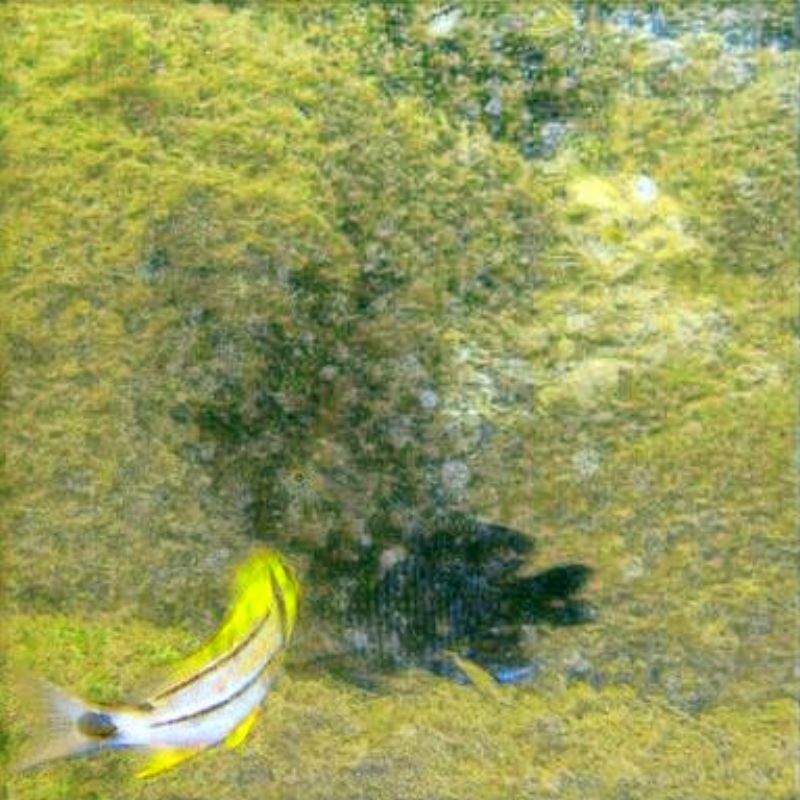}} 
    \subfloat{\includegraphics[width = 0.1\linewidth]{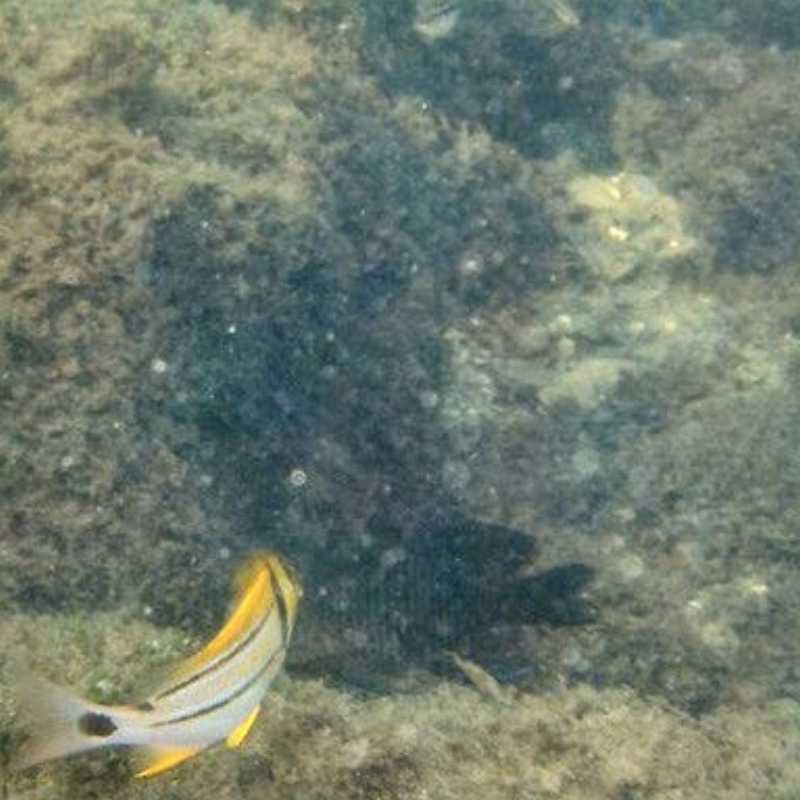}}
    \subfloat{\includegraphics[width = 0.1\linewidth]{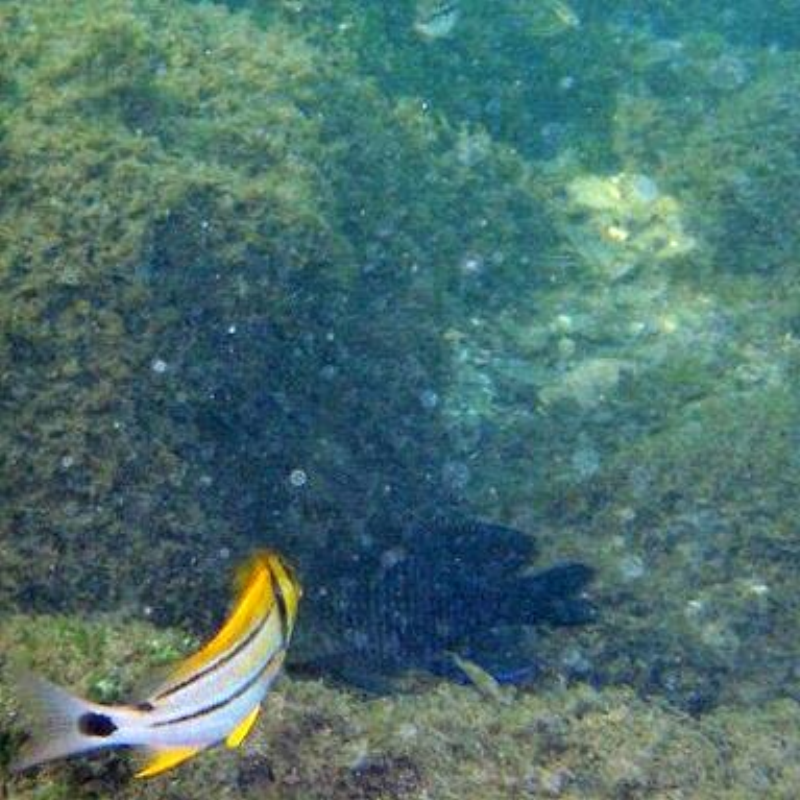}}
    \subfloat{\includegraphics[width = 0.1\linewidth]{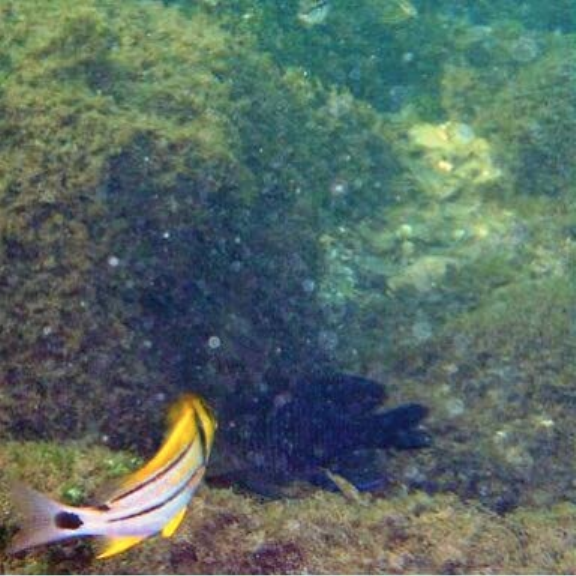}}
    \subfloat{\includegraphics[width = 0.1\linewidth]{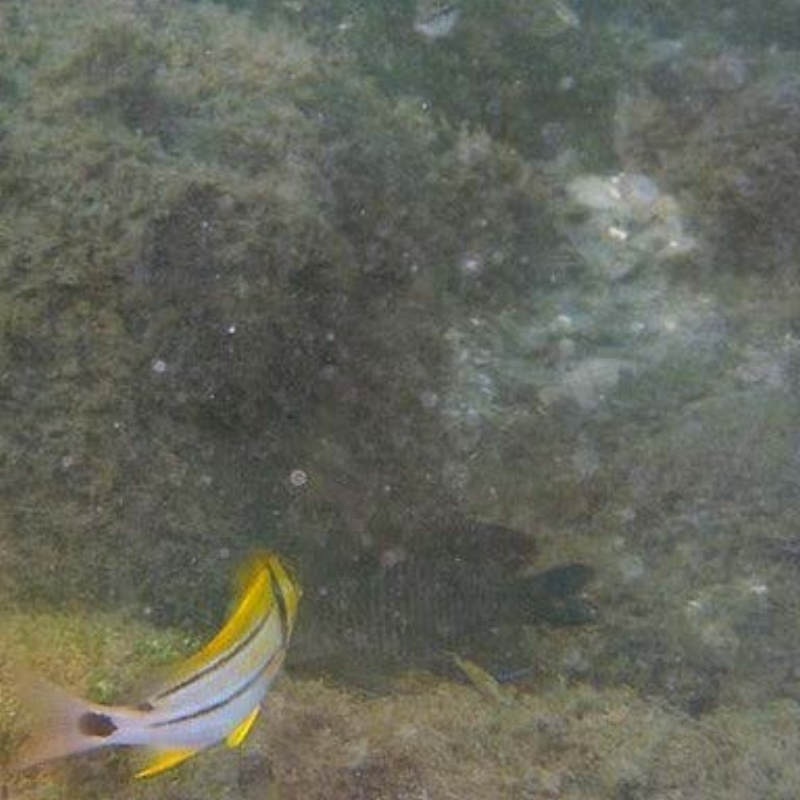}}
    \subfloat{\includegraphics[width = 0.1\linewidth]{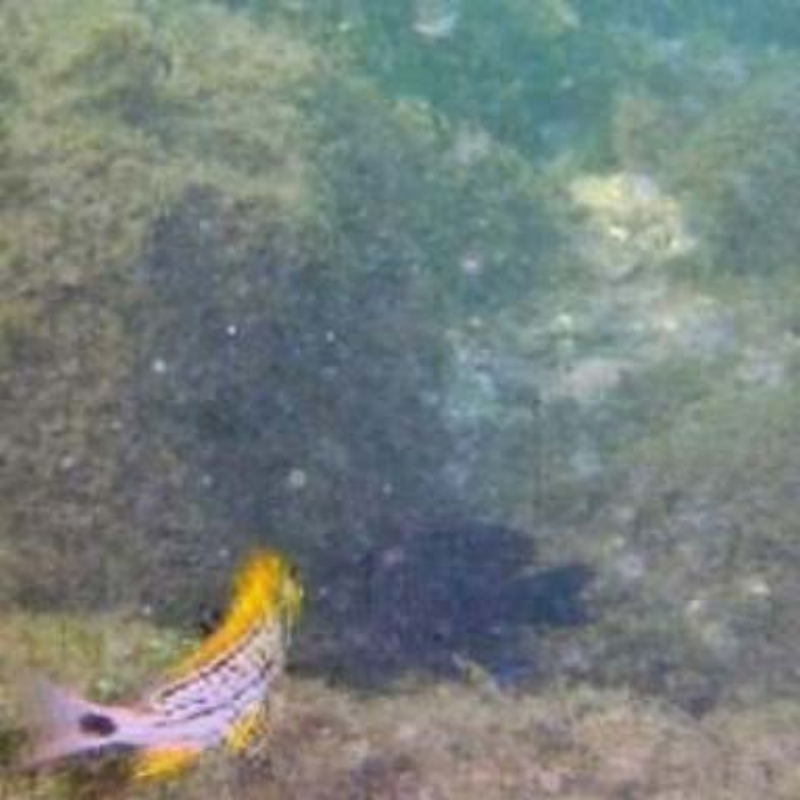}}
    \subfloat{\includegraphics[width = 0.1\linewidth]{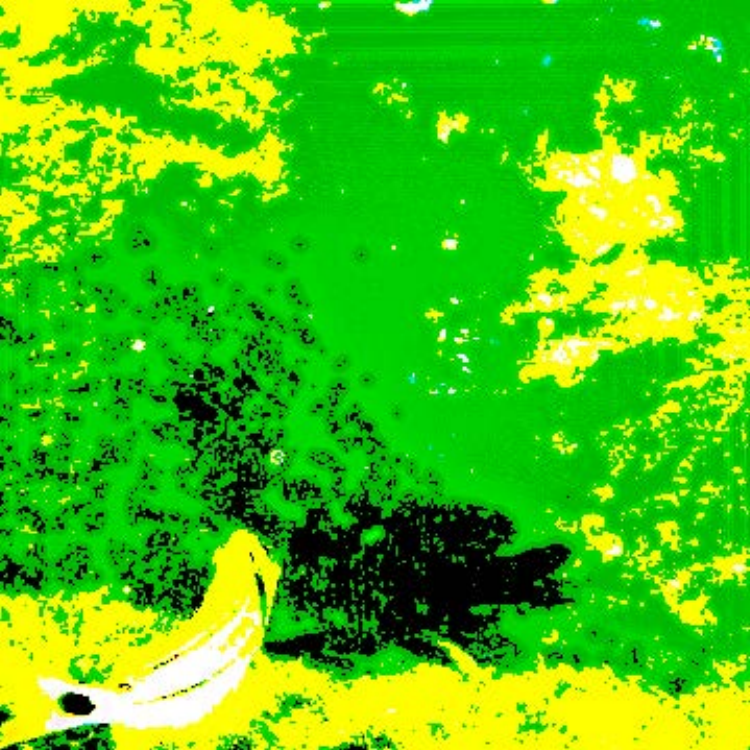}}\\
    % \subfloat{\includegraphics[width = 0.15\linewidth]{image/dataset/MSR_and_CC/PHISWID_room3.pdf}}
    % \subfloat{\includegraphics[width = 0.15\linewidth]{image/dataset/MSR_and_CC/PHISWID_room4.pdf}} 
    % \subfloat{\includegraphics[width = 0.15\linewidth]{image/dataset/MSR_and_CC/PHISWID_room5.pdf}}\\
    \vspace{-0.14in}%\caption*{Restoration results by Transformer.}%\vspace{-0.1in}
    % \subfloat{\includegraphics[width = 0.15\linewidth]{image/dataset/DeepWaveNet/Deepwavenet_205.pdf}} 
    % \subfloat{\includegraphics[width = 0.15\linewidth]{image/dataset/DeepWaveNet/Deepwavenet_206.pdf}} 
    % \subfloat{\includegraphics[width = 0.15\linewidth]{image/dataset/DeepWaveNet/Deepwavenet_207.pdf}}
    \subfloat{\includegraphics[width = 0.1\linewidth]{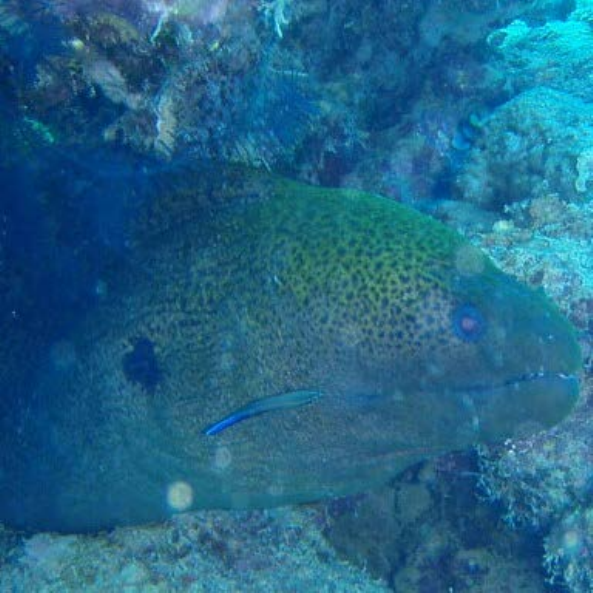}} 
    \subfloat{\includegraphics[width = 0.1\linewidth]{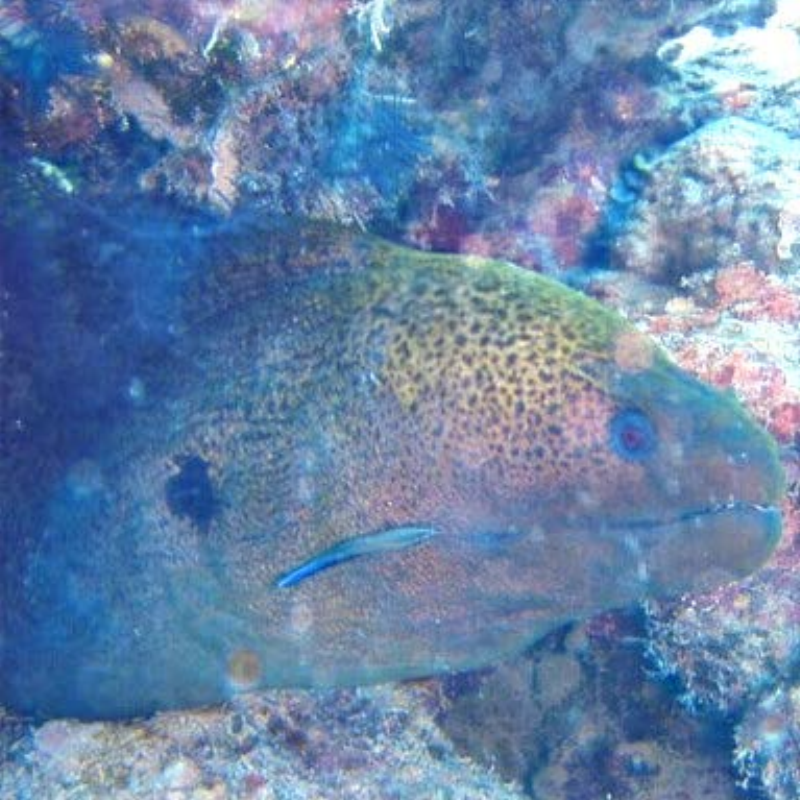}}
    \subfloat{\includegraphics[width = 0.1\linewidth]{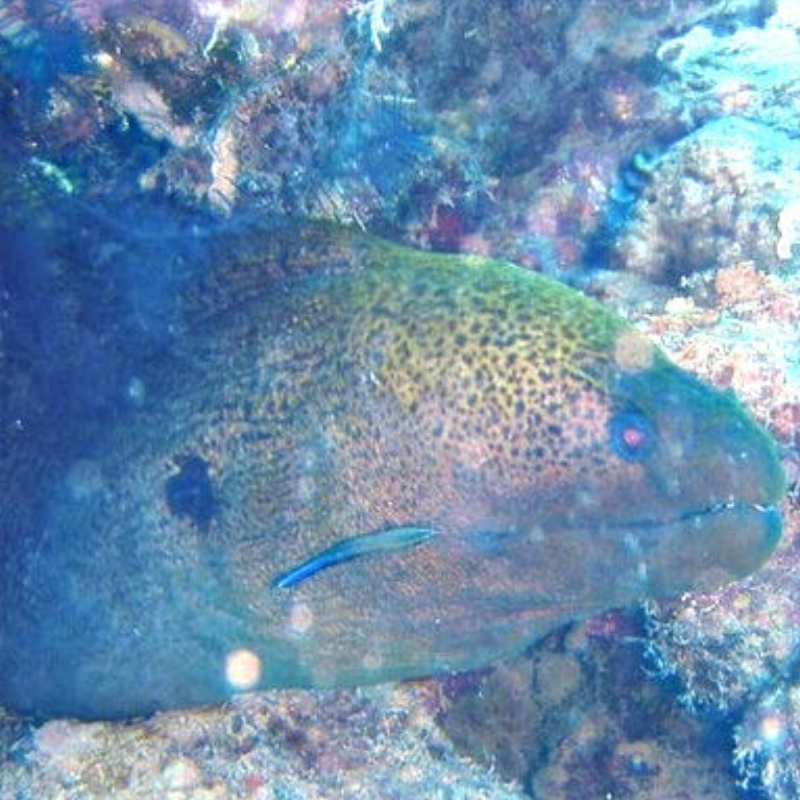}}
    \subfloat{\includegraphics[width = 0.1\linewidth]{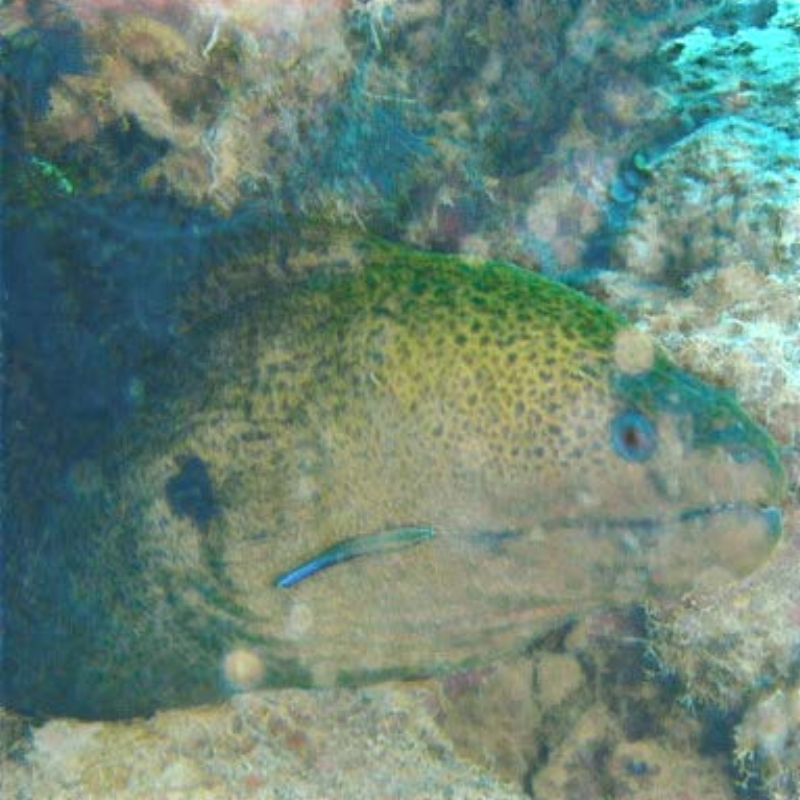}} 
    \subfloat{\includegraphics[width = 0.1\linewidth]{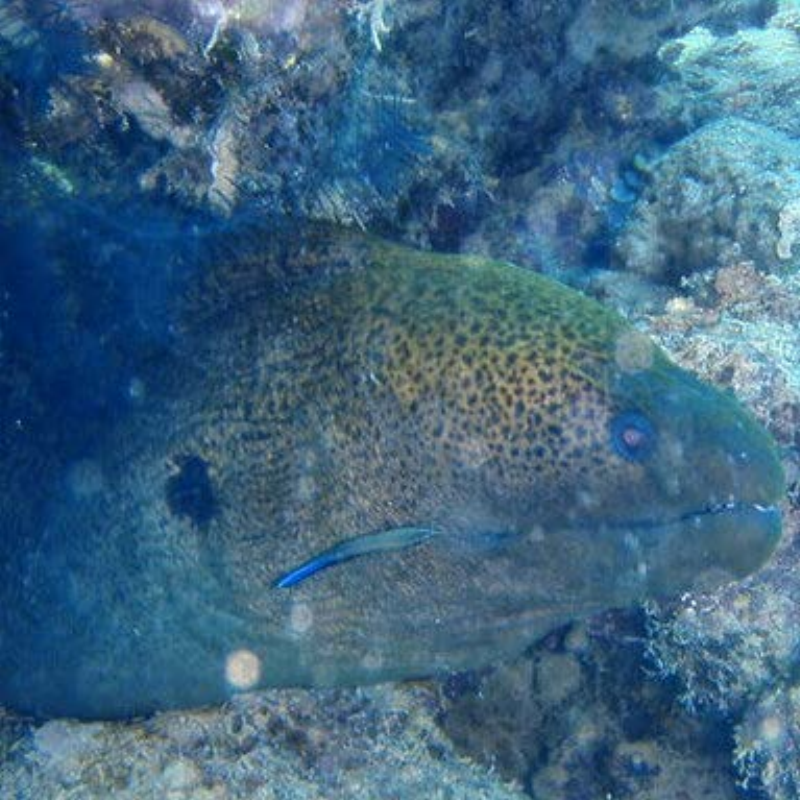}}
    \subfloat{\includegraphics[width = 0.1\linewidth]{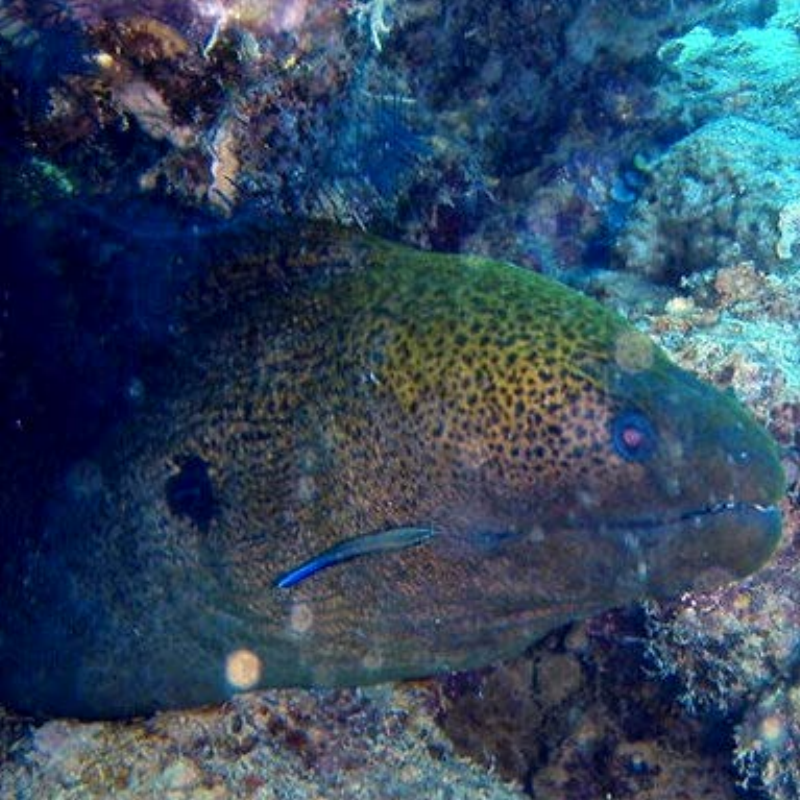}}
    \subfloat{\includegraphics[width = 0.1\linewidth]{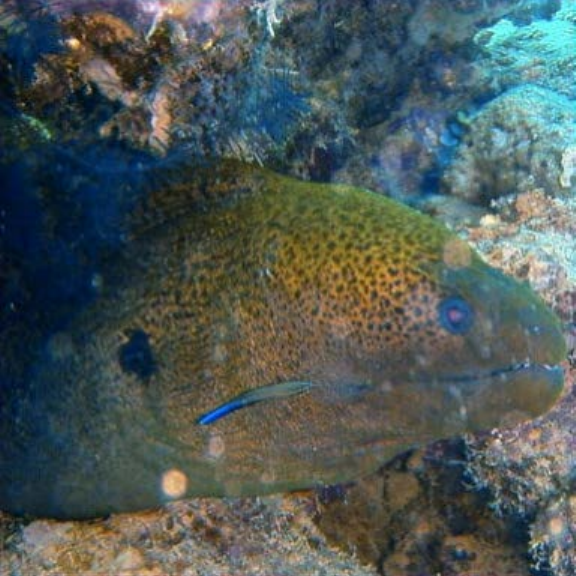}}
    \subfloat{\includegraphics[width = 0.1\linewidth]{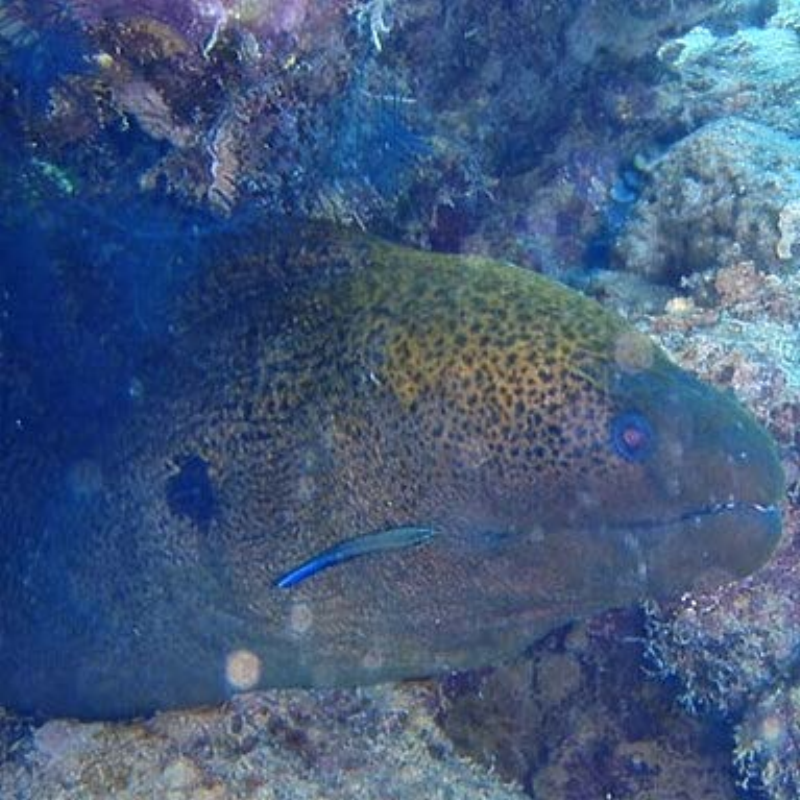}}
    \subfloat{\includegraphics[width = 0.1\linewidth]{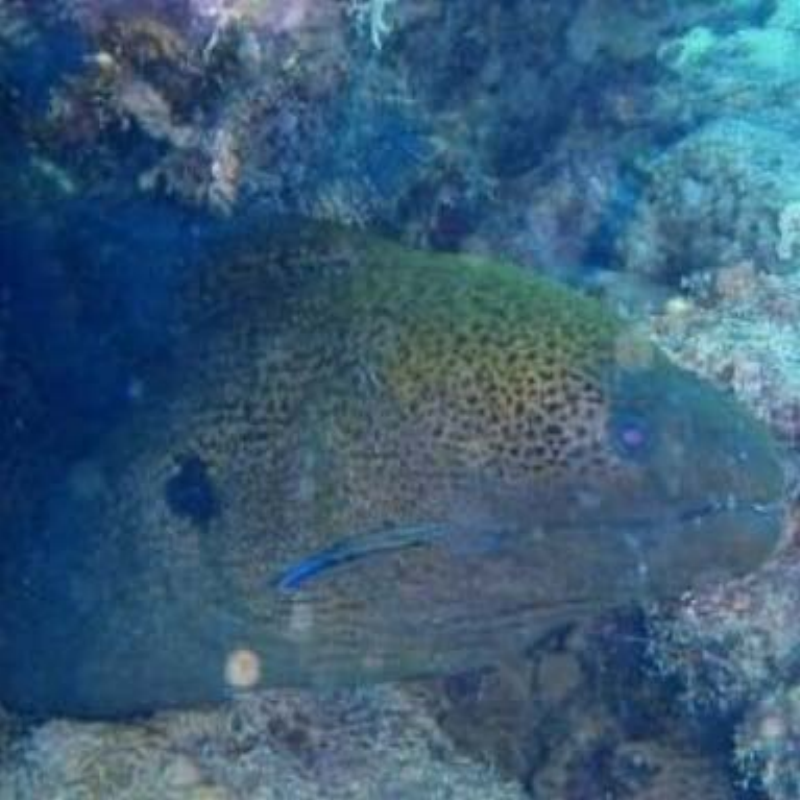}}
    \subfloat{\includegraphics[width = 0.1\linewidth]{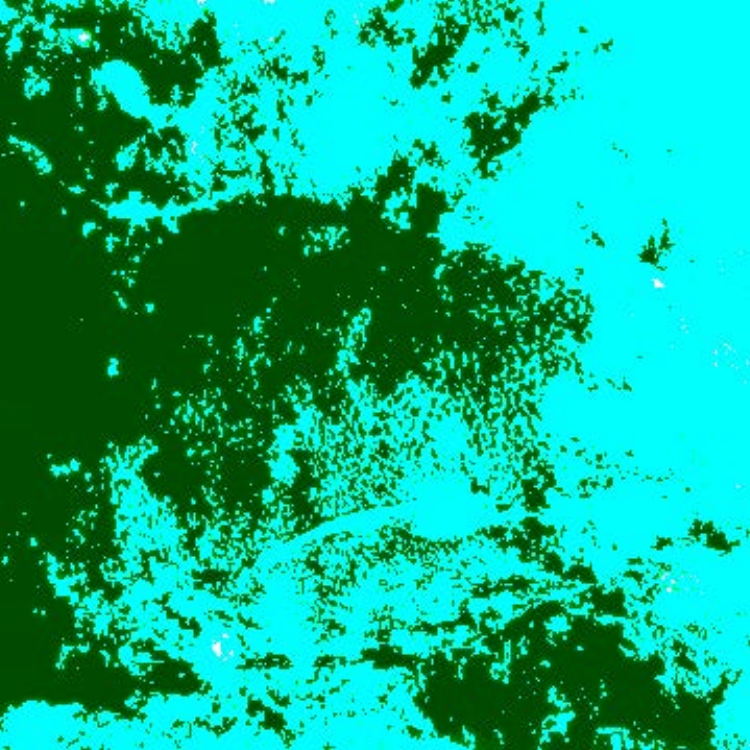}}\\
    % \subfloat{\includegraphics[width = 0.15\linewidth]{image/dataset/DeepWaveNet/Deepwavenet_208.pdf}}
    % \subfloat{\includegraphics[width = 0.15\linewidth]{image/dataset/DeepWaveNet/Deepwavenet_209.pdf}} \\
    \vspace{-0.14in}%\caption*{Restoration results by Deep WN.}%\vspace{-0.1in}
    % \subfloat{\includegraphics[width = 0.15\linewidth]{image/dataset/Waternet/waternet_205.pdf}} 
    % \subfloat{\includegraphics[width = 0.15\linewidth]{image/dataset/Waternet/waternet_206.pdf}} 
    % \subfloat{\includegraphics[width = 0.15\linewidth]{image/dataset/Waternet/waternet_207.pdf}}
    \subfloat{\includegraphics[width = 0.1\linewidth]{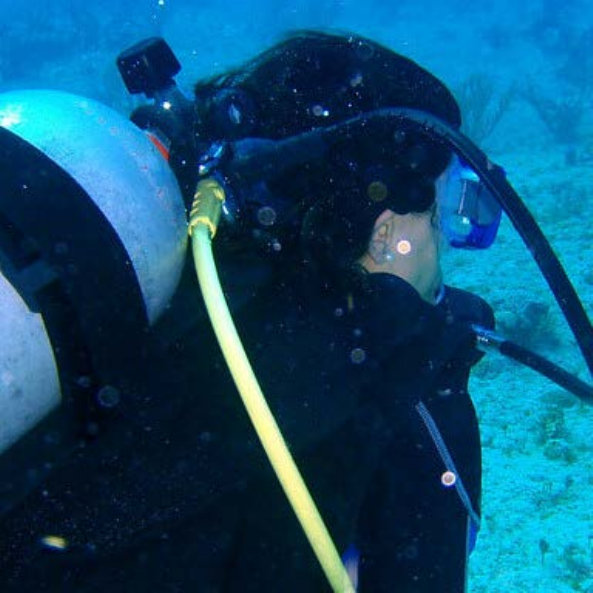}} 
    \subfloat{\includegraphics[width = 0.1\linewidth]{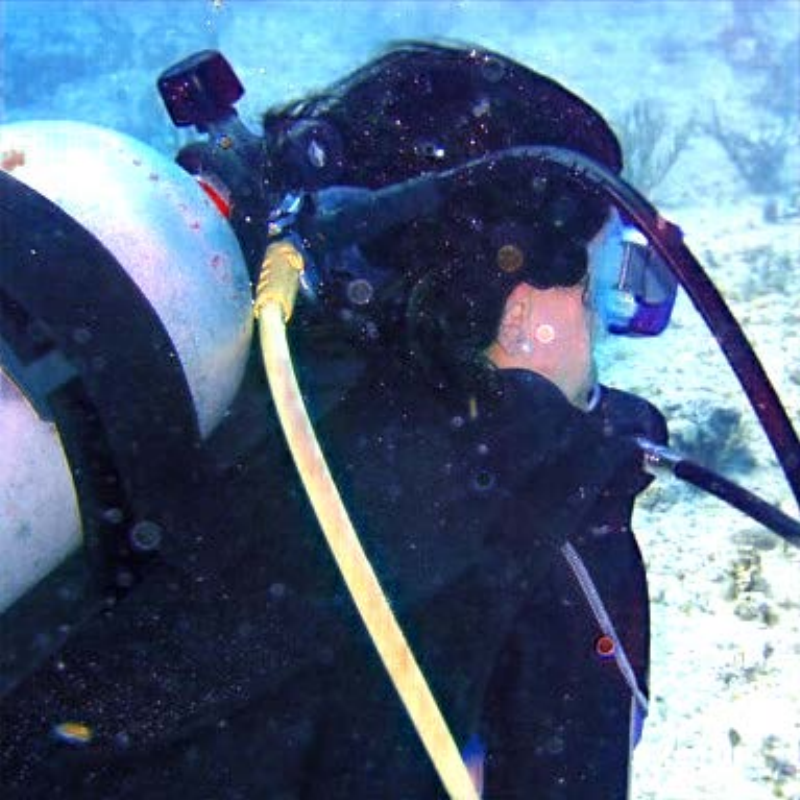}}
    \subfloat{\includegraphics[width = 0.1\linewidth]{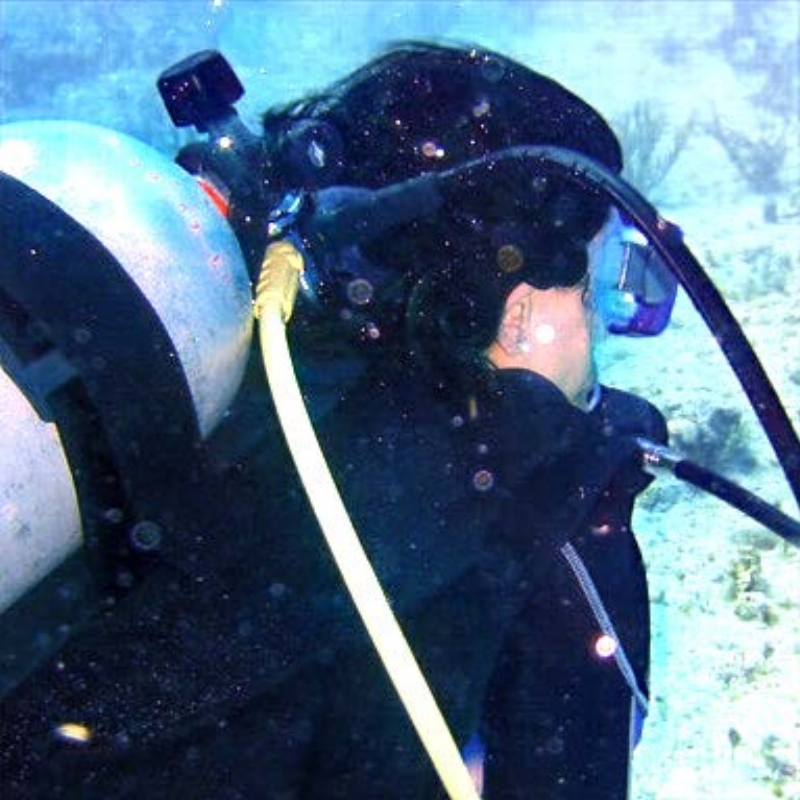}}
    \subfloat{\includegraphics[width = 0.1\linewidth]{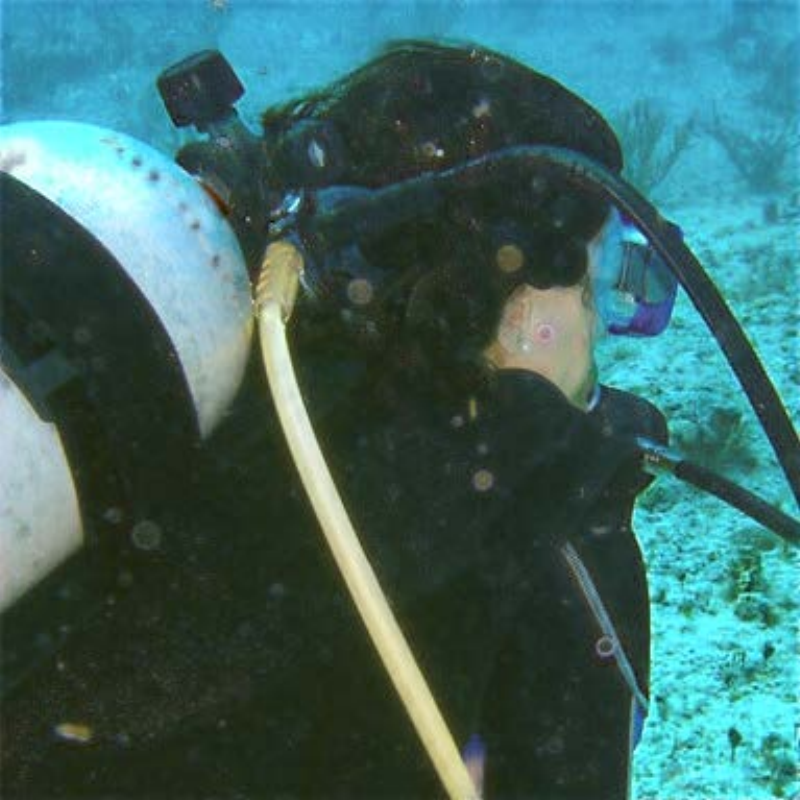}} 
    \subfloat{\includegraphics[width = 0.1\linewidth]{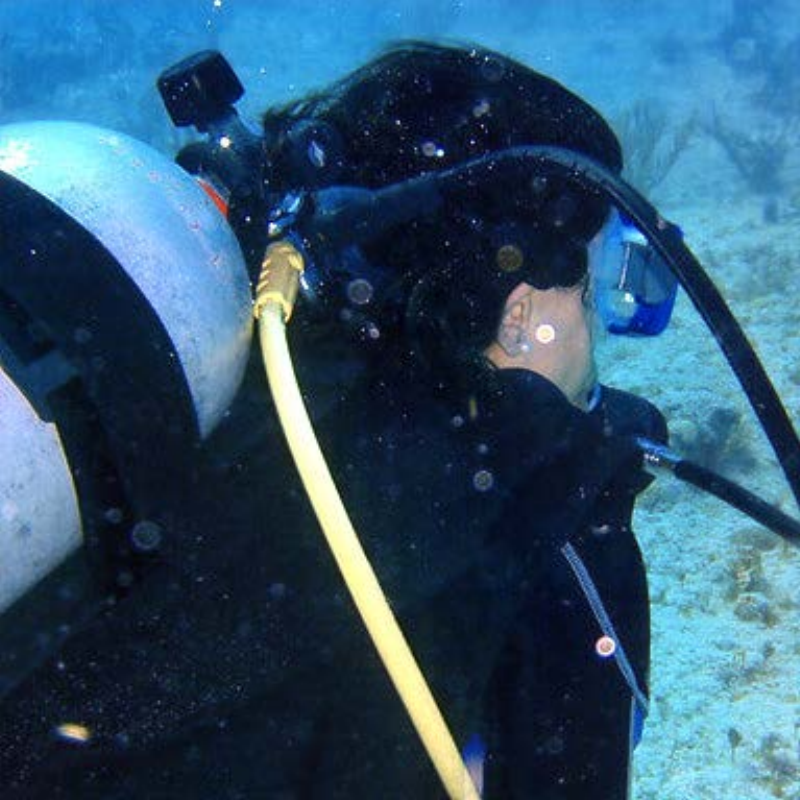}}
    \subfloat{\includegraphics[width = 0.1\linewidth]{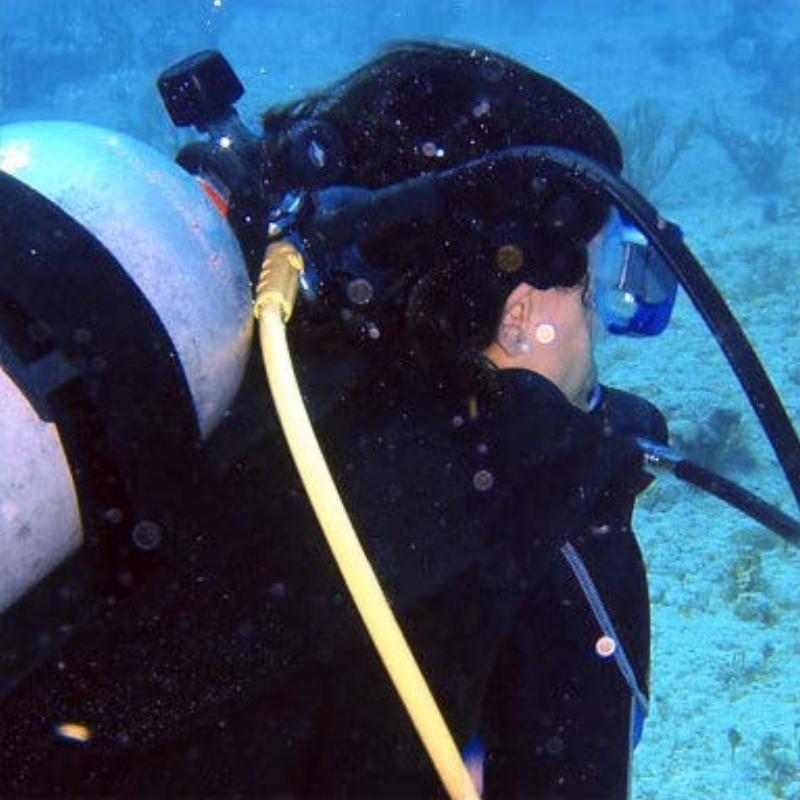}}
    \subfloat{\includegraphics[width = 0.1\linewidth]{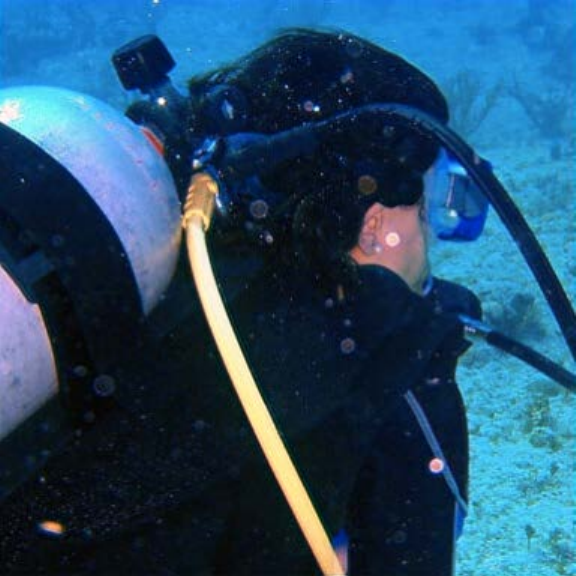}}
    \subfloat{\includegraphics[width = 0.1\linewidth]{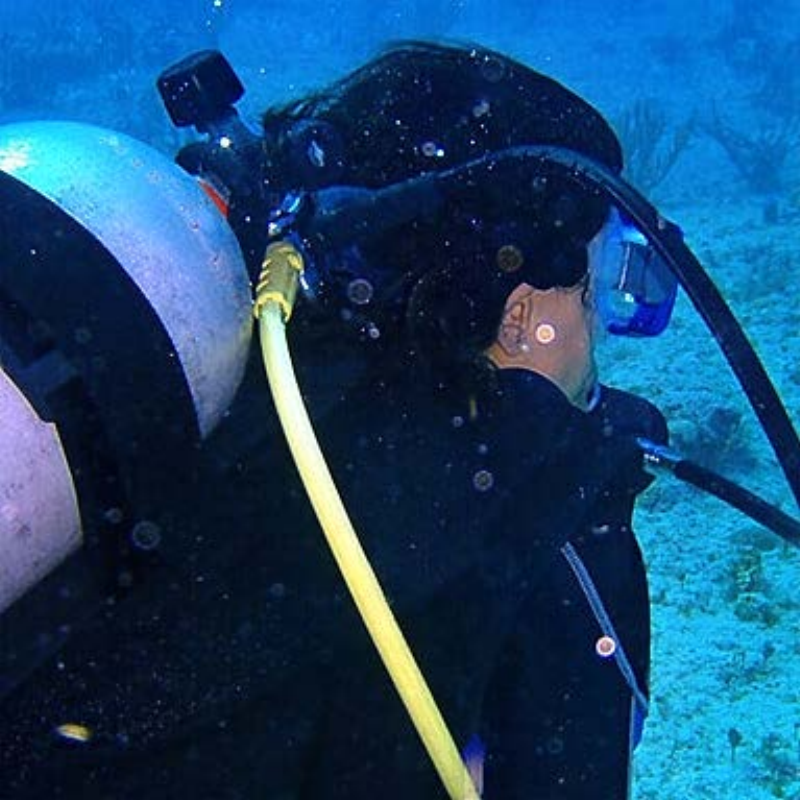}}
    \subfloat{\includegraphics[width = 0.1\linewidth]{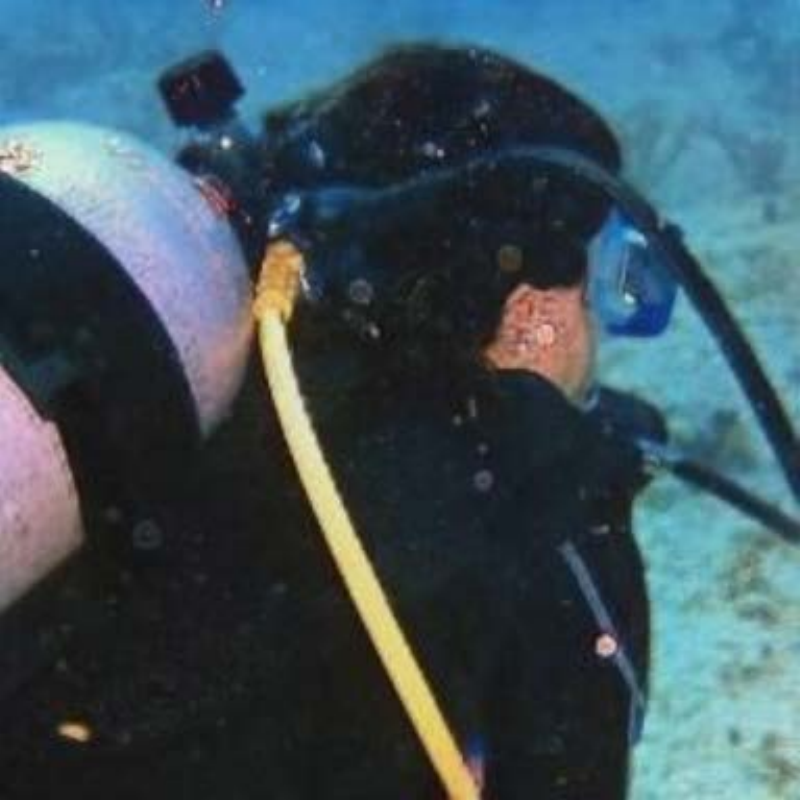}}
    \subfloat{\includegraphics[width = 0.1\linewidth]{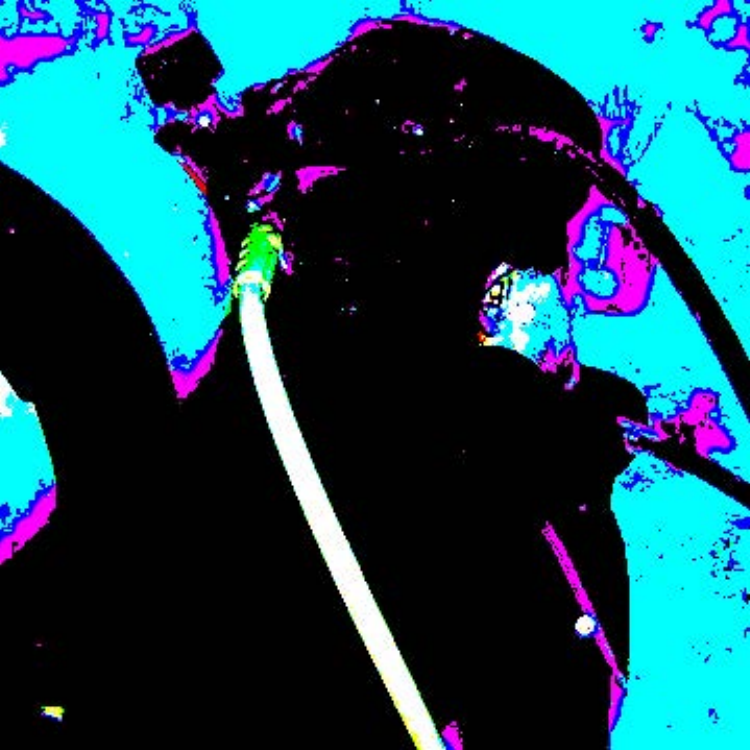}}\\
    % \subfloat{\includegraphics[width = 0.15\linewidth]{image/dataset/Waternet/waternet_209.pdf}} \\
    \vspace{-0.14in}%\caption*{Restoration results by WaterNet.}
    \setcounter{subfigure}{0}
    \subfloat[Real images]{\includegraphics[width = 0.1\linewidth]{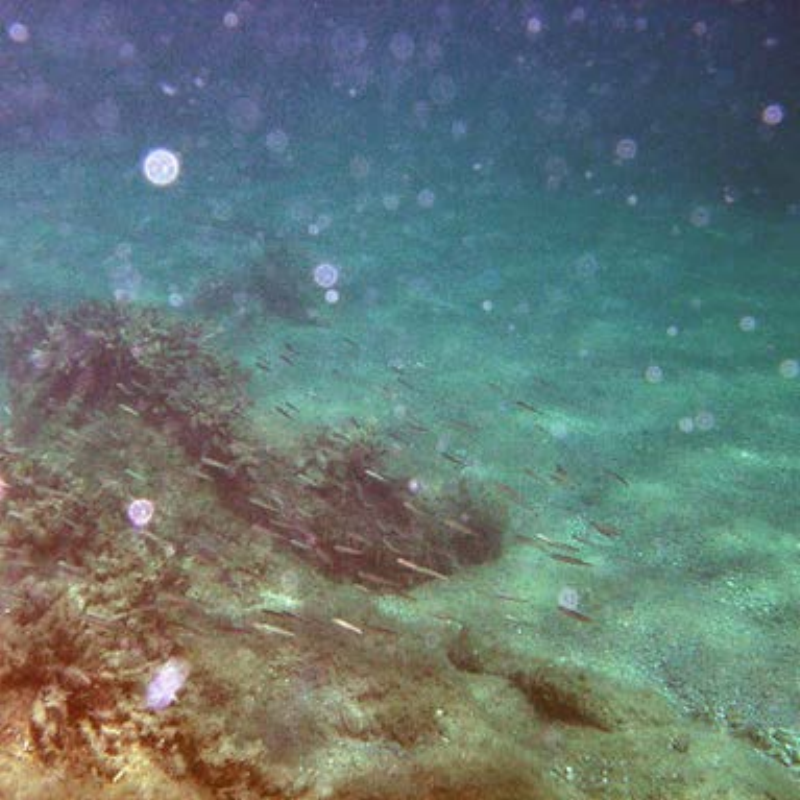}}
    \subfloat[Trans(P)]{\includegraphics[width = 0.1\linewidth]{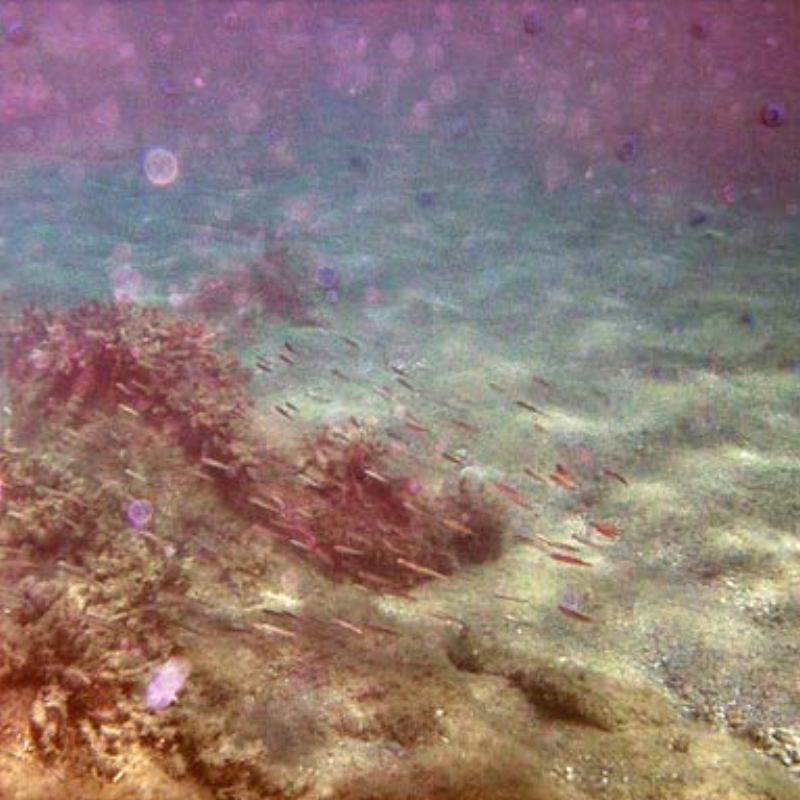}}
    \subfloat[Trans(P')]{\includegraphics[width = 0.1\linewidth]{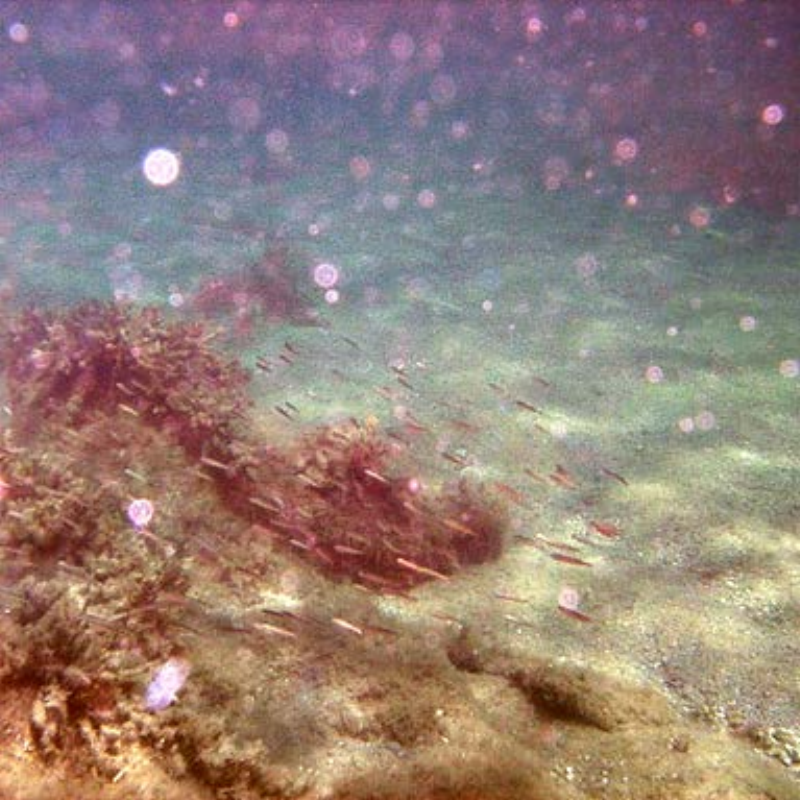}}
    \subfloat[DWN(P)]{\includegraphics[width = 0.1\linewidth]{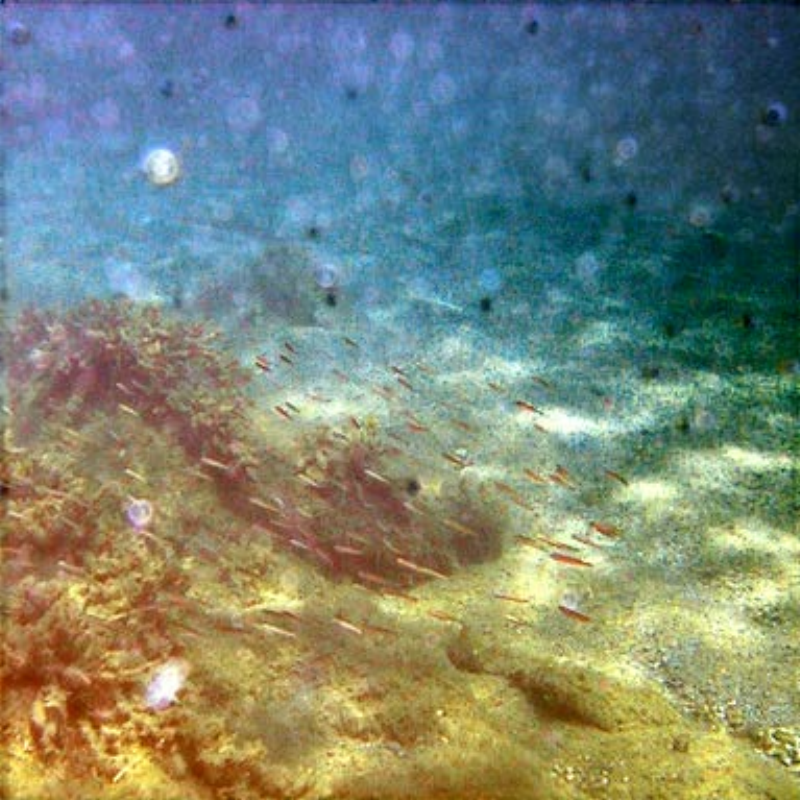}} 
    \subfloat[Trans(U)]{\includegraphics[width = 0.1\linewidth]{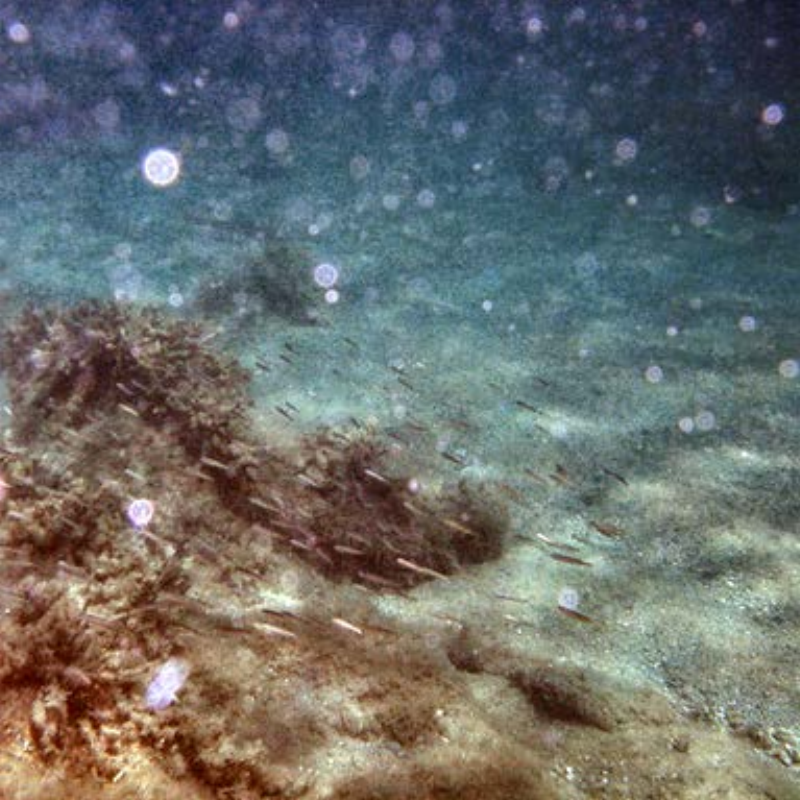}}
     \subfloat[DWN(U)]{\includegraphics[width = 0.1\linewidth]{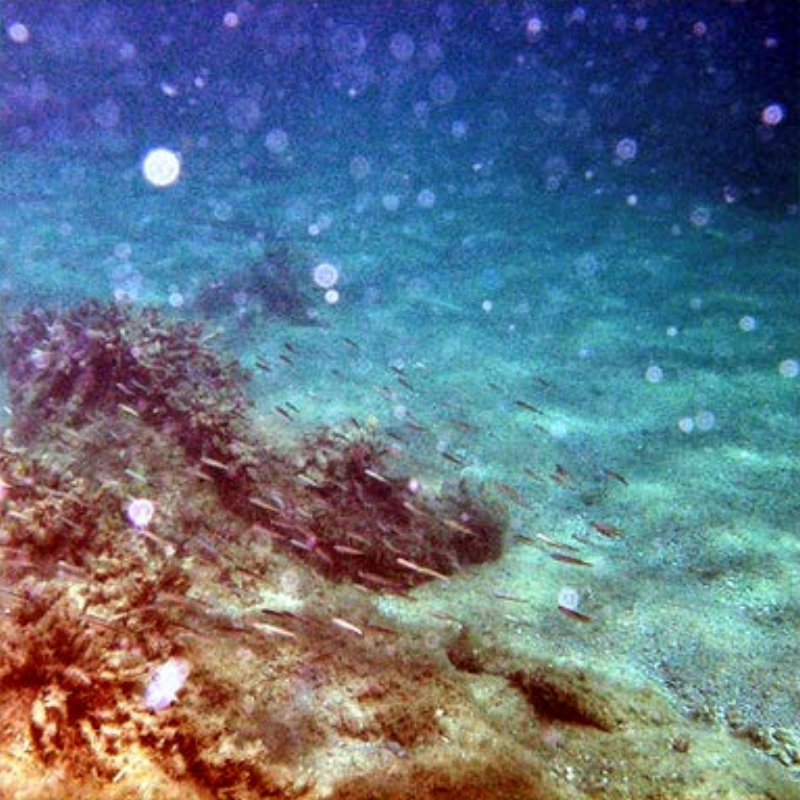}} 
     \subfloat[WN]{\includegraphics[width = 0.1\linewidth]{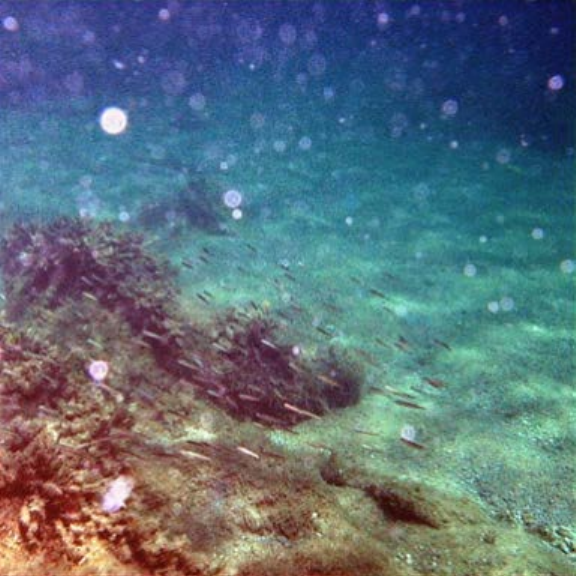}}
    \subfloat[Trans(L)]{\includegraphics[width = 0.1\linewidth]{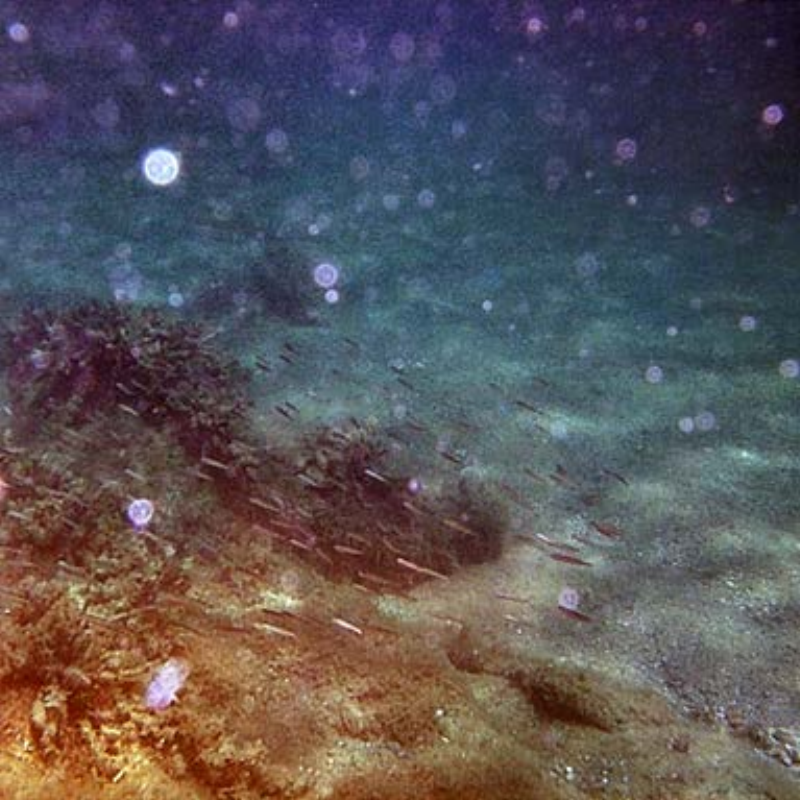}}
     \subfloat[U-shape]{\includegraphics[width = 0.1\linewidth]{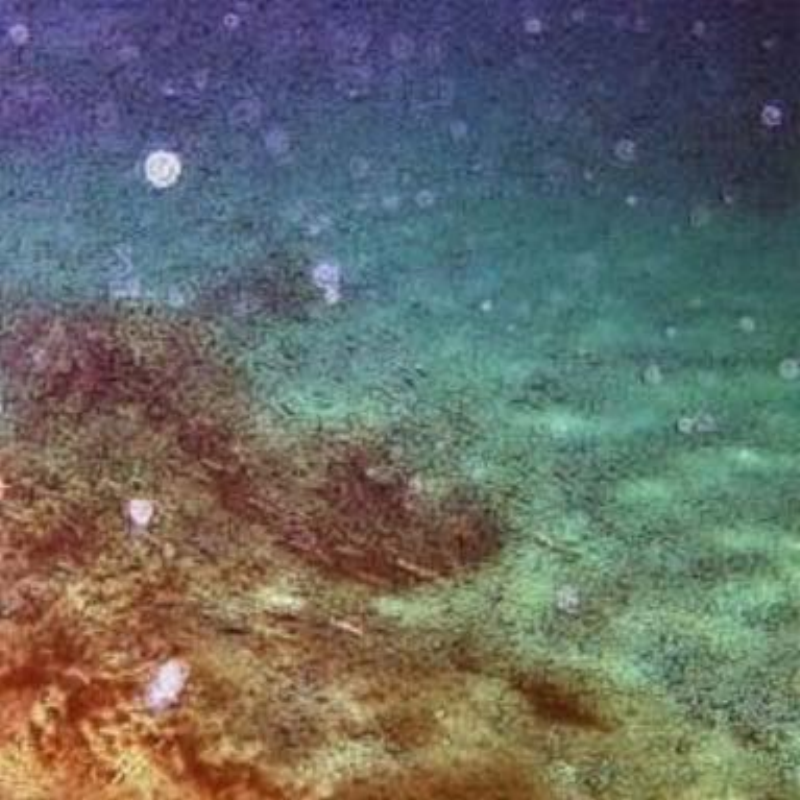}}
     \subfloat[HLRP]{\includegraphics[width = 0.1\linewidth]{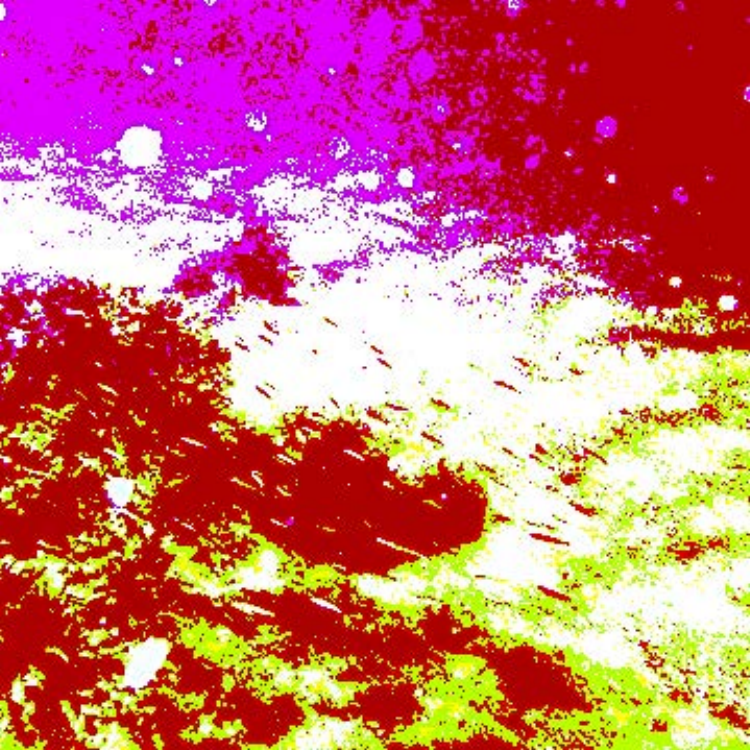}}\\
    \caption{Underwater image enhancement results for real-world images.}
    %From left to right: Synthesized test images, restoration results by Transformer-P, restoration results by Transformer-C, restoration results by Transformer-U, restoration results by Transformer-L, restoration results by Deep WN-U, restoration results by Deep WN-P, restoration results by WaterNet, restoration results by U-shape, and restoration results by HLRP.}
  \label{real_img_MSRCC}
\end{figure*}

\begin{figure*}[t]
%\vspace{-0.2in}
\centering
    \setcounter{subfigure}{0}
    \subfloat[Real images]{\includegraphics[width = 0.1\linewidth]{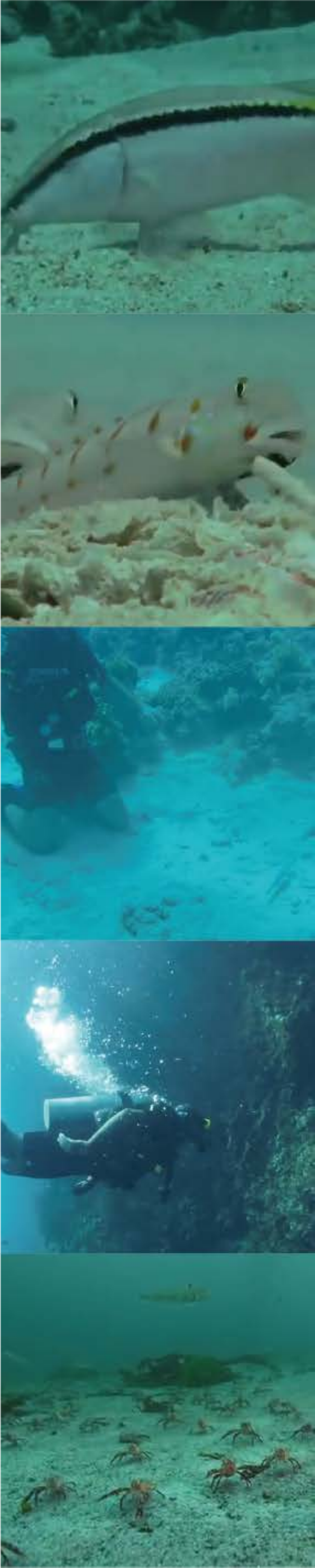}}
    \subfloat[Trans(P)]{\includegraphics[width = 0.1\linewidth]{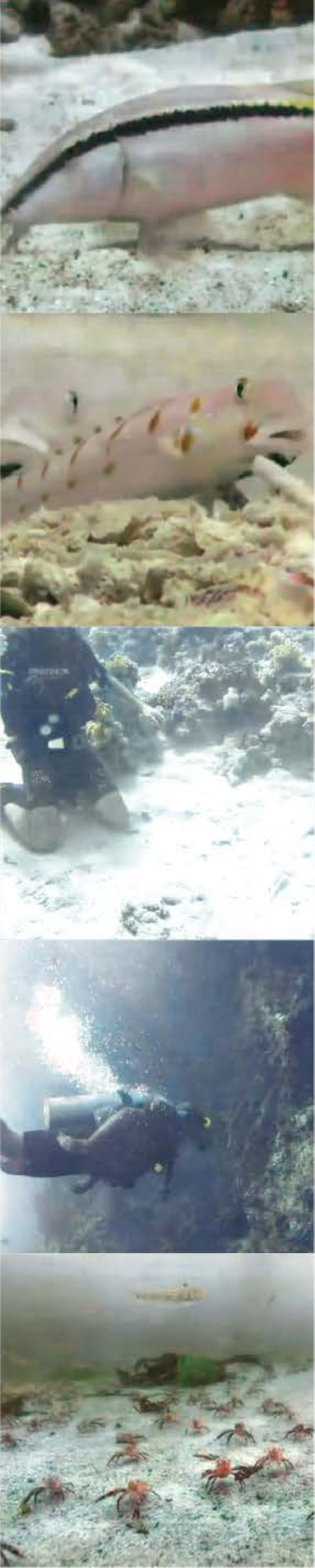}}
    \subfloat[Trans(P')]{\includegraphics[width = 0.1\linewidth]{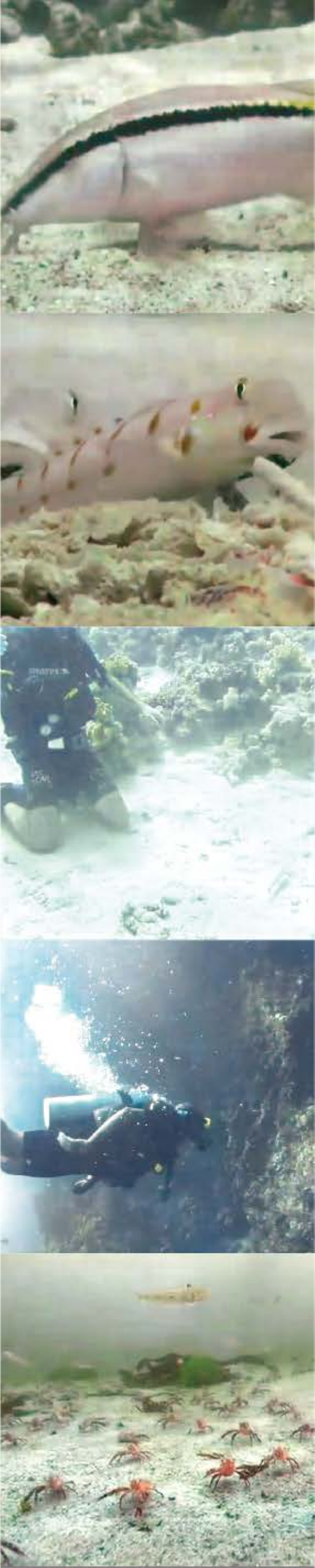}}
    \subfloat[DWN(P)]{\includegraphics[width = 0.1\linewidth]{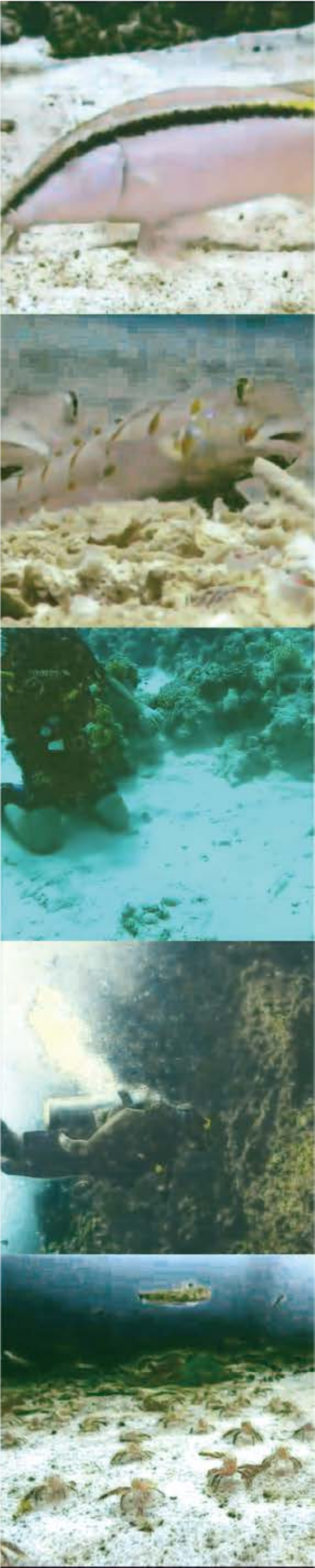}} 
    \subfloat[Trans(U)]{\includegraphics[width = 0.1\linewidth]{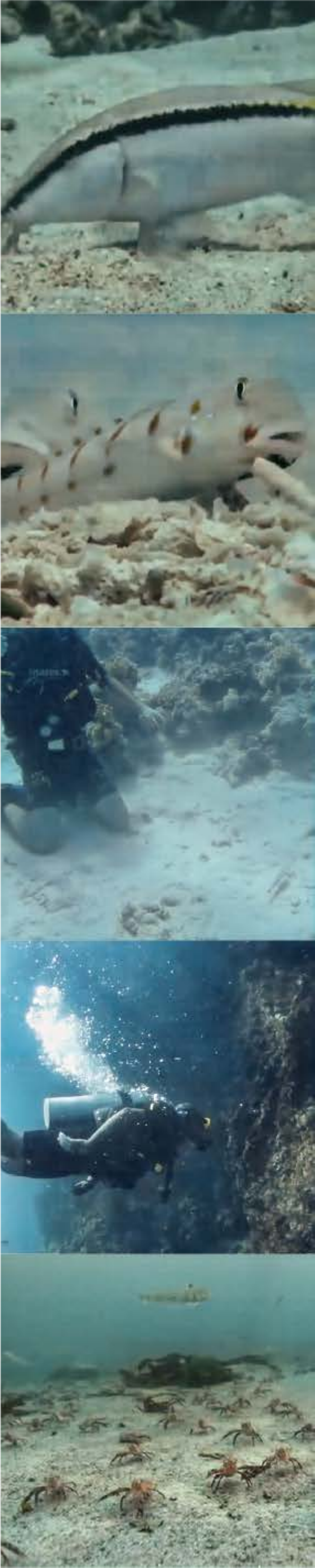}}
     \subfloat[DWN(U)]{\includegraphics[width = 0.1\linewidth]{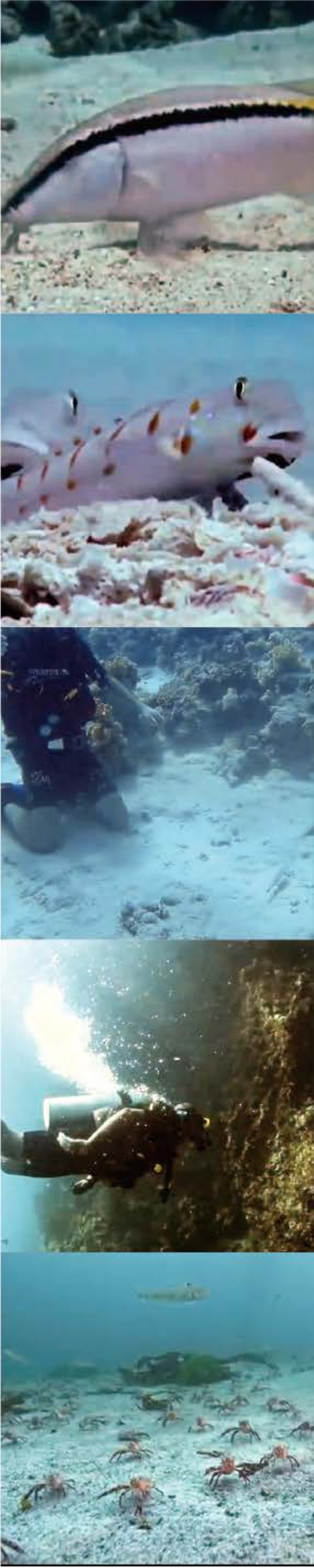}} 
     \subfloat[WN]{\includegraphics[width = 0.1\linewidth]{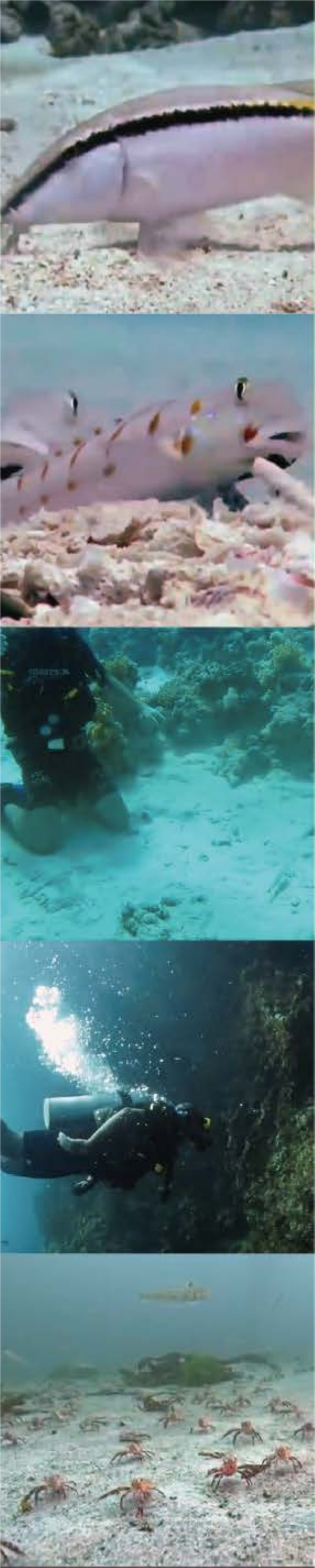}}
    \subfloat[Trans(L)]{\includegraphics[width = 0.1\linewidth]{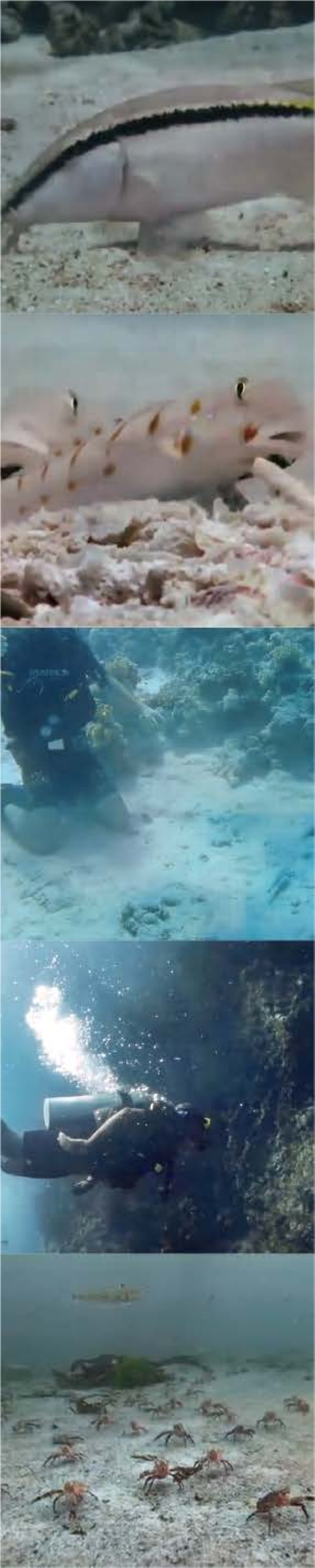}}
     \subfloat[U-shape]{\includegraphics[width = 0.1\linewidth]{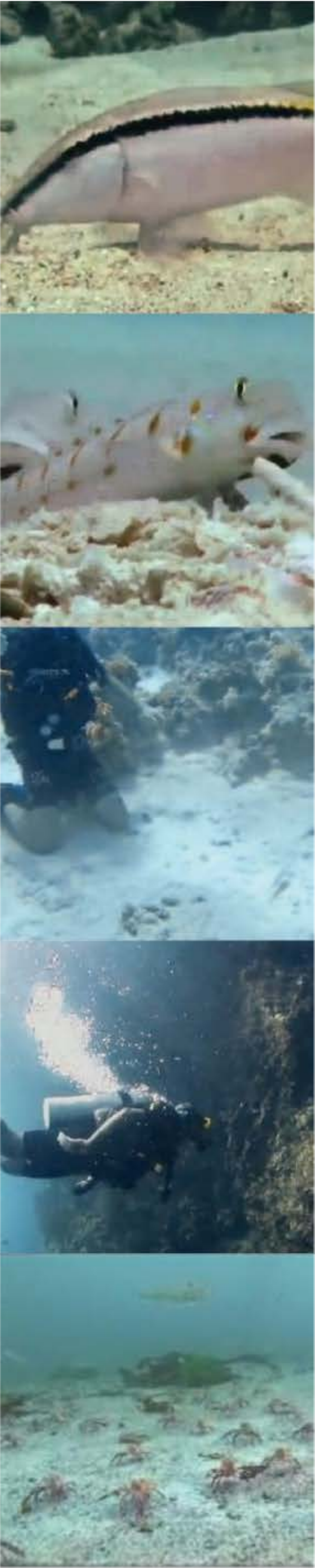}}
     \subfloat[HLRP]{\includegraphics[width = 0.1\linewidth]{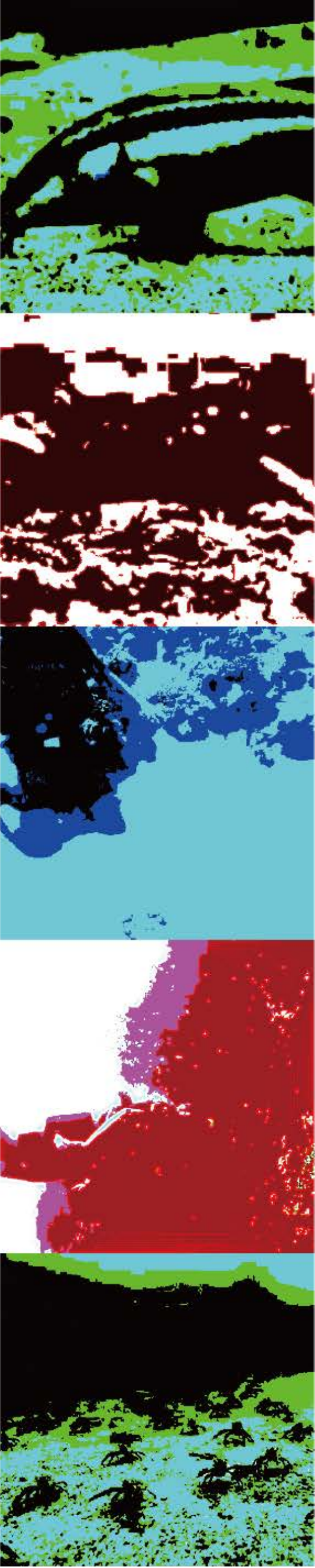}}\\
    \caption{Underwater image enhancement results for real-world images in UIEB challenges.}
  \label{realUIEB_img_MSRCC}
\end{figure*}

\begin{figure*}[t]
%\vspace{-0.2in}
\centering
    \subfloat{\includegraphics[width = 0.1\linewidth]{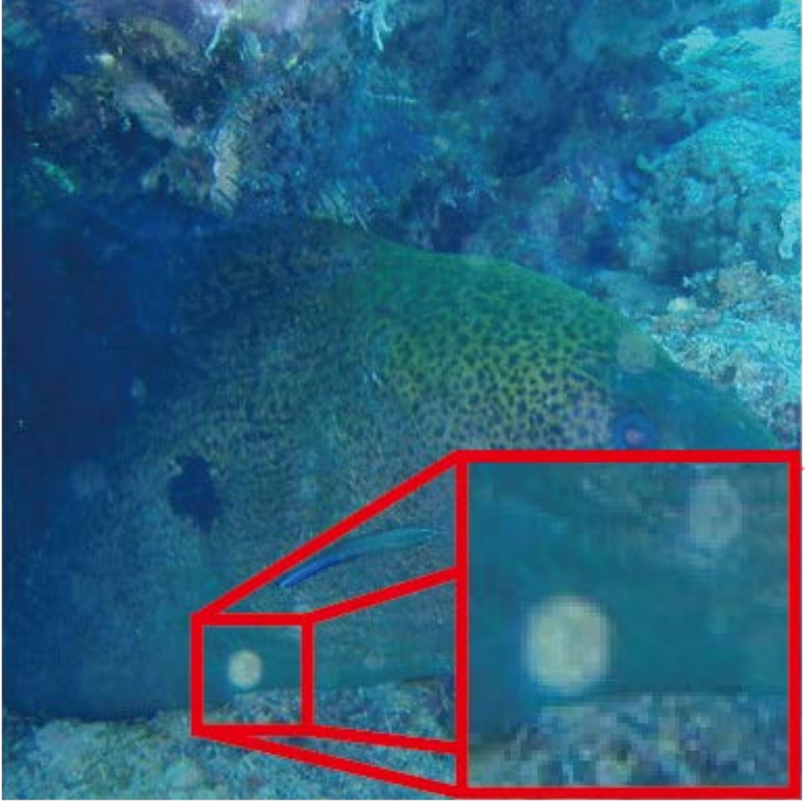}}
    \subfloat{\includegraphics[width = 0.1\linewidth]{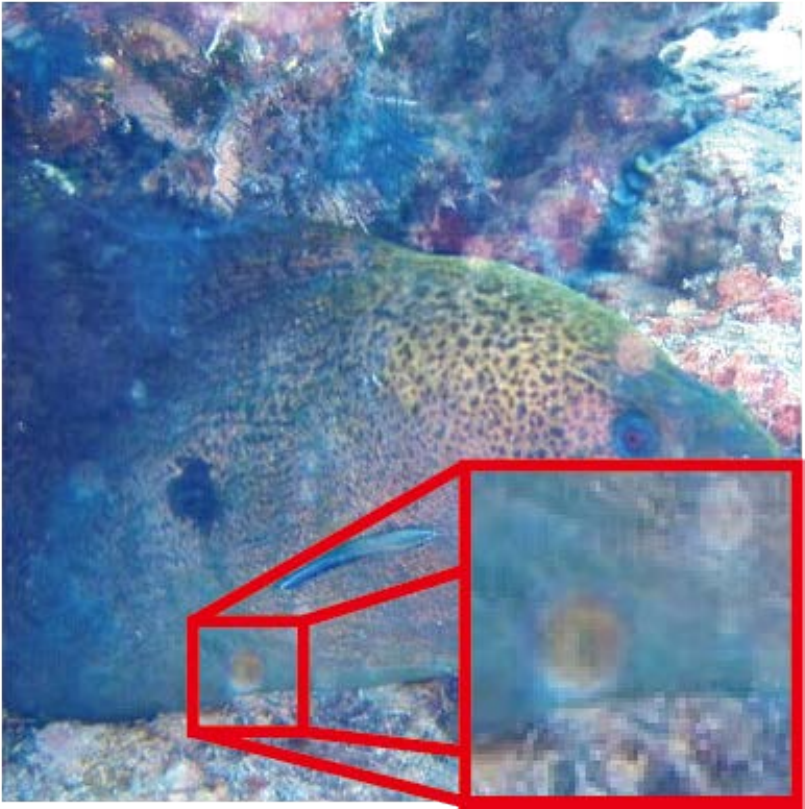}}
    \subfloat{\includegraphics[width = 0.1\linewidth]{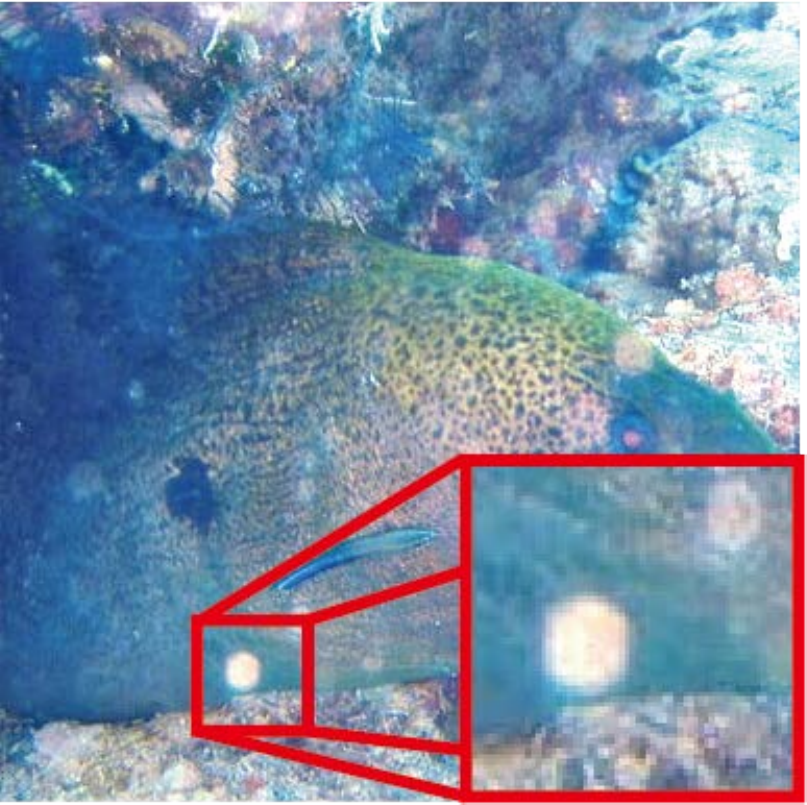}}
    \subfloat{\includegraphics[width = 0.1\linewidth]{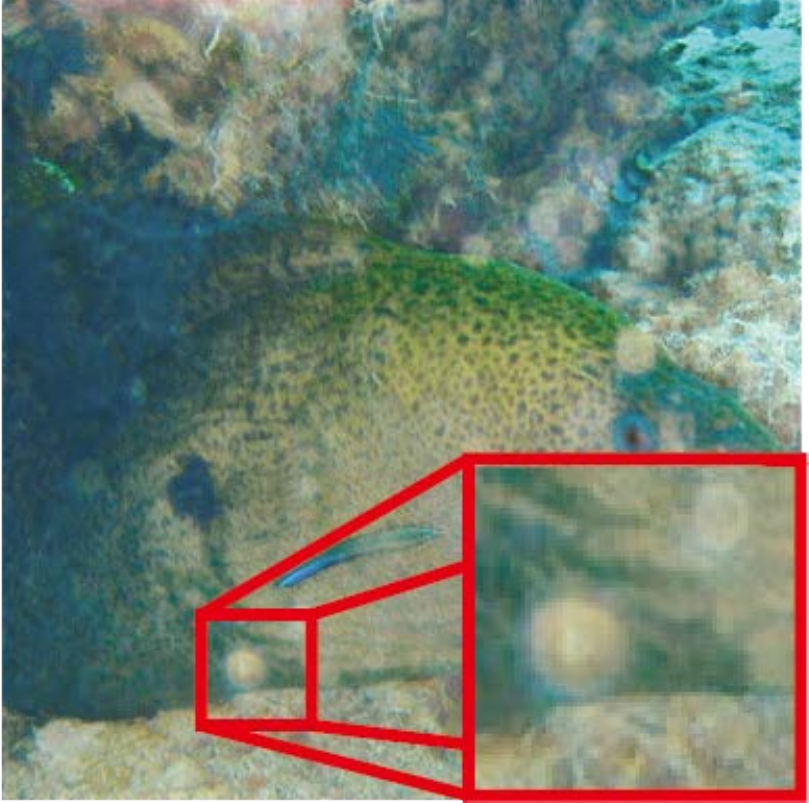}} 
    \subfloat{\includegraphics[width = 0.1\linewidth]{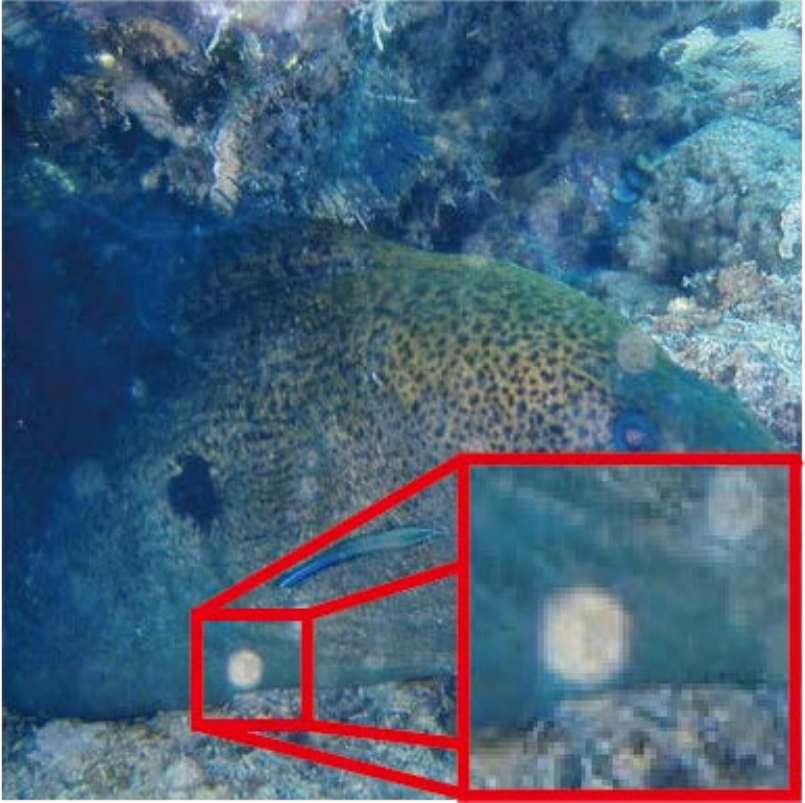}}
     \subfloat{\includegraphics[width = 0.1\linewidth]{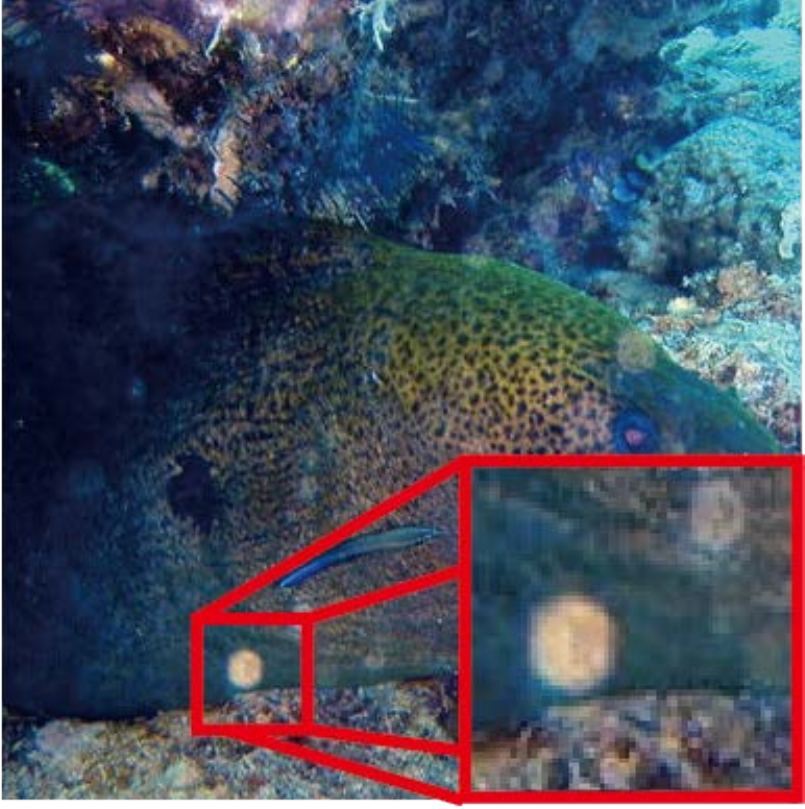}} 
     \subfloat{\includegraphics[width = 0.1\linewidth]{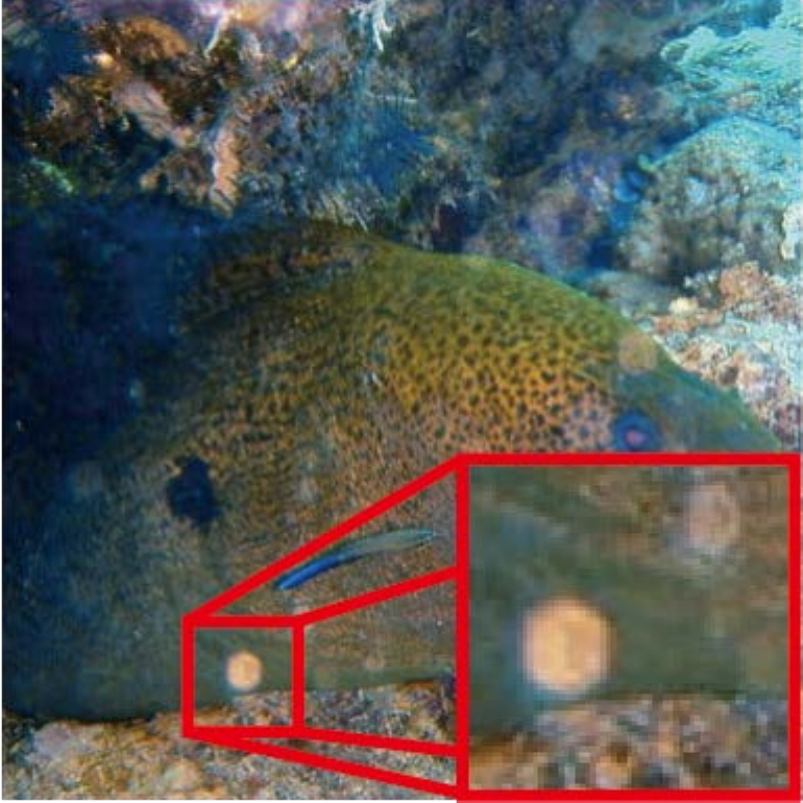}}
    \subfloat{\includegraphics[width = 0.1\linewidth]{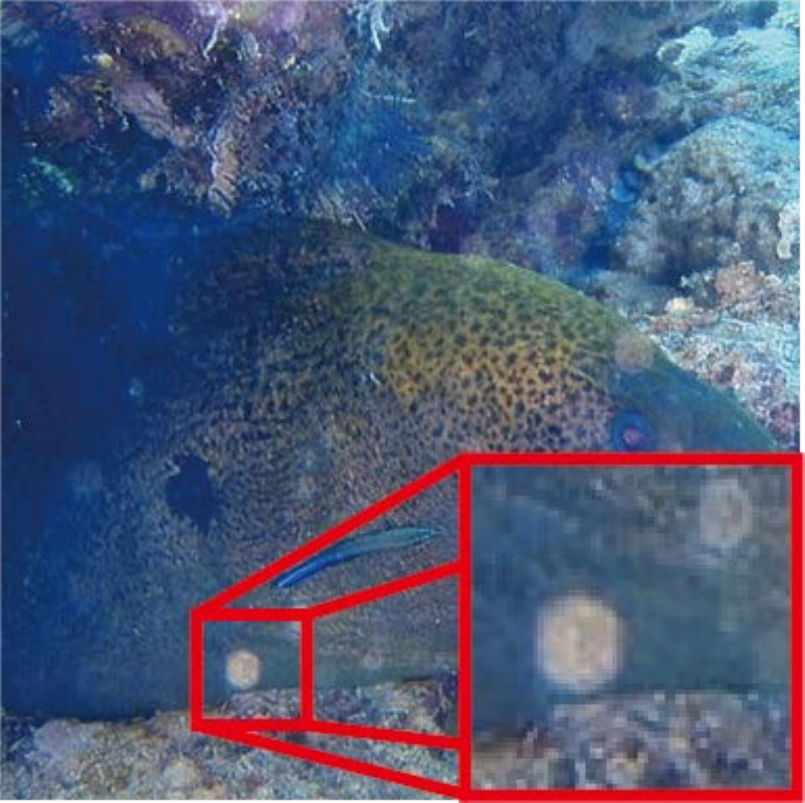}}
     \subfloat{\includegraphics[width = 0.1\linewidth]{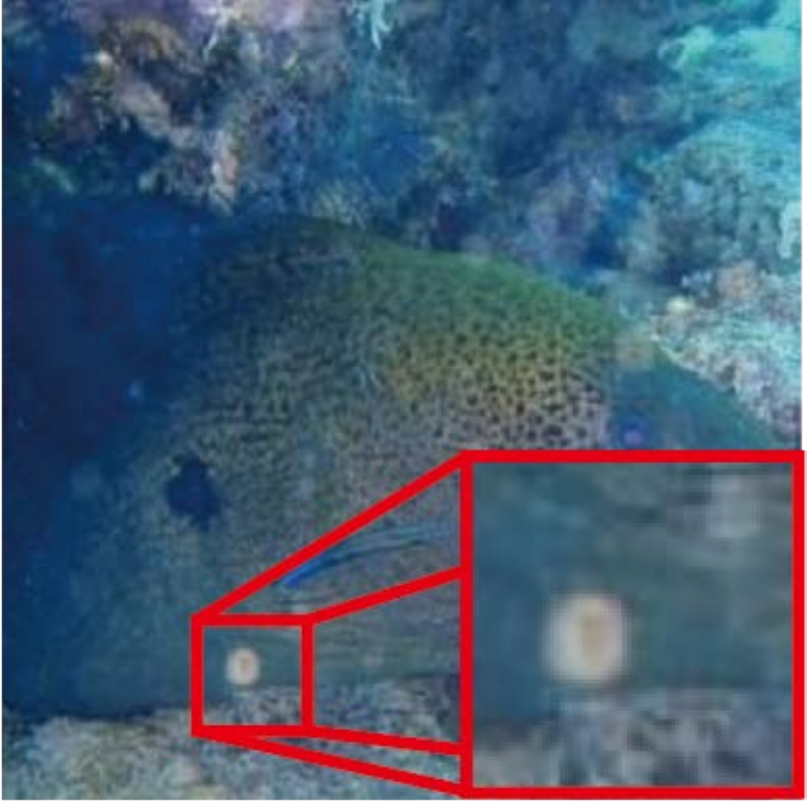}}
     \subfloat{\includegraphics[width = 0.1\linewidth]{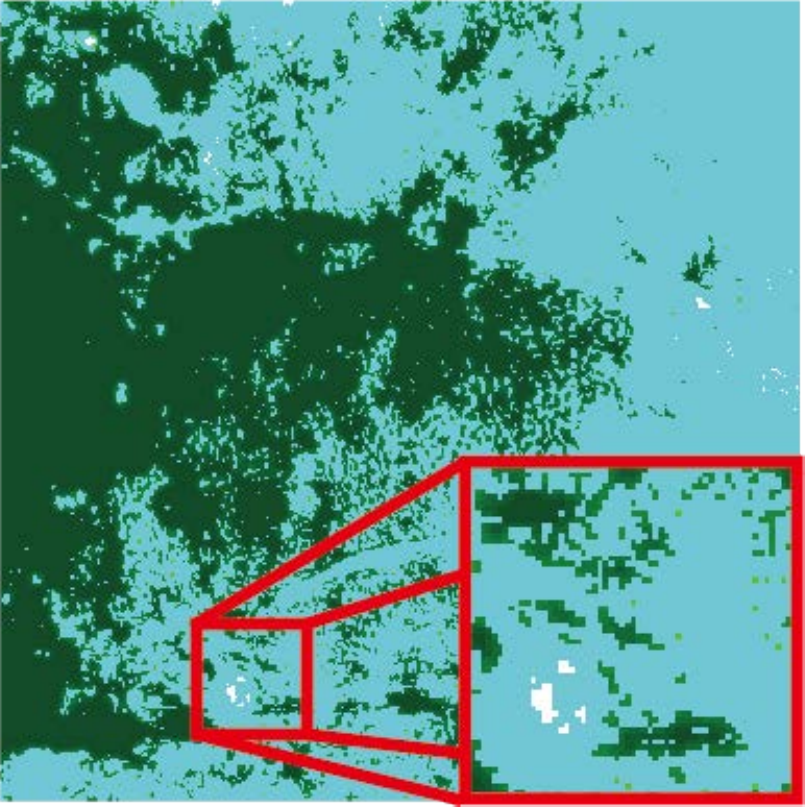}}\\\vspace{-0.15in}
    \setcounter{subfigure}{0}
    \subfloat[Real images]{\includegraphics[width = 0.1\linewidth]{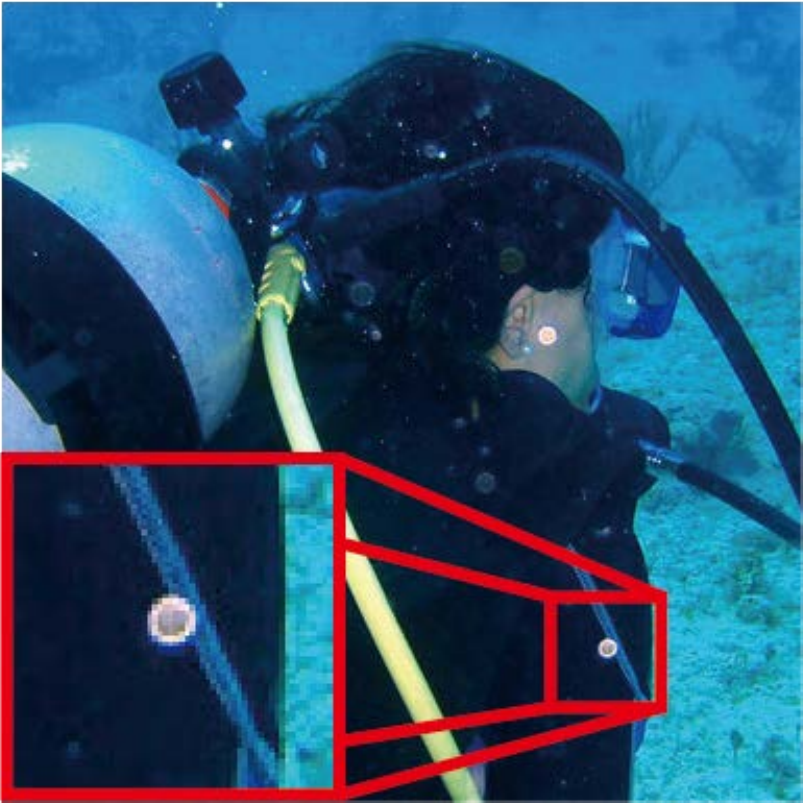}}
    \subfloat[Trans(P)]{\includegraphics[width = 0.1\linewidth]{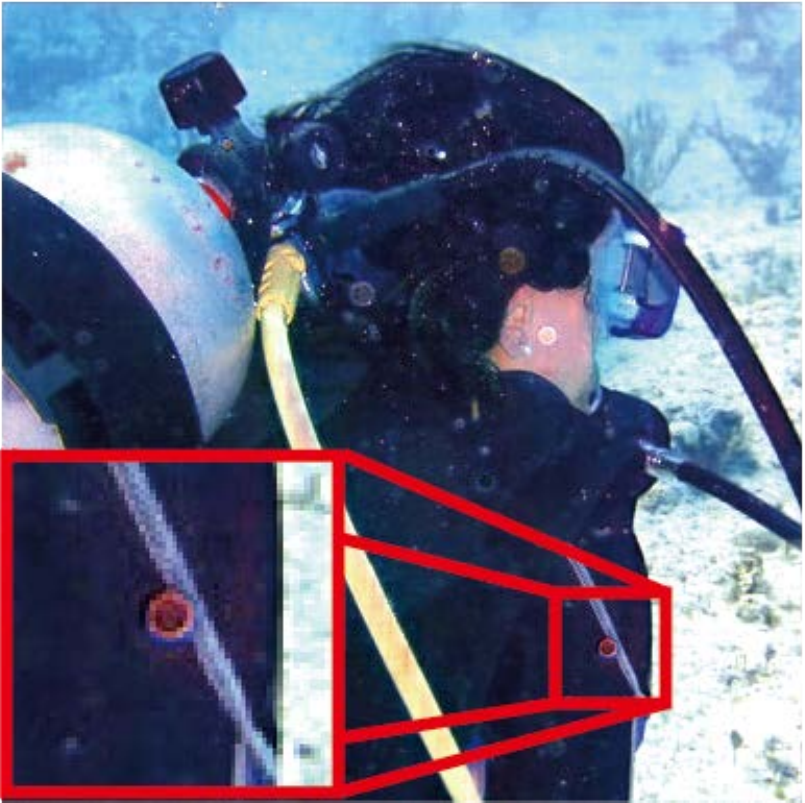}}
    \subfloat[Trans(P')]{\includegraphics[width = 0.1\linewidth]{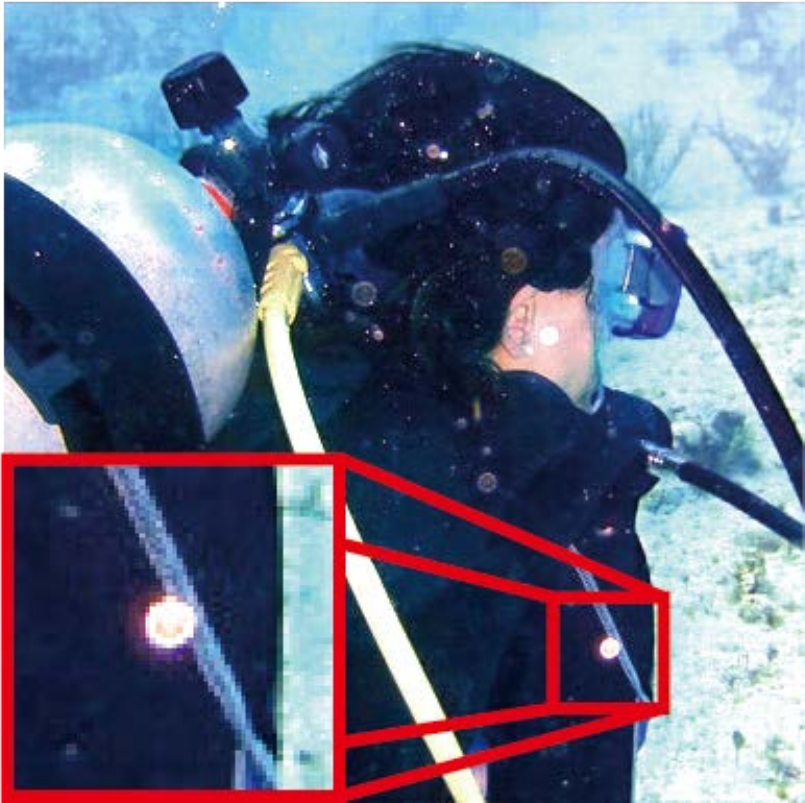}}
    \subfloat[DWN(P)]{\includegraphics[width = 0.1\linewidth]{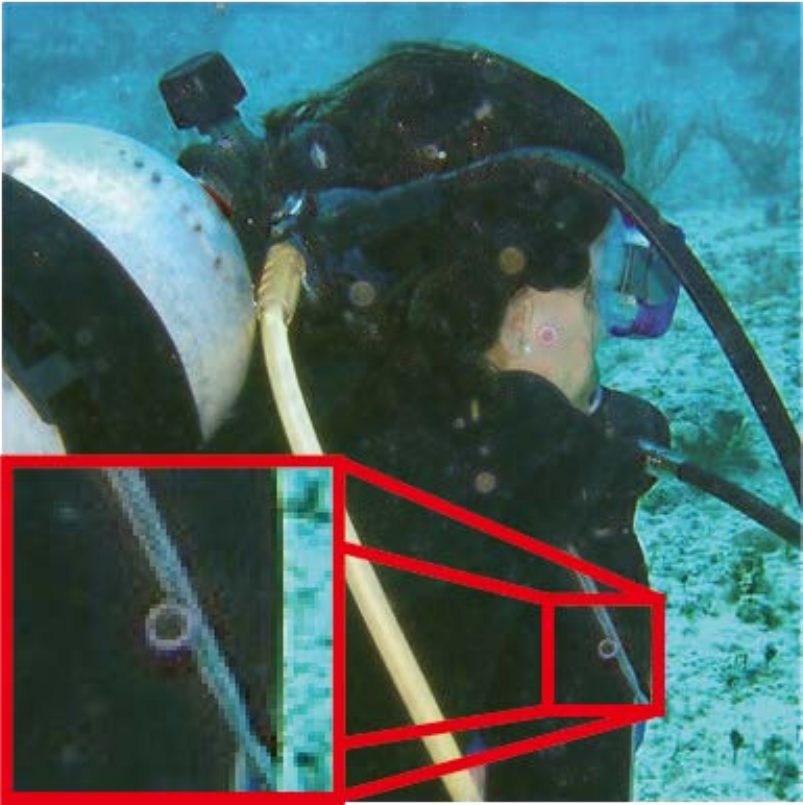}} 
    \subfloat[Trans(U)]{\includegraphics[width = 0.1\linewidth]{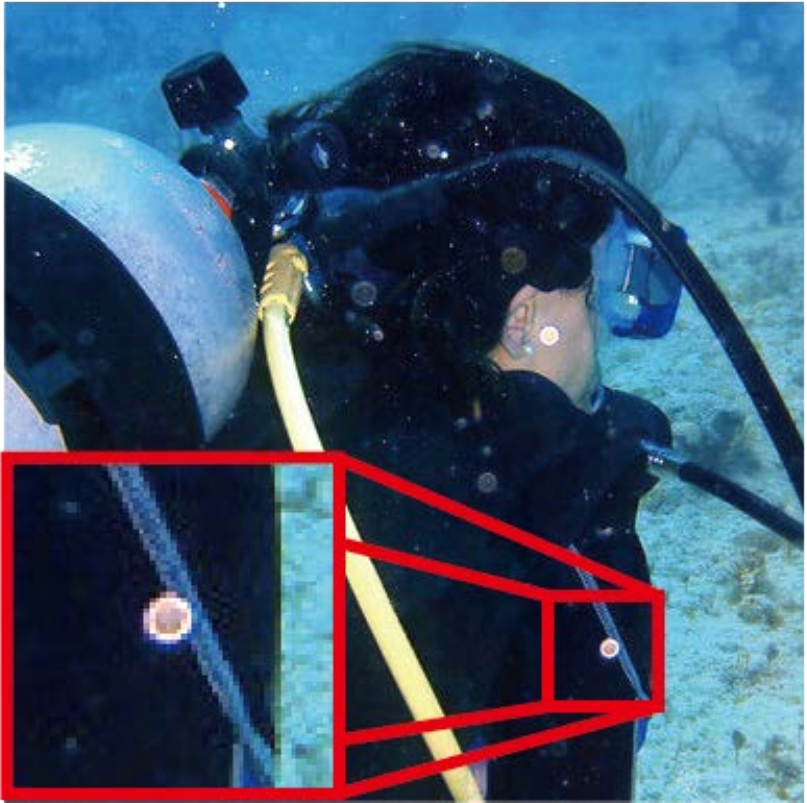}}
     \subfloat[DWN(U)]{\includegraphics[width = 0.1\linewidth]{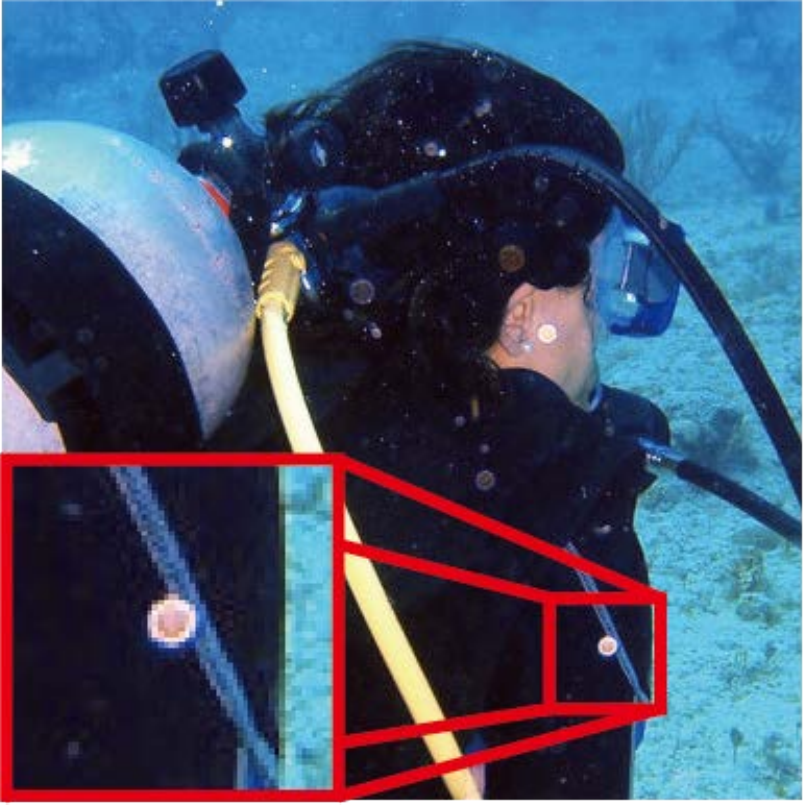}} 
     \subfloat[WN]{\includegraphics[width = 0.1\linewidth]{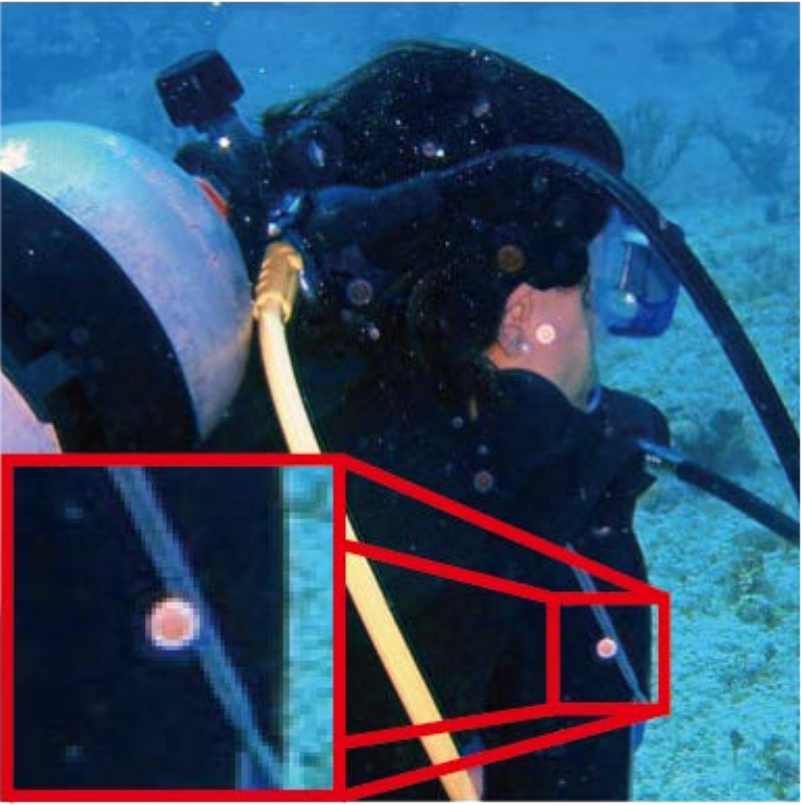}}
    \subfloat[Trans(L)]{\includegraphics[width = 0.1\linewidth]{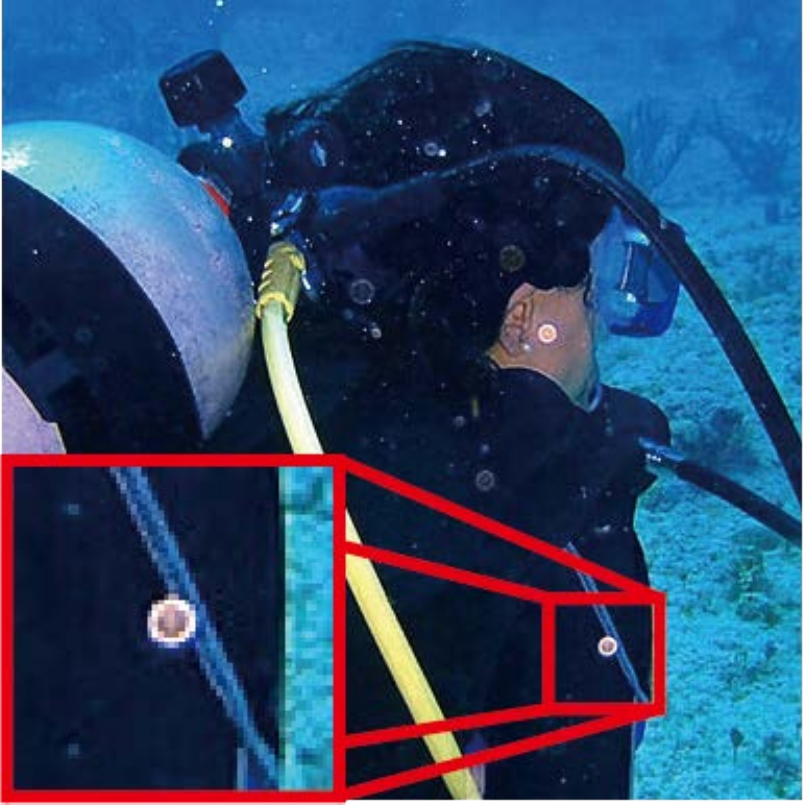}}
     \subfloat[U-shape]{\includegraphics[width = 0.1\linewidth]{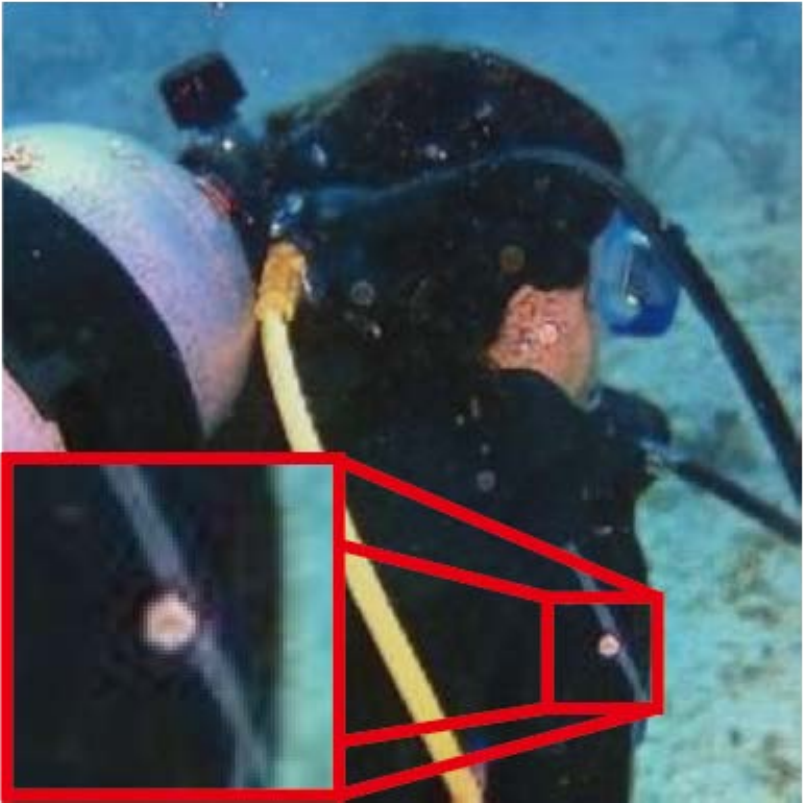}}
     \subfloat[HLRP]{\includegraphics[width = 0.1\linewidth]{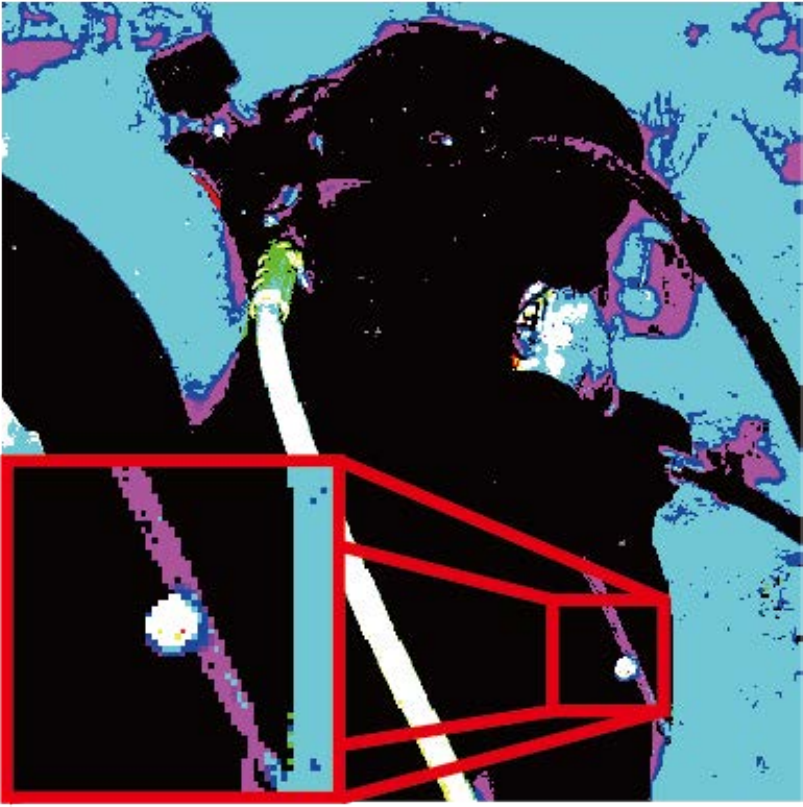}}\\
    \caption{Enlarged portion of marine snow removal.}
  \label{en_real_snow}
\end{figure*}

\begin{table}[t]
  \centering
  \caption{Average NIQE over real images.}
  \label{real_niqe}
  \begin{tabular}{l|cc} \hline
  %& \multicolumn{2}{c}{Dataset 2}\\\hline
  Method & NIQE$\downarrow$ \\ \hline\hline
  Trans(P) & 4.07\\
  Trans(P') &  4.01\\
  DWN(P) & \bf{3.88} \\\hline
  Trans(U) & 4.30 \\
  DWN(U) & 4.21 \\
  WN & 4.64\\\hline
  Trans(L) & 4.57\\
  U-shape & 4.97 \\\hline
  HLRP & NaN\\\hline
  Original image& 4.62 \\\hline
  \end{tabular}
\end{table}

% \begin{figure*}[t]
% %\vspace{-0.2in}
% \centering
%     \setcounter{subfigure}{0}
%     \subfloat[Test images\\in PHISWID]{\includegraphics[width = 0.2\linewidth]{image/testimg.pdf}}
%     \subfloat[Transformer(P)]{\includegraphics[width = 0.2\linewidth]{image/PHISWIDtransformer.pdf}}
%     \subfloat[Transformer(U)]{\includegraphics[width = 0.2\linewidth]{image/UIEBtransformer.pdf}}
%     \subfloat[Transformer(L)]{\includegraphics[width = 0.2\linewidth]{image/LSUItransformer.pdf}}
%     \caption{PHISWID results using transformer. We use super-resolution to visualize clearly.}
%     %From left to right: Synthesized test images, restoration results by Transformer-P, restoration results by Transformer-C, restoration results by Transformer-U, restoration results by Transformer-L, restoration results by Deep WN-U, restoration results by Deep WN-P, restoration results by WaterNet, restoration results by U-shape, and restoration results by HLRP.}
%   \label{transformer_testresult}
% \end{figure*}

\section{Real-World Examples}
\label{sec:real_msr}
In this section, we present image enhancement results by methods trained by PHISWID for real-world underwater images.
We also discuss limitations.

\subsection{Real-world Underwater Image Enhancement}
%\subsubsection{Dataset 2: PHISWID dataset}
We compare the methods used in Section \ref{subsec:generalUWIE} through real-world underwater images.
% We use NIQE because it returns a low score when images are \textit{natural taken in atmospheric scenes}.

Fig. \ref{real_img_MSRCC} shows underwater image enhancement results for real-world images. All original images are collected from flickr\footnote{https://www.flickr.com/} under a Creative Commons AttributionNonCommercial-ShareAlike 2.0 Generic (CC BY-NC-SA 2.0) License.
We also show results for image taken from the UIEB challenging set \cite{li2019underwater} in Fig. \ref{realUIEB_img_MSRCC}.
We calculate the average NIQE of 66 real-world images, including Fig. \ref{real_img_MSRCC} and UIEB challenging sets, is summarized in Table \ref{real_niqe}.
% Fig. \ref{realUIEB_img_MSRCC} shows additional underwater image enhancement for real-world images of 60
% non-reference images in UIEB challenging set and objective performance (NIQE) are summarized in Table \ref{real_niqe}.
% We compare the methods used in Section \ref{subsec:generalUWIE}.
% using Transformer-P, Transformer-U, Transformer-L, Transformer-c, Deep WN-P, Deep WN-U, WaterNet, U-shape, and HLRP in the previous section (Table \ref{tab:compare}).
% In addition, Deep WN \cite{sharma2023wavelength} is applied to validate the regular underwater enhancement model works for marine snow removal.
% The other methods are taken from the original authors' GitHub repository.
% We also compare with Transformer trained with the dataset in \cite{ueda2019underwater}, i.e., synthesized images without marine snow (abbreviated as Transformer').
%The trained model of Deep WN is taken from the original authors' GitHub repository.
% using wavelength-based attributed deep neural network (WDN).
%

As shown in the figures, methods trained with PHISWID remove color shift sufficiently.
A significant difference between PHISWID and the other datasets can be observed in background regions. The PHISWID-based methods have a strong effect to remove the bluish color shift in the background.
%Since PHISWID (with and without marine snow) is designed to enhance underwater images as if they are shot in an atmospheric situation, the bluish regions of the resulting images are removed.
%In contrast, the enhanced images by methods trained by UIEB and LSUI still contain blue regions, especially in background.

The PHISWID-based methods also remove marine snow artifacts and small bubbles similar to marine snow artifacts. Fig. \ref{en_real_snow} shows enlarged portions of the enhancement results.
In contrast, Trans(P') and the methods trained with the existing datasets fail to remove the artifacts.
\subsection{Limitations}
\label{sec:limitations}
Although our PHISWID are carefully designed based on physical measurement processes, we still assume many parameters and settings as shown in Table~\ref{tab:params}: This is one of the limitations.

As observed in Fig. \ref{real_img_MSRCC}, thick marine snow artifacts still remain in the restored images.
The reason for this could be two-fold: First, the architectures of the neural networks are not specifically designed for underwater image enhancement with marine snow artifacts.
% MSR although we slightly customized its structure.
% A neural network designed specifically for marine snow removal would improve the performance.
Second, our proposed model classifies marine snow artifacts into two representative types.
% Marine snow artifacts not fitted to these types are not well removed.
%This limitation is described in the next section.
% based on synthesizing marine snow artificially into real underwater images, some limitations exist.
% First, the dataset is designed for removing marine snow artifacts and does not aim at achieving other underwater image enhancements such as color correction.
% The design of a simultaneous restoration of MSR and color correction is an interesting area of future study.
% The Transformer model trained on the PHISMID shows good results for removing marine snow. However, the dataset has some problems. 
Our modeling based on this classification could be incomplete since particles causing marine snow have various structures \cite{trudnowska2021marine}.
% does not include all the different types of marine snow that can happen.
%Therefore, the current dataset does not result in removing all marine snow artifacts.
% as shown in Fig. \ref{real_img_MSR}.
% Specifically, Transformer does not work well on images where the marine snow is larger or denser than the examples it trained on.

\begin{figure}[t]
%\vspace{-0.2in}
\centering
    \subfloat{\includegraphics[width = 0.2\linewidth]{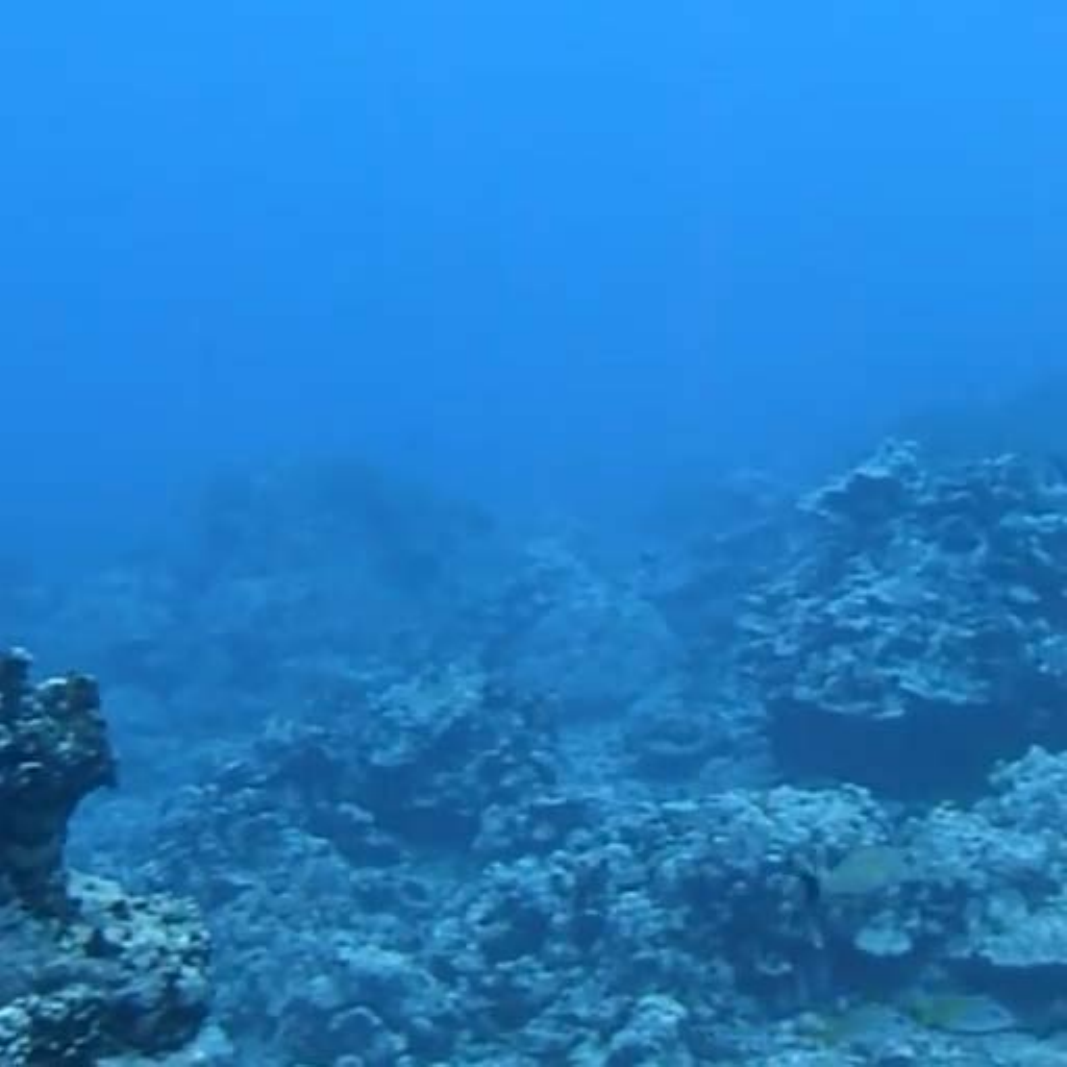}} 
    \subfloat{\includegraphics[width = 0.2\linewidth]{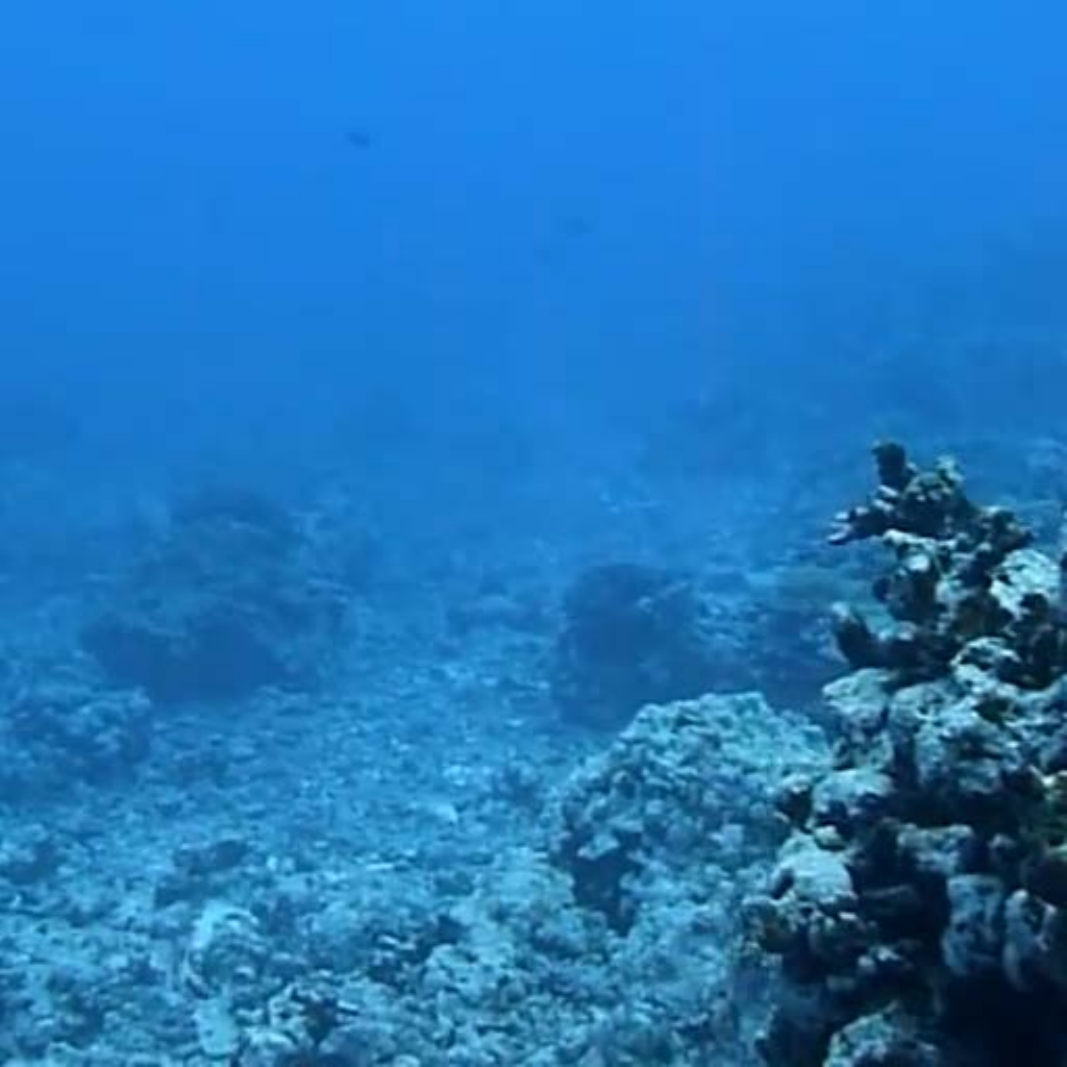}}
    \subfloat{\includegraphics[width = 0.2\linewidth]{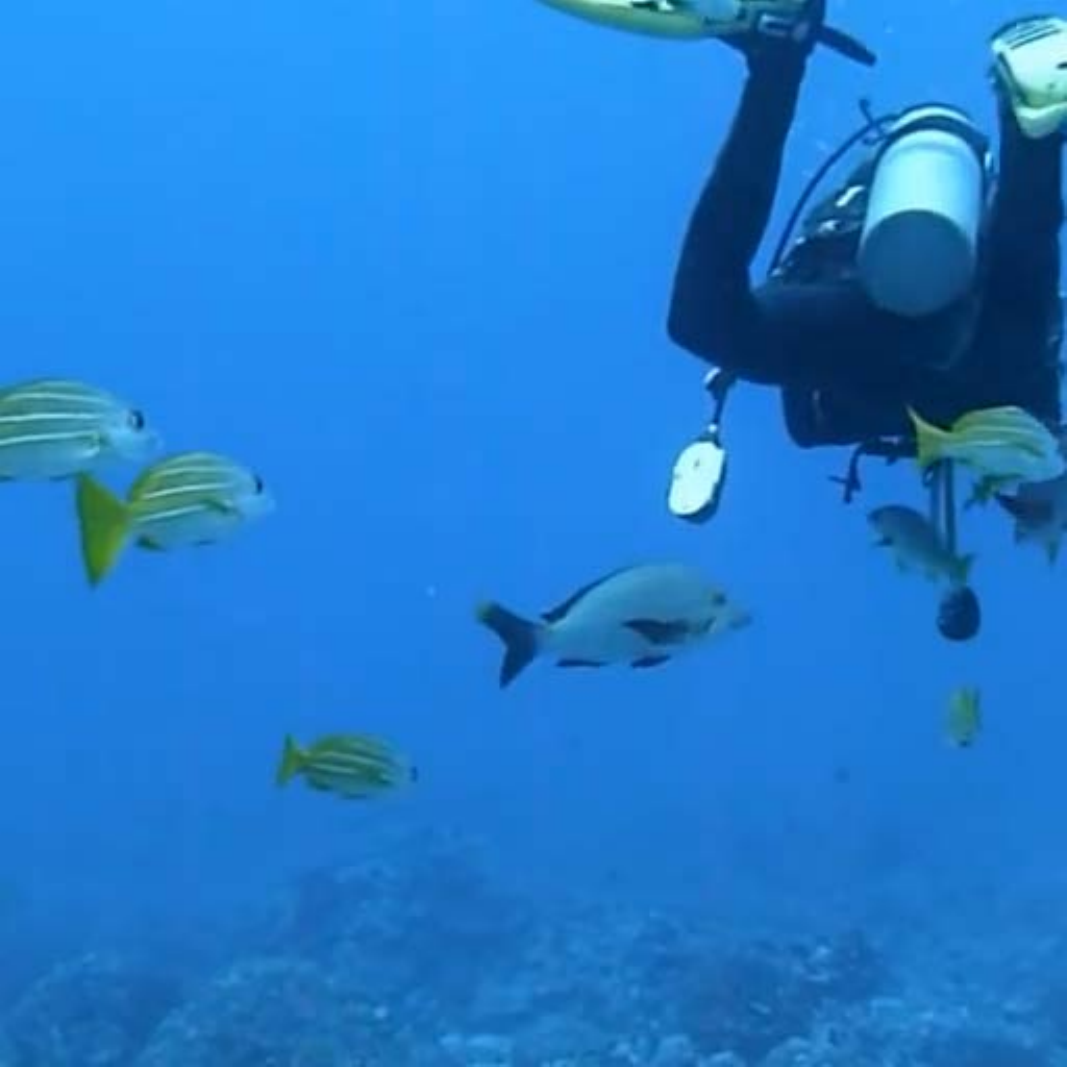}}\\\vspace{-0.15in}
    \subfloat{\includegraphics[width = 0.2\linewidth]{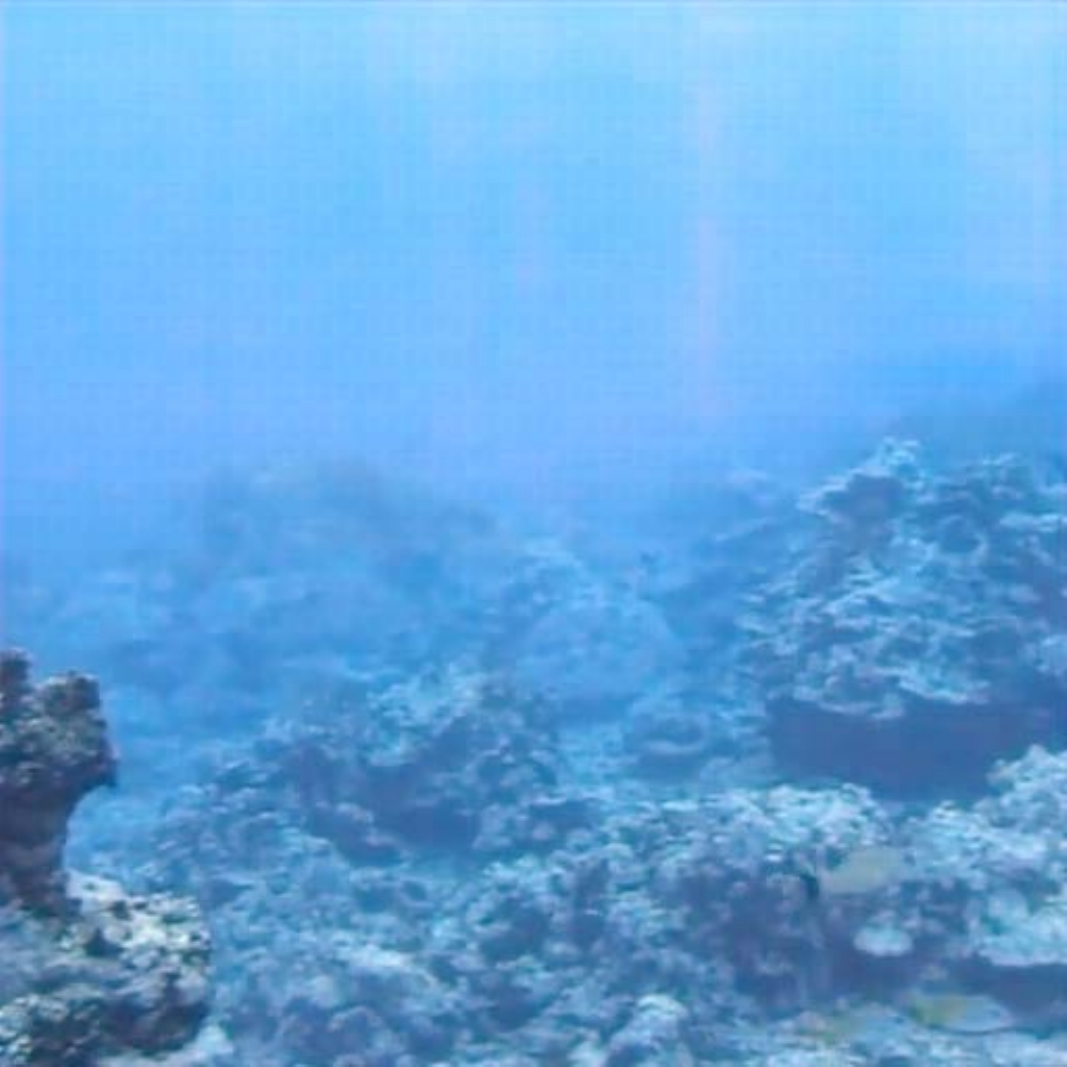}} 
    \subfloat{\includegraphics[width = 0.2\linewidth]{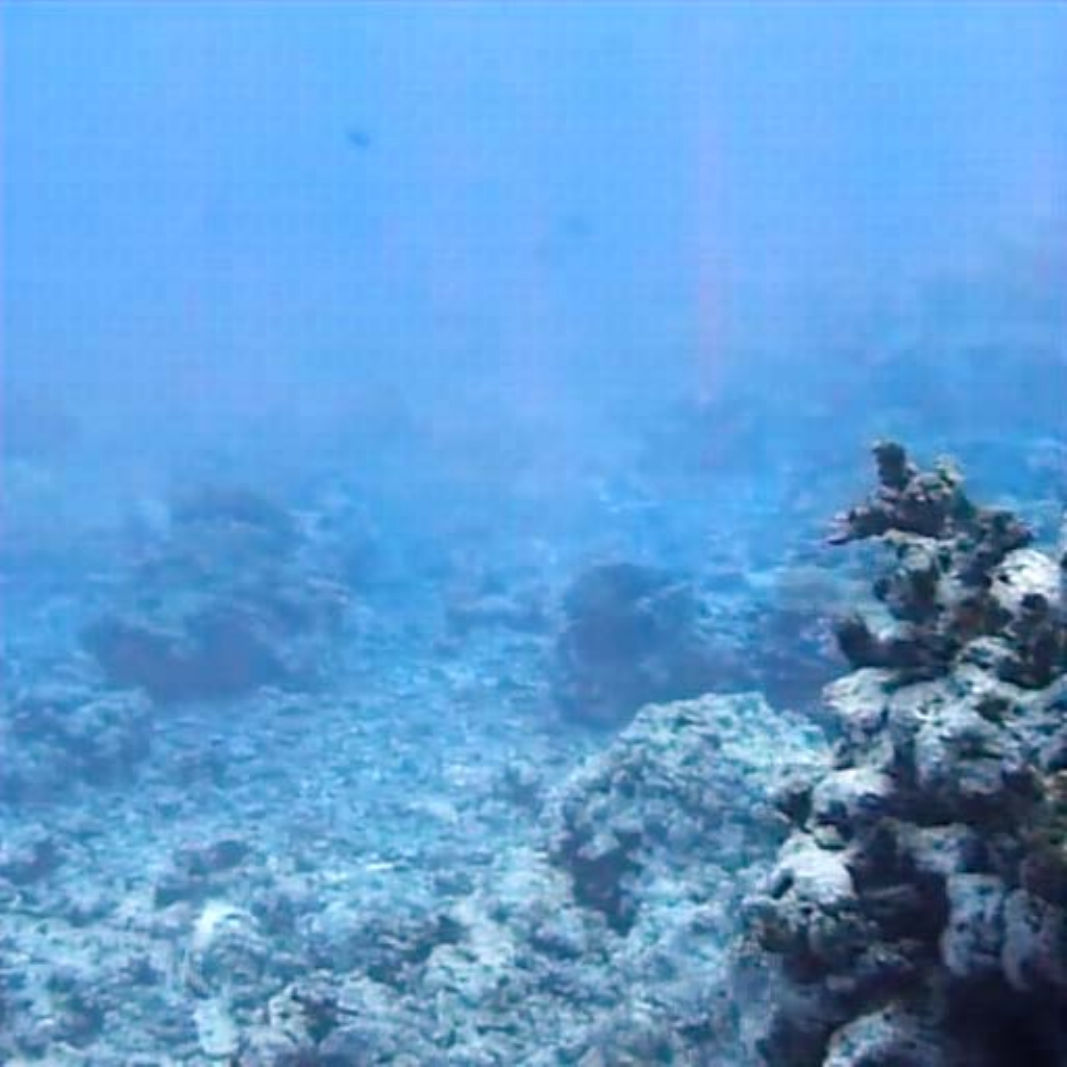}}
    \subfloat{\includegraphics[width = 0.2\linewidth]{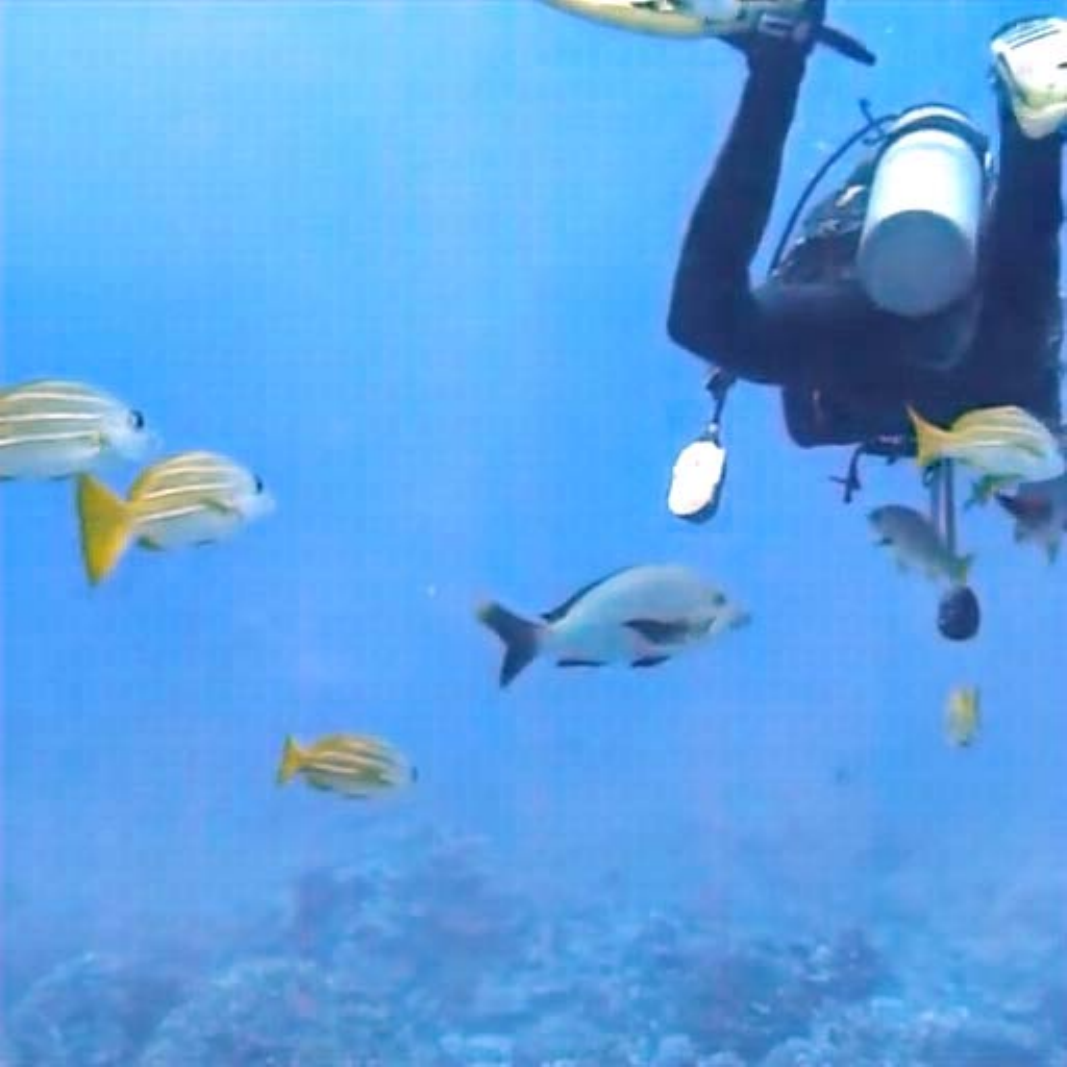}}
    \caption{Limitations. Top: Original images. Bottom: Restoration
results by Trans(P).}
  \label{limitation}
\end{figure}

Depths beyond our assuming ranges also affect the overall qualities.
% The generalizability of the dataset also requires more evaluation. 
If the light scattering is very different from the training images, color shift, particularly in the distant backgrounds, could not be well restored as shown in Fig. \ref{limitation}. A possible reason for this low performance is that the RGB-D images used in PHISWID are recorded in indoor situations which results in limited horizontal distances (i.e., depth) in scenes.
% the depths recorded in the indoor RGB-D images are limited.
Therefore it may not appropriately restore color shifts of distant backgrounds.

Our future work includes to update PHISWID for covering broader distance and degradation patterns.

% In summary - the current benchmark dataset for marine snow removal is limited in what types of marine snow and lighting variations it includes. So when Transformer is applied to new images that are different, the performance gets worse.To improve, the datasets need to expand to capture more diversity in marine snow size, density, and lighting conditions. This will make the Transformer model robust to more types of images, and work better in practice. In addition, simultaneously applying color correction can sometimes degrade image quality. 
% These are future works.
\section{Conclusions}
\label{sec:conclusion}
In this paper, we propose PHISWID for underwater image enhancement.
% PHISMID is designed for marine snow removal, which is overlooked artifacts in underwater images, and PHISWID is general underwater image enhancement containing marine snow artifacts.
The proposed dataset has a wide variety of underwater image degradation including overlooked effects of marine snow.
We physically modeled color shift and marine snow artifacts, and synthesized them in atmospheric RGB-D images.
% Two dataset, the removal of marine snow and marine snow and color shift were also proposed.
The first benchmarking is performed with PHISWID.
The benchmark suggested that methods aided our dataset can improve underwater image enhancement objectively and subjectively.

%\IEEEpubidadjcol

\appendices

% % you can choose not to have a title for an appendix
% % if you want by leaving the argument blank
% \section{}
\section{Detailed Derivation of The Jaffe-Macglamery Model}
\label{app:model}
We describe detailed derivations of the Jaffe-MacGlamery model introduced in Section \ref{light scattering}.

The following is the original equation in the main paper:
\begin{equation}
    E_{\mathrm{d}}(x, y)=E_{\mathrm{I}}\left(x, y, 0\right)e^{-\beta_c R} M\left(x, y\right) \cdot \frac{\cos ^4 \theta T_l}{4 f_n}\left[\frac{R-F_l}{R}\right]^2.
\label{originalequ}
\end{equation}
By substituting angles into \eqref{originalequ}, the equation is rewritten as:
\begin{equation}
\label{app:model_direct1}
E_{\mathrm{d}}(x, y)=E_{\mathrm{I}}\left(x, y, 0\right)e^{-\beta_c R} M\left(x, y\right) \cdot \frac{T_l}{4 f_n}\left[\frac{R-F_l}{R}\right]^2,
\end{equation}
where $E_{\mathrm{I}}\left(x, y, 0\right)$ is defined as \cite{mcglamery}:
\begin{equation}
\label{ei}
    E_{\mathrm{I}}\left(x, y, 0\right) = A' \cos \gamma \frac{e^{-\beta_c R}}{R^2}*p + A'\cos \gamma\frac{e^{-\beta_c R}}{R^2},
\end{equation}
% where p is defined as
% \begin{equation}
%     p=(e^{-GR_s} - e^{-cR_s})F^{-1}(e^{-BR_s f})
% \end{equation}
% and $F^{-1}$ is an inverse Fourier transform. Further transformations are as follows:
% \begin{equation}
%     E_{\mathrm{I}}\left(x^{\prime}, y^{\prime}, 0\right) = BP(0,0)\frac{e^{-cR_s}}{R_S^2}*(e^{-GR_s} - e^{-cR_s})\sqrt{2\pi}\delta(t-iBR_s) + BP(0,0)\frac{e^{-cR_s}}{R_S^2}
% \end{equation}
% in which $BP(0,0)$ is a constant.
% By letting $BP(0,0) = A'$, the above equation 
% \begin{equation}
% \label{ei}
%     E_{\mathrm{I}}\left(x^{\prime}, y^{\prime}, 0\right) = A'\frac{e^{-cR}}{R^2}*p + A'\frac{e^{-cR}}{R^2},
% \end{equation}
where $A'$ is a constant.
% and it comes from \eqref{}.
Substituting \eqref{ei} into \eqref{app:model_direct1}, we obtain:
% \begin{equation}
%     E_{\mathrm{d}}(x, y)=(A'\frac{e^{-cR}}{R^2}*p + A'\frac{e^{-cR}}{R^2})e^{-c R} M\left(x, y\right) \cdot \frac{\cos ^4 \theta T_l}{4 f_n}\left[\frac{R-F_l}{R}\right]^2
% \end{equation}
\begin{equation}
\begin{aligned}
    E_{\mathrm{d}}(x, y) = &(A'\frac{e^{-\beta_c R}}{R^2}*p +A'\frac{e^{-\beta_c R}}{R^2}))e^{-\beta_c R}\\
    &\times M\left(x, y\right)\frac{\cos ^4 \theta T_l}{4 f_n}\left[\frac{R-F_l}{R}\right]^2,
\end{aligned}
\end{equation}
where $M\left(x, y\right)$, $T_l$, and $f_n$ are constant. By merging all the above constant values, we finally obtain the following:
%To further summarize is following the equation:
\begin{equation}
    E_{\mathrm{d}}(x, y)=(A\frac{e^{-2\beta_c R}}{R^2}*p + A\frac{e^{-2\beta_c R}}{R^2})\left[\frac{R-F_l}{R}\right]^2.
\end{equation}

\section{Parameters for Marine Snow Models}
\label{app:param}
% These are the parameters used in the modeling of marine snow in 
Table \ref{tab:params} summarizes the parameters used for modeling marine snow artifacts.

\section{Transformer Structure}
\label{app:unet}
\begin{figure*}[t]
\centering
\includegraphics[width=\linewidth]{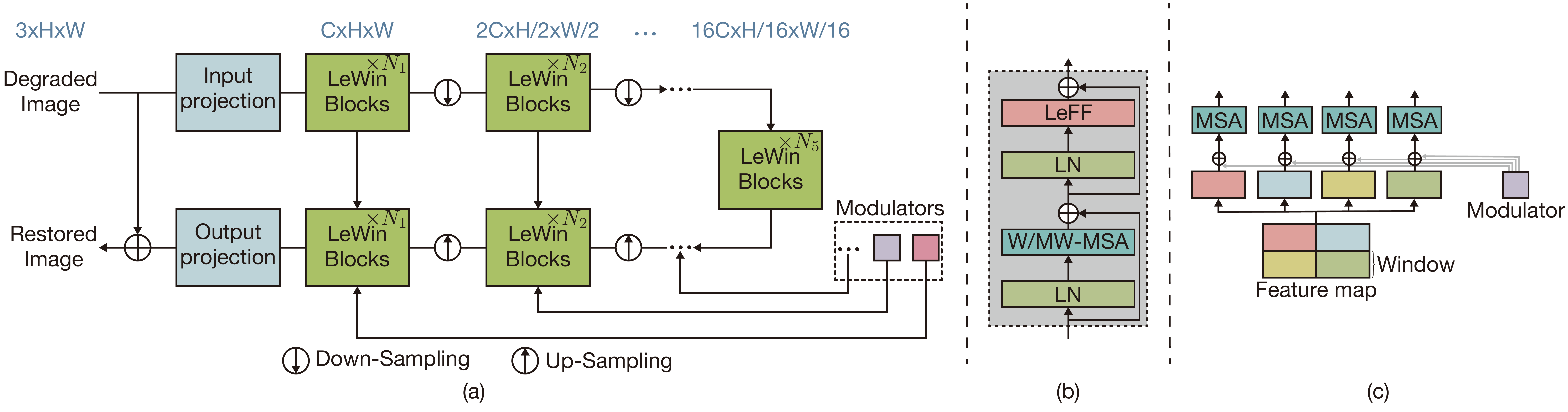}
\caption{(a) Uformer structure \cite{wang_2022_CVPR}. (b) LeWin block. (c) Illustration of how the modulators modulate the W-MSAs in each LeWin Transformer block (MW-MSA in (b)).}
\label{model}
\end{figure*}

The structure of Transformer \cite{wang_2022_CVPR} used for our underwater image enhancement experiments in Sections \ref{sec:benchmark} and \ref{sec:real_msr} are illustrated in Fig. \ref{model}. 
% This is basically the same as the original Transformer but we slightly changed the number of convolution layers and that of channels in each layer. 
The hyperparameters are summarized in Table \ref{para-trans}.

\begin{table*}[!t]
     \caption{Parameter settings for marine snow synthesis.}%where $\mathcal{U}(v_{\min}, v_{\max})$ is a continuous uniform distribution between $v_{\min}$ and $v_{\max}$.
    \label{tab:params}
    \centering
    \begin{tabular}{c|c|c|c}
    \hline
    Type & Parameter & Setting & Corresponding equation\\
    & & &  or subsection\\\hline
    H & \{$d_{-1,\text{horiz}}$, $d_{0,\text{horiz}}$, $d_{1,\text{horiz}}$, $d_{2,\text{horiz}}$\}  &   \{$0$, $64$, $128$, $192$\} & (5)\\
    % $d_0$ & $64$  & (5)\\
    % $d_1$ & $128$ & (5)\\
    % $d_2$  & $192$ & (5)\\
    H & \{$\sigma_1$, $\sigma_2$, $\sigma_3$, $\sigma_4$\} & \{7, 5, 3, 3\} & Sec. 3.2\\
    % $g(x,y;\sigma_2)$ & $\sigma_2$ = 5 & Sec. 3.2\\
    % $g(x,y;\sigma_3),g(x,y;\sigma_4)$ & $\sigma_3, \sigma_4$ = 3 & Sec. 3.2\\
    H & \{$a_1$, $a_2$, $a_3$, $a_4$\} & \{80, 100,150,200\} & (6)\\
    % $a_2$  & 100 & (6)\\
    % $a_3$ & 150 & (6)\\
    % $a_4$ & 200 & (6)\\
    H & threshold & 80 & (6)\\\hline
    V & \{$d_{-1,\text{horiz}}$, $d_{0,\text{horiz}}$, $d_{1,\text{horiz}}$, $d_{2,\text{horiz}}$\}  &   \{$0$, $64$, $128$, $192$\} & (5)\\
    % $E'_1$ in \eqref{4}     &  $192 \leq $matrix elements\\
    % $E'_2$ in \eqref{4}     & $128 \leq$ matrix elements $< 192$\\
    % $E'_3$ in \eqref{4}     & $64 \leq$ matrix elements $< 128$\\
    % $E'_4$ in \eqref{4}     &  matrix elements $< 64$\\
    V &\{ $\sigma_1$, $\sigma_2$, $\sigma_3$, $\sigma_4$ \} & \{7, 5, 4, 4\} & Sec. 3.2\\
    % $g(x,y;\sigma_2)$  & $\sigma_1$ = 5 & Sec. 3.2\\
    % $g(x,y;\sigma_3),g(x,y;\sigma_4)$  & $\sigma_3, \sigma_4$ = 4 & Sec. 3.2\\
    V & $\sigma$ & 0.2 & (7)\\
    V & \{$a'_1$, $a'_2$, $a'_3$, $a'_4$\}  & \{70, 80, 120, 150\} & (7)\\
    % $a'_2$  & 80 & (7)\\
    % $a'_3$  & 120 & (7)\\
    % $a'_4$ & 150 & (7)\\
    V & threshold & 28 & (7)\\\hline
    \end{tabular}
\end{table*}

% \begin{table}[t]
%   \centering
%   \caption{Parameter settings for Transformer.}
%   \label{para-unet}
%   \begin{tabular}{l|r} \hline
%   %& \multicolumn{2}{c}{Dataset 2}\\\hline
%   Parameter & Setting \\ \hline
%   Learning rate & 0.001 \\
%   Number of epochs & 150 \\
%   % Weight initializing method & \bf{23.52} \\\hline
%   Initialized bias values & 0 \\\hline
%   \end{tabular}
% \end{table}

\begin{table}[t]
  \centering
  \caption{Parameter settings for Transformer.}
  \label{para-trans}
  \begin{tabular}{l|r} \hline
  %& \multicolumn{2}{c}{Dataset 2}\\\hline
  Parameter & Setting \\ \hline
  Learning rate & 0.0001 \\
  Number of epochs & 60 \\
  % Weight initializing method & \bf{23.52} \\\hline
  Initialized bias values & 0 \\\hline
  \end{tabular}
\end{table}

\ifCLASSOPTIONcaptionsoff
  \newpage
\fi



%\bibliographystyle{IEEEtran}


% \end{thebibliography}
\bibliographystyle{IEEEtran}
\bibliography{egbib_new2}

% \end{IEEEbiography}


% % insert where needed to balance the two columns on the last page with
% %\newpage



%\vfill


\end{document}